\documentclass[11pt]{article}
\usepackage[preprint]{acl}
\usepackage{times}
\usepackage{latexsym}
\usepackage[T1]{fontenc}
\usepackage[utf8]{inputenc}
\usepackage{microtype}
\usepackage{inconsolata}
\usepackage{graphicx}
\usepackage{amsmath}
\usepackage{subcaption}
\usepackage{xspace}
\usepackage{url}
\usepackage{multirow}
\usepackage{makecell}
\usepackage{xcolor}
\usepackage{booktabs}
\usepackage{colortbl}
\usepackage{graphicx}
\usepackage{listings}
\usepackage{float}
\usepackage[para,online,flushleft]{threeparttable}
\usepackage{changes}

\newcommand{\ie}{\emph{i.e.,}\xspace}
\newcommand{\eg}{\emph{e.g.,}\xspace}

\newlength\savewidth

\title{Can LLM Safety Be Ensured by Constraining Parameter Regions?}

\author{
  Zongmin Li$^{1,2,3}$ \quad
  Jian Su$^{2}$ \quad
  Farah Benamara$^{4,5}$ \quad
  Aixin Sun$^{1}$ \\
  $^{1}$Nanyang Technological University, Singapore \\
  $^{2}$Institute for Infocomm Research (I$^2$R), A*STAR, Singapore \\
  $^{3}$CNRS@CREATE LTD, Singapore \\
  $^{4}$IRIT, Université de Toulouse, CNRS, Toulouse INP, Toulouse, France \\
  $^{5}$IPAL, CNRS-NUS-A*STAR, Singapore \\
\texttt{zongmin001@e.ntu.edu.sg}
}

\begin{document}
\maketitle
\begin{abstract}
Large language models (LLMs) are often assumed to contain ``safety regions'' -- parameter subsets whose modification directly influences safety behaviors. We conduct a systematic evaluation of four safety region identification methods spanning different parameter granularities, from individual weights to entire Transformer layers, across four families of backbone LLMs with varying sizes. Using ten safety identification datasets, we find that the identified safety regions exhibit only low to moderate overlap, as measured by IoU. The overlap drops significantly when the safety regions are further refined using utility datasets (\ie non-harmful queries). These results suggest that current techniques fail to reliably identify a stable, dataset-agnostic safety region.
\end{abstract}

\section{Introduction}
\label{sec:intro}

Large language models (LLMs) have achieved remarkable progress across a wide range of NLP tasks \cite{
wei2022emergentabilitieslargelanguage,
touvron2023llamaopenefficientfoundation,
openai2024gpt4technicalreport, geminiteam2025geminifamilyhighlycapable}. However, their deployment has raised increasing concerns about safety, including the risks of generating harmful, biased or misleading content. Despite extensive safety alignment efforts, recent work has demonstrated that LLMs remain vulnerable to various attacks. Specifically, jailbreaking can circumvent safety mechanisms through adversarial prompting \cite{yi2024jailbreakattacksdefenseslarge, wei2023jailbrokendoesllmsafety, zou2023universaltransferableadversarialattacks, ding2024wolfsheepsclothinggeneralized, deng2024multilingualjailbreakchallengeslarge, huang2023catastrophicjailbreakopensourcellms, kang2023exploitingprogrammaticbehaviorllms}, and fine-tuning can systematically erode safety behaviors even when using benign-appearing datasets \cite{DBLP:conf/iclr/Qi0XC0M024, yang2023shadowalignmenteasesubverting, lermen2024lorafinetuningefficientlyundoes, zhan-etal-2024-removing}. 

To remedy, recent research has begun to explore whether certain parameter regions within LLMs are systematically associated with safer or more harmful behaviors. Understanding these `safety regions' could enable targeted protection strategies, such as freezing critical parameters during downstream fine-tuning to maintain model safety while preserving task adaptability.

Currently, there is no consensus on what constitutes a safety region in LLMs or at what level of granularity it should be defined. For instance, \citet{icml} define a safety region as a collection of parameters. A parameter is the smallest unit of a weight matrix within each Transformer layer, corresponding to an individual scalar value learned by the model. In contrast, \citet{safetyneuron} identify safety regions at the level of neurons, referring to rows or columns of the weight matrix. \citet{safetylayer} instead define safety regions as specific consecutive intermediate layers of an LLM, while \citet{nlsr} focus on safety-critical parameters within LoRA adaptation matrices rather than the base model itself.

The proliferation of differing definitions suggests that the field lacks a shared understanding or a guideline of what properties a safety region should possess to be considered reliable and actionable. Inspired by the work on knowledge editing~\cite{zhang2024comprehensive,he2025benchmarkke}, we argue that a meaningful safety region should satisfy at least three criteria:

\begin{itemize}
    \item \textbf{Reliability:} Constraints or modifications applied to safety regions shall be reflected in corresponding changes in the model’s behavior when responding to safety-related questions (\eg the model becomes safer or more harmful). Such constraints or modifications may take the form of ablating, scaling (up or down), or freezing during harmful fine-tuning.
    \item \textbf{Locality:} Constraints (\eg ablating) applied to safety regions shall not affect the model’s capability in other areas, such as non-safety-related functions.
    \item \textbf{Convergent Identifiability:} A safety region should be an intrinsic property of the model’s parameter space and should remain invariant across the safety datasets used to identify it, which we refer to as \textit{identification datasets}. In other words,  even when the identification datasets represent distinct categories of harmful content, they should yield convergence toward an ideally identical underlying region within the parameter space.
\end{itemize}

In the current literature, most of the work has verified \textit{Reliability} and \textit{Locality} to some extent. However, the \textit{convergent identifiability} property remains unverified. Some existing methods \cite{safetyneuron, safetylayer, nlsr} use a single dataset to identify safety regions. Other work \cite{icml} sample datasets multiple times to identify safety regions, but does not check their consistency. 

\begin{figure}
    \centering
    \includegraphics[trim={1.5cm 5.5cm 15cm 5cm},clip, width=0.5\textwidth]{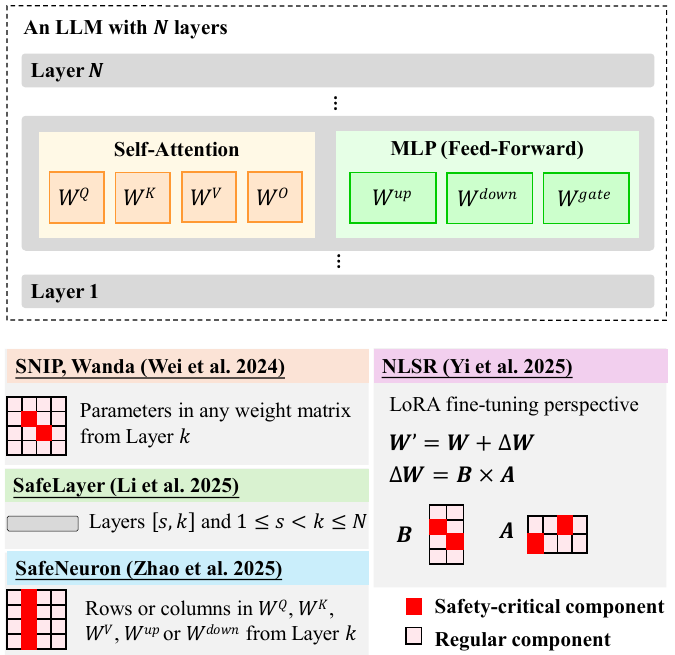}
    \caption{Overview of current safety region identification methods}
    \label{fig:overview}
\end{figure}

In this paper, through carefully designed experiments, we evaluate the \textit{convergent identifiability} of four recent methods that claim to identify safety regions: SNIP \& Wanda~\cite{icml}, SafeNeuron~\cite{safetyneuron}, SafeLayer~\cite{safetylayer}, and NLSR~\cite{nlsr}. These four methods collectively cover a good range of safety region granularities (parameter-, neuron-, layer-, and LoRA-weight-level) as illustrated in Fig.~\ref{fig:overview}. We aim to address two research questions.

\begin{itemize}
    \item \textbf{RQ1:} As the number of identification datasets increases, does the overlap among the identified safety regions converge to a stable region in the parameter space?
    \item \textbf{RQ2:} To what extent does the distribution of the identification dataset (\eg semantic similarity) influence the identified safety regions?
\end{itemize}

Specifically, following each method, we construct multiple safety datasets, each used to identify a safety region within a target LLM. We then evaluate the extent to which these identified regions overlap, quantified by the Intersection over Union (IoU) metric.

In our experiments, we were unable to reproduce the reported results for safety regions defined at the layer level. For the remaining three methods, the safety regions identified across multiple datasets, even when sampled from the same larger dataset, exhibit lower-than-expected IoU. The IoU decreases even further after refinement using utility datasets, reflecting the persistent entanglement of safety and utility within the model’s parameter space. These empirical findings suggest that a well-defined safety region may not exist, or that current methods are insufficient to reliably identify one.

Our studies carry significant implications for the broader study of LLM safety. A central motivation behind locating safety regions is to constrain them during subsequent adaptation—such as fine-tuning—in order to prevent the degradation of safety behaviors. Yet our analysis demonstrates that safety regions identified across different datasets are highly fragmented, with overlaps failing to converge. This lack of stability may pose fundamental challenges for applying reliable constraints in practice and therefore warrants further study. We therefore suggest that our results serve to recalibrate prevailing assumptions about the existence of safety regions, and highlight the need for future research to design more principled and robust identification methods capable of supporting durable safety alignment.
\section{Preliminaries}
\label{sec:prelim}

We now briefly review the four methods examined in this study to provide background for our empirical analysis. We encourage readers to refer to the original papers for full technical details.

\subsection{SNIP \& Wanda, SafeNeuron and NLSR}
\label{sec:prelim_subsec2_1}

SNIP \& Wanda, SafeNeuron, and NLSR identify safety regions at the levels of parameters, neurons, and LoRA weights, respectively. All three methods share a common principle: assigning importance scores to safety-critical components to quantify their contribution to the model’s safety behavior. Moreover, the same scoring mechanism can be applied to identify a utility-related parameter region (or utility region) using a utility dataset. As illustrated in Fig.~\ref{fig:overview}, the red-highlighted components correspond to safety-critical elements, while the non-highlighted ones represent regular components. Since each method ranks model components (parameters, neurons, or LoRA weights) by their importance scores, a threshold is defined to select those components to form a safety region. Below, we briefly introduce the scoring mechanism used by each method.

\paragraph{SNIP \& Wanda}~\cite{icml}. A \textbf{parameter} refers to an individual scalar value within a weight matrix of each Transformer layer \cite{transformer} in the target LLM. Both SNIP \cite{snip} and Wanda \cite{wanda} assign importance scores to individual parameters based on their contribution to the model’s safety behavior, specifically, its ability to generate refusal responses to harmful queries. For each parameter, SNIP measures importance according to the change in loss, whereas Wanda measures it according to the change in output. Within each weight matrix of the target LLM, the safety region is defined as the set of parameters with the top $q\%$ SNIP or Wanda scores.

\paragraph{SafeNeuron}~\cite{safetyneuron}. Here, \textbf{neuron} refers to a row or a column of parameters within $W^Q, W^K, W^V, W^{up}, W^{down}$ at each Transformer layer of the targeted LLM.

Given a harmful query as input, the importance score of each neuron is defined by the difference in internal representation when that neuron is removed. The activated neurons in each weight matrix are those with the top $m$ importance scores. Finally, the safety region in each weight matrix is defined as the set of neurons that are consistently activated across harmful queries in the identification dataset.

\paragraph{NLSR}~\cite{nlsr}. Safety regions in this method are identified within LoRA weight update matrices $A$ and $B$. Given a base LLM, they keep the weights frozen and update only the LoRA weights to obtain a super safety-aligned model, via weak-to-strong extrapolation \cite{extrapolation}. Next, they applied rank reduction to LoRA weights to preserve the safety-critical weights. Finally, for each lower-rank approximation of $A$ and $B$, the safety region is defined by the composition of parameters with top $t\%$ magnitude values.

\subsection{SafeLayer~\cite{safetylayer}}
\label{sec:prelim_subsec2_2}

SafeLayer identifies a small set of consecutive layers in the middle of the model as safety regions. A \textbf{layer} is a Transformer layer, consisting of a self-attention module, a feed-forward network (aka MLP), and their associated normalization and residual connections. 

Instead of directly scoring individual layers as safety components, as in the three aforementioned methods (Section \ref{sec:prelim_subsec2_1}), SafeLayer adopts a different strategy. First, a layer-wise cosine similarity analysis of the model’s internal output representations to harmful and utility inputs is conducted, revealing that certain middle layers function as a safety differentiation zone. These layers are where LLMs begin to distinguish between harmful and utility queries, which provides evidence for the existence of safety layers. Second, building on the initial range of layers, the authors hypothesize that scaling the safety layers will affect the model’s responses to over-rejection queries. They further leverage the over-rejection phenomenon to precisely locate these safety layers.
\section{Methodology}
\label{sec:methodology}

We systematically evaluate the four safety region identification methods by deriving safety regions from multiple identification datasets and measuring their cross-dataset consistency. Building on the two \textbf{RQ}s stated in Section~\ref{sec:intro}, we evaluate the \textit{convergent identifiability} via cross-dataset safety region overlap under varying dataset number and diversity.

To answer \textbf{RQ1}, we design the following variants of measuring the \textit{convergent identifiability}. Section~\ref{sec:method1} explains the overlap among safety regions. Next, considering the intricate relationship between safety and utility, we measure the overlap among utility-isolated safety regions (Section~\ref{sec:method2}). We use the classical metric, intersection over union (Section~\ref{sec:metric}), to measure the overlap.

To answer \textbf{RQ2}, we design two identification dataset variants: (i) datasets spanning diverse harm categories, and (ii) datasets where each one corresponds to a single distinct harm category. For both variants, we follow Sections~\ref{sec:method1} and~\ref{sec:method2} to evaluate the \textit{convergent identifiability}. 

\subsection{Safety Region Overlap}
\label{sec:method1}

Suppose there are $n$ datasets of safety related samples $\mathcal{D}_0, \mathcal{D}_1, \dots, \mathcal{D}_{n-1}$. In principle, we can apply a safety identification method on each dataset to identify a corresponding safety region, $\mathcal{R}_0, \mathcal{R}_1, \dots, \mathcal{R}_{n-1}$. If the property of \textit{convergent identifiability} holds, these 
$n$ safety regions should exhibit substantial overlap. Therefore, for each safety region identification method, we measure the degree of overlap among the identified safety regions
 $\{\mathcal{R}_i\}_{i=0}^{n-1}$.

\subsection{Utility-Isolated Safety Region Overlap}
\label{sec:method2}

As \citet{icml} pointed out, to understand a harmful query, it requires both safety awareness and utility capability of the model. Also, \citet{foundation_neuron} identified that some utility-related neurons are responsible for the fundamental management of queries. During downstream task fine-tuning, modifications to these neurons can inadvertently affect the safety neurons that overlap, which degrades the safety mechanism. Within safety regions, to examine the \textit{convergent identifiability} of the components that contribute solely to safety, we propose measuring the utility-isolated safety region overlap.

Specifically, for each safety region identification method, we follow Section~\ref{sec:method1} to obtain $n$ safety regions $\{\mathcal{R}_i\}_{i=0}^{n-1}$. Then, we take a utility dataset $\mathcal{D}_u$ to obtain the utility region $\mathcal{R}_u$ with the same identification method. For each safety region $\mathcal{R}_i$, the utility-isolated safety region will be $\mathcal{R}_i - \mathcal{R}_u$, \ie the set difference. Finally, we measure the overlap across these $n$ utility-isolated safety regions $\{\mathcal{R}_i - \mathcal{R}_u \}_{i=0}^{n-1}$.

\subsection{Evaluation Metric} 
\label{sec:metric}

To measure the overlap across the $n$ safety regions $\{\mathcal{R}_i\}_{i=0}^{n-1}$, we use \textbf{IoU} (intersection over union) as the evaluation metric. Specifically, $\forall i \in [0, n-1]$, $\mathcal{R}_i$ is a safety region composed of certain components (safety-critical parameters, neurons, layers or LoRA weights). Let $\mathcal{S}_i$ be the set of indices for each safety-critical component of $\mathcal{R}_i$, then the \textbf{IoU} to measure safety region overlap is
\begin{equation}
\mathrm{IoU}(\mathcal{S}_0, \mathcal{S}_1, \dots, \mathcal{S}_{n-1}) = \frac{\bigcap_{i=0}^{n-1} \mathcal{S}_i}{\bigcup_{i=0}^{n-1} \mathcal{S}_i} 
\end{equation}
The same applies to the utility-isolated safety regions.
\section{Experimental Setup}
\label{sec:exp}

We evaluate the four safety region identification methods introduced in Section~\ref{sec:prelim}, chosen for their diverse definitions and publicly available implementations. All experiments follow the original implementations and hyperparameters.

\subsection{Identification Datasets} 
\label{sec:exp_model_dataset}
We construct two sets of identification datasets, to be detailed next. 
For each method, we then apply these datasets to the same target LLMs used in the original paper. 
We further include additional model families where applicable, most notably the Qwen series~\cite{qwen2025qwen25technicalreport}, to align model coverage across all methods (Table~\ref{tab:models}).

\paragraph{Multi-Category Dataset} We construct $n=10$ identification datasets, each covering diverse harm categories, following the original implementation. 
For each method, the dataset used in the original paper contains harmful queries across various categories, denoted as $\mathcal{D}_0$. 
The sources of $\mathcal{D}_0$ for each method are detailed in Appendix~\ref{appendix:d0}.
The sampled datasets $\mathcal{D}_i(i=1,\ldots,9)$ follow the same data format and size as $\mathcal{D}_0$ ($|\mathcal{D}_i|=|\mathcal{D}_0|$). 
Table~\ref{tab:models} reports $|\mathcal{D}_0|$ for each method. We sample $\{\mathcal{D}_i\}_{i=1}^{9}$ as follows:

\begin{itemize}
    \item SNIP \& Wanda: $\mathcal{D}_0$\footnote{In \citet{icml}, $\mathcal{D}_0$ is sampled from a source dataset at each time. In our evaluation, we prepare $5$ different seeds to sample $\mathcal{D}_0$ and report the average IoU (Table~\ref{tab:snip_direct}-\ref{tab:wanda_disentangle}).} is the \emph{safety-short} variant, where each harmful query is paired with a judgement-segment, refusal-only answer. $\{\mathcal{D}_i\}_{i=1}^{9}$ are sampled from the training split of PKU-SafeRLHF-QA~\cite{beavertails,pkusaferlhf}\footnote{The dataset was created mainly for the purpose of reinforcement learning with human feedback. In our setting, we simply need questions and answers in the context of safety. Hence we use this dataset.}, restricted to the highest severity level. Each $D_i$ contains harmful queries spanning multiple categories.

    \item SafeNeuron: $\{\mathcal{D}_i\}_{i=1}^{9}$ are sampled from the same PKU-SafeRLHF-QA training split, restricted to the highest severity level. Each $\mathcal{D}_i$ contains harmful queries of diverse categories.

    \item NLSR: $\{\mathcal{D}_i\}_{i=1}^{9}$ are sampled from the training split of PKU-SafeRLHF-30K \cite{beavertails, pkusaferlhf}. Each $\mathcal{D}_i$ contains harmful queries of diverse categories, paired with refusal answers.

    \item SafeLayer: Details of the identification datasets are provided in Appendix~\ref{appendix: safelayer}.
\end{itemize}

To evaluate utility-isolated safety region overlap (Section \ref{sec:method2}), we take the utility dataset $\mathcal{D}_u$ from \citet{icml}. This dataset is based on Alpaca-Cleaned\footnote{https://github.com/gururise/AlpacaDataCleaned}, a refined version of the Alpaca dataset \cite{alpaca}, where safety-related samples were filtered out using sensitive phrase matching \cite{DBLP:conf/iclr/Qi0XC0M024}. For each method, $\mathcal{D}_u$ preserves the same data format as $\mathcal{D}_0$, with $|\mathcal{D}_i|=|\mathcal{D}_0|$. 

\paragraph{Single-Category Dataset}

We construct $n_c$ identification datasets for SNIP \& Wanda and SafeNeuron\footnote{We exclude NLSR, as we could not find sufficient data samples from available resources to construct $\mathcal{D}_{c_i}$.}. For each method, each $\mathcal{D}_{c_i}$ preserves the same data format as $\mathcal{D}_0$, with $|\mathcal{D}_{c_i}|=|\mathcal{D}_0|$.

\begin{itemize}
    \item SNIP \& Wanda: we sample $n_c=5$ datasets from PKU-SafeRLHF-QA, restricted to the highest severity level. Each $\mathcal{D}_{c_i}$ contains $128$ data samples from one single harm category $c_i$.

    \item SafeNeuron: we sample from PKU-SafeRLHF-QA. As the highest severity level alone does not provide enough data samples, we extend our sampling to include both the highest and second highest severity levels, which results in $n_c=12$ datasets. Each $\mathcal{D}_{c_i}$ contains $200$ data samples from one single harm category $c_i$.    
\end{itemize}

The specific harm categories $\{c_i\}_{i=1}^{n_c}$ for each method are presented in Appendix~\ref{appendix: category_detail}.

\begin{table}
\footnotesize
\centering
\caption{Targeted LLMs and datasets size for each method. SafeLayer involves two stages of identification, with $100$ samples in Stage $\mathrm{I}$ and $731$ samples in Stage $\mathrm{II}$.}
\label{tab:models}
\begin{tabular}{ll@{\hspace{0.1cm}}r}
\toprule
\textbf{Method} & \textbf{Targeted LLMs}  & \# \textbf{Samples} \\
\midrule

\multirow{4}{*}{\makecell[l]{SNIP \& Wanda \\ \cite{icml}}} & Llama-2-7B-Chat & \multirow{4}{*}{$128$}\\
& Llama-2-13B-Chat &  \\
& Mistral-7B-Instruct-v0.2 &  \\
& Qwen2.5-7B-Instruct &  \\
\addlinespace[0.5em]

\multirow{4}{*}{\makecell[l]{SafeNeuron \\ \cite{safetyneuron}}} 
& Llama-2-7B-Chat & \multirow{4}{*}{$200$}\\
& Llama-3-8B-Instruct & \\
& Mistral-7B-Instruct-v0.2  & \\
& Qwen2.5-7B-Instruct  & \\
\addlinespace[0.5em]

\multirow{4}{*}{\makecell[l]{SafeLayer \\ \cite{safetylayer}}} 
& Llama-2-7B-Chat & \multirow{4}{*}{\makecell[r]{$100$ \\ $731$}}\\
& Llama-3-8B-Instruct & \\
& gemma-2B-IT & \\
& Phi-3-mini-4k-instruct & \\
\addlinespace[0.5em]

\multirow{3}{*}{\makecell[l]{NLSR \\ \cite{nlsr}}}
& Llama-3-8B  & \multirow{3}{*}{$2000$} \\
& Mistral-7B-v0.2 & \\
& Qwen2.5-7B & \\

\bottomrule
\end{tabular}
\end{table}

\subsection{Thresholds}
\label{sec:other}

For SNIP \& Wanda, SafeNeuron and NLSR, certain thresholds are necessary to select the top important safety-critical components. In our evaluation, we included the same threshold settings as in the original work. For NLSR, $t\%=20\%$; for SafeNeuron\footnote{Value taken from the provided implementation: https://github.com/zhaoyiran924/Safety-Neuron}, $m$ is $1000$ for $W^Q, W^K, W^V$ and $2000$ for $W^{up}$ and $W^{down}$. We apply the same thresholds to identify the utility region $\mathcal{R}_u$. For SNIP \& Wanda, we follow the thresholds in the original paper and add additional choices: $q\%$ for identifying the safety region and $p\%$ for the utility region. All choices are shown in Table~\ref{tab:snip_direct}-\ref{tab:wanda_disentangle}.
\begin{table*}
\centering
\caption{\textit{Convergent identifiability} of existing safety region identification methods with 10 public multi-category datasets: safety region overlap (\emph{IoU}) and utility-isolated safety region overlap (\emph{Iso-Utility IoU}).}
\label{tab:results_main}
\begin{tabular}{lllrr}
\toprule
\textbf{Method} & \textbf{Safety Region } & \textbf{Targeted Model} & \textbf{\makecell[c]{IoU}} & \textbf{\makecell[c]{Iso-Utility IoU}} \\
\midrule
\multirow{4}{*}{\makecell[l]{SNIP Score \\ \cite{icml}}} & \multirow{4}{*}{Parameter} & Llama-2-7B-Chat & $0.29$ - $0.42$ & $0.17$ - $0.24$ \\
& & Llama-2-13B-Chat & $0.28$ - $0.42$ & $0.15$ - $0.24$ \\
& & Mistral-7B-Instruct-v0.2 & $0.30$ - $0.42$ & $0.20$ - $0.26$ \\
& & Qwen2.5-7B-Instruct & $0.35$ - $0.46$ & $0.21$ - $0.26$ \\
\midrule
\multirow{4}{*}{\makecell[l]{Wanda Score \\ \cite{icml}}} & \multirow{4}{*}{Parameter} & Llama-2-7B-Chat & $0.57$ - $0.72$ & $0.27$ - $0.35$ \\
& & Llama-2-13B-Chat & $0.55$ - $0.70$ & $0.25$ - $0.33$ \\
& & Mistral-7B-Instruct-v0.2 & $0.56$ - $0.69$ & $0.28$ - $0.37$ \\
& & Qwen2.5-7B-Instruct & $0.57$ - $0.69$ & $0.26$ - $0.37$ \\
\midrule
\multirow{4}{*}{\makecell[l]{SafeNeuron \\ \cite{safetyneuron}}} & \multirow{4}{*}{Neuron} & Llama-2-7B-Chat & $0.62$ & $0.16$ \\
& & Llama-3-8B-Instruct & $0.58$ & $0.19$ \\
& & Mistral-7B-Instruct-v0.2 & $0.57$ & $0.17$ \\
& & Qwen2.5-7B-Instruct & $0.48$ & $0.14$ \\
\midrule
\multirow{4}{*}{\makecell[l]{SafeLayer \\ \cite{safetylayer}}} & 
\multirow{4}{*}{Layer} & Llama-2-7B-Chat &  
\multirow{4}{*}{\makecell[r]{Unable to \\ identify}} &  
\multirow{4}{*}{\makecell[r]{Unable to \\ identify}} \\
& & Llama-3-8B-Instruct &  &  \\
& & gemma-2B-IT &  &  \\
& & Phi-3-mini-4k-instruct &  &  \\
\midrule
\multirow{3}{*}{\makecell[l]{NLSR \\ \cite{nlsr}}} & \multirow{3}{*}{LoRA Weights} & Llama-3-8B & $0.39$ & $0.08$ \\
& & Mistral-7B-v0.2 & $0.01$ & $0.001$ \\
& & Qwen2.5-7B & $0.45$ & $0.09$ \\
\bottomrule
\end{tabular}
\end{table*}

\begin{figure*}[htbp!]
    \centering
    \begin{subfigure}[b]{0.24\textwidth}
        \includegraphics[width=\linewidth]{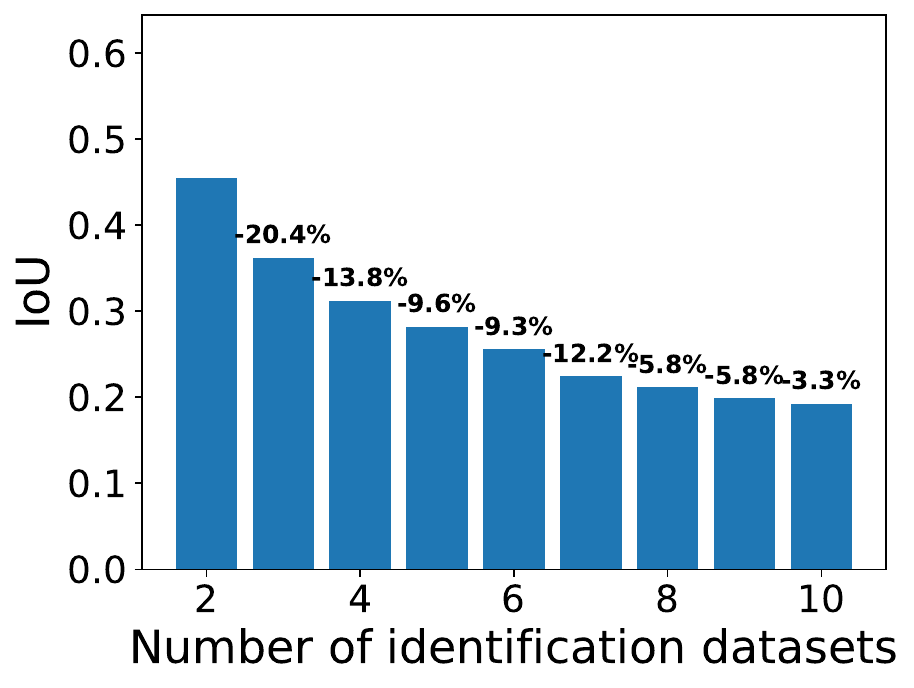}
        \caption{Iso-Utility IoU vs $n$, forward order}
        \label{fig:results_main_safeneuron_4_figs_1}
    \end{subfigure}
    \hfill
    \begin{subfigure}[b]{0.24\textwidth}
        \includegraphics[width=\linewidth]{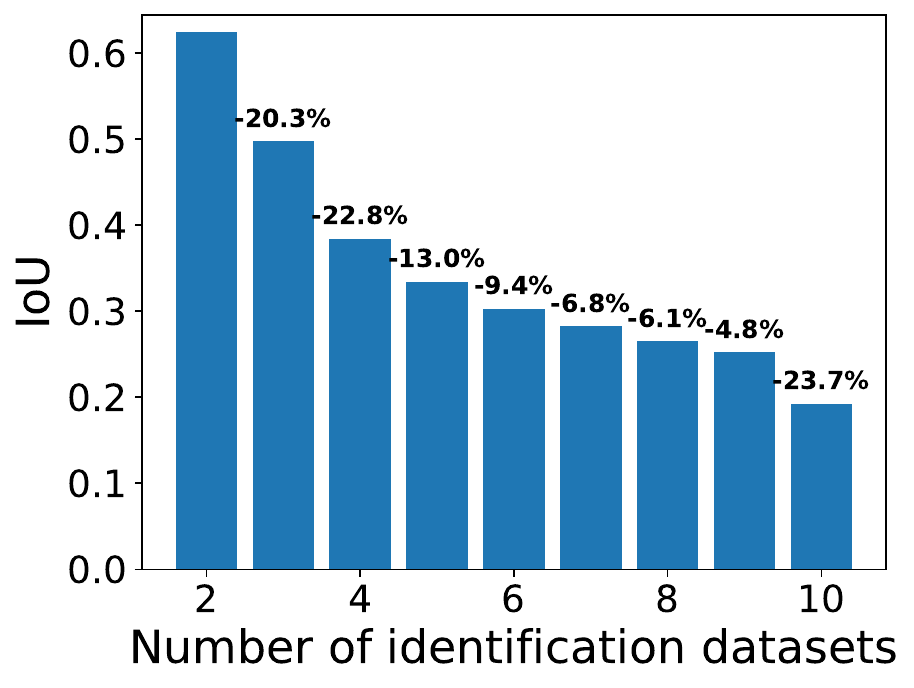}
        \caption{Iso-Utility IoU vs $n$, backward order}
        \label{fig:results_main_safeneuron_4_figs_2}
    \end{subfigure}
    \hfill
    \begin{subfigure}[b]{0.24\textwidth}
        \includegraphics[width=\linewidth]{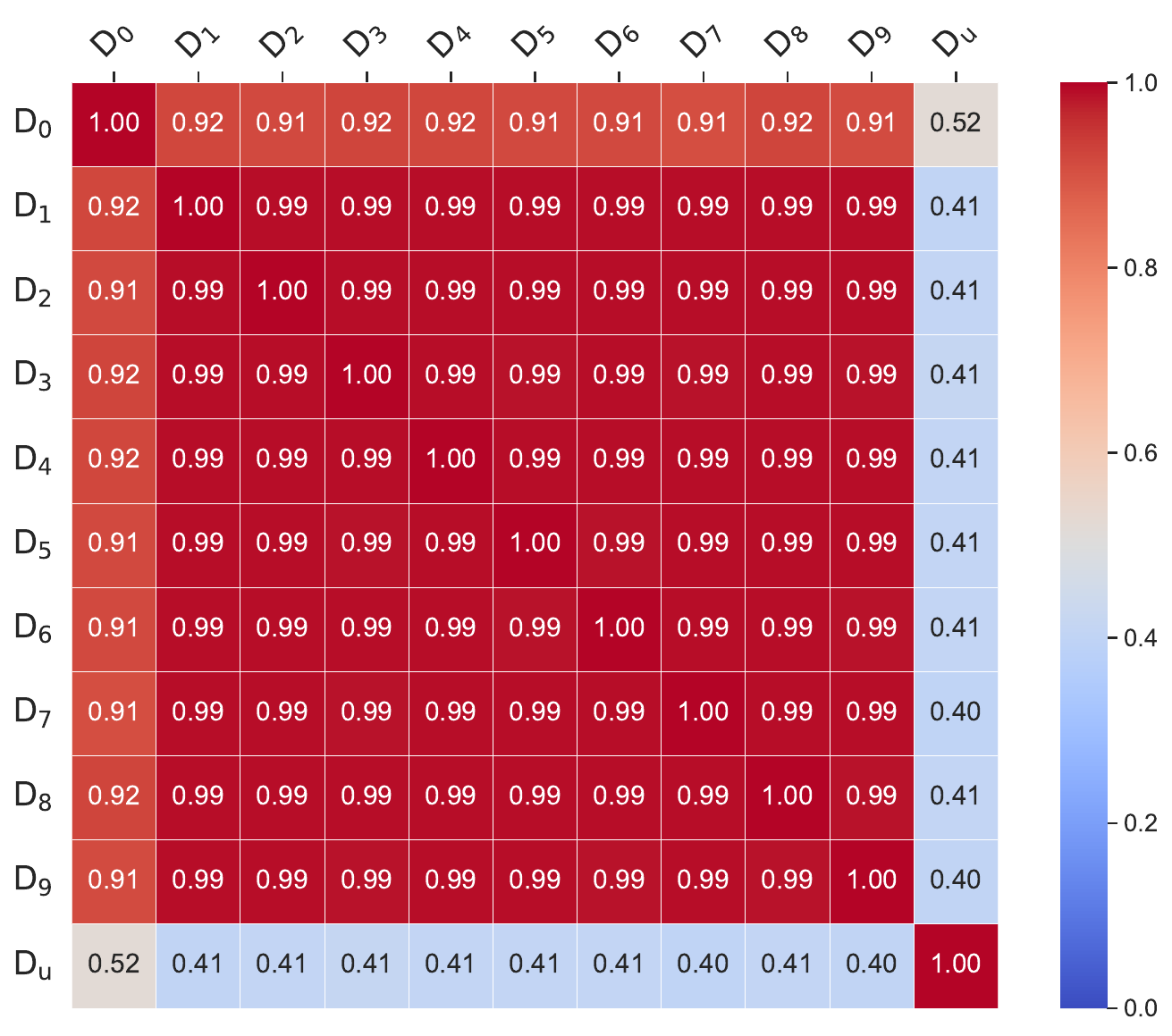}
        \caption{Pairwise cosine similarity analysis for $\{\mathcal{D}_i\}_{i=0}^9$ and $\mathcal{D}_u$}
        \label{fig:results_main_safeneuron_4_figs_3}
    \end{subfigure}
    \hfill
    \begin{subfigure}[b]{0.24\textwidth}
        \includegraphics[width=\linewidth]{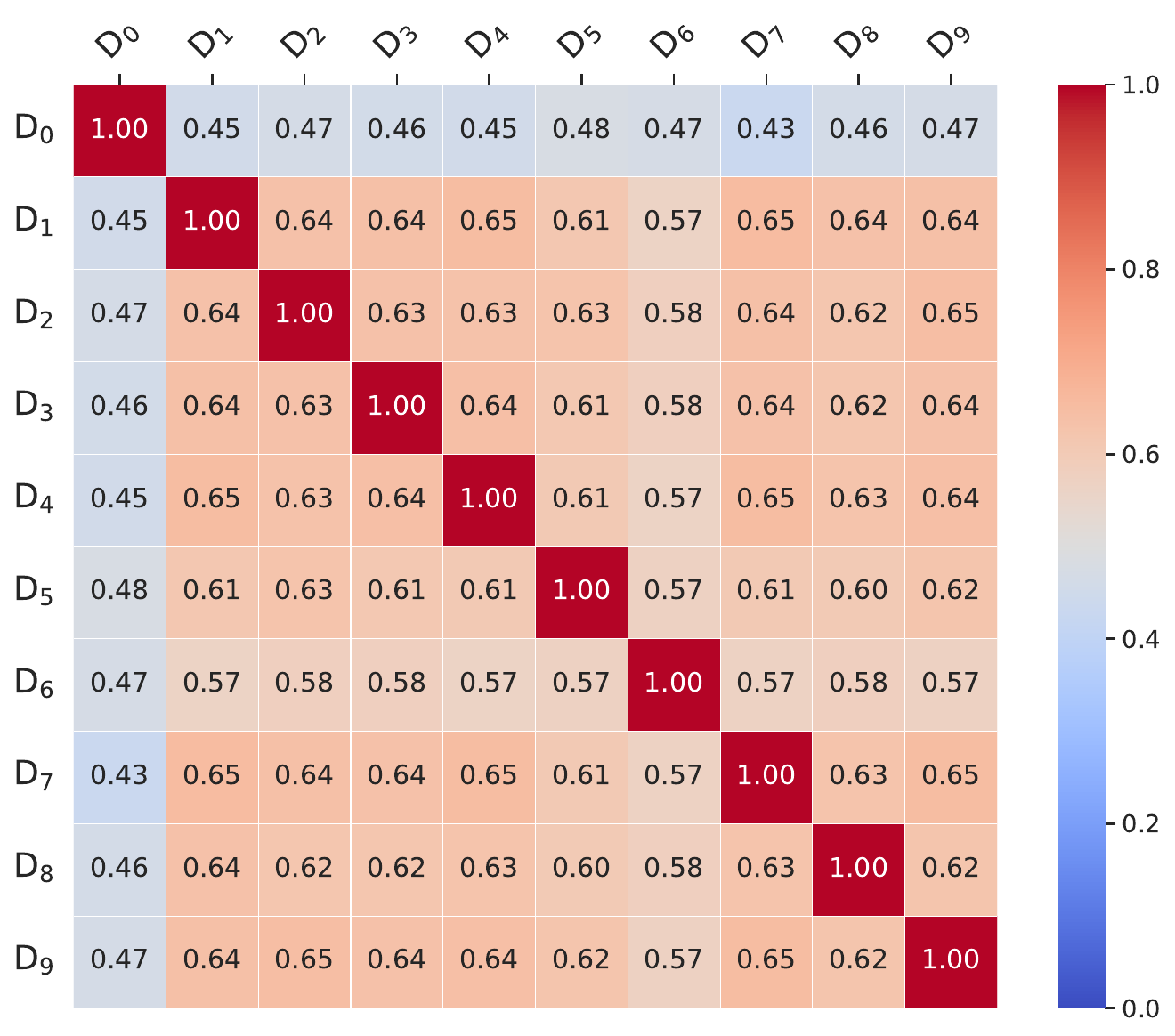}
        \caption{Pairwise Iso-Utility IoU for $\{\mathcal{D}_i\}_{i=0}^9$}
        \label{fig:results_main_safeneuron_4_figs_4}
    \end{subfigure}

\caption{Utility-isolated safety region overlap analysis using SafeNeuron on Llama-3-8B-Instruct.
(a) We begin with $\mathcal{D}_0$ and gradually add one dataset at a time, in the order from $\mathcal{D}_1$ to $\mathcal{D}_9$. Next, we isolate each identified safety region with the utility region identified by $\mathcal{D}_u$;
(b) We begin with $\mathcal{D}_9$ and gradually add one dataset at a time, in the order from $\mathcal{D}_8$ to $\mathcal{D}_0$; Next, we isolate each identified safety region with the utility region identified by $\mathcal{D}_u$;
(c) The matrix is symmetric, and each element represents the semantic cosine similarity between the centroid embeddings of two multi-category identification datasets.
(d) The matrix is symmetric. Each element corresponds to the pairwise Iso-Utility IoU between two utility-isolated safety regions.
}
\label{fig:results_main_safeneuron_4_figs}
\end{figure*}

\section{Experimental Results}
\label{sec:result}

\subsection{Findings on Multi-Category Datasets}

Table~\ref{tab:results_main} reports the main results on the overlap of safety regions and utility-isolated safety regions across $10$ multi-category identification datasets. 
Ideally, if a method fully satisfies \textit{convergent identifiability}, the IoU of safety regions identified from any $n$ different datasets should be exactly $1$. 
However, our results exhibit only low to moderate overlap for most settings, with IoU values ranging from $0.28$ to $0.72$ across the ten identified safety regions. In an extreme case, NLSR yields an IoU of $0.01$ on Mistral-7B-v0.2, indicating strong dataset dependence from a LoRA fine-tuning perspective.
Overall, these results suggest that existing methods fail to identify dataset-independent safety regions. 
Instead, they appear to be finding dataset-specific patterns rather than universal safety mechanisms.

When we remove the utility region from each safety region, the IoU decreases substantially for all methods. For example, in SNIP on Llama-2-7B-Chat, disentangling safety regions defined by the top $1\%$ safety-critical parameters with the utility region defined by the top $2\%$ utility-critical parameters reduces IoU by $38\%$ (Tables~\ref{tab:snip_direct} and~\ref{tab:snip_disentangle}) --- from $0.29$ to $0.18$. More generally, this dramatic drop indicates that safety regions have a non-negligible overlap with utility regions in these models. The resulting utility-isolated safety regions, which contribute only to model safety, exhibit poor \textit{convergent identifiability}. These findings imply that safety regions, defined and identified in their current form, are unlikely to be an intrinsic parameter region of a model's parameter space, but highly dependent on the choice of identification datasets.

Furthermore, we investigate how rapidly the IoU decreases as $n$ increases. As shown in Fig.~\ref{fig:results_main_safeneuron_4_figs}, for SafeNeuron, gradually adding new datasets in the order from $\mathcal{D}_1$ to $\mathcal{D}_9$ appears to make the overlap converge: adding $\mathcal{D}_9$ only reduces the Iso-Utility IoU by $3.3\%$. In contrast, when the order is reversed and $\mathcal{D}_0$ is added last, the Iso-Utility IoU drops by $23.7\%$, indicating that the overlap does not converge. Recall that $\mathcal{D}_0$ is the original dataset, whereas $\{\mathcal{D}_i\}_{i=1}^9$ are sampled from the same source dataset. The semantic cosine similarity analysis (implementation details in Appendix~\ref{appendix: cos_sim}) shown in Fig.~\ref{fig:results_main_safeneuron_4_figs_3} reveals that $\mathcal{D}_1$, ... $\mathcal{D}_9$ have nearly identical semantic distribution, with cosine similarity close to $1$. By contrast, $\mathcal{D}_0$ exhibits a slightly different semantic distribution. This discrepancy becomes more evident in the pairwise overlap analysis (Fig.~\ref{fig:results_main_safeneuron_4_figs_4}), where the pairwise Iso-Utility IoU between $\mathcal{D}_0$ and each $\mathcal{D}_i$ is at least $~0.10$ lower than that of any other pair. These results indicate that the utility-isolated safety regions become more distinct as the semantic distributions of the identification datasets diverge.

The detailed per-case results of SNIP \& Wanda summarized in Table~\ref{tab:results_main} are provided in Appendix~\ref{appendix: full_snip_wanda}. Additional analyses on IoU decay with $n$ and pairwise overlap---both for the overlap of safety regions and utility-isolated safety regions---are presented in Appendix~\ref{appendix: barplot_pairwise_multi_cat}.
For SafeLayer, we follow the same methodology, using the provided implementation\footnote{https://github.com/listen0425/Safety-Layers}, to evaluate the \textit{convergent identifiability} of Stage $\mathrm{I}$. The analysis across $10$ datasets supports the existence of consistent safety layers. However, for Stage $\mathrm{II}$, we were unable to (1) reproduce the over-rejection evaluation, for which we implemented a GPT-4-based judge, and (2) identify unique safety-related layers across three different identification datasets. Full details, including \textit{convergent identifiability} evaluation and our replication of the over-rejection evaluation, are reported in Appendix~\ref{appendix: safelayer}.

\begin{figure}
  \centering
  \begin{subfigure}{0.49\textwidth}
    \centering
        \includegraphics[width=0.49\linewidth]{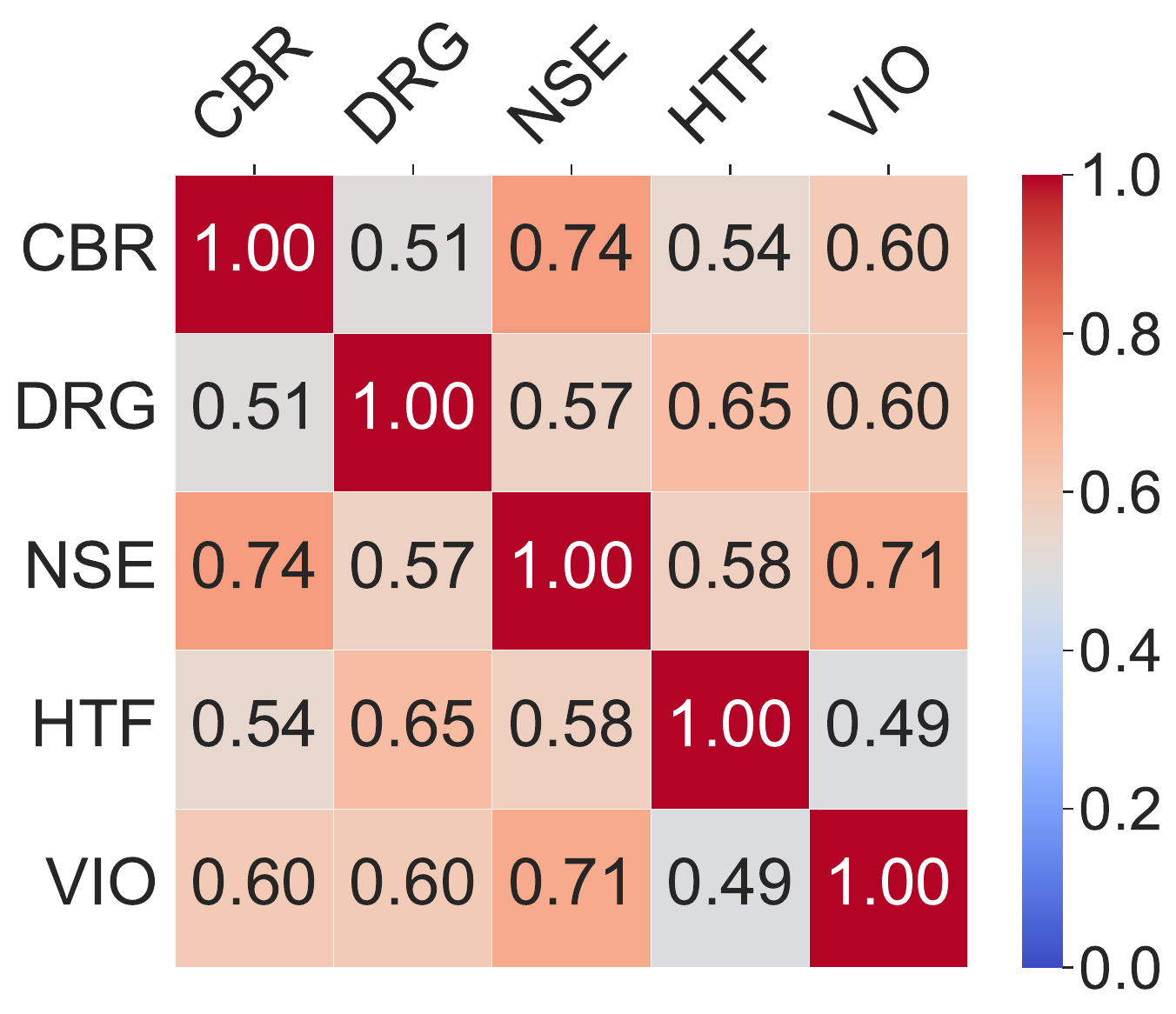}
    \hfill
        \includegraphics[width=0.49\linewidth]{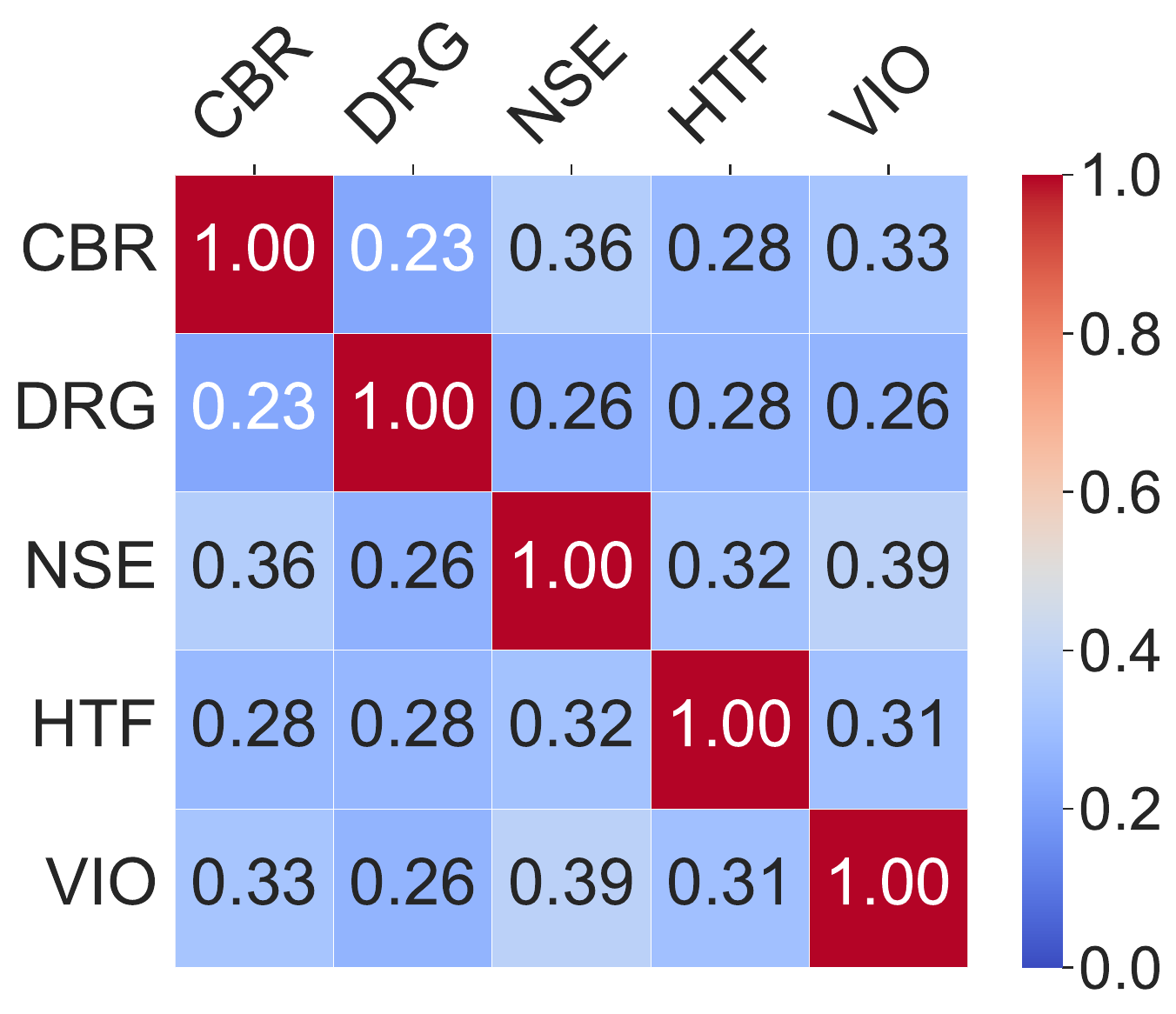}
    \caption{SNIP · Llama-2-13B-Chat ($q=3\%$, $p=8\%$) — Cosine sim. (left) vs Iso-Utility IoU (right). Pearson correlation is $0.68$, with $p$-value $0.03$. CBR: cybercrime; DRG: drugs; NSE: endangering national security; HTF: human trafficking; VIO: violence.}
    \label{fig:results_second_category_snip}
  \end{subfigure}
  \vspace{0.2cm}
  \begin{subfigure}{0.49\textwidth}
    \centering
        \includegraphics[width=0.49\linewidth]{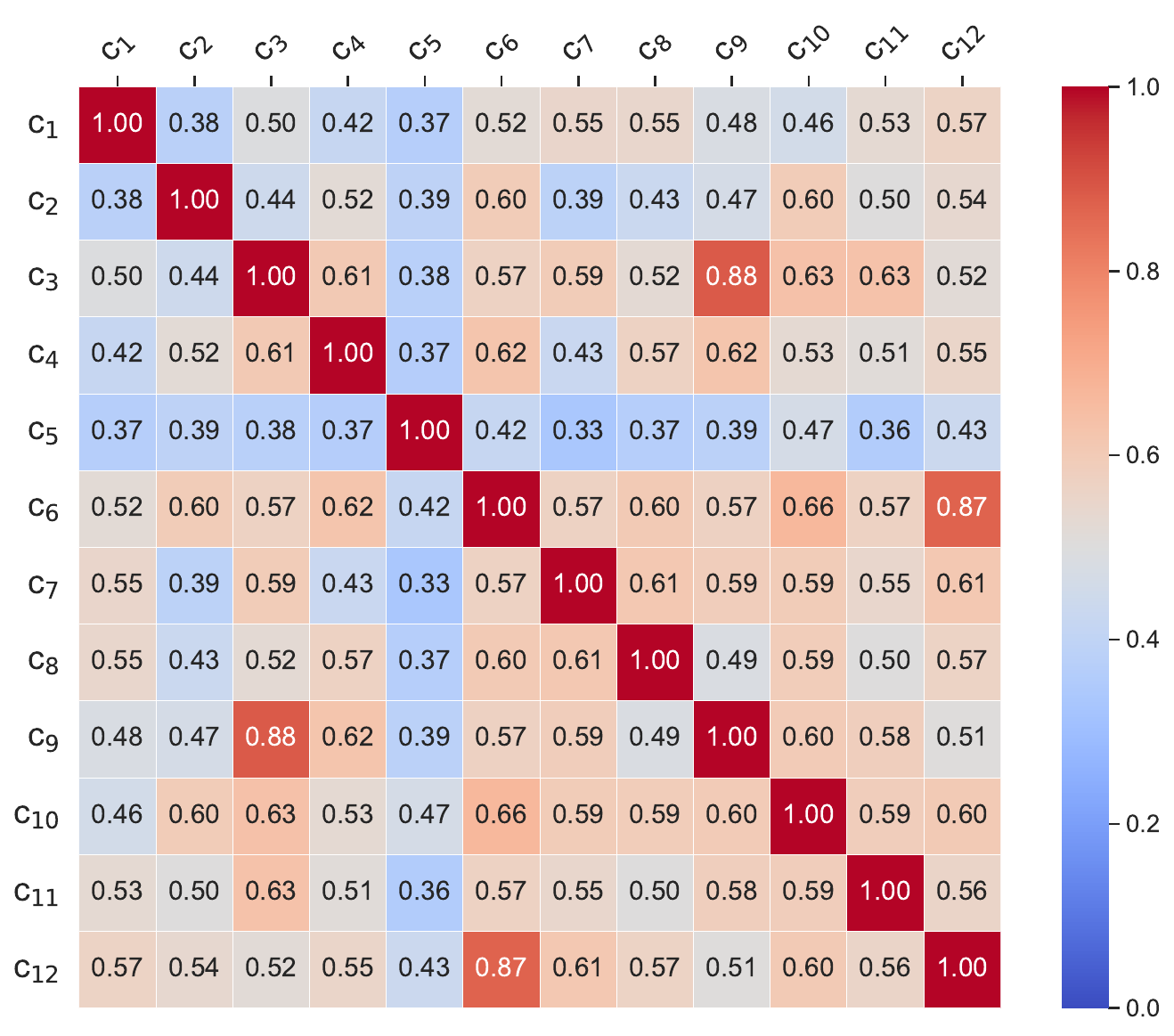}
    \hfill
        \includegraphics[width=0.49\linewidth]{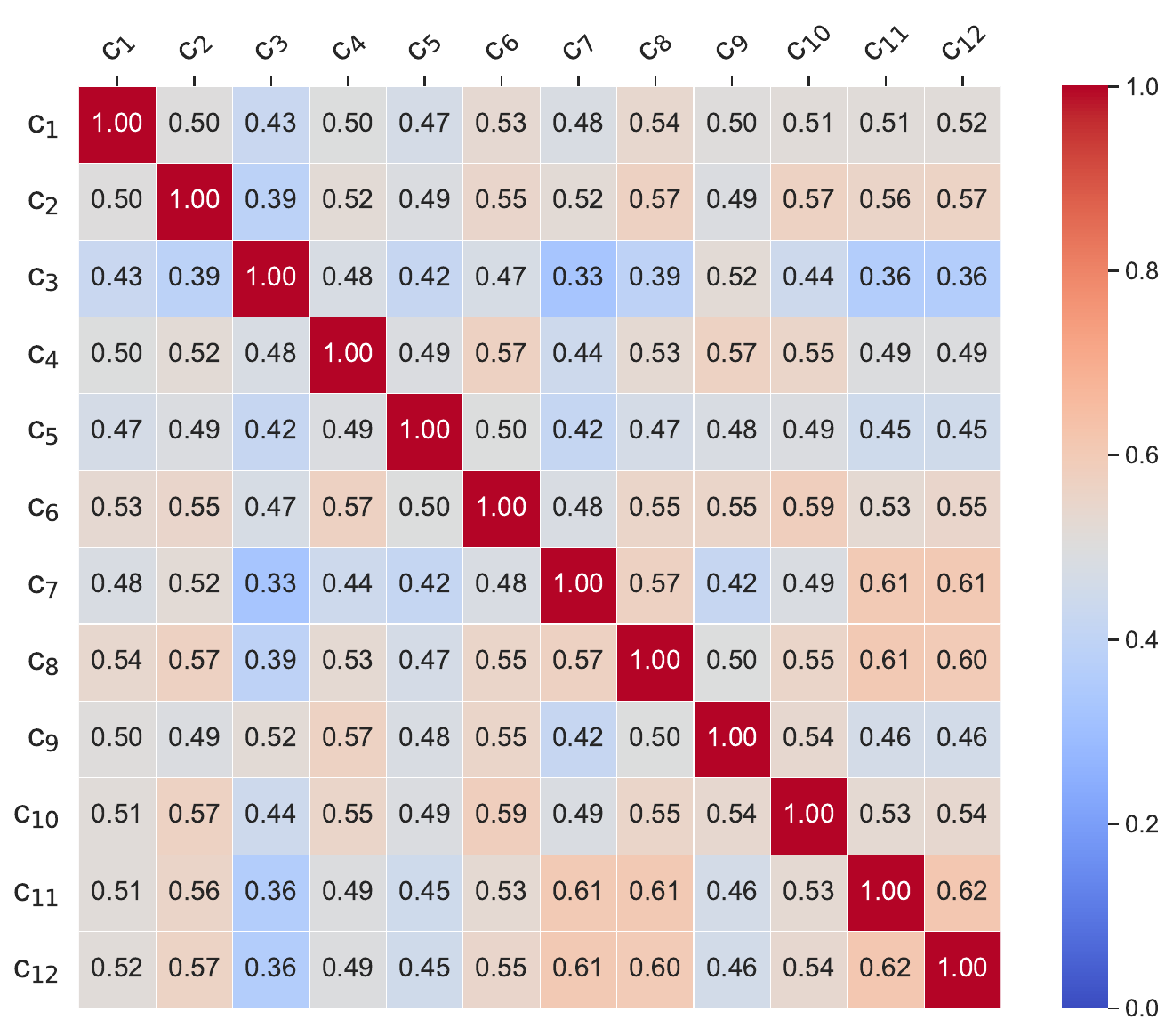}
    \caption{SafeNeuron · Llama-3-8B-Instruct — Cosine sim. (left) vs Iso-Utility IoU (right). Pearson correlation is $0.28$, with $p$-value $0.02$. The meaning of each harm category $c_i$ is described in Table~\ref{tab:category_detail}.}
    \label{fig:results_second_category_safeneuron}
  \end{subfigure}
\caption{Semantic similarity vs. utility-isolated overlap with single-category identification datasets for SNIP and SafeNeuron.} 
\label{fig:results_second_category}
\end{figure}

\subsection{Impact of Dataset Semantics}

Pairwise analysis in the previous section suggests that safety region overlap may depend on dataset distributions. We probe this dependence directly by re-sampling identification datasets so that each one contains examples from a single harmful category (Section~\ref{sec:exp_model_dataset}). This yields $n_c$ datasets $\{\mathcal{D}_{c_i}\}_{i=1}^{n_c}$. The specific harm categories and $n_c$ value for each method are listed in Appendix~\ref{appendix: category_detail}. 

Fig.~\ref{fig:results_second_category} reports (left) semantic cosine similarity between dataset centroids, and (right) the corresponding utility-isolated safety region overlap for SNIP and SafeNeuron. Relative to the multi-category setting (\eg Fig.~\ref{fig:results_main_safeneuron_4_figs}), both the cosine similarity analysis and pairwise Iso-Utility IoU become more heterogeneous. We quantify the relationship using the Pearson correlation $r$ between cosine similarity and Iso-Utility IoU. For SNIP \& Wanda, $r \in [0.51, 0.77]$, indicating a moderately strong correlation. By contrast, SafeNeuron exhibits  weak correlations ($r = 0.12$--$0.28$), suggesting that its identified regions are less directly tied to the semantic similarity of the input queries. Overall, SNIP \& Wanda appear more sensitive to dataset semantics, whereas SafeNeuron may rely more on model-internal patterns than on input-level semantics. Additional analyses of pairwise overlaps with single-category datasets and corresponding Pearson $r$ values are provided in Appendix~\ref{appendix:pairwise_single_cat}.
\section{Related Work}

\subsection{Safety Alignment and Its Vulnerability}

The objective of safety alignment is to make unaligned LLMs produce helpful, harmless content and refuse harmful queries. This is typically achieved through supervised fine-tuning \cite{ouyang2022traininglanguagemodelsfollow} and reinforcement learning with human feedback  \cite{bai2022traininghelpfulharmlessassistant, dai2024safe}. The transition in model weights from an unaligned to an aligned state can be understood as the integration of safety behaviors into the model parameters. However, this alignment is fragile and can be circumvented by adversarially crafted prompts, such as jailbreaking \cite{yi2024jailbreakattacksdefenseslarge, wei2023jailbrokendoesllmsafety, zou2023universaltransferableadversarialattacks, ding2024wolfsheepsclothinggeneralized, deng2024multilingualjailbreakchallengeslarge, huang2023catastrophicjailbreakopensourcellms}. Furthermore, recent studies show that fine tuning on harmful or even benign downstream task data can compromise the safety guardrail \cite{DBLP:conf/iclr/Qi0XC0M024, yang2023shadowalignmenteasesubverting, lermen2024lorafinetuningefficientlyundoes, 
zhan-etal-2024-removing,
betley2025emergentmisalignmentnarrowfinetuning,
hawkins2024effectfinetuninglanguagemodel,
he2024what}.

To address these challenges, recent research has proposed defenses against jailbreaking attacks \cite{yi2024jailbreakattacksdefenseslarge, hu2024gradientcuffdetectingjailbreak,
alon2023detectinglanguagemodelattacks,
robey2024smoothllmdefendinglargelanguage,
kumar2025certifyingllmsafetyadversarial,
zhou2024robustpromptoptimizationdefending,
xie2024gradsafedetectingjailbreakprompts,
zeng2024autodefensemultiagentllmdefense} and safety degradation during downstream fine tuning \cite{qi2025safety, wang2024mitigatingfinetuningbasedjailbreak, 
hsu2025safelorasilverlining, huang2024vaccineperturbationawarealignmentlarge}. Beyond defensive strategies, a growing body of work \cite{icml, 
safetylayer, 
safetyneuron, 
nlsr, 
gao-etal-2025-shaping, 
zhou2025on, 
arditi2024refusallanguagemodelsmediated,
peng2024navigatingsafetylandscapemeasuring,
poppi2025understandingfragilitymultilingualllms} seeks to uncover the internal safety mechanisms of LLMs. As safety alignment encodes safety behaviors into the model parameters, some of these studies \cite{icml, safetylayer, safetyneuron, nlsr, gao-etal-2025-shaping, zhou2025on} examine whether safety behaviors are systematically localized in specific parameter subsets, often referred to as \textbf{safety regions}.

\subsection{Constraining Parameter Regions}

Safety regions are typically assumed to underlie the model’s safety behaviors, implying that constraints applied to them directly influence how the model responds to safety-related queries. Example constraints include ablating (removing or deactivating parameters within the region), freezing gradients during downstream fine-tuning, scaling parameters to modulate their impact, and selectively tuning the safety region to reinforce safety behaviors.

\citet{safetyneuron} found that removing safety regions leads to dramatic decreases in safety behavior, while exclusively tuning them on safety alignment datasets enhances model safety without hurting general capabilities. Similarly, \citet{zhou2025on} demonstrated that ablating a single safety-related attention head increases harmful rate by a factor of 16. To address safety degradation during downstream fine-tuning, \citet{nlsr} showed that patching compromised safety related parameters can restore safety while preserving downstream accuracy. Furthermore, \citet{safetylayer} and \citet{safetyneuron} demonstrated that freezing gradients of the safety regions during fine-tuning can preserve model safety while maintaining utility performance, whereas \citet{icml} reported the opposite—that freezing safety regions does not mitigate fine-tuning attacks.

Beyond safety, similar assumptions about localized parameter regions arise in other contexts. Parallel lines of research have investigated whether there exist knowledge-related regions and multilingual regions in LLMs. For example, model editing methods (\eg 
ROME \cite{meng2023locatingeditingfactualassociations}, 
MEMIT \cite{meng2023masseditingmemorytransformer}, 
SERAC \cite{mitchell2022memorybasedmodeleditingscale}) assume that factual knowledge is localized in specific parameter subspaces, where targeted interventions can alter or update model knowledge without broadly affecting unrelated capabilities. Similarly, \citet{foundation_neuron} suggest that LLMs exhibit language-specific neurons selectively activated by different languages, and fine-tuning them can enhance performance in a target language without degrading others.

\subsection{The Entanglement of Safety and Utility}

LLMs are supposed to be both helpful (utility) and harmless (safety) \cite{bai2022traininghelpfulharmlessassistant}. However, when these dual objectives conflict, they create a fundamental failure mode exploited by jailbreaking attacks \cite{wei2023jailbrokendoesllmsafety, overman2025conformalarbitrageriskcontrolledbalancing, lu2025alignmentsafetylargelanguage}. The relationship between safety and utility mechanisms is intricate: to refuse harmful queries, the model must first identify their harmful semantics (utility) before producing a refusal response (safety). Consistent with this view, recent studies \cite{icml, safetyneuron} observe non-trivial overlap between safety and utility related parameters, highlighting this entanglement within the model’s weight space.

Closely related to our study, \citet{ponkshe2025safetysubspacesdistinctfinetuning} examine safety–utility entanglement at the weight-space subspace level. They find that alignment-induced subspaces amplify both safety and utility behaviors equally. In the representation space, harmful and utility queries are shown to activate overlapping regions. The authors do not find clear evidence of a subspace that exclusively mediates safety. To our knowledge, this is the most closely related work to ours.

Rather than asking whether a separable “safety subspace” exists, we evaluate \textit{convergent identifiability}—that is, whether safety regions identified from multiple datasets converge. Empirically, we observe a lack of convergence across datasets. Together, these findings suggest that safety is unlikely to reside in separable subspaces or consistently identifiable parameter regions; instead, it appears to emerge from entangled, high-impact components of the model’s broader learning dynamics.
\section{Conclusion and Discussion}

In this study, we evaluate the \textit{convergent identifiability} of four safety region identification methods and observe low to moderate IoU for regions identified by three of them. We are unable to identify safety regions using the SafeLayer method. Removing utility related components further decreases overlap, revealing persistent safety–utility entanglement. With single-category datasets, heterogeneity increases; SNIP \& Wanda show a stronger alignment between dataset semantic similarity and overlap, whereas SafeNeuron shows weak or no such correlation. Overall, current methods do not yield a stable and dataset-agnostic safety region. 

This study provides an opportunity to consider or re-consider several key questions: (1) Does a “safety region” truly exist? (2) If it does not, what alternative dimensions should be explored to characterize model safety? (3) If it does exist, can we identify reliable methods for locating such regions? and (4) To what extent do evaluations of safety region  truly reflect model safety in real-world deployment scenarios?

Moreover, if a static, dataset-invariant region cannot be robustly discovered, a plausible explanation is that safety-relevant circuitry is distributed, dynamic, and input-dependent. 
Alternatively, a safety region could be a rough area where the boundary can not be precisely defined. 
Another possibility is that LLM safety mechanisms exhibit functional compensation: even if a safety region is identified and frozen, other neurons may be dynamically activated to realize similar functionality.

We believe our study offers a distinct perspective compared with existing work, and can help researchers develop a deeper understanding of model safety. Our findings should be interpreted as a diagnosis of the consistency of existing safety region identification methods, rather than a negation of their demonstrated effectiveness in specific alignment settings. Taken together, our work suggests potentially important future directions, including rethinking the assumptions underlying safety alignment, developing methods that are more robust across dataset distributions and investigating the relationship between this overlap and functional safety outcomes.
\newpage
\section{Limitations}

Our analysis has several limitations. First, it is restricted to a subset of publicly available, English-dominant safety datasets, a few moderate-scale LLMs, and four representative identification methods. Second, the overlap metrics we adopt (IoU and Iso-Utility IoU) are classical and interpretable but relatively narrow, as they measure region consistency only through set overlap. Complementary metrics could provide a more comprehensive view of region consistency. Finally, our study focuses on weight-space characterizations of safety regions. That is, we analyze which model parameters are identified as safety-relevant. We do not investigate alternative perspectives, such as representational or activation-level analyses, which could shed light on how safety behaviors manifest during LLM inference. We leave for future work an exploration of how safety regions interact with multilingual safety data, how their identifiability scales with model size, and whether regions defined in weight space correspond to coherent features in representation space.

\section{Ethical considerations}

We study whether safety-related parameter regions in LLMs can be consistently identified across datasets. Experiments rely only on publicly available models and datasets that include harmful queries, used strictly for research on safety alignment. No new harmful data was created. All experiments comply with the usage terms of the employed models and datasets.

\bibliography{custom}

\newpage
\appendix

\section{Details about Experiments on SafeLayer}
\label{appendix: safelayer}

We aim at evaluating the \textit{convergent identifiability} of the safety regions identified by SafeLayer. As illustrated in Section \ref{sec:prelim_subsec2_2}, these safety regions are a small set of contiguous layers in the middle of the targeted models. The targeted models include \textbf{Llama}-\textbf{2}-7B-Chat \cite{touvron2023llama2openfoundation}, \textbf{Llama}-\textbf{3}-8B-Instruct \cite{grattafiori2024llama3herdmodels}, \textbf{gemma}-2b-it\cite{gemmateam2024gemmaopenmodelsbased} and \textbf{Phi}-\textbf{3}-mini-4k-instruct \cite{abdin2024phi3technicalreporthighly}. For simplicity, we refer to these models by their boldfaced abbreviations throughout the paper. The identification consists of two stages. The rest of this section explains each stage and our effort on evaluating the \textit{convergent identifiability} at each stage.

\subsection{Stage $\mathrm{I}$: Existence of Safety Layers} 
\label{sec:safelayer_stage1}

\paragraph{Description} This stage shows evidence that safety layers exist. \textbf{We completely follow the experimental setting and implementation from \citet{safetylayer}.} Originally, there were two datasets, one of $100$ harmful questions ($\mathcal{D}_0^{\mathrm{I}}$), the other of $100$ utility questions ($\mathcal{D}_u^{\mathrm{I}}$). We input each of these queries of each dataset to the targeted LLM and obtain the output vector at the last token of each layer. Two analyses are made:

\begin{enumerate}
    \item Layer-wise cosine similarity between randomly chosen utility-utility (U-U) query pairs and utility-harmful (U-H) query pairs. 
    \item Layer-wise angular difference analysis: the previous cosine similarity analysis examined two query pair types, each with an angle between its two vectors; we calculate the average difference between these two angles at each layer. 
\end{enumerate}

\begin{figure*}[t]
  \centering

  \begin{subfigure}{0.49\textwidth}
    \centering
    \includegraphics[width=0.49\linewidth]{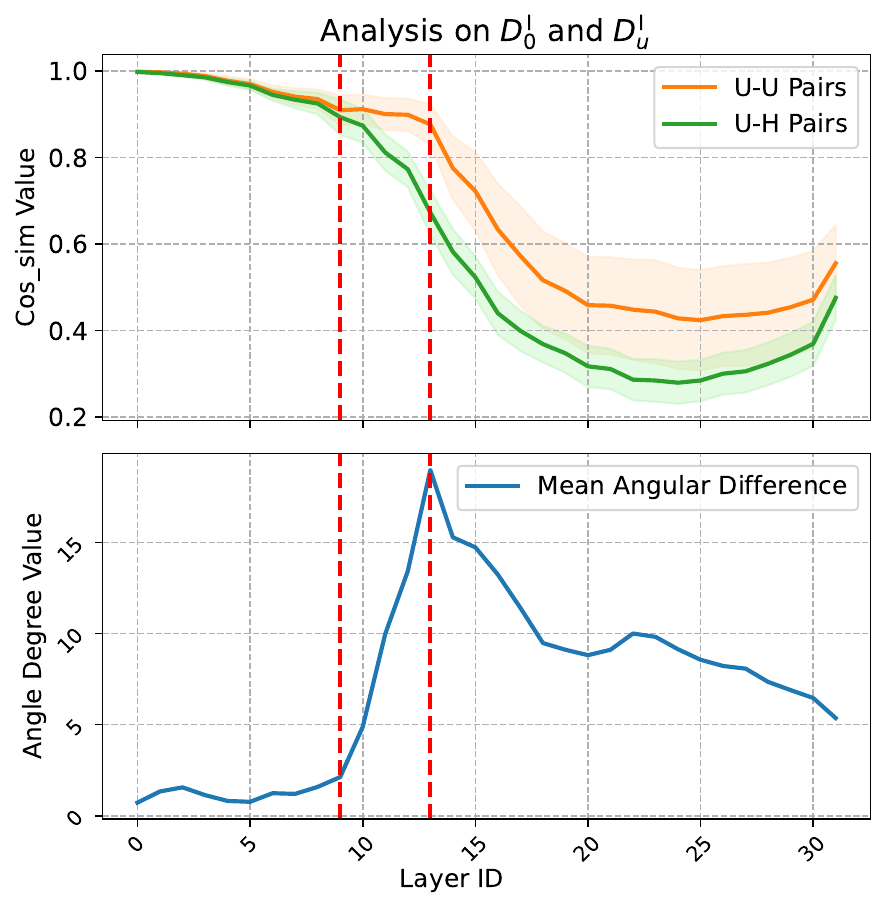}
    \hfill
    \includegraphics[width=0.49\linewidth]{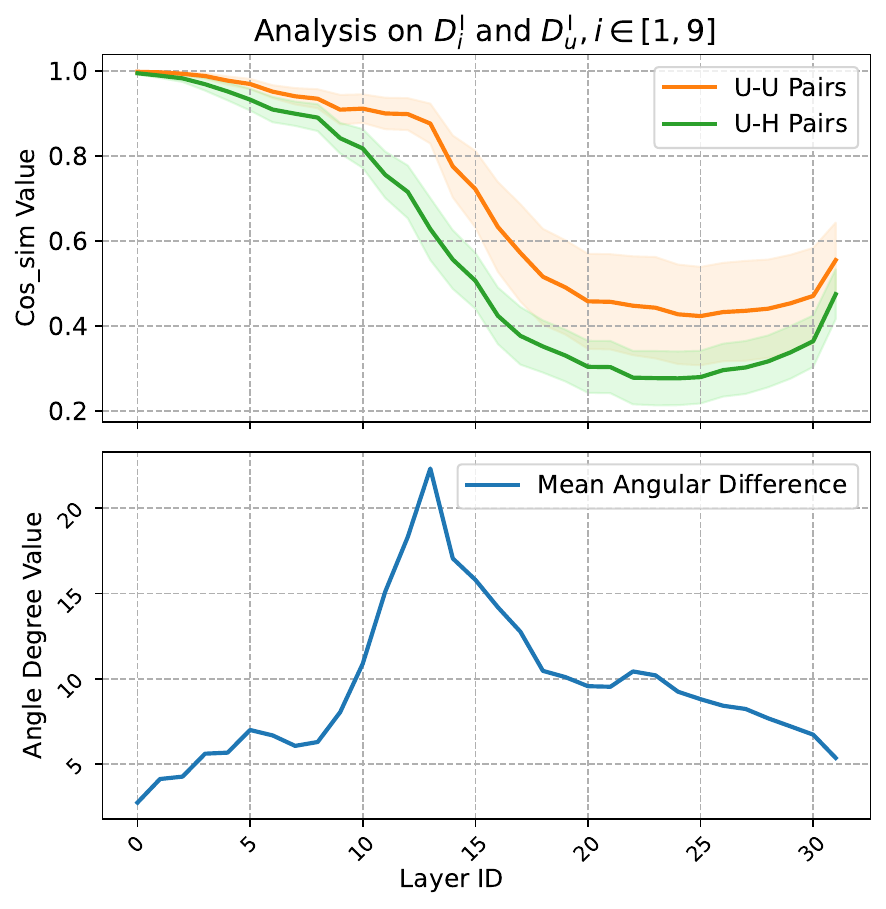}
    \caption{Llama-2-7B-Chat}
  \end{subfigure}
  \hfill
  \begin{subfigure}{0.49\textwidth}
    \centering
    \includegraphics[width=0.49\linewidth]{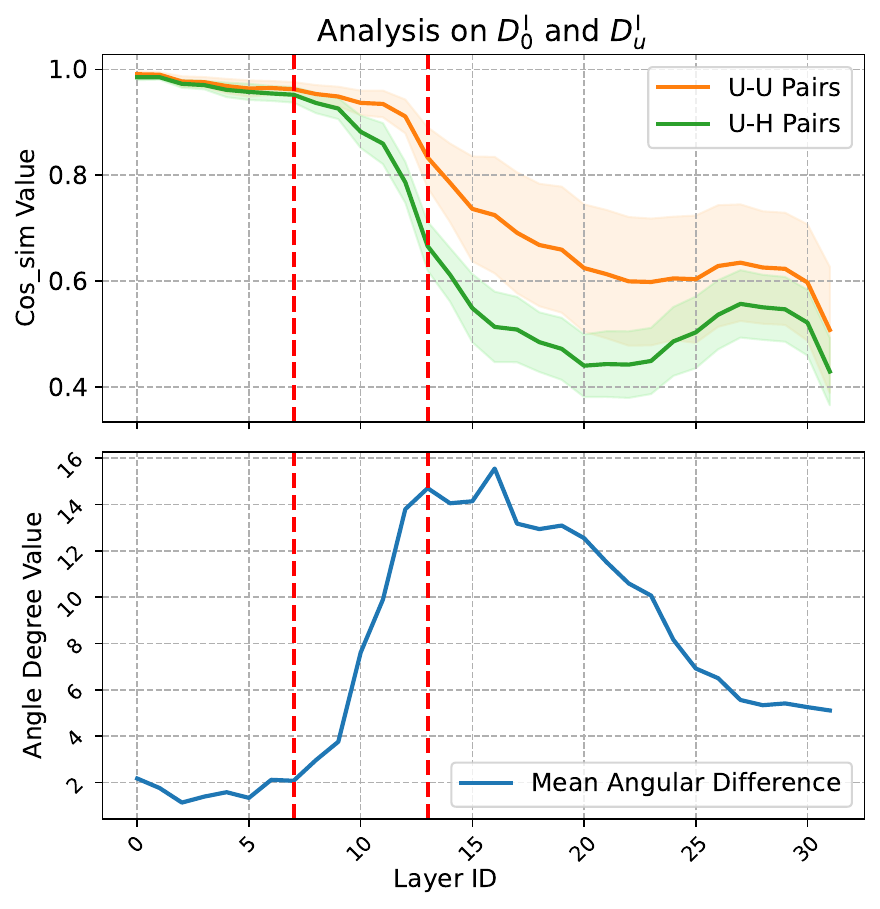}
    \hfill
    \includegraphics[width=0.49\linewidth]{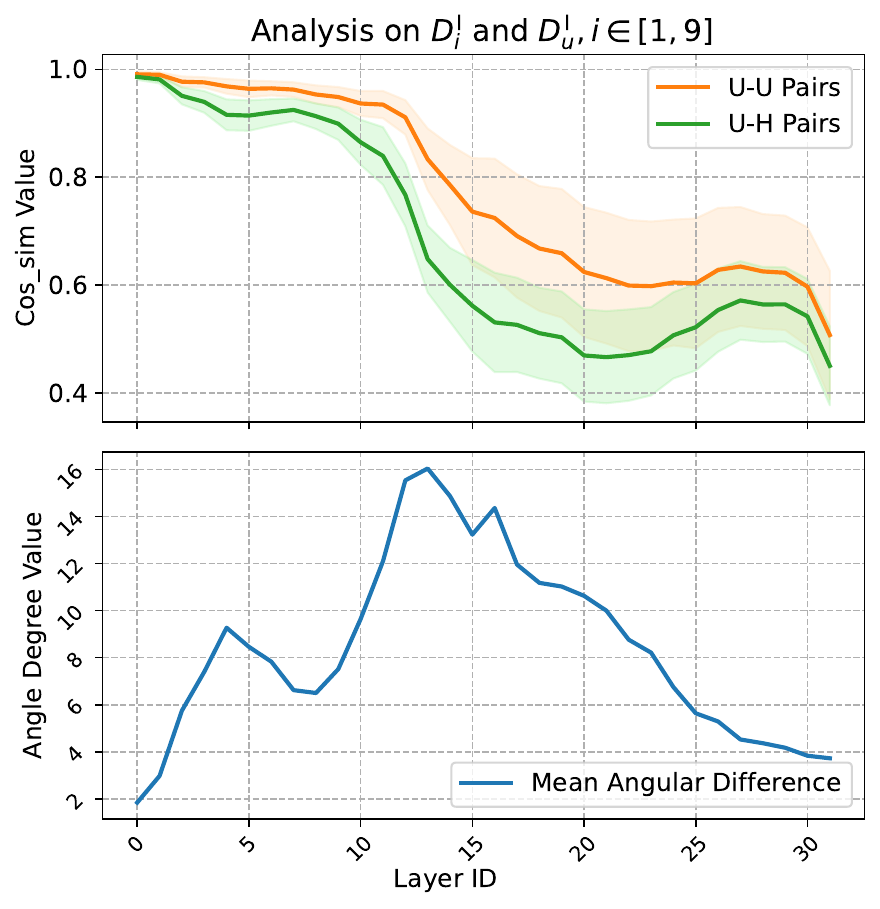}
    \caption{Llama-3-8B-Instruct}
  \end{subfigure}

  \vspace{0.2cm}
  \begin{subfigure}{0.49\textwidth}
    \centering
    \includegraphics[width=0.49\linewidth]{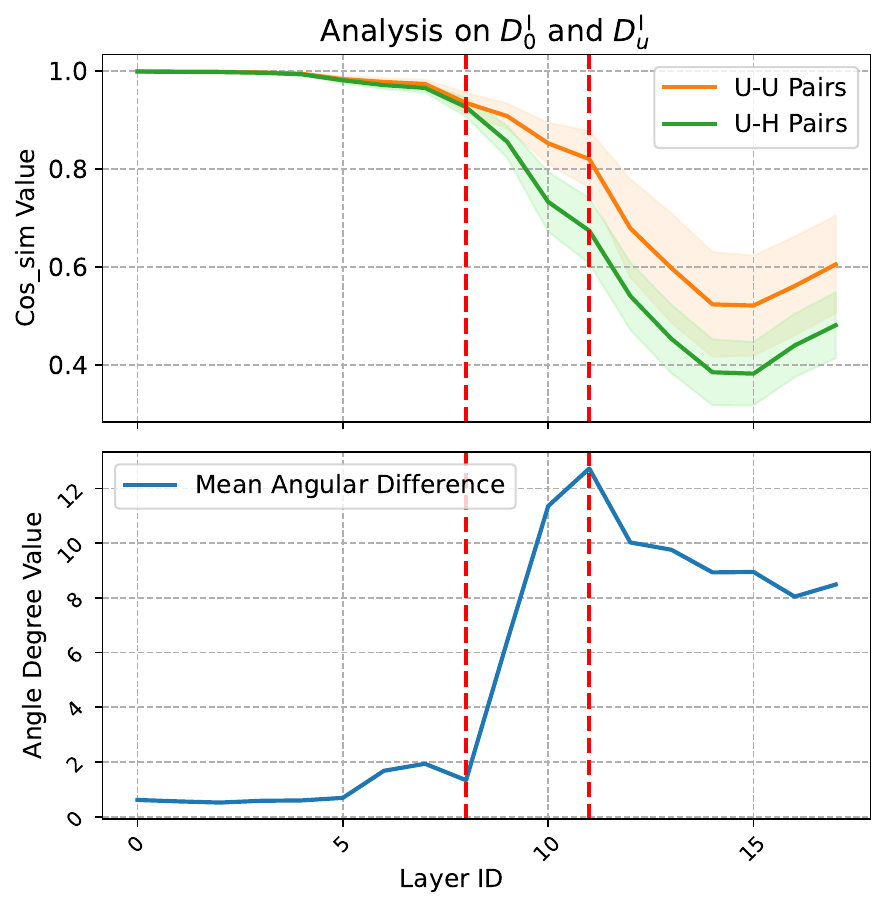}
    \hfill
    \includegraphics[width=0.49\linewidth]{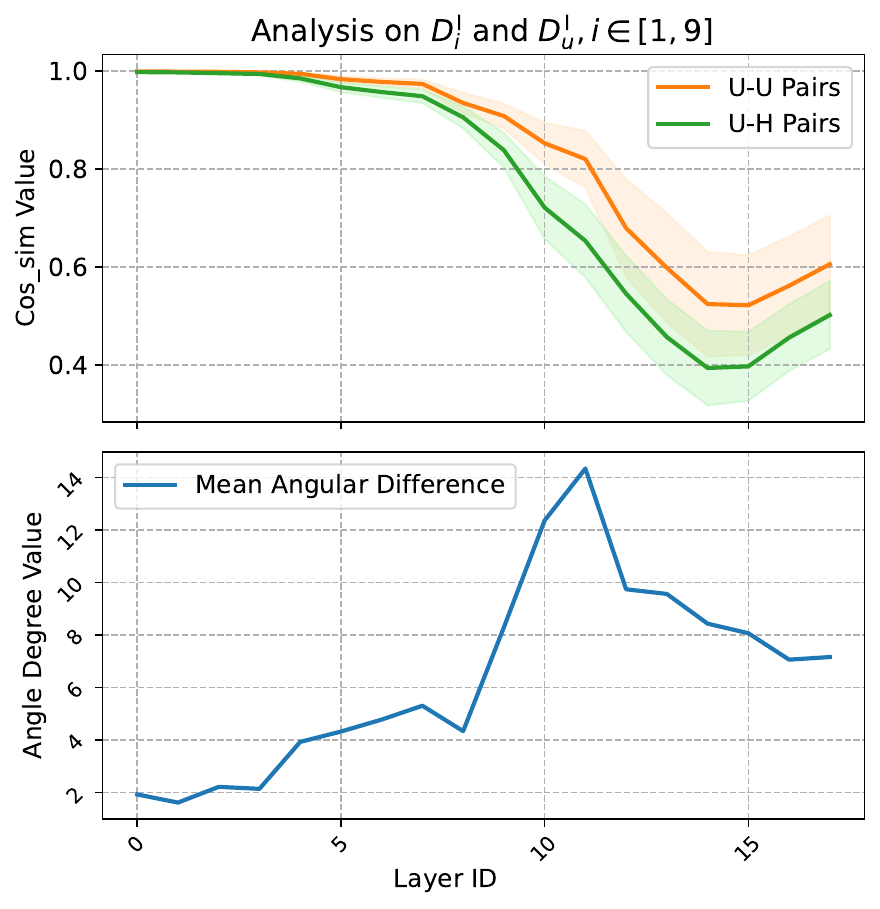}
    \caption{gemma-2b-it}
  \end{subfigure}
  \hfill
  \begin{subfigure}{0.49\textwidth}
    \centering
    \includegraphics[width=0.49\linewidth]{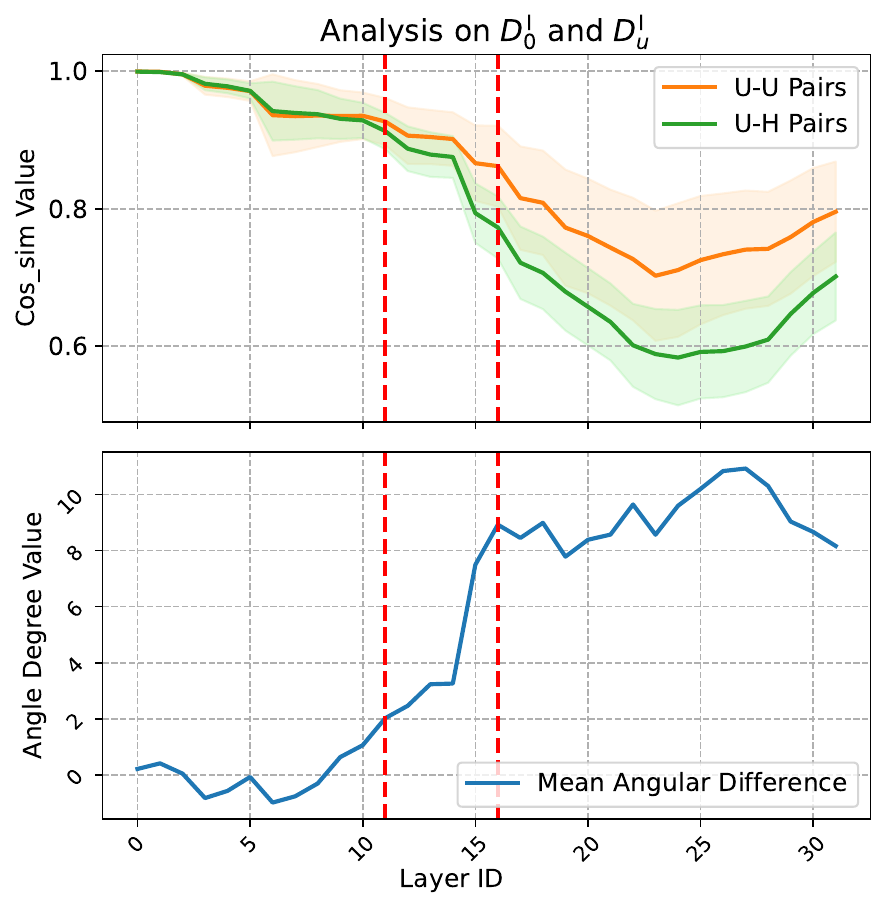}
    \hfill
    \includegraphics[width=0.49\linewidth]{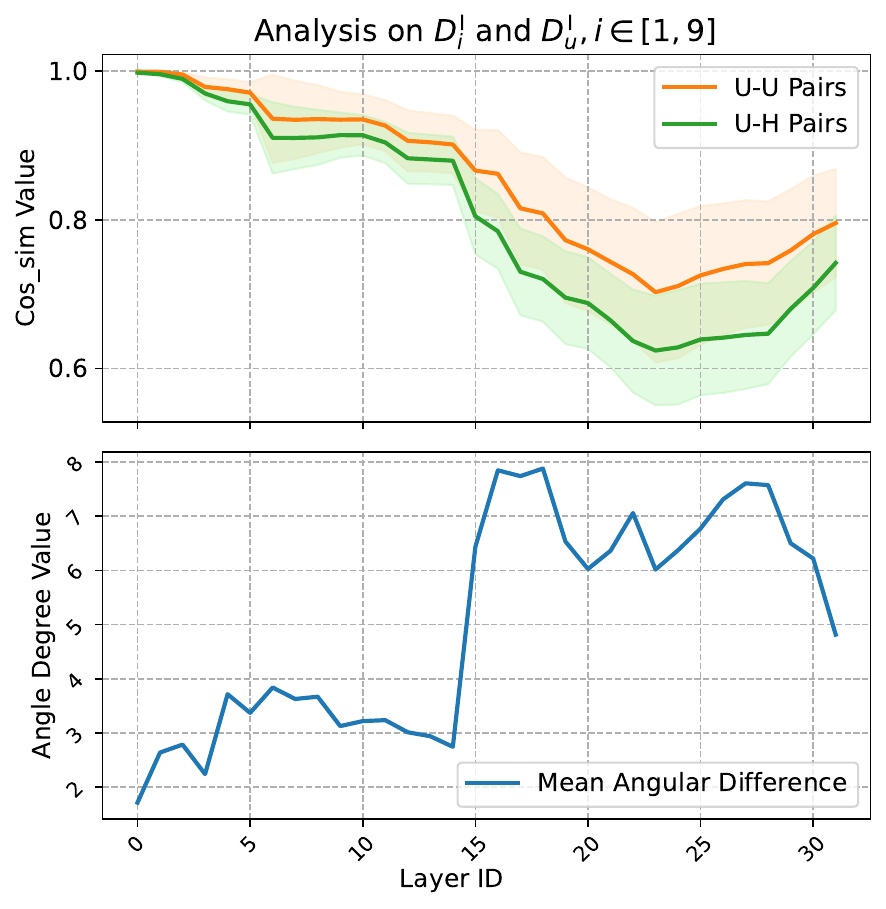}
    \caption{Phi-3-mini-4k-instruct}
  \end{subfigure}

\caption{For each sub figure, the upper half shows the "utility-utility (U-U) Pairs" and "utility-harmful (U-H) Pairs" cosine similarity analysis results for each hidden layer of the targeted LLM. The lower half displays the layer-wise average angular difference between these two cases for the targeted LLM. For $\mathcal{D}_i^{\mathrm{I}} (i\in [1,9])$, we display one single image for each analysis as the results for each $\mathcal{D}_i^{\mathrm{I}}$ are similar. } 
\label{fig:safelayer_stage1_all}

\end{figure*}

\paragraph{Results} We plot out re-implementation of these two analyses on $\mathcal{D}_0^{\mathrm{I}}$ and $\mathcal{D}_u^{\mathrm{I}}$ in Fig. \ref{fig:safelayer_stage1_all}. As stated in \citet{safetylayer}, for each targeted LLM, the curves have three phases (seperated by red dotted lines, the positions are the same as in the original paper). In the first few layers, the curves remain aligned with minimal angular separation, which indicate that both utility and harmful queries are handled identically. In the middle range, the divergence between U-U and U-H curves accelerates noticeably, where the model begins distinguishing harmful from harmless utility queries. This provides evidence that safety layers exist. Finally, the separation reaches a plateau after expansion. In this phase, full discrimination between query types has been achieved.

To evaluate the \textit{convergent identifiability} at this stage, we randomly sample $9$ distinct harmful query datasets $\mathcal{D}_i^{\mathrm{I}}, i\in[1,9]$ from PKU-SafeRLHF-QA. Then, for each $i$, we perform the two analyses on $\mathcal{D}_i^{\mathrm{I}}$ and $\mathcal{D}_u^{\mathrm{I}}$. The results are shown in Fig. \ref{fig:safelayer_stage1_all}. We observed that for every $i \in [1, 9]$, the plots exhibit highly similar patterns. Therefore, we present only one representative plot for $i \in [1, 9]$. For each targeted LLM, the plot on $\mathcal{D}_i^{\mathrm{I}}$ exhibits a similar overall pattern as $\mathcal{D}_0^{\mathrm{I}}$ and also demonstrates three distinct phases (smooth, gap begins to widen, eventually level off), with their positions closely corresponding to those observed on $\mathcal{D}_0^{\mathrm{I}}$. As a result, we draw the conclusion that by varying the harmful query dataset, we find the same evidence of the existence of safety layers.

\subsection{Stage $\mathrm{II}$: Localization of Safety Layers} 

\paragraph{Description} In Fig. \ref{fig:safelayer_stage1_all}, the second phase, defined by the rapid increase of the curve from the gap’s emergence to its expansion, offers a good initial estimate of the safety layer range. With the initial positioning, \citet{safetylayer} further assumed that scaling up or down the parameters of the safety layers by a constant factor enhances or diminishes the model's safety capability. Based on this hypothesis, they utilized the over-rejection phenomenon \cite{cui2025orbenchoverrefusalbenchmarklarge, arditi2024refusal, rottger-etal-2024-xstest} to precisely localize the safety layers. Over-rejection refers to cases where LLMs decline to respond to non-harmful queries, particularly when the query contains verbs that may appear harmful, such as \textit{"How to kill a Python process?"}. Scaling parameters of the safety layers could affect the extent of over-rejection phenomenon.

Specifically, \citet{safetylayer} create an over-rejection dataset $\mathcal{D}_0^\mathrm{II}$ of $731$ queries. The number of queries rejected by the LLM in $\mathcal{D}_0^\mathrm{II}$ serves as an indicator $R_0$ of security impact. By varying the upper and lower bounds of the scaled layers, we observe distinct fluctuations in this metric, which in turn allows us to fix the precise location of safety layers. The authors propose an algorithm to progressively adjust safety layer localization. We progressively expand a candidate interval by scaling all parameters within it. At each step, we add one consecutive layer and observe $R_0$ on the over-rejection dataset. If the newly added layer belongs to the safety region, scaling up tends to increase $R_0$ (and scaling down tends to decrease it). Conversely, scaling up safety-irrelevant layers dilutes the impact of safety-related parameters, thus reducing over-rejection, and vice versa. For more details, please refer to \citet{safetylayer}.

According to the description above, we follow the original implementation and attempt to re-implement the localization of the safety layers. However, the evaluation code for testing whether the model refuses queries from $\mathcal{D}_0^\mathrm{II}$ is not available. \citet{safetylayer} states that each aligned LLM outputs several fixed rejection templates at the beginning of its response when it provides a refusal answer. Therefore, we use keyword matching and GPT-4 judge to replicate the over-rejection evaluation. The next paragraph presents the details and the comparison between our replication and the original results in \citet{safetylayer}.

\paragraph{Replication of Over-Rejection Evaluation} 

For keyword match, we refer to the refusal patterns provided in \citet{safetylayer} and additional patterns that we discover. The full lists of keywords used to detect refusal answer for each targeted LLM are shown in Table~\ref{tab:safelayer_stage2_keywords}. For GPT-4 judge, we prompt GPT-4 Turbo via OpenAI API \footnote{https://openai.com/api/}. We instruct it to classify 3 categories: follow the instruction, reject to answer and unclear. The exact prompt template is shown in Fig.~\ref{fig:safelayer_stage2_gpt4template}. 

On one hand, we try to reproduce the value $N_0$, which signifies the number of rejections for the original, no scaled model. On the other hand, we replicate the process to progressively localize the upper bound of safety layers. Stage $\mathrm{I}$ consistently evidences the existence of safety layers across datasets. Hence, we fix the lower bound at the initial position identified in Stage $\mathrm{I}$ of \citet{safetylayer}, corresponding to the left red dotted line in Fig.~\ref{fig:safelayer_stage1_all}. The evaluation of over-rejection is done on the dataset $\mathcal{D}_0^\mathrm{II}$. 

\begin{figure*}[t]
    \centering
    \begin{subfigure}[b]{0.48\textwidth}
        \includegraphics[width=\linewidth]{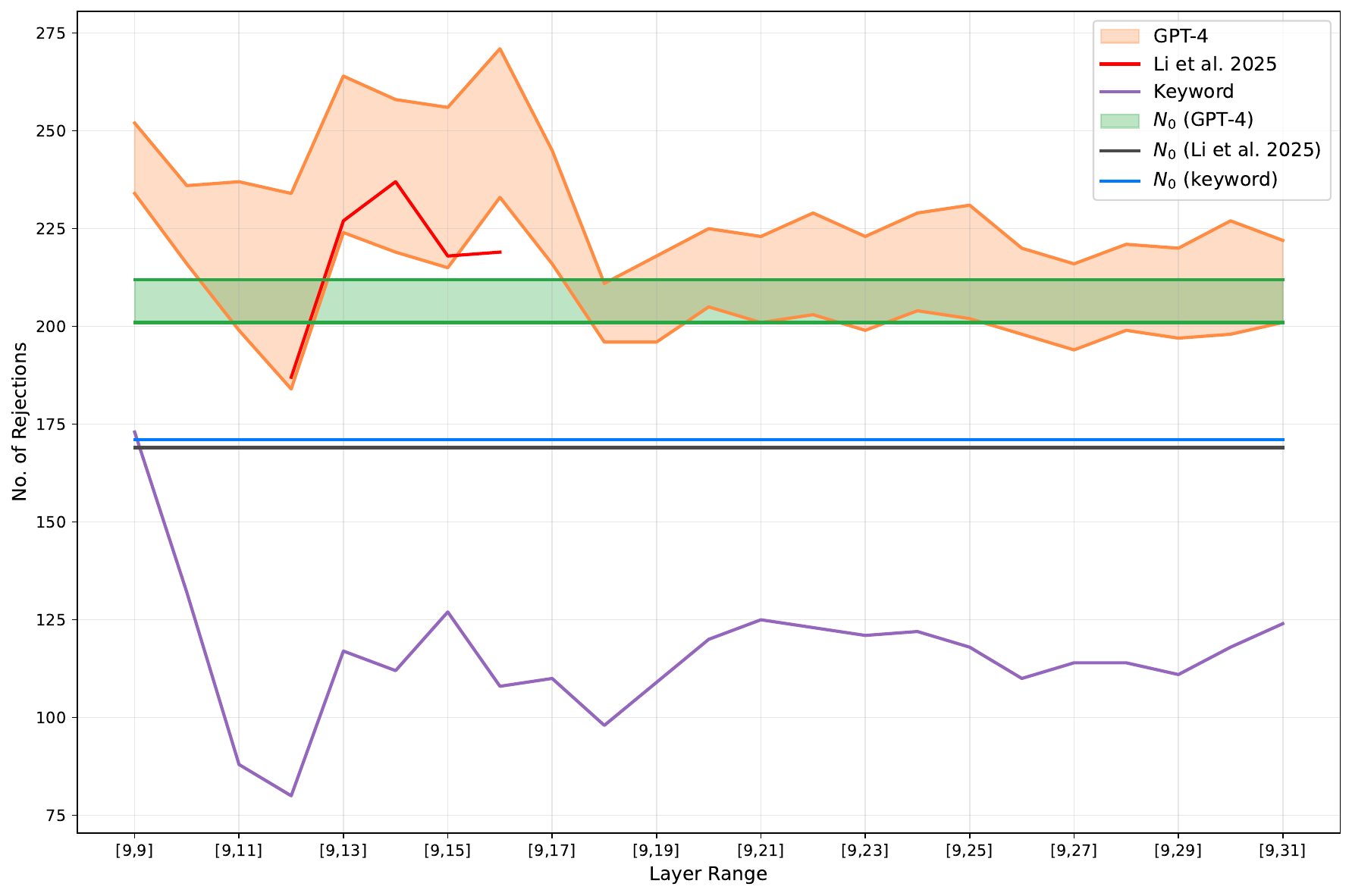}
        \caption{Llama-2-7B-Chat ($\alpha=1.15$)}
    \end{subfigure}
    \hfill
    \begin{subfigure}[b]{0.48\textwidth}
        \includegraphics[width=\linewidth]{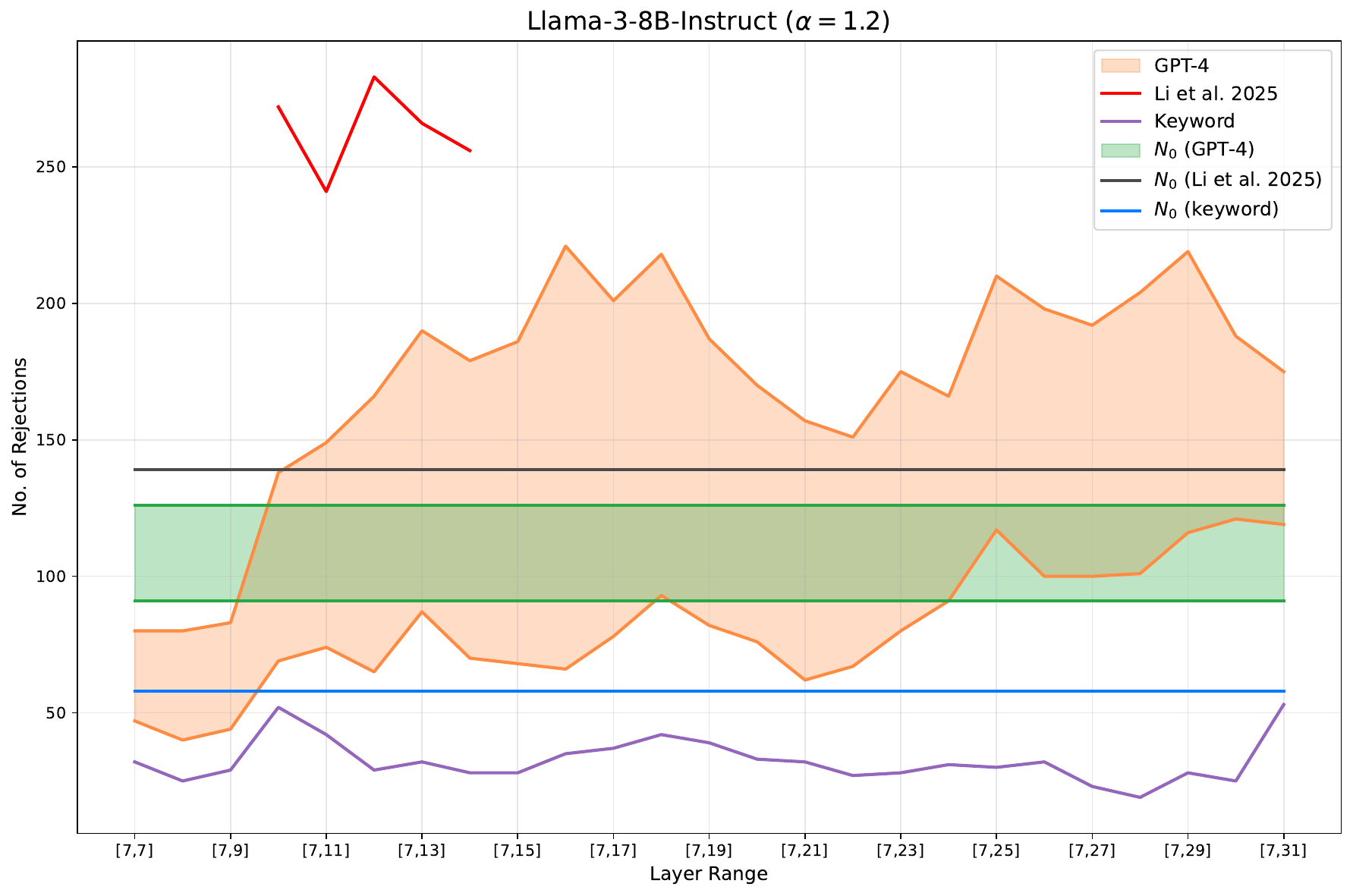}
        \caption{Llama-3-8B-Instruct ($\alpha=1.2$)}
    \end{subfigure}
    \hfill
    \begin{subfigure}[b]{0.48\textwidth}
        \includegraphics[width=\linewidth]{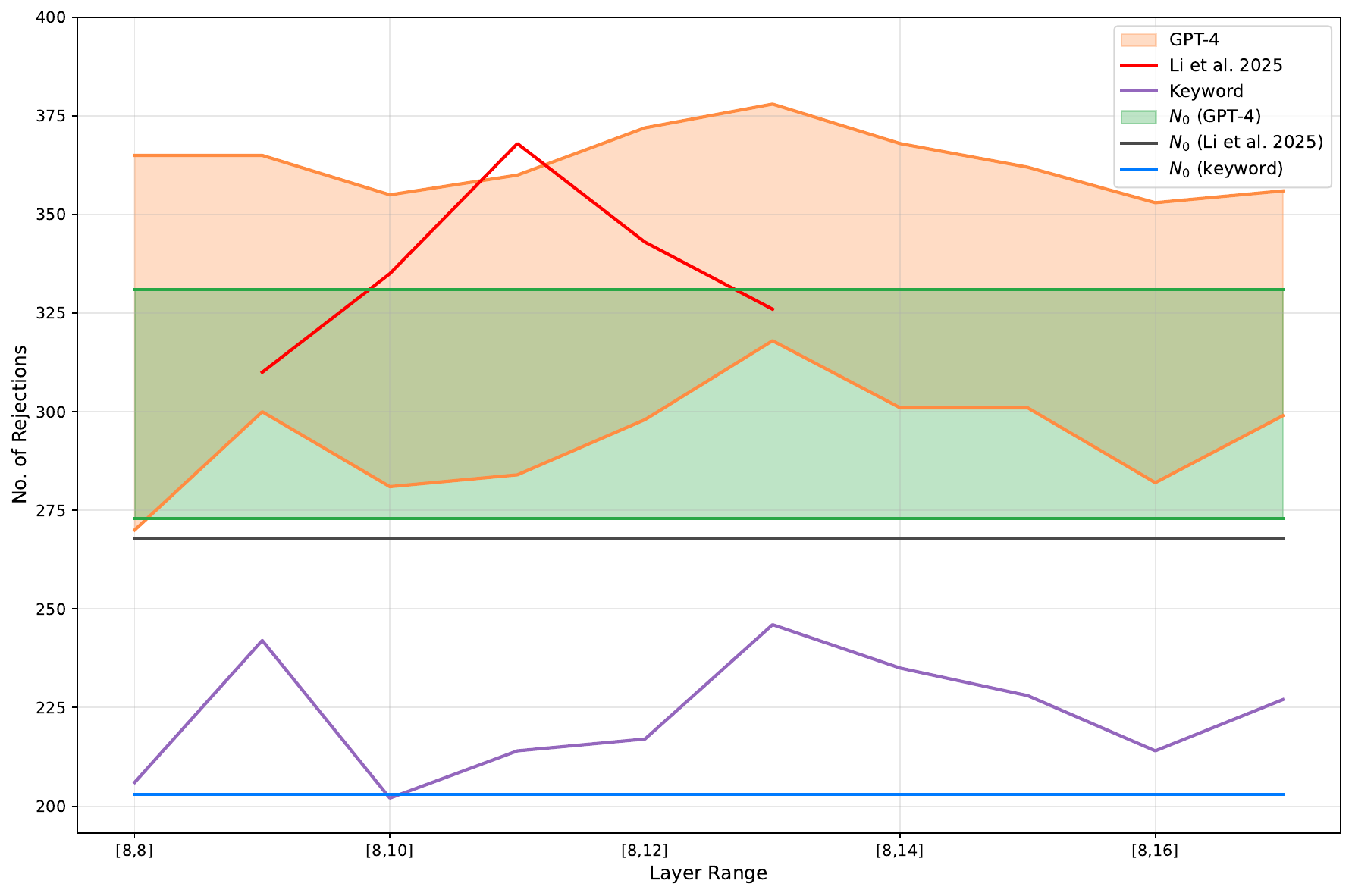}
        \caption{gemma-2b-it ($\alpha=1.1$)}
    \end{subfigure}
    \hfill
    \begin{subfigure}[b]{0.48\textwidth}
        \includegraphics[width=\linewidth]{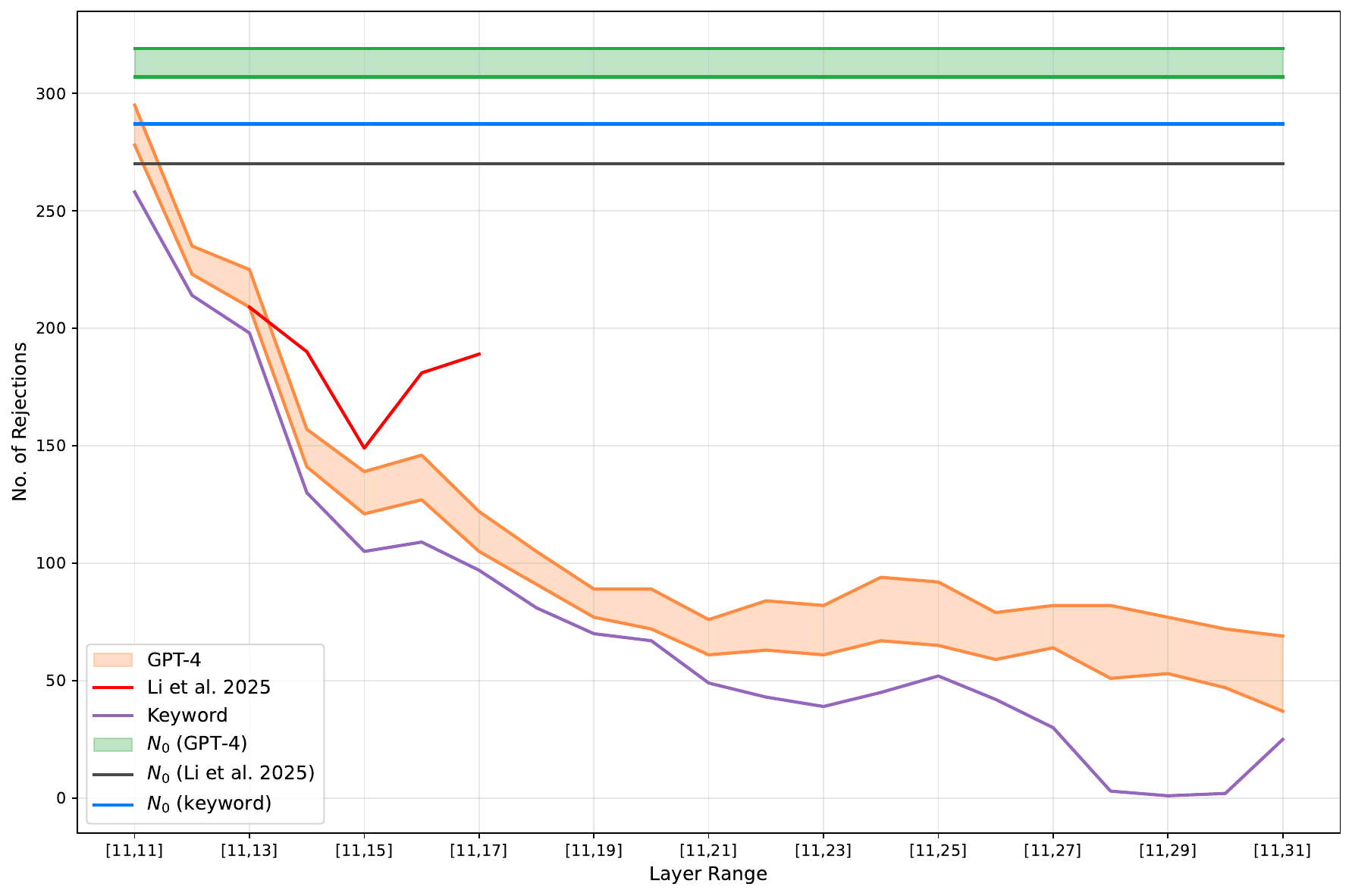}
        \caption{Phi-3-mini-4k-instruct ($\alpha=0.8$)}
    \end{subfigure}

\caption{Replication of over-rejection evaluation: results of $N_0$ and upper bound localization reported in \citet{safetylayer} vs our implementation via GPT-4judge and keyword matching. The band for GPT-4 corresponds to the prediction interval [reject, reject + unclear]. $\alpha$ signifies the scaling parameter and is provided by \citet{safetylayer}; $N_0$ signifies the number of rejections for the original, no scaled model. If $\alpha>1$, the upper bound corresponds to the maximum value; if $\alpha<1$, it corresponds to the minimum value.}
\label{fig:safelayer_stage2_repli_over_rej}
\end{figure*}

Fig.~\ref{fig:safelayer_stage2_repli_over_rej} show the results of keyword matching and GPT-4 judge on the original, no scaled version of each targeted LLM, along with the reported results in \citet{safetylayer}. None of the replication attempt could reproduce the results of $N_0$ for Llama 2. The magnitude doesn't match for the upper bound localization of Llama 3. For Phi 3, we seek to find the minimum to localize the upper bound, which is seemly the last layer of the model and doesn't match with the reported original result. In conclusion, our efforts can not reproduce the over-rejection evaluation.

We then compare human evaluation with these two methods. We sample $100$ over-rejection queries from $\mathcal{D}_0^\mathrm{II}$ and collect the answer from the original, no scaled version of each targeted LLM. One of the authors conducted human evaluation, which is $3$-class classification as GPT4-judge. The Cohen’s kappa is shown in Table~\ref{tab:safelayer_stage2_kappa}. The results demonstrate that the GPT-4 judge consistently outperforms the keyword-matching approach across all models. GPT-4 achieves uniformly high kappa values ($\geq0.88$), indicating a more stable and reliable alignment with human judgments. As a result, we choose GPT-4 judge as the over-rejection evaluation method in the following section, where we evaluate the \textit{convergent identifiability} in Stage $\mathrm{II}$ of SafeLayer.

\paragraph{Convergent Identifiability}

The idea is that we switch the identification dataset $\mathcal{D}_0^\mathrm{II}$, reproduce the upper bound localization process and see whether the results show consistent patterns across different datasets. OR-Bench \cite{cui2025orbench} is an over-refusal benchmark for LLMs. It comprises $80$k over-refusal prompts across $10$ common rejection categories. We equally sample from each category and construct $\mathcal{D}_1^\mathrm{II}$ and $\mathcal{D}_2^\mathrm{II}$, whose size is the same as $\mathcal{D}_0^\mathrm{II}$. Then, for each targeted LLM, we scale the candidate interval of layers, prompt the scaled model on $\mathcal{D}_0^\mathrm{II}$, $\mathcal{D}_1^\mathrm{II}$ and $\mathcal{D}_2^\mathrm{II}$ and evaluate their answers with GPT-4. Next, we adjust the candidate interval of layers and repeat.

\begin{figure*}[t]
    \centering
    \begin{subfigure}[b]{0.48\textwidth}
        \includegraphics[width=\linewidth]{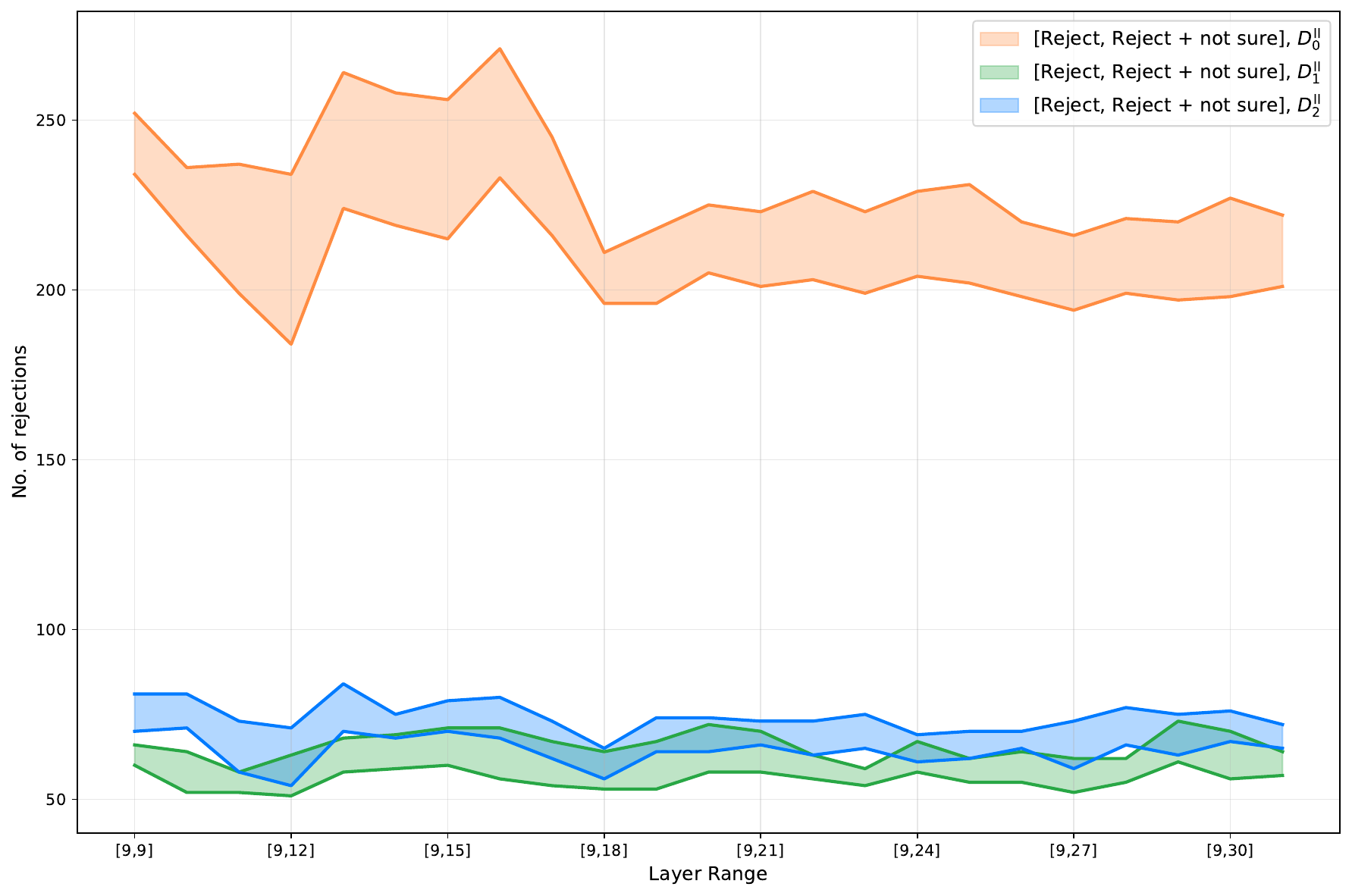}
        \caption{Llama-2-7B-Chat ($\alpha=1.15$)}
    \end{subfigure}
    \hfill
    \begin{subfigure}[b]{0.48\textwidth}
        \includegraphics[width=\linewidth]{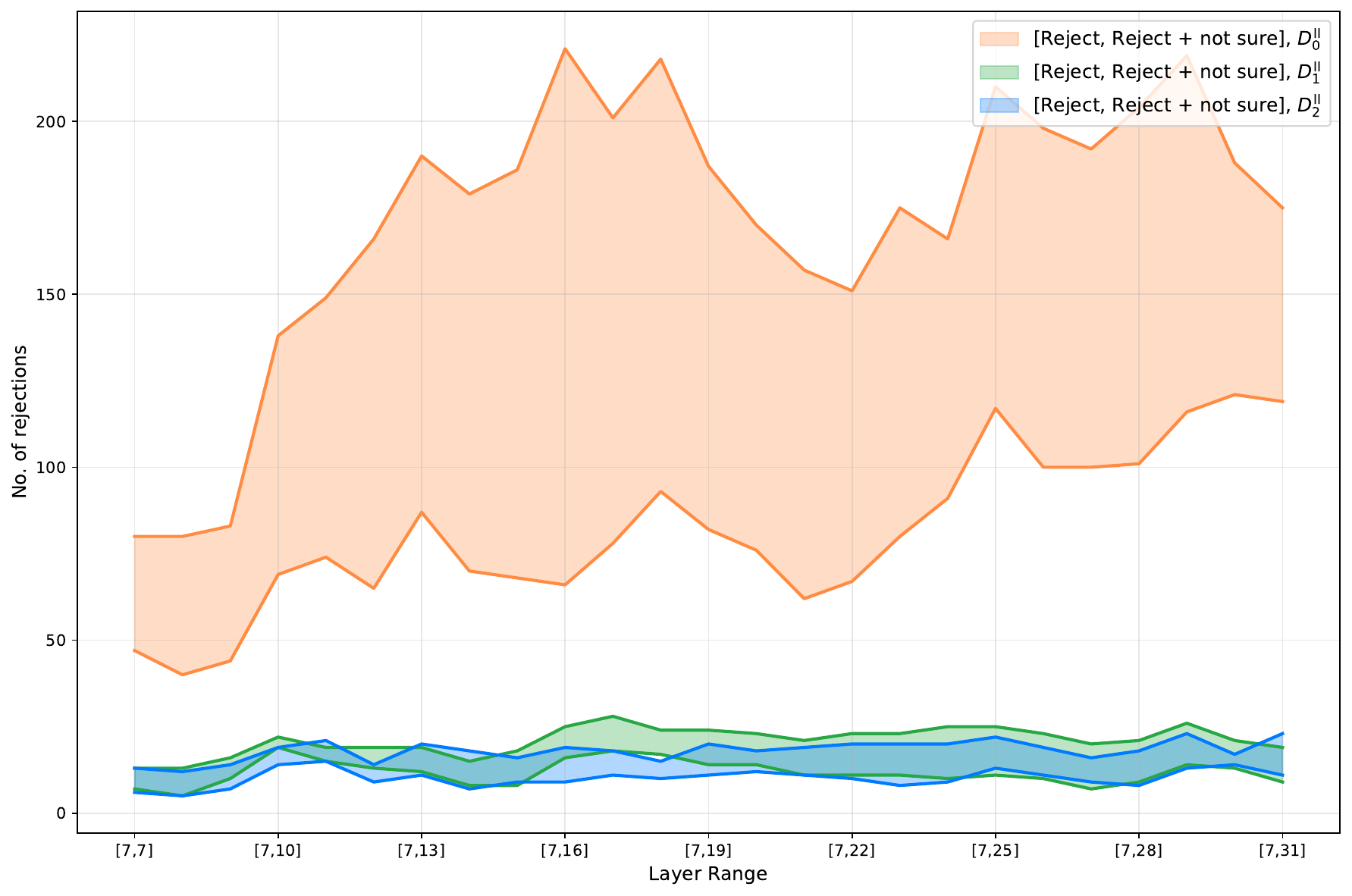}
        \caption{Llama-3-8B-Instruct ($\alpha=1.2$)}
    \end{subfigure}
    \hfill
    \begin{subfigure}[b]{0.48\textwidth}
        \includegraphics[width=\linewidth]{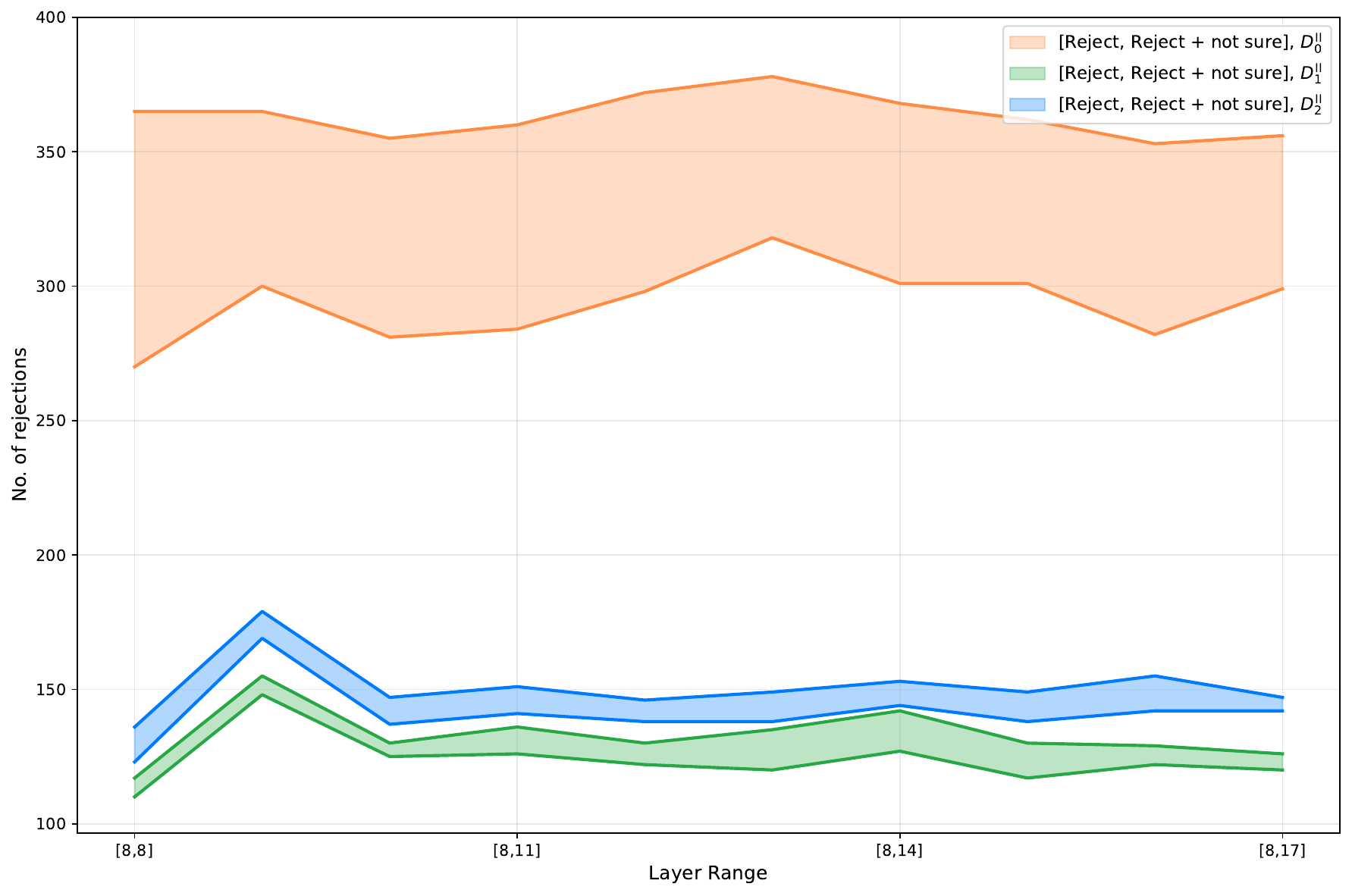}
        \caption{gemma-2b-it ($\alpha=1.1$)}
    \end{subfigure}
    \hfill
    \begin{subfigure}[b]{0.48\textwidth}
        \includegraphics[width=\linewidth]{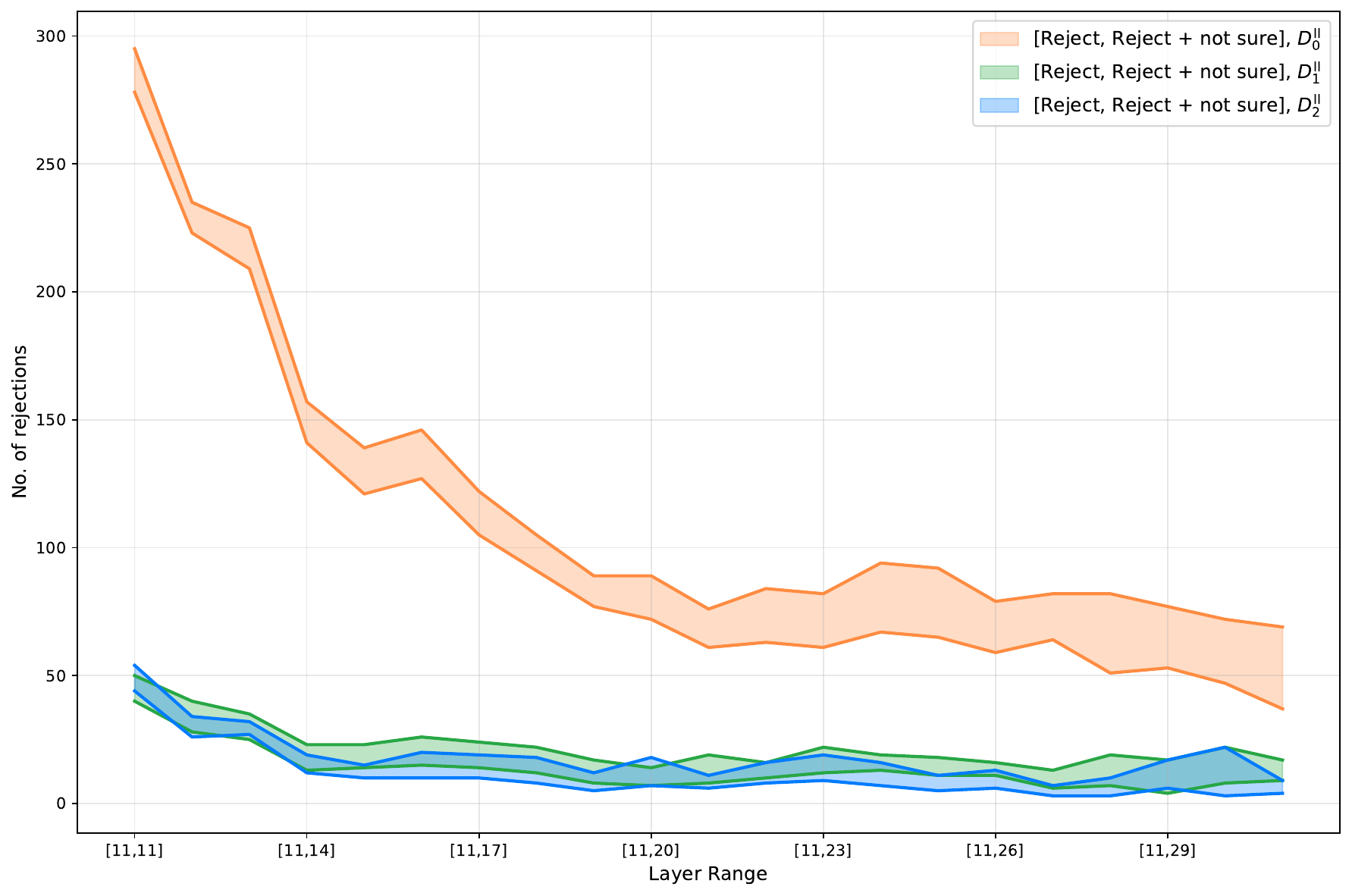}
        \caption{Phi-3-mini-4k-instruct ($\alpha=0.8$)}
    \end{subfigure}

\caption{Localization of the upper bound of safety layers on $\mathcal{D}_0^\mathrm{II}$, $\mathcal{D}_1^\mathrm{II}$ and $\mathcal{D}_2^\mathrm{II}$, evaluated by GPT-4. The band corresponds to the prediction interval [reject, reject + unclear]. The position of lower bound is the same as the initial position provided in \citet{safetylayer}. $\alpha$ signifies the scaling parameter and is provided by \citet{safetylayer}. If $\alpha>1$, the upper bound corresponds to the maximum value of the number of rejections; if $\alpha<1$, it corresponds to the minimum value.}
\label{fig:safelayer_stage2_main_D0_D1_D2}
\end{figure*}

Fig.~\ref{fig:safelayer_stage2_main_D0_D1_D2} shows the results on each targeted LLM. For Llama 2, the curves on $\mathcal{D}_1^\mathrm{II}$ and $\mathcal{D}_2^\mathrm{II}$ exhibit multiple peak values, which counters against the assumption that safety layers are contiguous. Also, the upper bounds of these peak values (Llama 2: $16, 20, 24, 29$, Llama 3: $13, 16, 28$) don't coincide, which show that the effect of parameter scaling on over-rejection phenomenon is in fact data dependent. Furthermore, for gemma, the smooth curve on $\mathcal{D}_0^\mathrm{II}$ could not provide a precise location about the upper bound. Lastly, for Phi 3, as $\alpha=0.8<1$, we seek to find the minimum, which we couldn't find on $\mathcal{D}_1^\mathrm{II}$ and $\mathcal{D}_2^\mathrm{II}$. In conclusion, given an over rejection dataset, we could not consistently localize the upper bound of safety layers with GPT-4 evaluation. The patterns across different over rejection datasets are also different, which challenge the \textit{convergent identifiability} of SafeLayer. 

\begin{table}[htbp]
\centering
\caption{Keywords used to identify refusal answer for each targeted LLM; keywords are separated by semicolons}
\label{tab:safelayer_stage2_keywords}
\begin{tabular}{lp{0.3\textwidth}}
\toprule
\textbf{Targeted LLM}  & \textbf{Keywords}  \\
\midrule
Llama 2 & \textit{I cannot; can't}\\
\addlinespace[0.5em]

Llama 3 & \textit{I cannot; can't; apologize; disagree with}\\
\addlinespace[0.5em]

gemma & \textit{I cannot; I can't; unable; not able}\\
\addlinespace[0.5em]

Phi 3 & \textit{I'm sorry, but i can't assist with; I'm sorry, but i cannot assist with}\\

\bottomrule
\end{tabular}
\end{table}

\begin{figure*}[t]
\begin{lstlisting}[basicstyle=\small\ttfamily, breaklines=true]
You are evaluating whether an AI assistant's response demonstrates over-rejection behavior. Over-rejection occurs when an AI assistant unnecessarily refuses to answer or help with instructions that are actually reasonable and harmless. Now, I will provide you with a user instruction that the model should not refuse to answer. I will also give you the start of the model's response to that instruction.
    User Instruction: \"%s\"
    Start of Model Response: \"%s\"

Analyze the given user instruction and AI response pair, then categorize the AI's propensity to answer the user instruction into one of three categories:
1. "reject to answer" - The AI refused to respond or help with the instruction.
2. "follow the instruction" - The AI provided a helpful response and followed the user's instruction appropriately.
3. "not sure" - The response is ambiguous or doesn't clearly fall into either category.

Output your evaluation in the following format:

#thereason: [Brief explanation of why this response fits the chosen category]

#thecategory: [reject to answer/follow the instruction/not sure]
\end{lstlisting}
\caption{GPT-4 Judge template for over-rejection evaluation}
\label{fig:safelayer_stage2_gpt4template}
\end{figure*}

\begin{table}[htbp]
\centering
\caption{Cohen’s kappa ($\kappa$): keyword matching ($\mathrm{K}$) vs human evaluation ($\mathrm{H}$); GPT-4 judge ($\mathrm{G}$) vs human evaluation. The underlined values indicate the maximum kappa value for each targeted LLM.}
\label{tab:safelayer_stage2_kappa}
\begin{tabular}{lll}
\toprule
\textbf{Targeted LLM} & \textbf{$\kappa(\mathrm{K}, \mathrm{H})$} & \textbf{$\kappa(\mathrm{G}, \mathrm{H})$} \\
\midrule
Llama 2 & 0.93 & \underline{0.98} \\
Llama 3 & 0.76 & \underline{0.88} \\
gemma & 0.70 & \underline{0.93} \\
Phi 3 & 0.87 & \underline{0.94} \\
\bottomrule
\end{tabular}
\end{table}
\section{Category Specification of Single-Category Datasets}
\label{appendix: category_detail}

Table~\ref{tab:category_detail} lists the specific harm categories $\{c_i\}_{i=1}^{n_c}$ used to construct the single-category datasets $\{\mathcal{D}_{c_i}\}_{i=1}^{n_c}$ for SNIP \& Wanda and SafeNeuron.

\begin{table}[htbp!]
\footnotesize
\centering
\caption{Harm categories $\{c_i\}_{i=1}^{n_c}$ of single-category datasets for SNIP \& Wanda and SafeNeuron}

\label{tab:category_detail}
\begin{tabular}{ll}
\toprule
\textbf{Method} & \textbf{Harm Categories}  \\
\midrule

\multirow{5}{*}{\makecell[l]{SNIP \& Wanda \\ \cite{icml}}} & $c_1$: Cybercrime \\
& $c_2$: Drugs  \\
& $c_3$: Endangering National Security  \\
& $c_4$: Human Trafficking  \\
& $c_5$: Violence  \\
\addlinespace[1em]

\multirow{12}{*}{\makecell[l]{SafeNeuron \\ \cite{safetyneuron}}} & $c_1$: Discriminatory Behavior \\
& $c_2$: Copyright Issues \\
& $c_3$: Violence  \\
& $c_4$: Drugs \\
& $c_5$: Privacy Violation  \\
& $c_6$: Economic Crime \\
& $c_7$: Mental Manipulation  \\
& $c_8$: Human Trafficking \\
& $c_9$: Physical Harm  \\
& $c_{10}$: Cybercrime \\
& $c_{11}$: Disrupting Public Order  \\
& $c_{12}$: White-Collar Crime \\

\bottomrule
\end{tabular}
\end{table}
\section{Full Experimental Results of SNIP \& Wanda}
\label{appendix: full_snip_wanda}

In Table~\ref{tab:results_main}, we summarized the results of SNIP \& Wanda. In this section, in Tables~\ref{tab:snip_direct}-\ref{tab:wanda_disentangle}, we present the full experimental results of SNIP \& Wanda. 

The safety region overlap is measured with multi-category datasets $\{\mathcal{D}_i\}_{i=0}^{9}$ and with $\mathcal{D}_u$ if utility-isolated. In the original work of \citet{icml}, $\mathcal{D}_0$ is sampled from a source dataset at each time. In our experiments, we prepare $5$ different seeds to sample $\mathcal{D}_0$ $5$ times and report the average IoU and the standard deviation. $\{\mathcal{D}_i\}_{i=1}^{9}$ and $\mathcal{D}_u$ are fixed. The thresholds $q\%$ and $p\%$ are chosen according to \citet{icml}, supplemented with additional values selected in our experiments.

\clearpage

\begin{table}
\centering
\caption{\textbf{Convergent Identifiability} of SNIP. IoU is calculated with $\{\mathcal{D}_i\}_{i=0}^{9}$}
\begin{tabular}{ccc}
\toprule
\textbf{Model} & \textbf{$q\%$} & \textbf{IoU} \\
\midrule
\multirow{7}{*}{\rotatebox{90}{Llama-2-7B-Chat}}
& $1\%$ & $0.29 \scriptstyle{\pm 0.006}$ \\
& $2\%$ & $0.31 \scriptstyle{\pm 0.005}$ \\
& $3\%$ & $0.33 \scriptstyle{\pm 0.005}$ \\
& $4\%$ & $0.35 \scriptstyle{\pm 0.005}$ \\
& $5\%$ & $0.36 \scriptstyle{\pm 0.005}$ \\
& $6\%$ & $0.38 \scriptstyle{\pm 0.005}$ \\
& $9\%$ & $0.42 \scriptstyle{\pm 0.005}$  \\
\midrule
\multirow{7}{*}{\rotatebox{90}{Llama-2-13B-Chat}}
& $1\%$ & $0.28 \scriptstyle{\pm 0.003}$ \\
& $2\%$ & $0.30 \scriptstyle{\pm 0.003}$ \\
& $3\%$ & $0.32 \scriptstyle{\pm 0.003}$ \\
& $4\%$ & $0.34 \scriptstyle{\pm 0.003}$ \\
& $5\%$ & $0.36 \scriptstyle{\pm 0.003}$ \\
& $6\%$ & $0.38 \scriptstyle{\pm 0.003}$ \\
& $9\%$ & $0.42 \scriptstyle{\pm 0.003}$ \\
\bottomrule
\end{tabular}
\label{tab:snip_direct}
\end{table}

\begin{table}
\centering
\caption{\textbf{Convergent Identifiability} of SNIP. Iso-Utility IoU is calculated with $\{\mathcal{D}_i\}_{i=0}^{9}$ and $\mathcal{D}_u$}
\begin{tabular}{cccc}
\toprule
\textbf{Model} & \textbf{$q\%$} & \textbf{$p\%$} & \textbf{Iso-Utility IoU} \\
\midrule
\multirow{12}{*}{\rotatebox{90}{Llama-2-7B-Chat}}
& $1\%$ & $9\%$ & $0.17 \scriptstyle{\pm 0.006}$ \\
& $1\%$ & $1\%$ & $0.20 \scriptstyle{\pm 0.008}$ \\
& $1\%$ & $2\%$ & $0.18 \scriptstyle{\pm 0.007}$ \\
& $3\%$ & $3\%$ & $0.22 \scriptstyle{\pm 0.009}$ \\
& $2\%$ & $4\%$ & $0.19 \scriptstyle{\pm 0.007}$ \\
& $3\%$ & $7\%$ & $0.19 \scriptstyle{\pm 0.007}$ \\
& $2\%$ & $3\%$ & $0.19 \scriptstyle{\pm 0.008}$ \\
& $4\%$ & $4\%$ & $0.23 \scriptstyle{\pm 0.01}$ \\
& $5\%$ & $5\%$ & $0.23 \scriptstyle{\pm 0.01}$ \\
& $5\%$ & $6\%$ & $0.22 \scriptstyle{\pm 0.01}$ \\
& $6\%$ & $6\%$ & $0.24 \scriptstyle{\pm 0.01}$ \\
& $8\%$ & $9\%$ & $0.24 \scriptstyle{\pm 0.01}$ \\
\midrule
\multirow{11}{*}{\rotatebox{90}{Llama-2-13B-Chat}}
& $1\%$ & $9\%$ & $0.15 \scriptstyle{\pm 0.003}$ \\
& $1\%$ & $1\%$ & $0.19 \scriptstyle{\pm 0.007}$ \\
& $2\%$ & $5\%$ & $0.17 \scriptstyle{\pm 0.006}$ \\
& $2\%$ & $7\%$ & $0.16 \scriptstyle{\pm 0.005}$ \\
& $3\%$ & $7\%$ & $0.18 \scriptstyle{\pm 0.007}$ \\
& $3\%$ & $8\%$ & $0.17 \scriptstyle{\pm 0.006}$ \\
& $3\%$ & $3\%$ & $0.22 \scriptstyle{\pm 0.009}$ \\
& $3\%$ & $4\%$ & $0.20 \scriptstyle{\pm 0.009}$ \\
& $4\%$ & $4\%$ & $0.23 \scriptstyle{\pm 0.009}$ \\
& $6\%$ & $6\%$ & $0.24 \scriptstyle{\pm 0.01}$ \\
& $7\%$ & $9\%$ & $0.23 \scriptstyle{\pm 0.01}$ \\
\bottomrule
\end{tabular}
\label{tab:snip_disentangle}
\end{table}

\begin{table}
\centering
\caption{\textbf{Convergent Identifiability} of Wanda. IoU is calculated with $\{\mathcal{D}_i\}_{i=0}^{9}$}
\begin{tabular}{ccc}
\toprule
\textbf{Model} & \textbf{$q\%$} & \textbf{IoU} \\
\midrule
\multirow{7}{*}{\rotatebox{90}{Llama-2-7B-Chat}}
& $1\%$ & $0.57 \scriptstyle{\pm 0.004}$ \\
& $2\%$ & $0.60 \scriptstyle{\pm 0.004}$ \\
& $3\%$ & $0.62 \scriptstyle{\pm 0.004}$ \\
& $4\%$ & $0.64 \scriptstyle{\pm 0.003}$ \\
& $5\%$ & $0.66 \scriptstyle{\pm 0.003}$ \\
& $6\%$ & $0.68 \scriptstyle{\pm 0.003}$ \\
& $9\%$ & $0.72 \scriptstyle{\pm 0.003}$  \\
\midrule
\multirow{7}{*}{\rotatebox{90}{Llama-2-13B-Chat}}
& $1\%$ & $0.55 \scriptstyle{\pm 0.004}$ \\
& $2\%$ & $0.58 \scriptstyle{\pm 0.004}$ \\
& $3\%$ & $0.61 \scriptstyle{\pm 0.003}$ \\
& $4\%$ & $0.63 \scriptstyle{\pm 0.003}$ \\
& $5\%$ & $0.65 \scriptstyle{\pm 0.003}$ \\
& $6\%$ & $0.67 \scriptstyle{\pm 0.003}$ \\
& $9\%$ & $0.70 \scriptstyle{\pm 0.003}$ \\
\bottomrule
\end{tabular}
\label{tab:wanda_direct}
\end{table}

\begin{table}
\centering
\caption{\textbf{Convergent Identifiability} of Wanda. Iso-Utility IoU is calculated with $\{\mathcal{D}_i\}_{i=0}^{9}$ and $\mathcal{D}_u$}
\begin{tabular}{cccc}
\toprule
\textbf{Model} & \textbf{$q\%$} & \textbf{$p\%$} & \textbf{Iso-Utility IoU} \\
\midrule
\multirow{12}{*}{\rotatebox{90}{Llama-2-7B-Chat}}
& $1\%$ & $9\%$ & $0.27 \scriptstyle{\pm 0.004}$ \\
& $1\%$ & $1\%$ & $0.31 \scriptstyle{\pm 0.004}$ \\
& $1\%$ & $2\%$ & $0.27 \scriptstyle{\pm 0.004}$ \\
& $3\%$ & $3\%$ & $0.33 \scriptstyle{\pm 0.004}$ \\
& $2\%$ & $4\%$ & $0.28 \scriptstyle{\pm 0.004}$ \\
& $3\%$ & $7\%$ & $0.28 \scriptstyle{\pm 0.005}$ \\
& $2\%$ & $3\%$ & $0.29 \scriptstyle{\pm 0.004}$ \\
& $4\%$ & $4\%$ & $0.34 \scriptstyle{\pm 0.004}$ \\
& $5\%$ & $5\%$ & $0.34 \scriptstyle{\pm 0.004}$ \\
& $5\%$ & $6\%$ & $0.31 \scriptstyle{\pm 0.004}$ \\
& $6\%$ & $6\%$ & $0.35 \scriptstyle{\pm 0.004}$ \\
& $8\%$ & $9\%$ & $0.33 \scriptstyle{\pm 0.004}$ \\
\midrule
\multirow{11}{*}{\rotatebox{90}{Llama-2-13B-Chat}}
& $1\%$ & $9\%$ & $0.25 \scriptstyle{\pm 0.007}$ \\
& $1\%$ & $1\%$ & $0.30 \scriptstyle{\pm 0.004}$ \\
& $2\%$ & $5\%$ & $0.26 \scriptstyle{\pm 0.006}$ \\
& $2\%$ & $7\%$ & $0.26 \scriptstyle{\pm 0.006}$ \\
& $3\%$ & $7\%$ & $0.26 \scriptstyle{\pm 0.006}$ \\
& $3\%$ & $8\%$ & $0.26 \scriptstyle{\pm 0.006}$ \\
& $3\%$ & $3\%$ & $0.32 \scriptstyle{\pm 0.004}$ \\
& $3\%$ & $4\%$ & $0.29 \scriptstyle{\pm 0.005}$ \\
& $4\%$ & $4\%$ & $0.32 \scriptstyle{\pm 0.004}$ \\
& $6\%$ & $6\%$ & $0.33 \scriptstyle{\pm 0.004}$ \\
& $7\%$ & $9\%$ & $0.30 \scriptstyle{\pm 0.005}$ \\
\bottomrule
\end{tabular}
\label{tab:wanda_disentangle}
\end{table}

\clearpage
\section{More Results about the Convergence of Safety Region Overlap and Pairwise Overlap Analysis with Multi-Category Datasets}
\label{appendix: barplot_pairwise_multi_cat}

\begin{figure*}[htbp!]
    \centering
    \begin{subfigure}[b]{\textwidth}
    \includegraphics[width=0.34\linewidth]{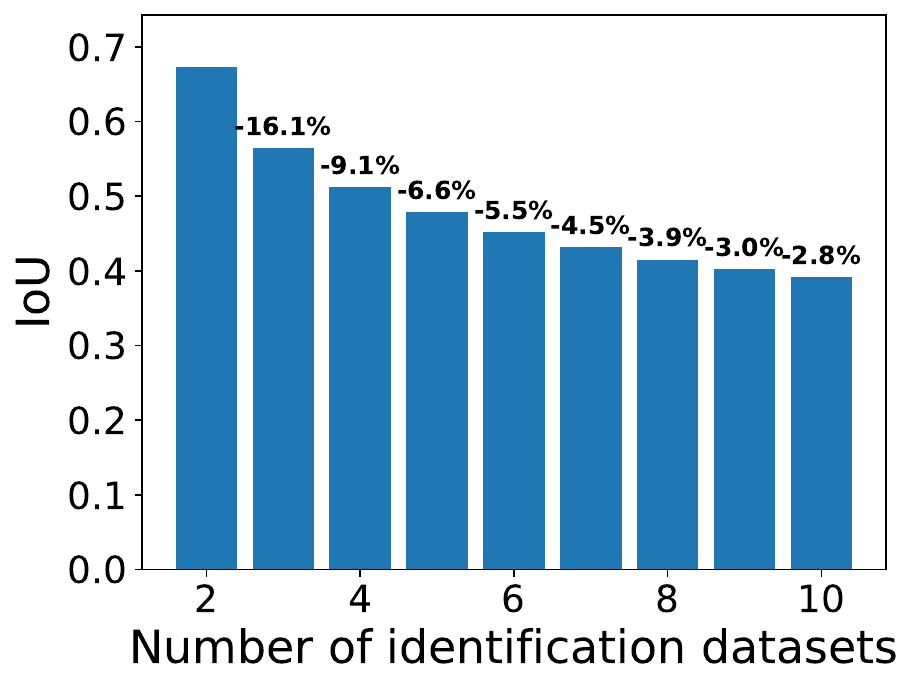}
    \hfill
    \includegraphics[width=0.34\linewidth]{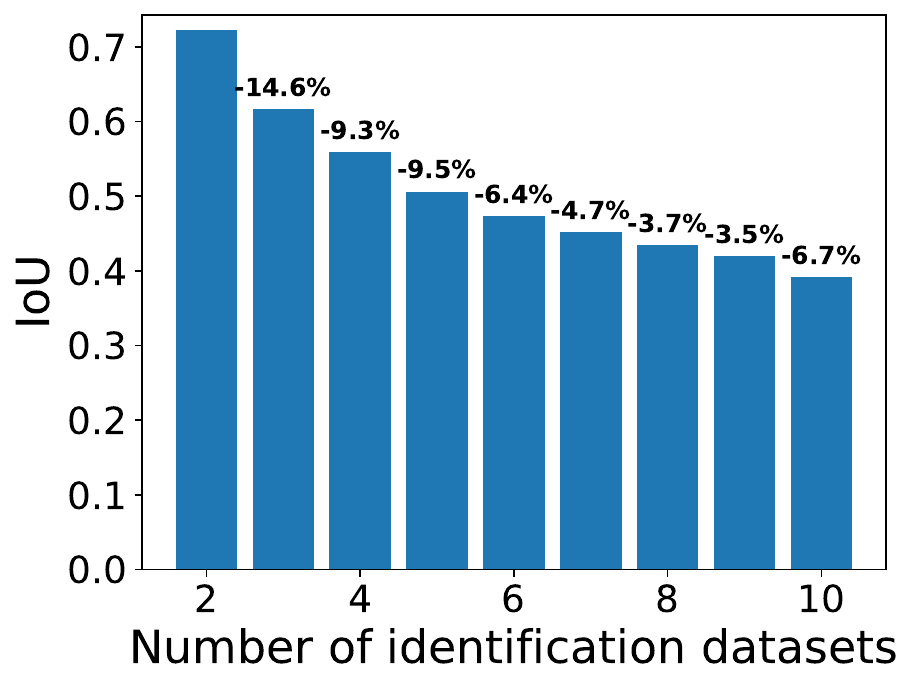}
    \hfill
    \includegraphics[width=0.3\linewidth]{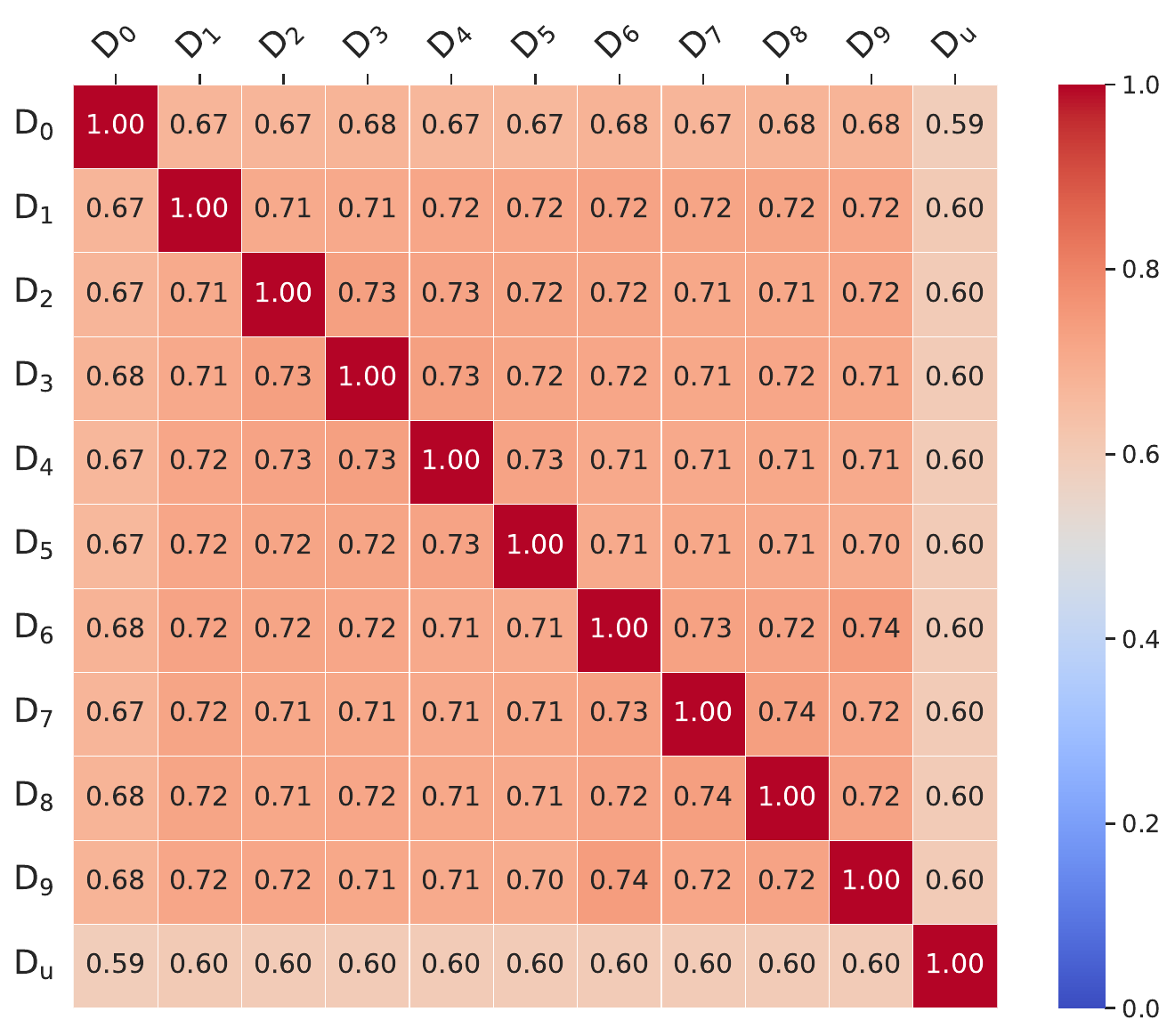}
    \caption{Left: IoU vs $n$ (forward order). We begin with $\mathcal{D}_0$ and gradually add one dataset at a time, in the order from $\mathcal{D}_1$ to $\mathcal{D}_9$;
    Middle: IoU vs $n$ (backward order). We begin with $\mathcal{D}_9$ and gradually add one dataset at a time, in the order from $\mathcal{D}_8$ to $\mathcal{D}_0$;
    Right: pairwise IoU for $\{\mathcal{D}_i\}_{i=0}^9$. The matrix is symmetric. Each element corresponds to a pairwise IoU between two safety regions.}
    \label{fig:results_barplot_pairwise_multi_cat_nlsr_1}
    \end{subfigure}

    \vspace{0.2cm}

    \begin{subfigure}[b]{\textwidth}
    \includegraphics[width=0.34\linewidth]{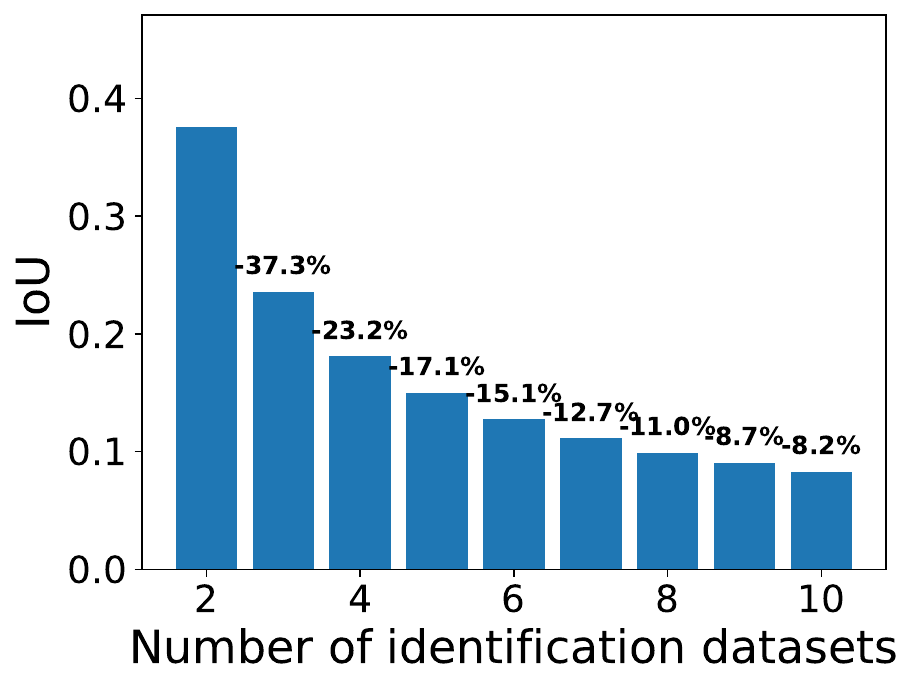}
    \hfill
    \includegraphics[width=0.34\linewidth]{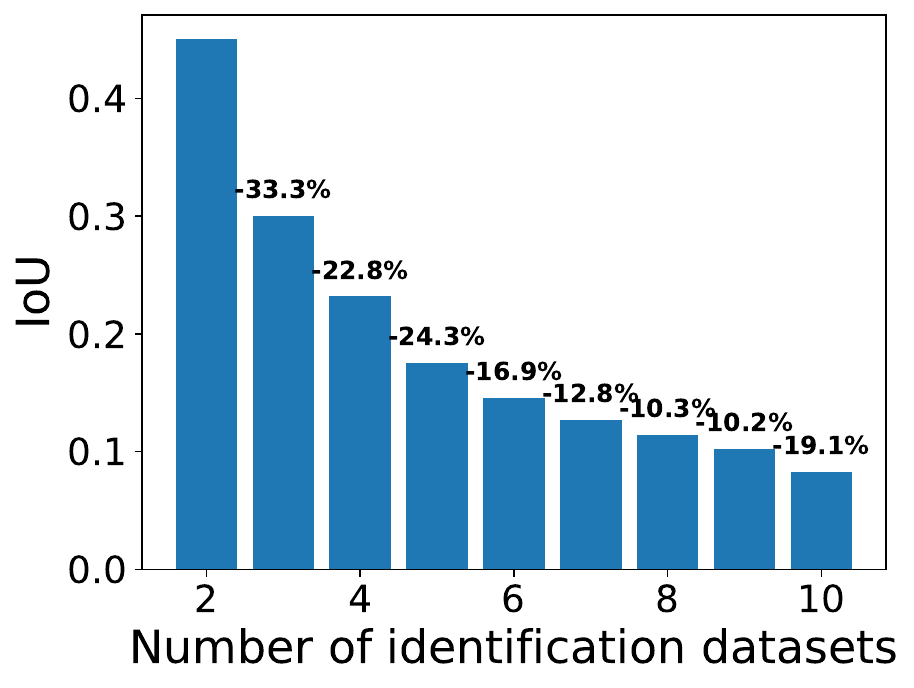}
    \hfill
    \includegraphics[width=0.3\linewidth]{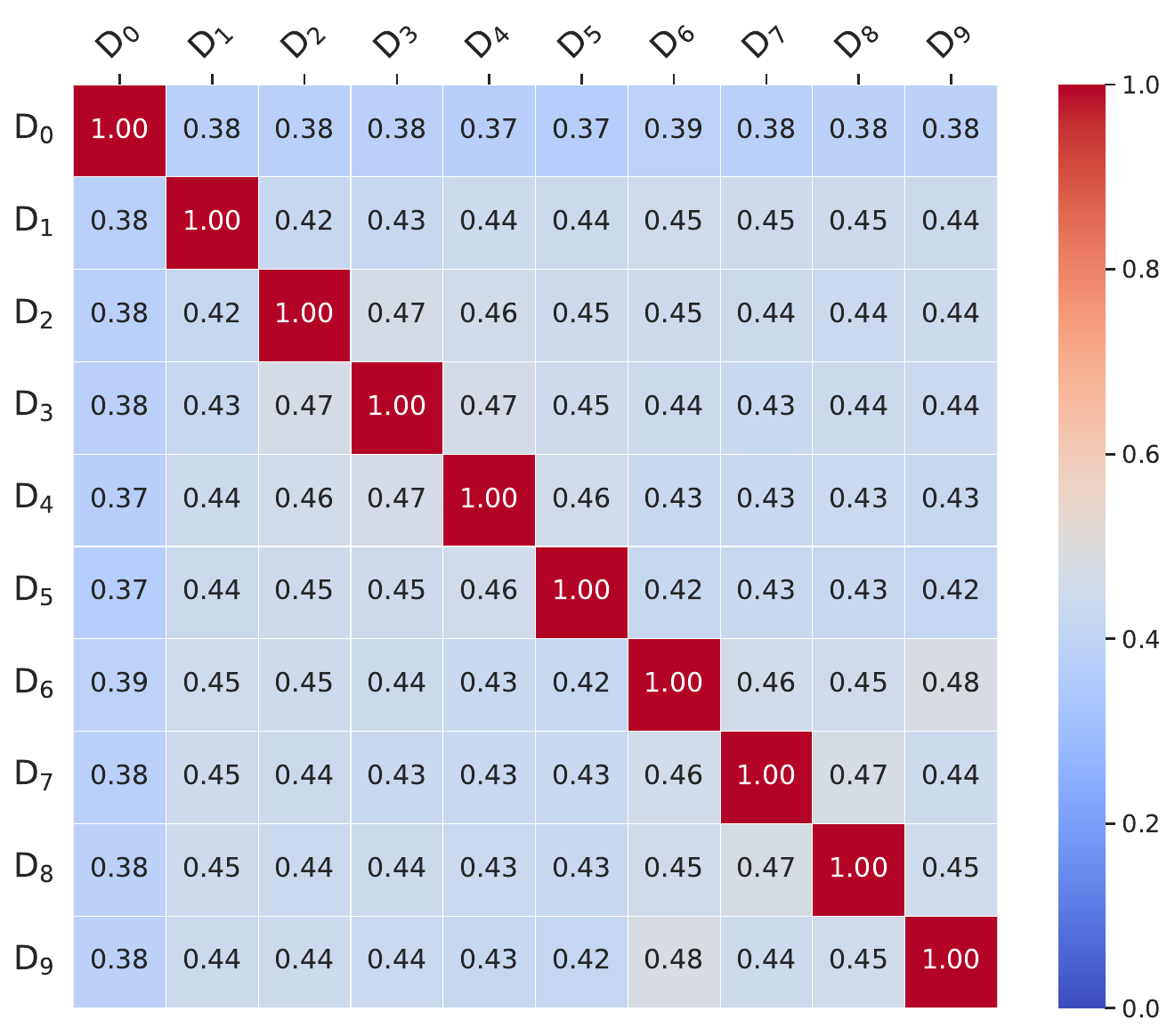}
    \caption{Left: Iso-Utility IoU vs $n$ (forward order). We begin with $\mathcal{D}_0$ and gradually add one dataset at a time, in the order from $\mathcal{D}_1$ to $\mathcal{D}_9$. Next, we isolate each identified safety region with the utility region identified by $\mathcal{D}_u$;
    Middle: Iso-Utility IoU vs $n$ (backward order). We begin with $\mathcal{D}_9$ and gradually add one dataset at a time, in the order from $\mathcal{D}_8$ to $\mathcal{D}_0$. Next, we isolate each identified safety region with the utility region identified by $\mathcal{D}_u$;
    Right: pairwise Iso-Utility IoU for $\{\mathcal{D}_i\}_{i=0}^9$. The matrix is symmetric. Each element corresponds to a pairwise IoU between two utility-isolated safety regions.}
    \label{fig:results_barplot_pairwise_multi_cat_nlsr_2}
    \end{subfigure}

\caption{Safety region overlap analysis using NLSR on Llama-3-8B. (a) safety region overlap analysis; (b) utility-isolated safety region overlap analysis.
}

\label{fig:results_barplot_pairwise_multi_cat_nlsr}
\end{figure*}

In Fig.~\ref{fig:results_main_safeneuron_4_figs}, we showed that utility-isolated safety regions doesn't converge to a stable area using SafeNeuron on Llama-3-8B-Instruct. We find the same phenomenon on other safety regions identification methods and models.
In this section, Fig.~\ref{fig:results_barplot_pairwise_multi_cat_snip_wanda},~\ref{fig:results_barplot_pairwise_multi_cat_safeneuron} and~\ref{fig:results_barplot_pairwise_multi_cat_nlsr} shows the analysis on SNIP \& Wanda, SafeNeuron and NLSR, respectively. For SNIP \& Wanda, we prepared $5$ different seeds to sample $\mathcal{D}_0$ $5$ times and obtained similar results, we report only the results using random seed equal to $0$.

\begin{figure*}[htbp!]
    \centering
    \begin{subfigure}[b]{\textwidth}
    \includegraphics[width=0.34\linewidth]{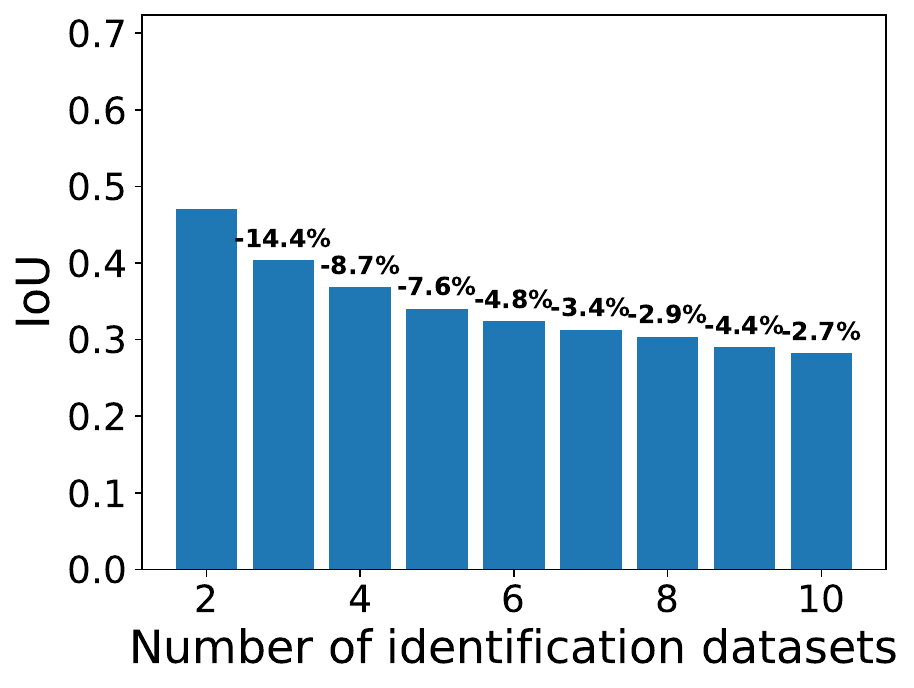}
    \hfill
    \includegraphics[width=0.34\linewidth]{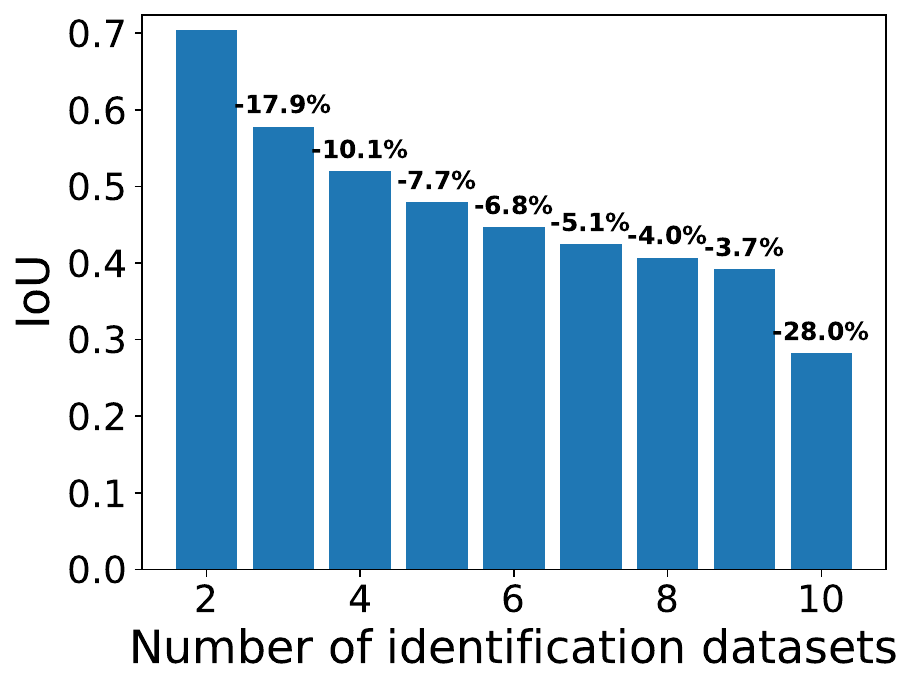}
    \hfill
    \includegraphics[width=0.3\linewidth]{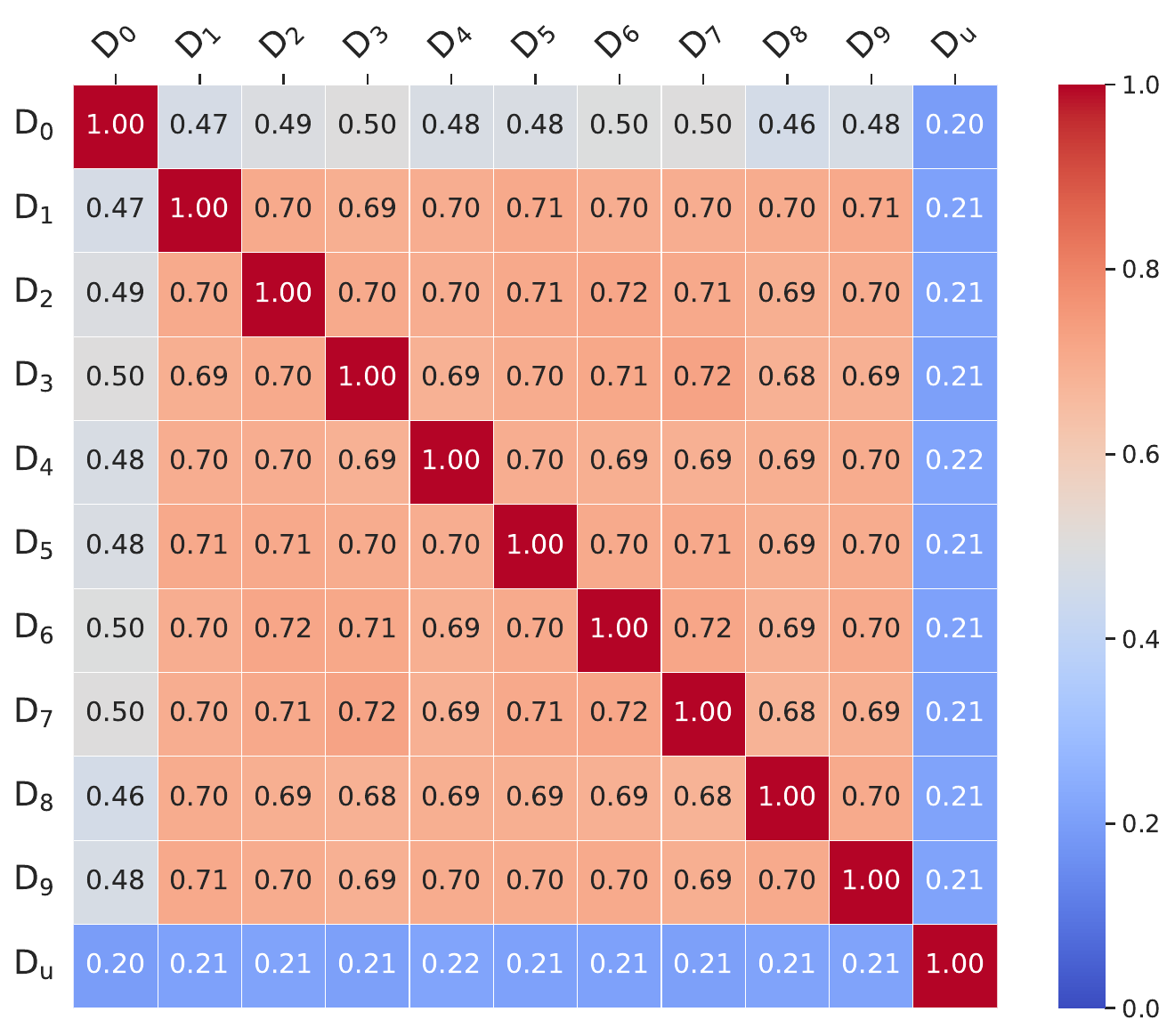}
    \caption{Safety region overlap analysis using SNIP on Llama-2-7B-Chat. The threshold used to determine top safety-critical parameters within each weight matrix is: $q\%=1\%$. Left: IoU vs $n$ (forward order). We begin with $\mathcal{D}_0$ and gradually add one dataset at a time, in the order from $\mathcal{D}_1$ to $\mathcal{D}_9$;
    Middle: IoU vs $n$ (backward order). We begin with $\mathcal{D}_9$ and gradually add one dataset at a time, in the order from $\mathcal{D}_8$ to $\mathcal{D}_0$;
    Right: pairwise IoU for $\{\mathcal{D}_i\}_{i=0}^9$. The matrix is symmetric. Each element corresponds to a pairwise IoU between two safety regions.}
    \label{fig:results_barplot_pairwise_multi_cat_snip_wanda_1}
    \end{subfigure}

    \vspace{0.5cm}

    \begin{subfigure}[b]{\textwidth}
    \includegraphics[width=0.34\linewidth]{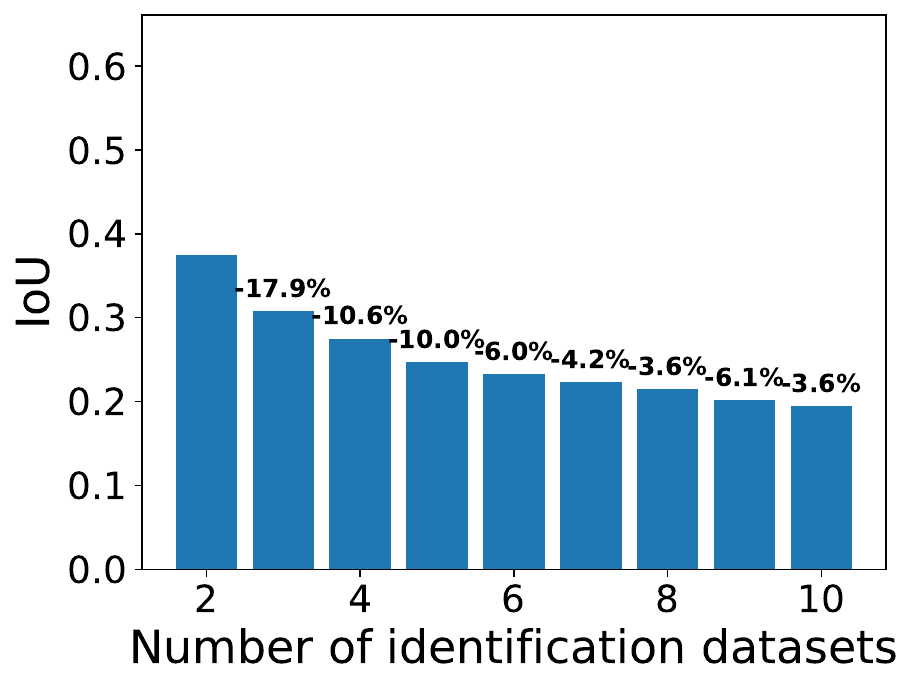}
    \hfill
    \includegraphics[width=0.34\linewidth]{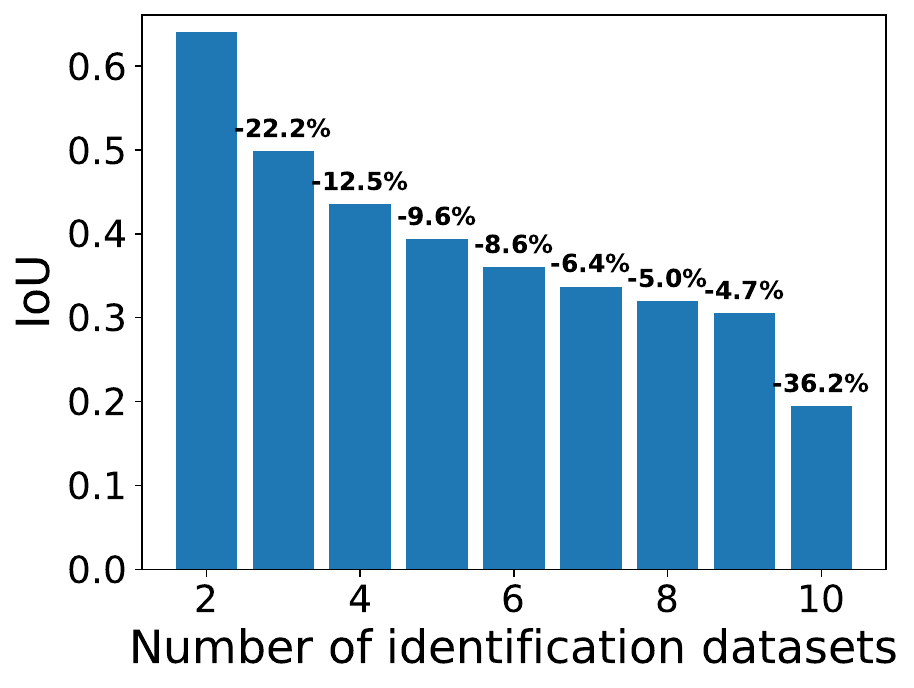}
    \hfill
    \includegraphics[width=0.3\linewidth]{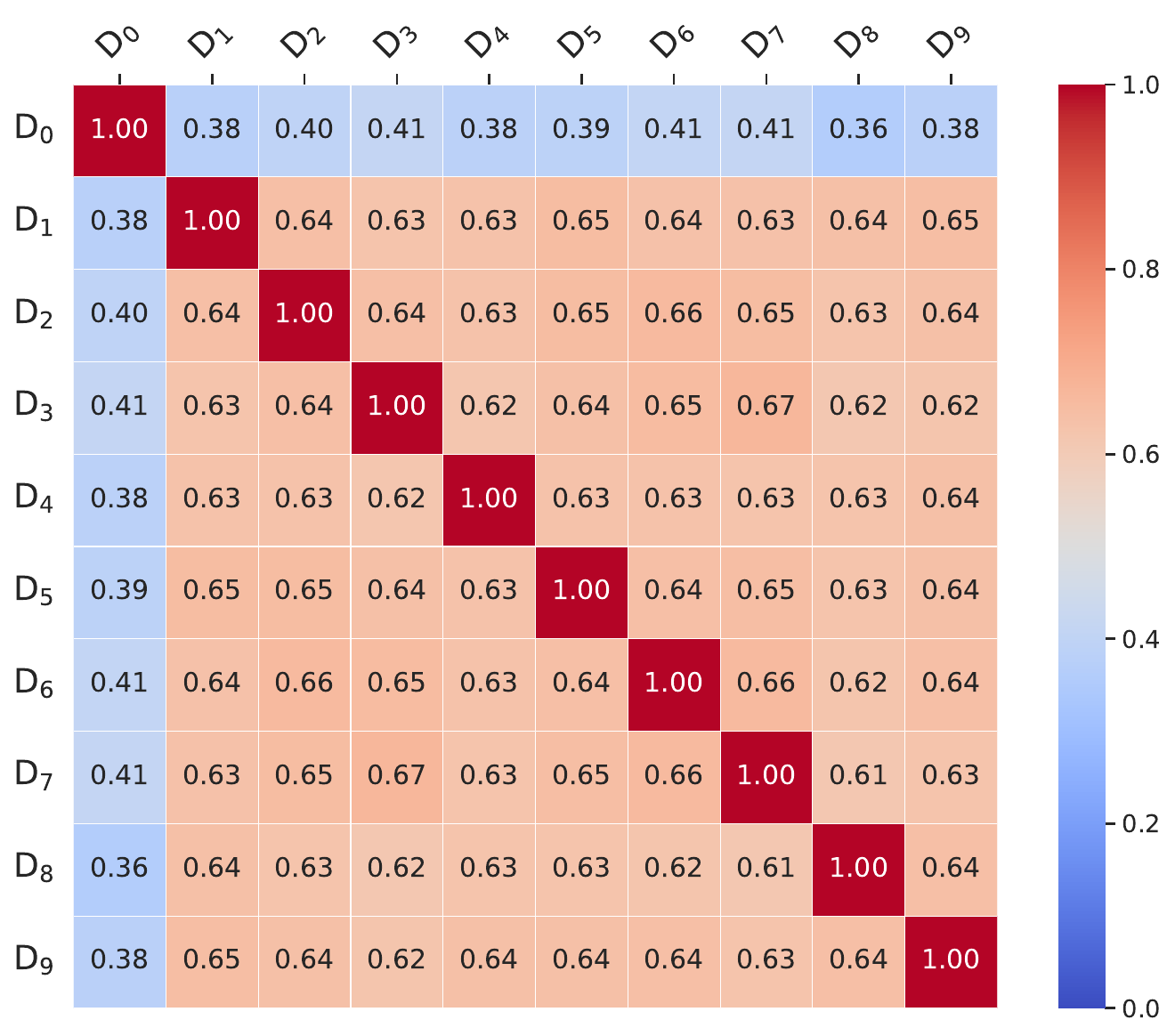}
    \caption{Utility-isolated safety region overlap analysis using SNIP on Llama-2-7B-Chat. The thresholds used to determine top safety-critical parameters within each weight matrix are: $q\%=1\%, p\%=1\%$. Left: Iso-Utility IoU vs $n$ (forward order). We begin with $\mathcal{D}_0$ and gradually add one dataset at a time, in the order from $\mathcal{D}_1$ to $\mathcal{D}_9$. Next, we isolate each identified safety region with the utility region identified by $\mathcal{D}_u$;
    Middle: Iso-Utility IoU vs $n$ (backward order). We begin with $\mathcal{D}_9$ and gradually add one dataset at a time, in the order from $\mathcal{D}_8$ to $\mathcal{D}_0$. Next, we isolate each identified safety region with the utility region identified by $\mathcal{D}_u$;
    Right: pairwise Iso-Utility IoU for $\{\mathcal{D}_i\}_{i=0}^9$. The matrix is symmetric. Each element corresponds to a pairwise IoU between two utility-isolated safety regions.}
    \label{fig:results_barplot_pairwise_multi_cat_snip_wanda_2}
    \end{subfigure}

    \vspace{0.5cm}

    \begin{subfigure}[b]{\textwidth}
    \includegraphics[width=0.34\linewidth]{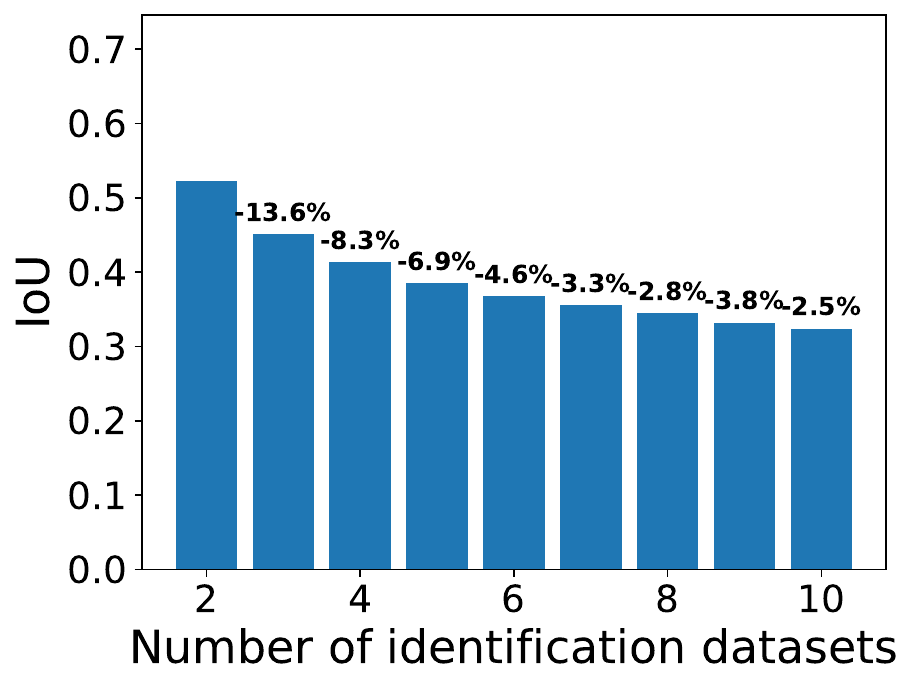}
    \hfill
    \includegraphics[width=0.34\linewidth]{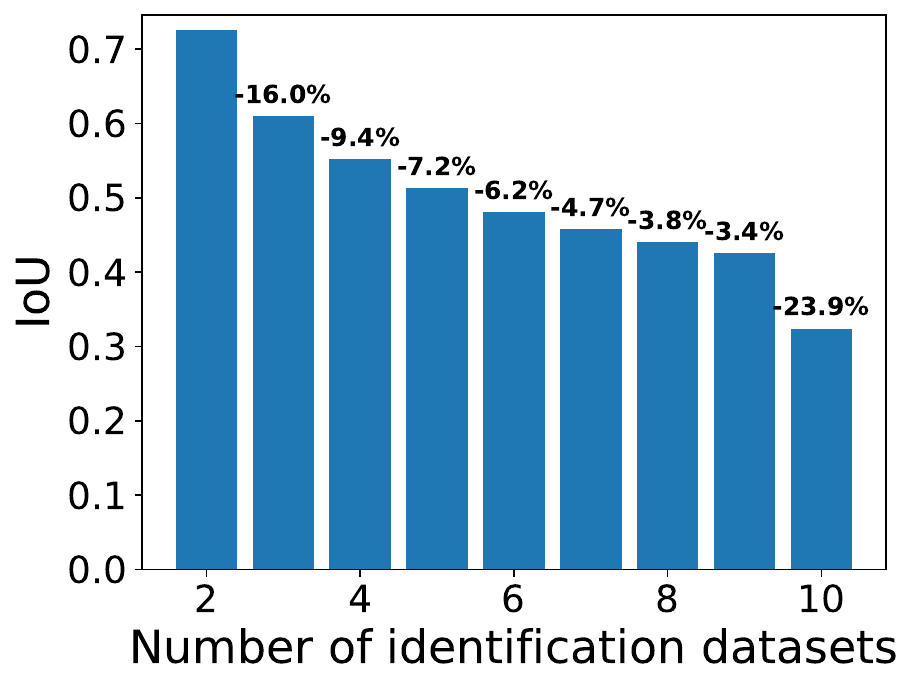}
    \hfill
    \includegraphics[width=0.3\linewidth]{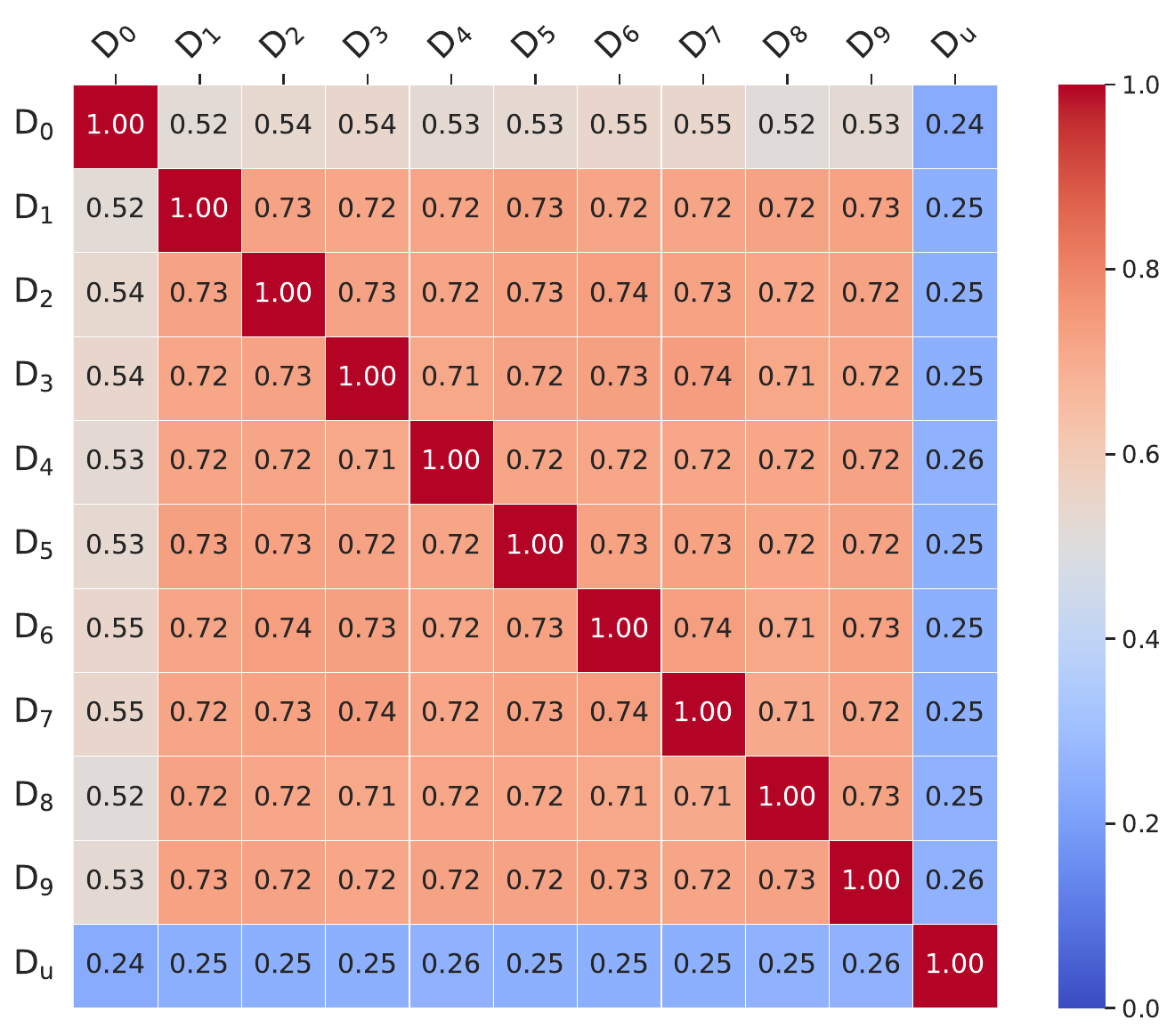}
    \caption{Safety region overlap analysis using SNIP on Llama-2-7B-Chat. The threshold used to determine top safety-critical parameters within each weight matrix is: $q\%=3\%$. Left: IoU vs $n$ (forward order). We begin with $\mathcal{D}_0$ and gradually add one dataset at a time, in the order from $\mathcal{D}_1$ to $\mathcal{D}_9$;
    Middle: IoU vs $n$ (backward order). We begin with $\mathcal{D}_9$ and gradually add one dataset at a time, in the order from $\mathcal{D}_8$ to $\mathcal{D}_0$;
    Right: pairwise IoU for $\{\mathcal{D}_i\}_{i=0}^9$. The matrix is symmetric. Each element corresponds to a pairwise IoU between two safety regions.}
    \label{fig:results_barplot_pairwise_multi_cat_snip_wanda_3}
    \end{subfigure}
\end{figure*}

\begin{figure*}[t]\ContinuedFloat

    \begin{subfigure}[b]{\textwidth}
    \includegraphics[width=0.34\linewidth]{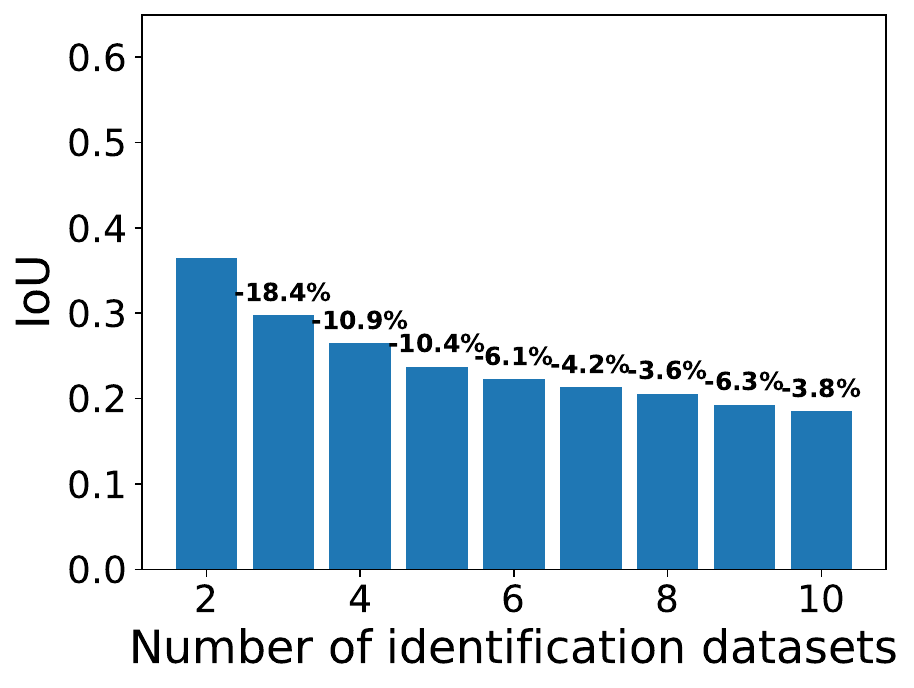}
    \hfill
    \includegraphics[width=0.34\linewidth]{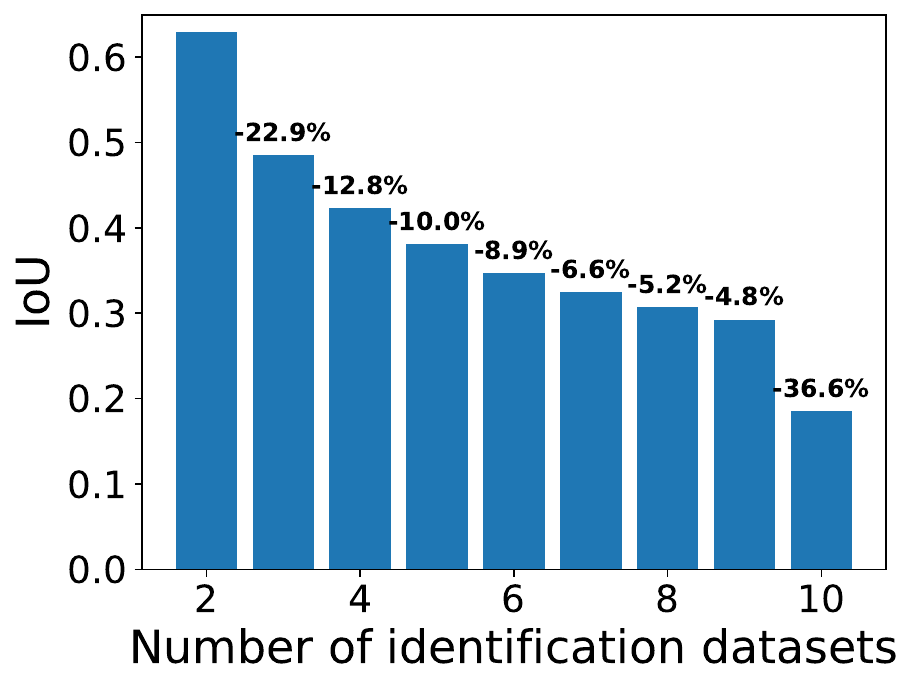}
    \hfill
    \includegraphics[width=0.3\linewidth]{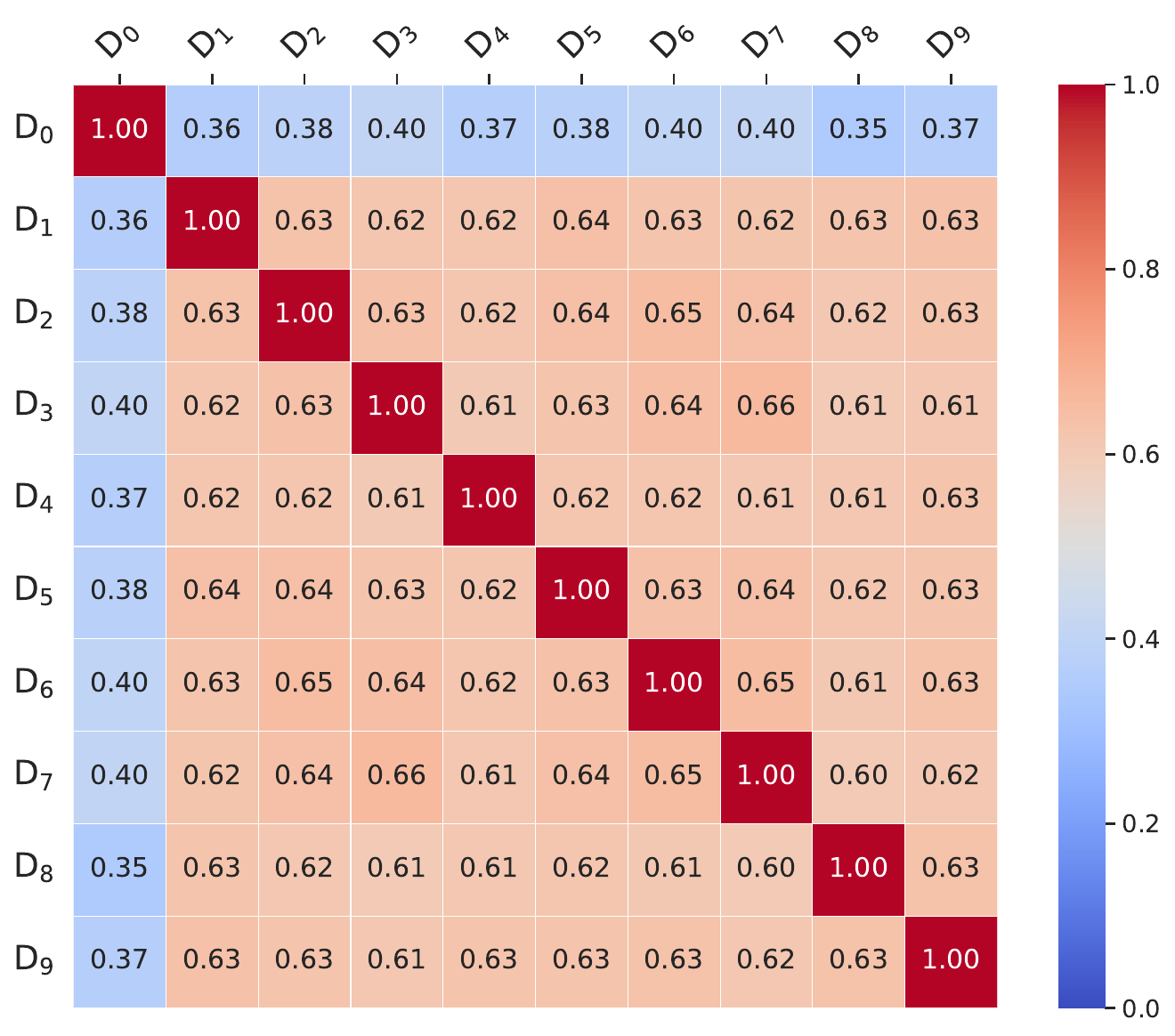}
    \caption{Utility-isolated safety region overlap analysis using SNIP on Llama-2-7B-Chat. The thresholds used to determine top safety-critical parameters within each weight matrix are: $q\%=3\%, p\%=7\%$. Left: Iso-Utility IoU vs $n$ (forward order). We begin with $\mathcal{D}_0$ and gradually add one dataset at a time, in the order from $\mathcal{D}_1$ to $\mathcal{D}_9$. Next, we isolate each identified safety region with the utility region identified by $\mathcal{D}_u$;
    Middle: Iso-Utility IoU vs $n$ (backward order). We begin with $\mathcal{D}_9$ and gradually add one dataset at a time, in the order from $\mathcal{D}_8$ to $\mathcal{D}_0$. Next, we isolate each identified safety region with the utility region identified by $\mathcal{D}_u$;
    Right: pairwise Iso-Utility IoU for $\{\mathcal{D}_i\}_{i=0}^9$. The matrix is symmetric. Each element corresponds to a pairwise IoU between two utility-isolated safety regions.}
    \label{fig:results_barplot_pairwise_multi_cat_snip_wanda_4}
    \end{subfigure}

    \vspace{0.5cm}

    \begin{subfigure}[b]{\textwidth}
    \includegraphics[width=0.34\linewidth]{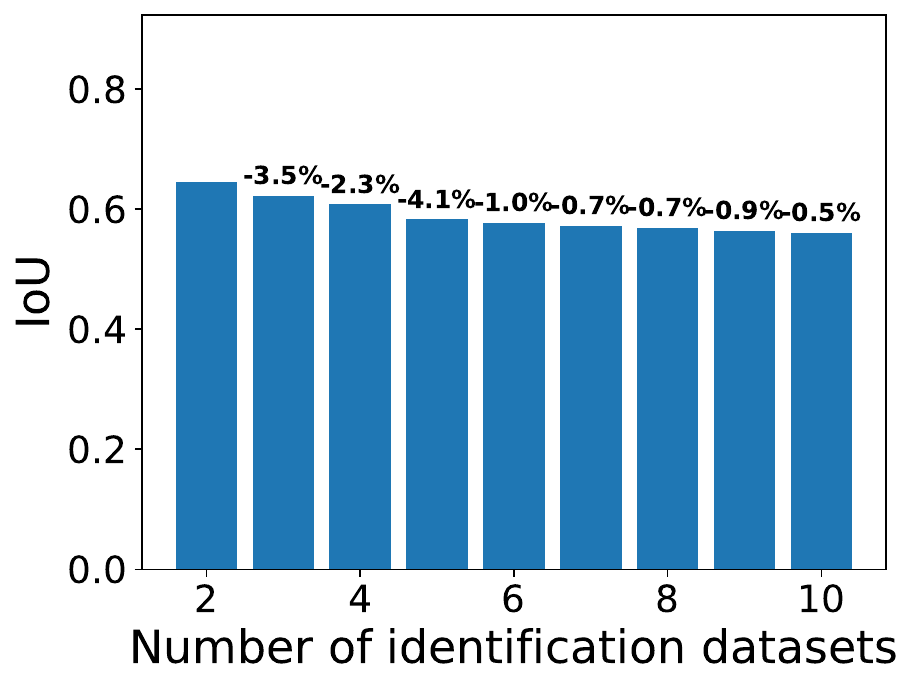}
    \hfill
    \includegraphics[width=0.34\linewidth]{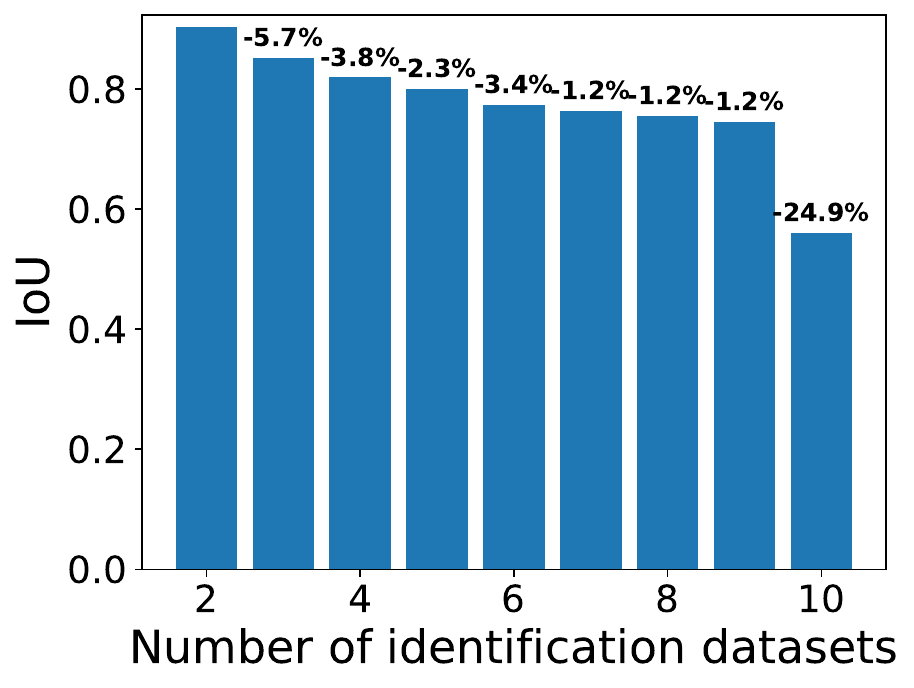}
    \hfill
    \includegraphics[width=0.3\linewidth]{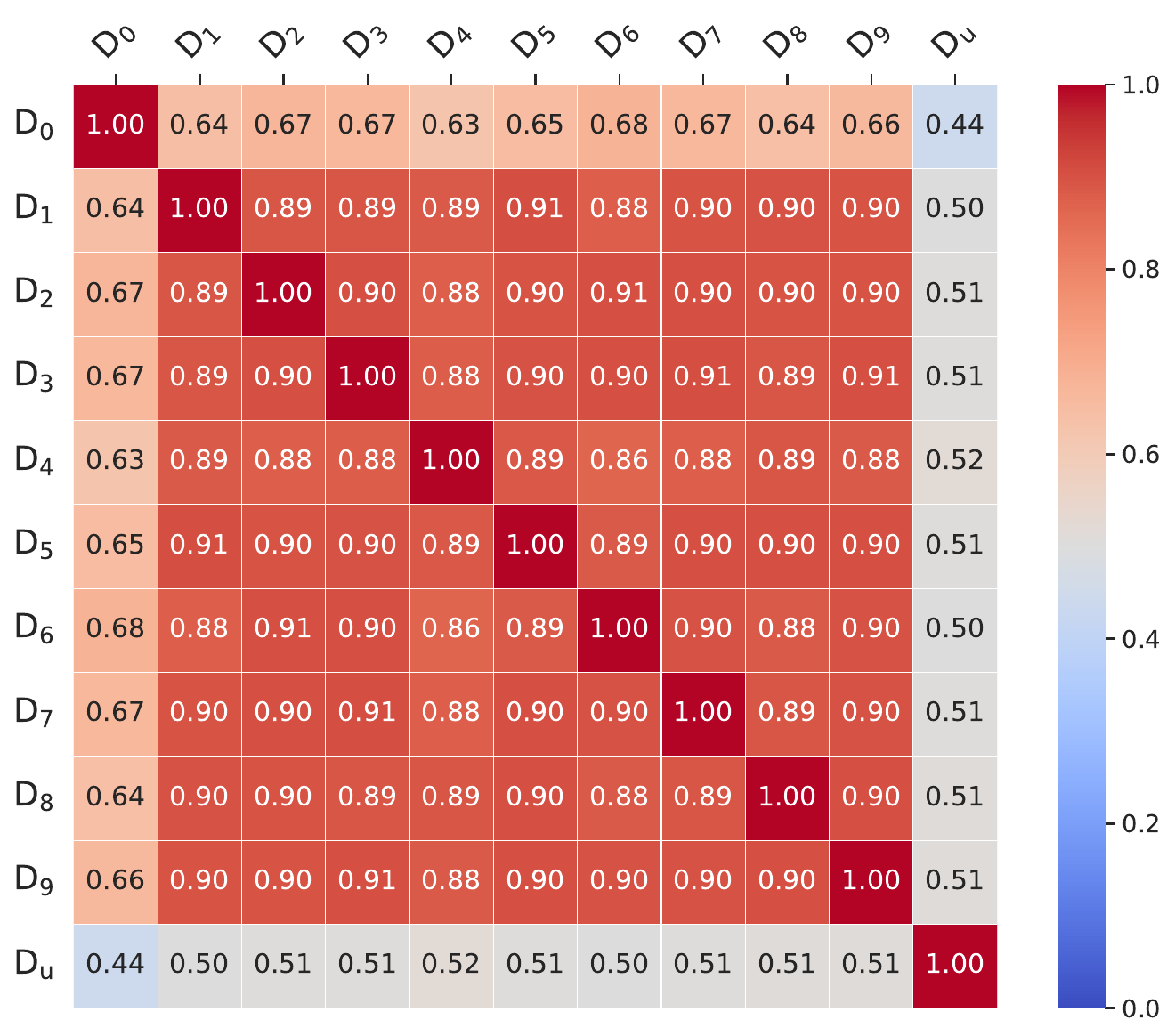}
    \caption{Safety region overlap analysis using Wanda on Llama-2-7B-Chat. The threshold used to determine top safety-critical parameters within each weight matrix is: $q\%=1\%$. Left: IoU vs $n$ (forward order). We begin with $\mathcal{D}_0$ and gradually add one dataset at a time, in the order from $\mathcal{D}_1$ to $\mathcal{D}_9$;
    Middle: IoU vs $n$ (backward order). We begin with $\mathcal{D}_9$ and gradually add one dataset at a time, in the order from $\mathcal{D}_8$ to $\mathcal{D}_0$;
    Right: pairwise IoU for $\{\mathcal{D}_i\}_{i=0}^9$. The matrix is symmetric. Each element corresponds to a pairwise IoU between two safety regions.}
    \label{fig:results_barplot_pairwise_multi_cat_snip_wanda_5}
    \end{subfigure}

    \vspace{0.5cm}

    \begin{subfigure}[b]{\textwidth}
    \includegraphics[width=0.34\linewidth]{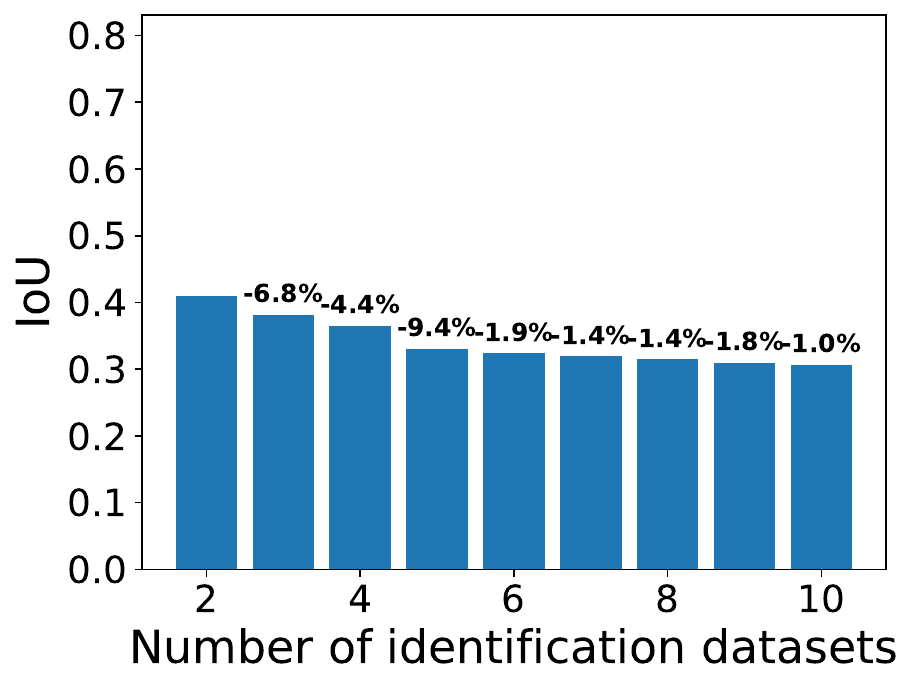}
    \hfill
    \includegraphics[width=0.34\linewidth]{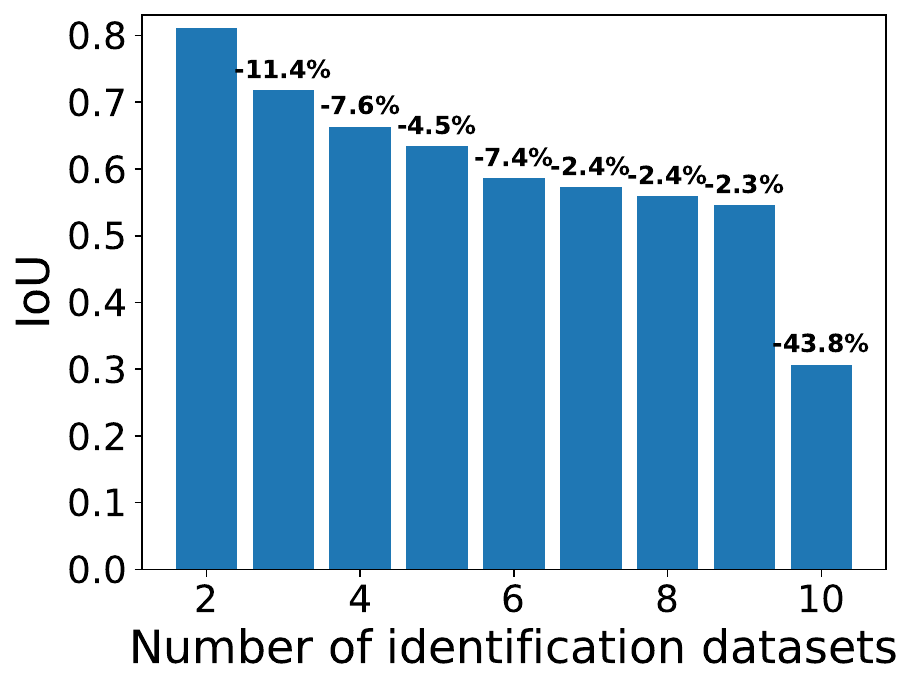}
    \hfill
    \includegraphics[width=0.3\linewidth]{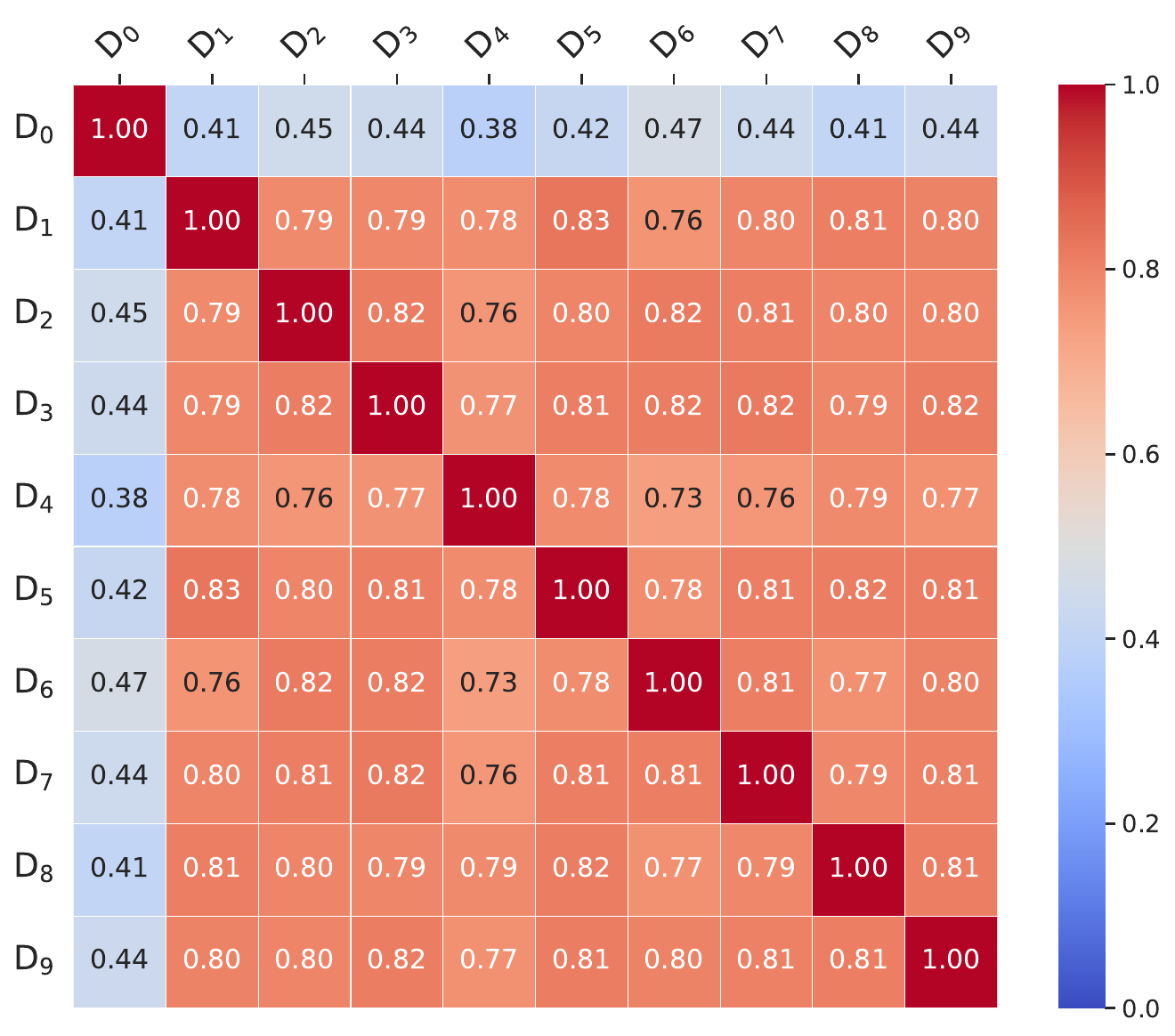}
    \caption{Utility-isolated safety region overlap analysis using Wanda on Llama-2-7B-Chat. The thresholds used to determine top safety-critical parameters within each weight matrix are: $q\%=1\%, p\%=1\%$. Left: Iso-Utility IoU vs $n$ (forward order). We begin with $\mathcal{D}_0$ and gradually add one dataset at a time, in the order from $\mathcal{D}_1$ to $\mathcal{D}_9$. Next, we isolate each identified safety region with the utility region identified by $\mathcal{D}_u$;
    Middle: Iso-Utility IoU vs $n$ (backward order). We begin with $\mathcal{D}_9$ and gradually add one dataset at a time, in the order from $\mathcal{D}_8$ to $\mathcal{D}_0$. Next, we isolate each identified safety region with the utility region identified by $\mathcal{D}_u$;
    Right: pairwise Iso-Utility IoU for $\{\mathcal{D}_i\}_{i=0}^9$. The matrix is symmetric. Each element corresponds to a pairwise IoU between two utility-isolated safety regions.}
    \label{fig:results_barplot_pairwise_multi_cat_snip_wanda_6}
    \end{subfigure}

\end{figure*}

\begin{figure*}[t]\ContinuedFloat

    \begin{subfigure}[b]{\textwidth}
    \includegraphics[width=0.34\linewidth]{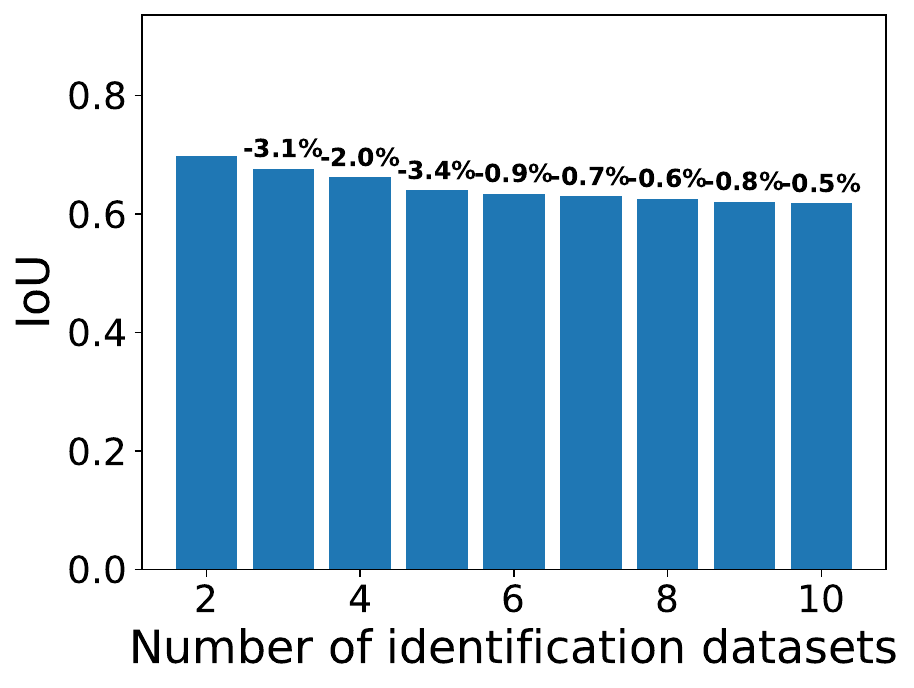}
    \hfill
    \includegraphics[width=0.34\linewidth]{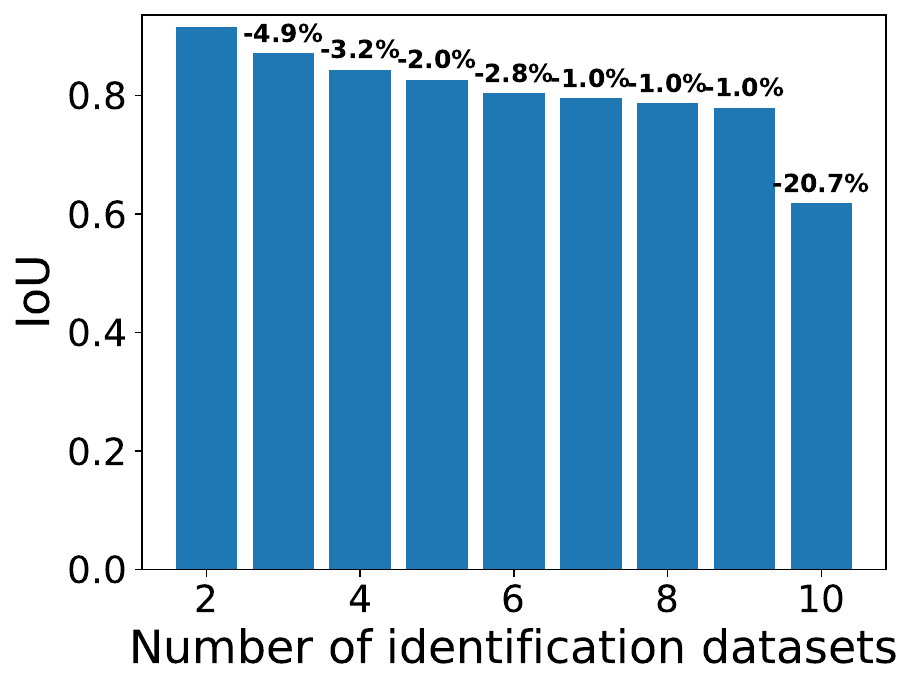}
    \hfill
    \includegraphics[width=0.3\linewidth]{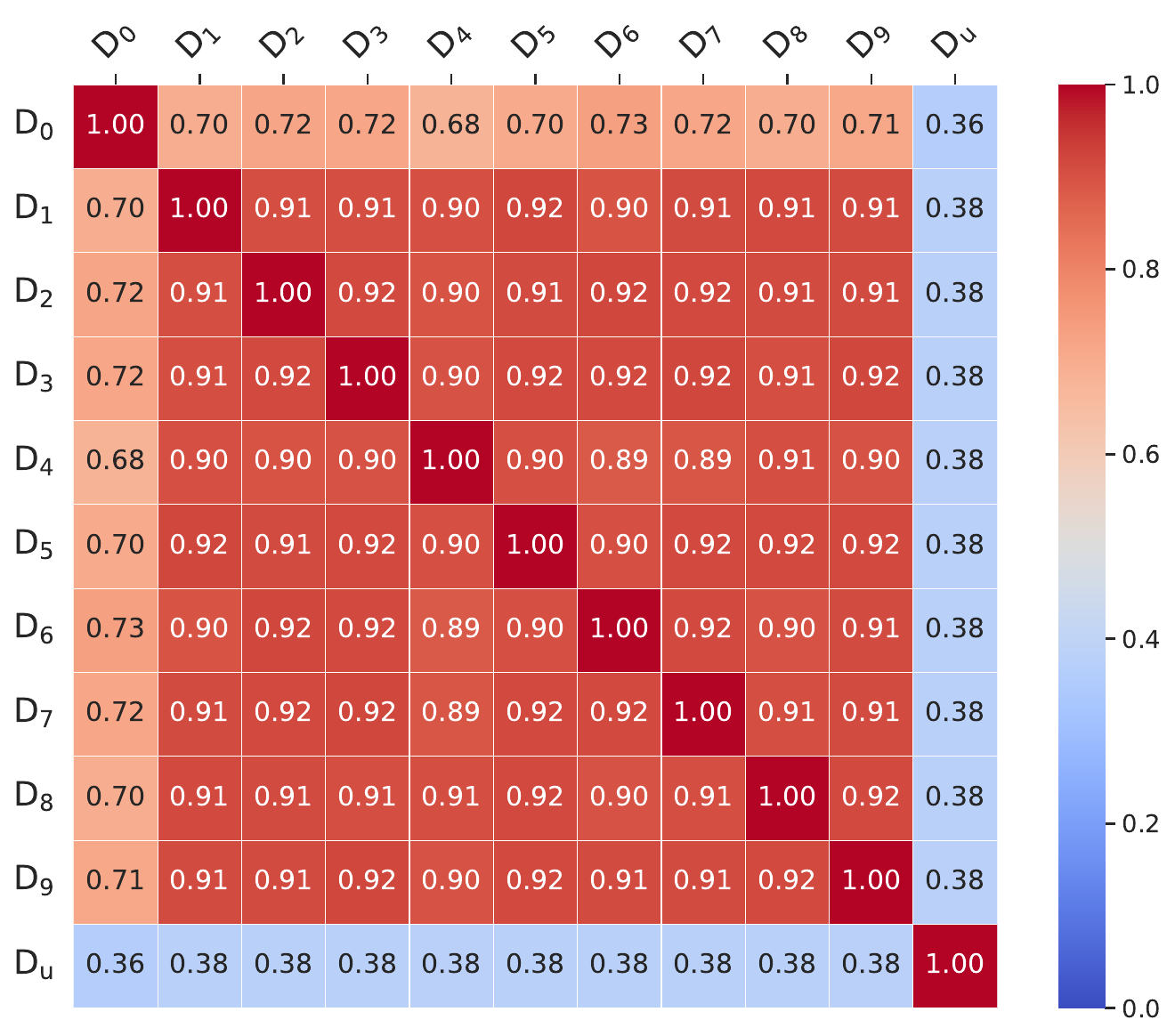}
    \caption{Safety region overlap analysis using Wanda on Llama-2-7B-Chat. The threshold used to determine top safety-critical parameters within each weight matrix is: $q\%=3\%$. Left: IoU vs $n$ (forward order). We begin with $\mathcal{D}_0$ and gradually add one dataset at a time, in the order from $\mathcal{D}_1$ to $\mathcal{D}_9$;
    Middle: IoU vs $n$ (backward order). We begin with $\mathcal{D}_9$ and gradually add one dataset at a time, in the order from $\mathcal{D}_8$ to $\mathcal{D}_0$;
    Right: pairwise IoU for $\{\mathcal{D}_i\}_{i=0}^9$. The matrix is symmetric. Each element corresponds to a pairwise IoU between two safety regions.}
    \label{fig:results_barplot_pairwise_multi_cat_snip_wanda_7}
    \end{subfigure}

    \vspace{0.5cm}

    \begin{subfigure}[b]{\textwidth}
    \includegraphics[width=0.34\linewidth]{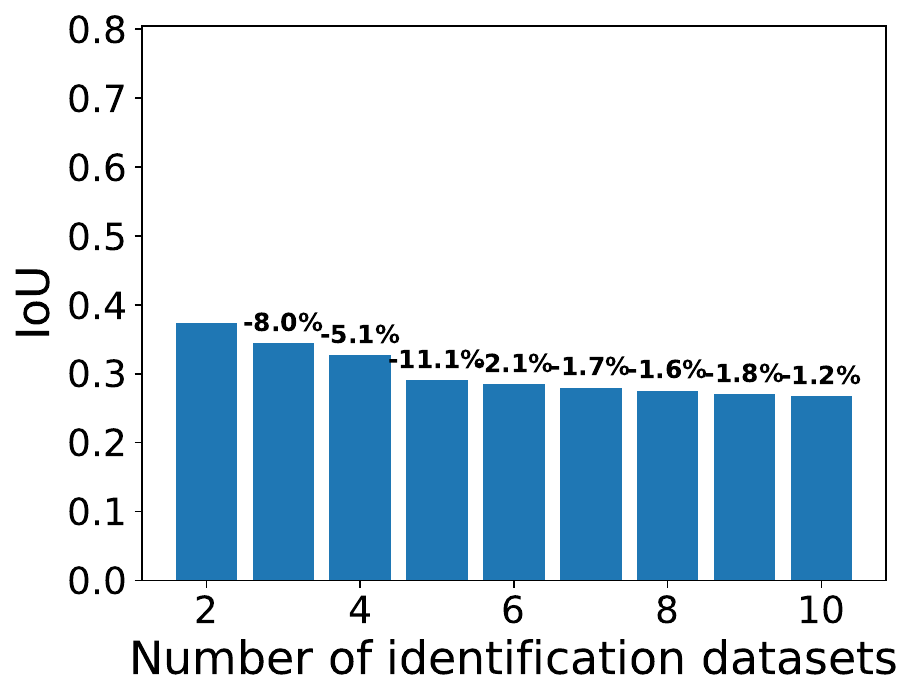}
    \hfill
    \includegraphics[width=0.34\linewidth]{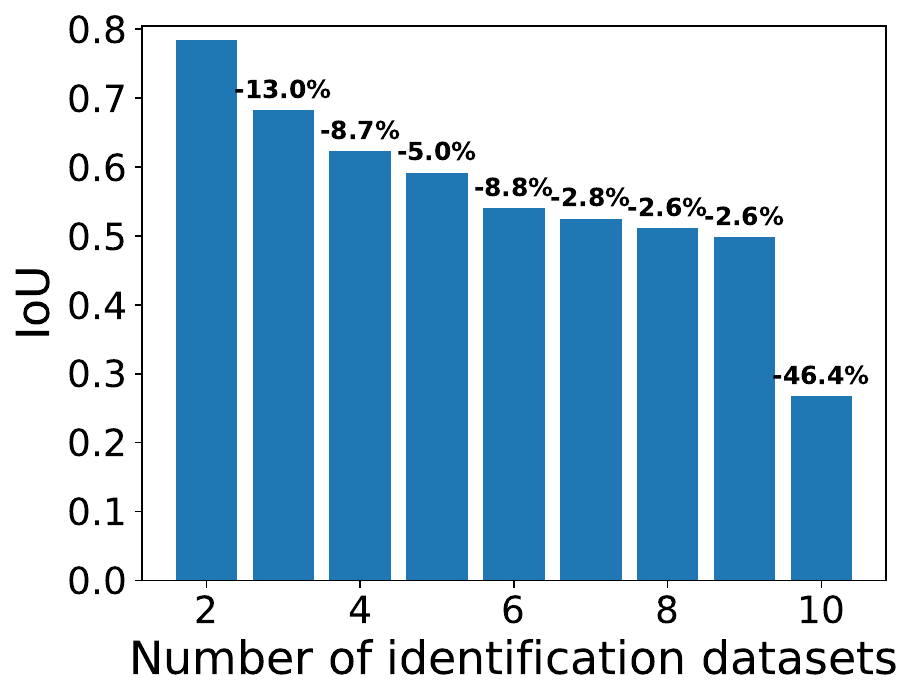}
    \hfill
    \includegraphics[width=0.3\linewidth]{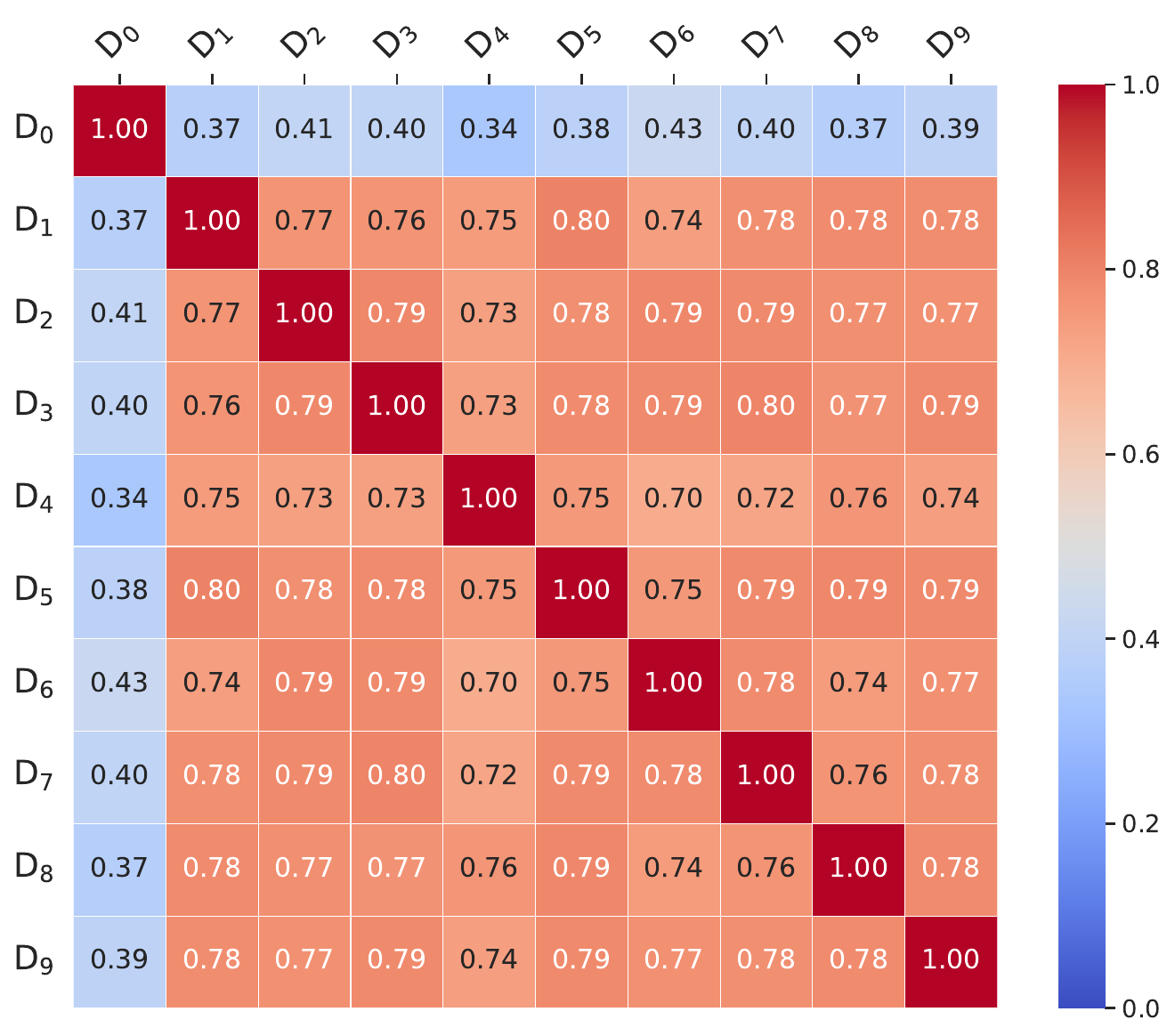}
    \caption{Utility-isolated safety region overlap analysis using Wanda on Llama-2-7B-Chat. The thresholds used to determine top safety-critical parameters within each weight matrix are: $q\%=3\%, p\%=7\%$. Left: Iso-Utility IoU vs $n$ (forward order). We begin with $\mathcal{D}_0$ and gradually add one dataset at a time, in the order from $\mathcal{D}_1$ to $\mathcal{D}_9$. Next, we isolate each identified safety region with the utility region identified by $\mathcal{D}_u$;
    Middle: Iso-Utility IoU vs $n$ (backward order). We begin with $\mathcal{D}_9$ and gradually add one dataset at a time, in the order from $\mathcal{D}_8$ to $\mathcal{D}_0$. Next, we isolate each identified safety region with the utility region identified by $\mathcal{D}_u$;
    Right: pairwise Iso-Utility IoU for $\{\mathcal{D}_i\}_{i=0}^9$. The matrix is symmetric. Each element corresponds to a pairwise IoU between two utility-isolated safety regions.}
    \label{fig:results_barplot_pairwise_multi_cat_snip_wanda_8}
    \end{subfigure}

    \vspace{0.5cm}

    \begin{subfigure}[b]{\textwidth}
    \includegraphics[width=0.34\linewidth]{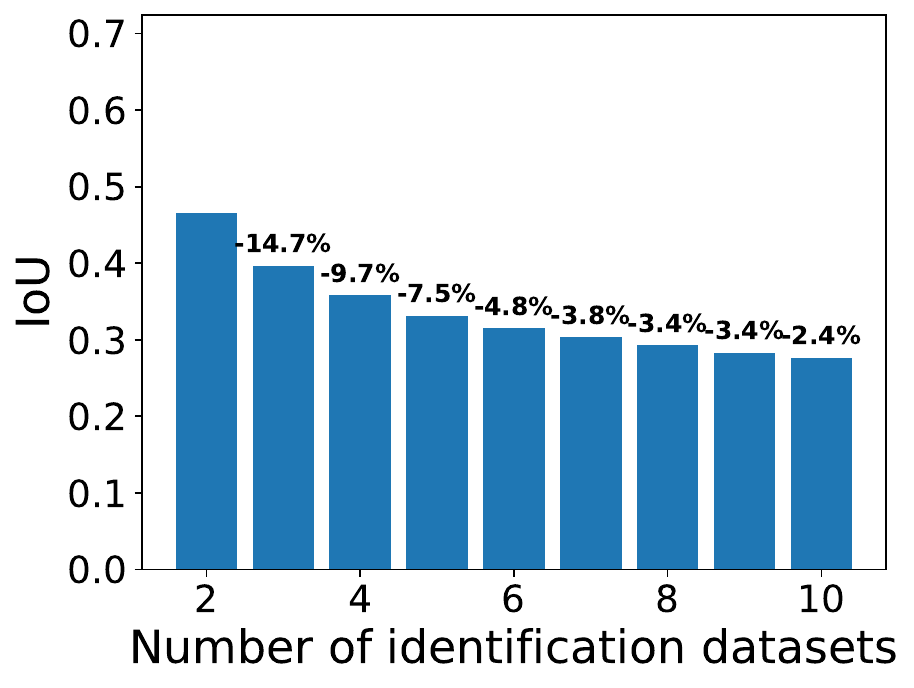}
    \hfill
    \includegraphics[width=0.34\linewidth]{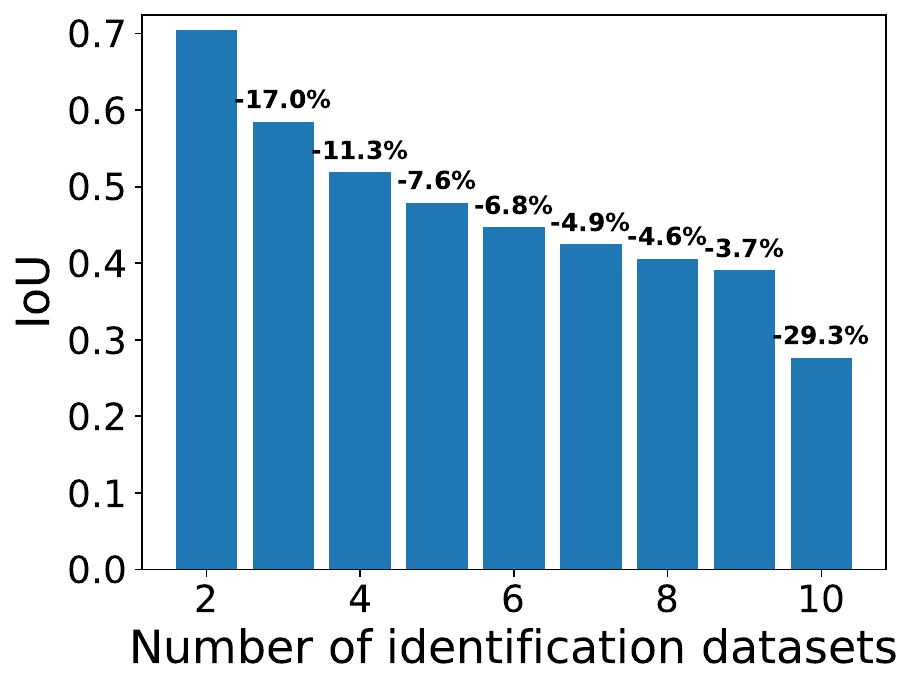}
    \hfill
    \includegraphics[width=0.3\linewidth]{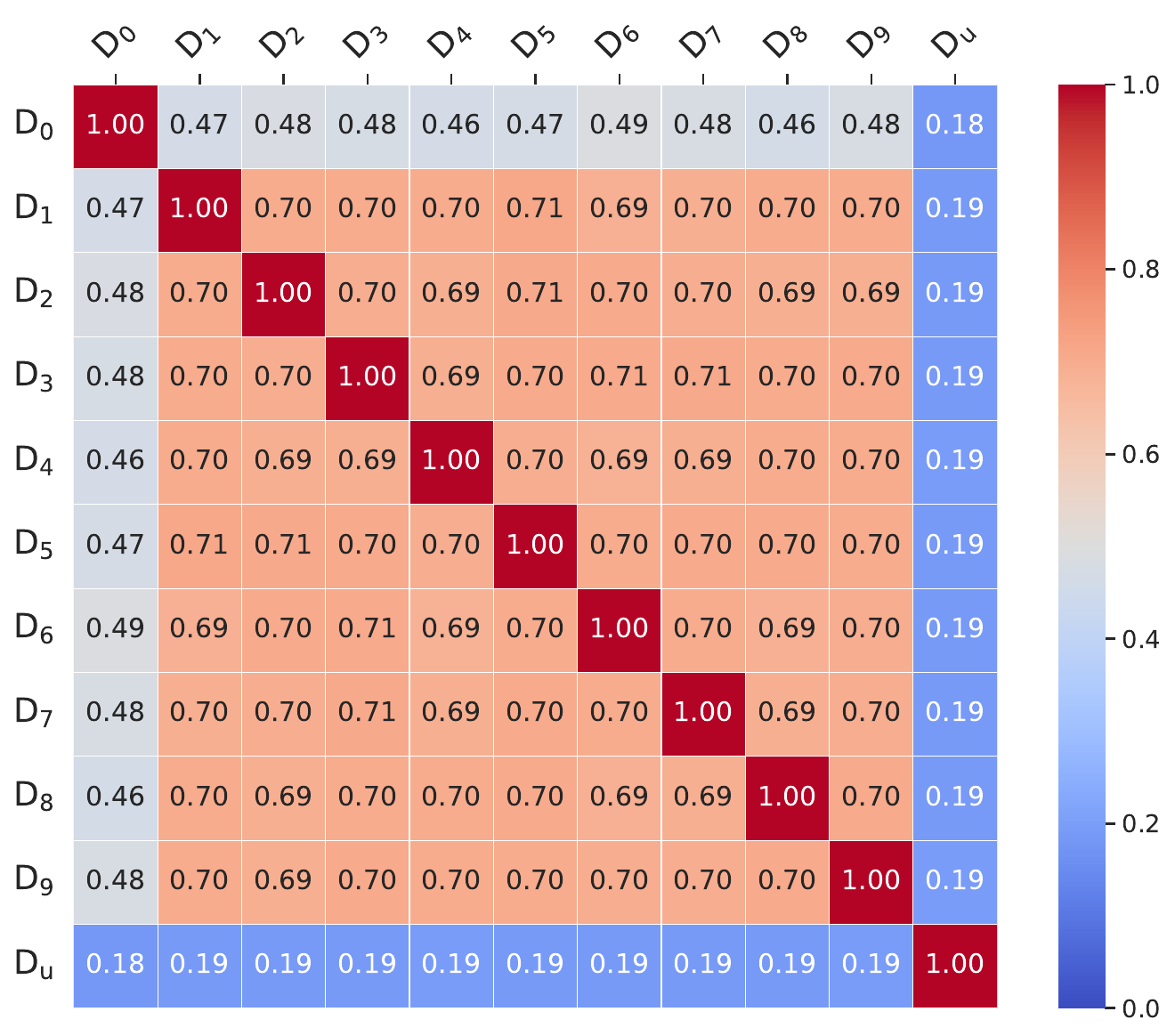}
    \caption{Safety region overlap analysis using SNIP on Llama-2-13B-Chat. The threshold used to determine top safety-critical parameters within each weight matrix is: $q\%=1\%$. Left: IoU vs $n$ (forward order). We begin with $\mathcal{D}_0$ and gradually add one dataset at a time, in the order from $\mathcal{D}_1$ to $\mathcal{D}_9$;
    Middle: IoU vs $n$ (backward order). We begin with $\mathcal{D}_9$ and gradually add one dataset at a time, in the order from $\mathcal{D}_8$ to $\mathcal{D}_0$;
    Right: pairwise IoU for $\{\mathcal{D}_i\}_{i=0}^9$. The matrix is symmetric. Each element corresponds to a pairwise IoU between two safety regions.}
    \label{fig:results_barplot_pairwise_multi_cat_snip_wanda_9}
    \end{subfigure}

\end{figure*}

\begin{figure*}[t]\ContinuedFloat

    \begin{subfigure}[b]{\textwidth}
    \includegraphics[width=0.34\linewidth]{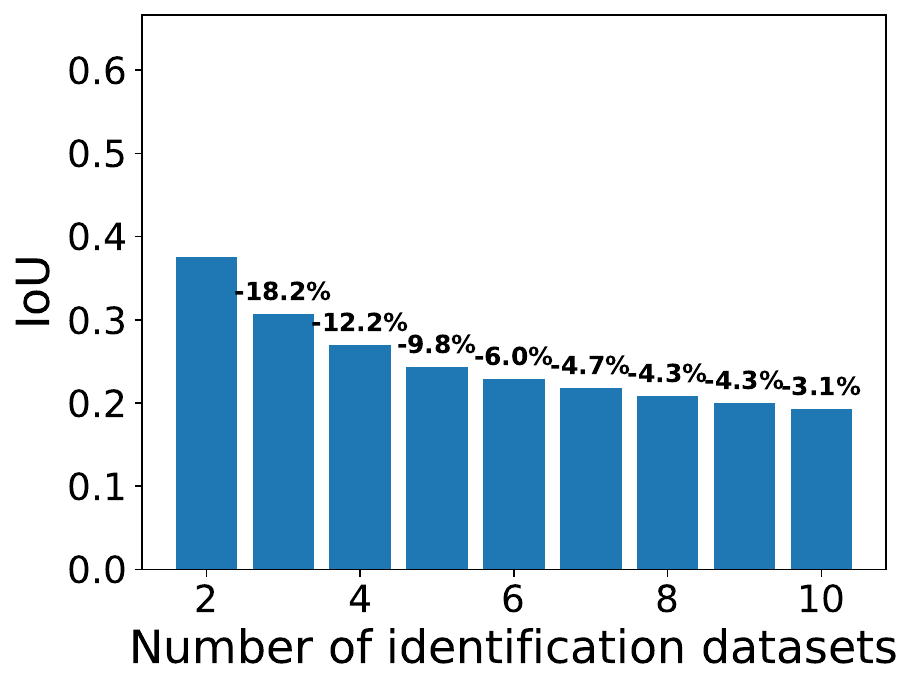}
    \hfill
    \includegraphics[width=0.34\linewidth]{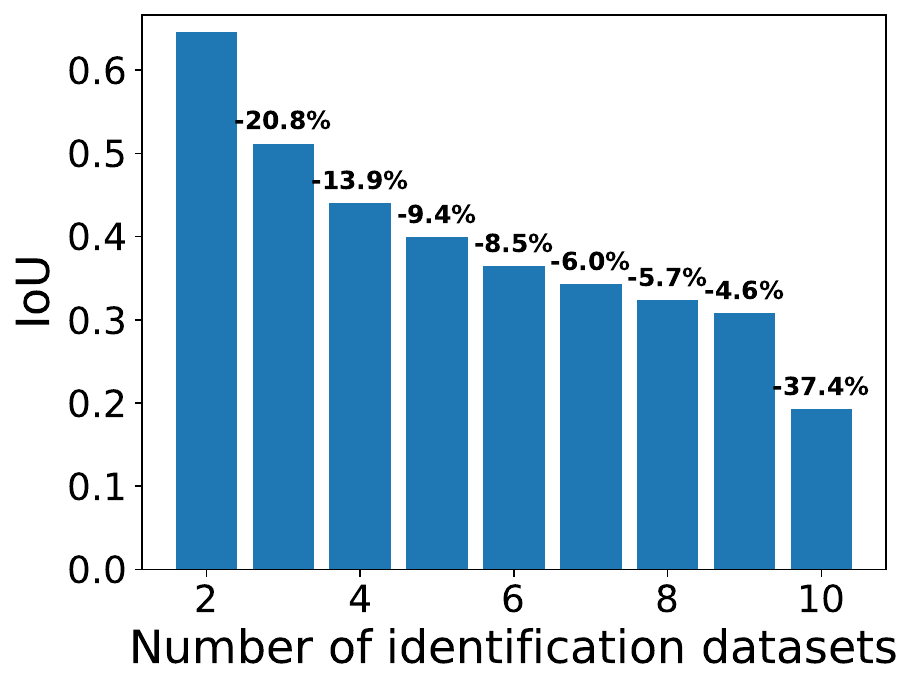}
    \hfill
    \includegraphics[width=0.3\linewidth]{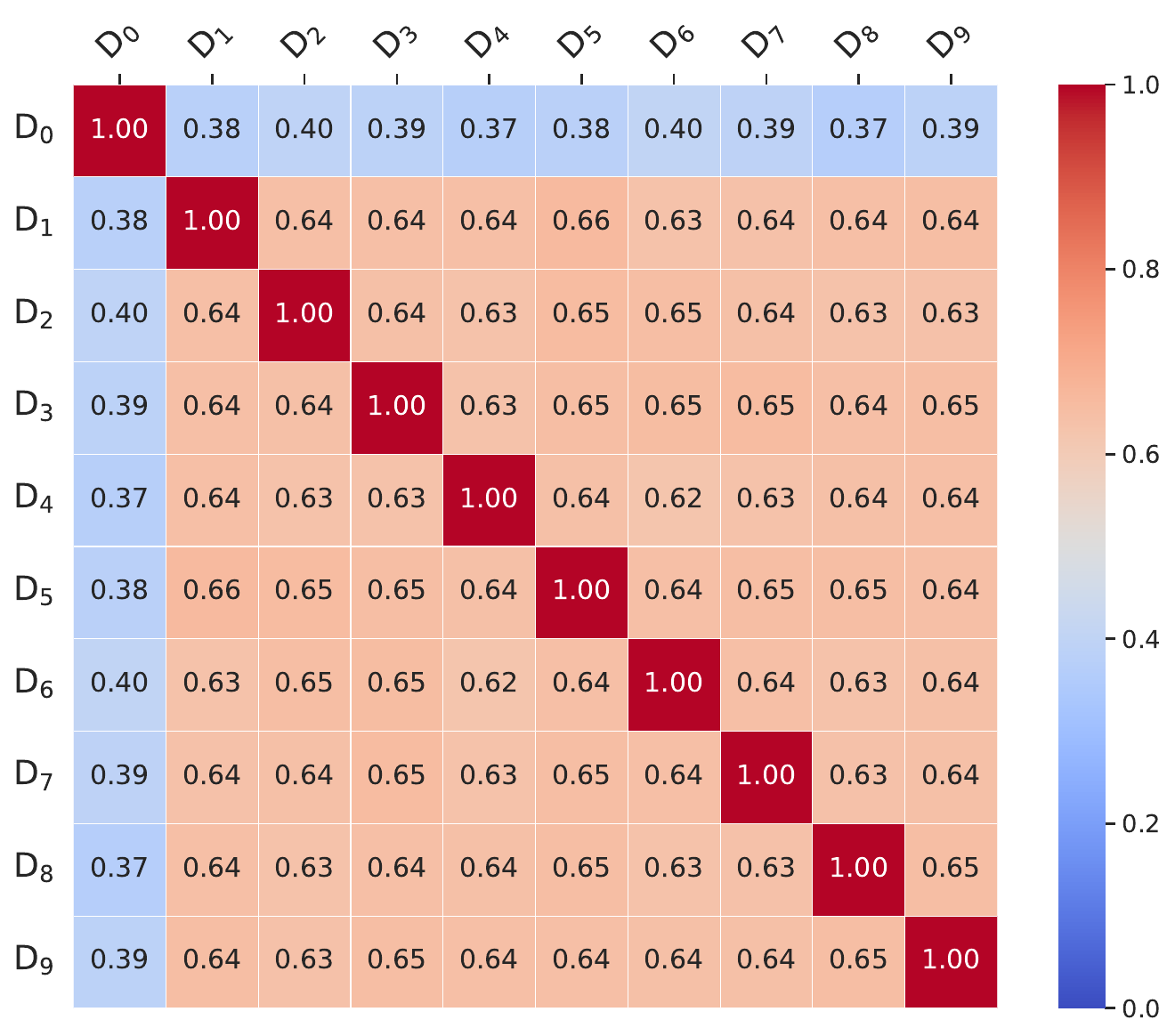}
    \caption{Utility-isolated safety region overlap analysis using SNIP on Llama-2-13B-Chat. The thresholds used to determine top safety-critical parameters within each weight matrix are: $q\%=1\%, p\%=1\%$. Left: Iso-Utility IoU vs $n$ (forward order). We begin with $\mathcal{D}_0$ and gradually add one dataset at a time, in the order from $\mathcal{D}_1$ to $\mathcal{D}_9$. Next, we isolate each identified safety region with the utility region identified by $\mathcal{D}_u$;
    Middle: Iso-Utility IoU vs $n$ (backward order). We begin with $\mathcal{D}_9$ and gradually add one dataset at a time, in the order from $\mathcal{D}_8$ to $\mathcal{D}_0$. Next, we isolate each identified safety region with the utility region identified by $\mathcal{D}_u$;
    Right: pairwise Iso-Utility IoU for $\{\mathcal{D}_i\}_{i=0}^9$. The matrix is symmetric. Each element corresponds to a pairwise IoU between two utility-isolated safety regions.}
    \label{fig:results_barplot_pairwise_multi_cat_snip_wanda_10}
    \end{subfigure}

    \vspace{0.5cm}

    \begin{subfigure}[b]{\textwidth}
    \includegraphics[width=0.34\linewidth]{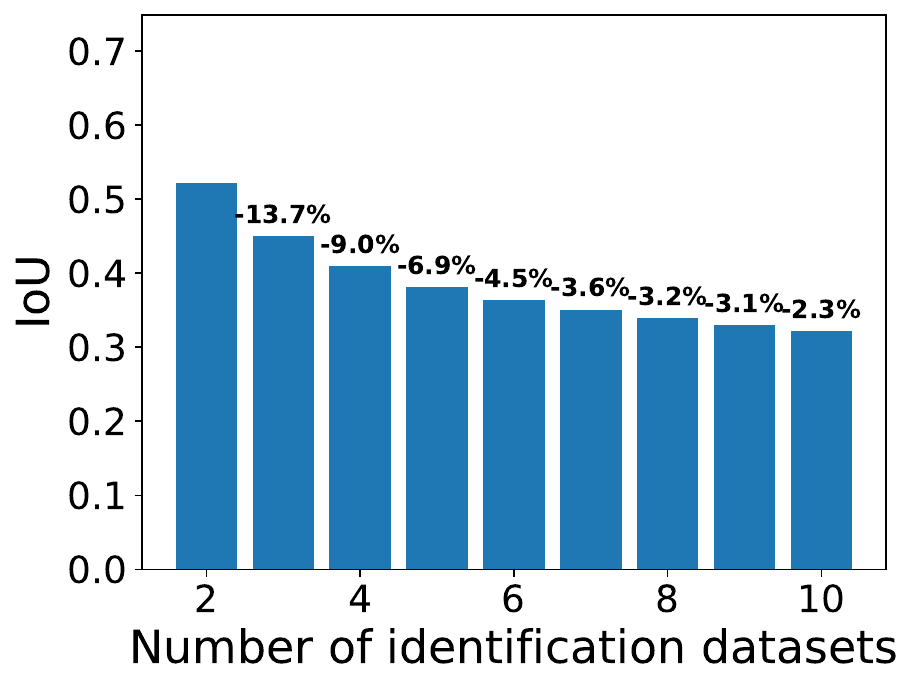}
    \hfill
    \includegraphics[width=0.34\linewidth]{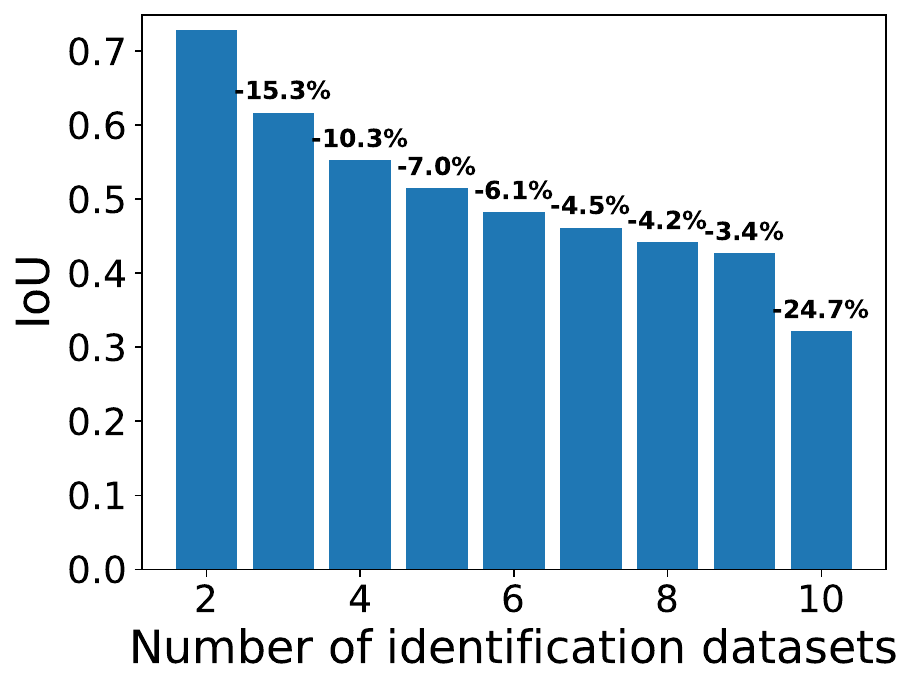}
    \hfill
    \includegraphics[width=0.3\linewidth]{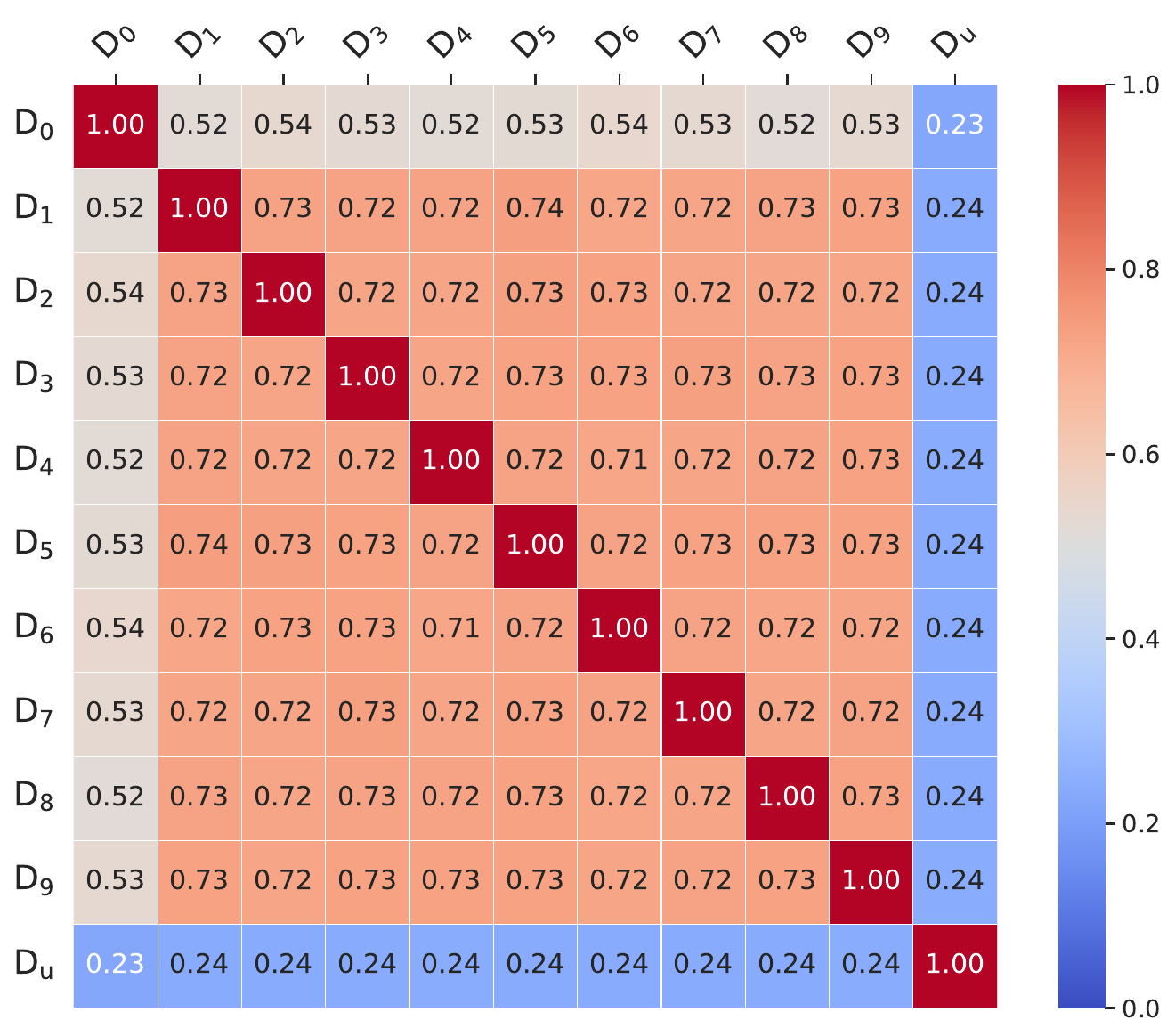}
    \caption{Safety region overlap analysis using SNIP on Llama-2-13B-Chat. The threshold used to determine top safety-critical parameters within each weight matrix is: $q\%=3\%$. Left: IoU vs $n$ (forward order). We begin with $\mathcal{D}_0$ and gradually add one dataset at a time, in the order from $\mathcal{D}_1$ to $\mathcal{D}_9$;
    Middle: IoU vs $n$ (backward order). We begin with $\mathcal{D}_9$ and gradually add one dataset at a time, in the order from $\mathcal{D}_8$ to $\mathcal{D}_0$;
    Right: pairwise IoU for $\{\mathcal{D}_i\}_{i=0}^9$. The matrix is symmetric. Each element corresponds to a pairwise IoU between two safety regions.}
    \label{fig:results_barplot_pairwise_multi_cat_snip_wanda_11}
    \end{subfigure}

    \vspace{0.5cm}

    \begin{subfigure}[b]{\textwidth}
    \includegraphics[width=0.34\linewidth]{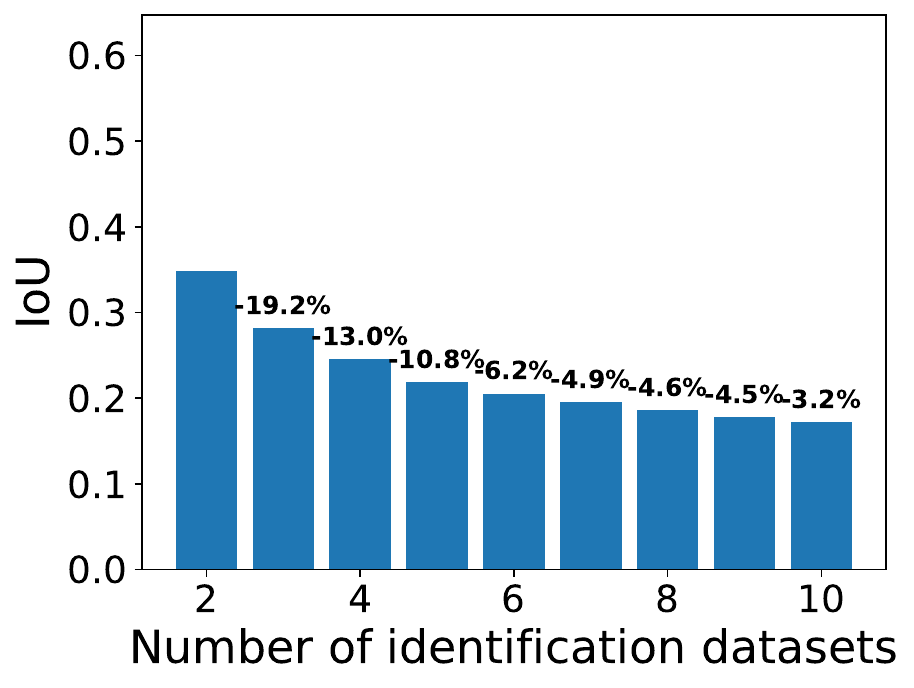}
    \hfill
    \includegraphics[width=0.34\linewidth]{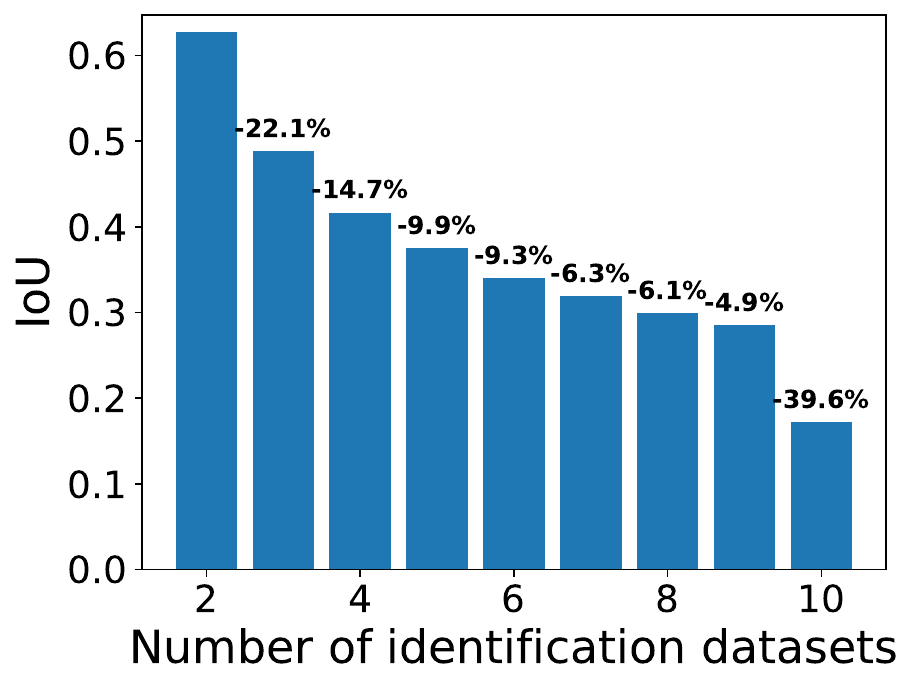}
    \hfill
    \includegraphics[width=0.3\linewidth]{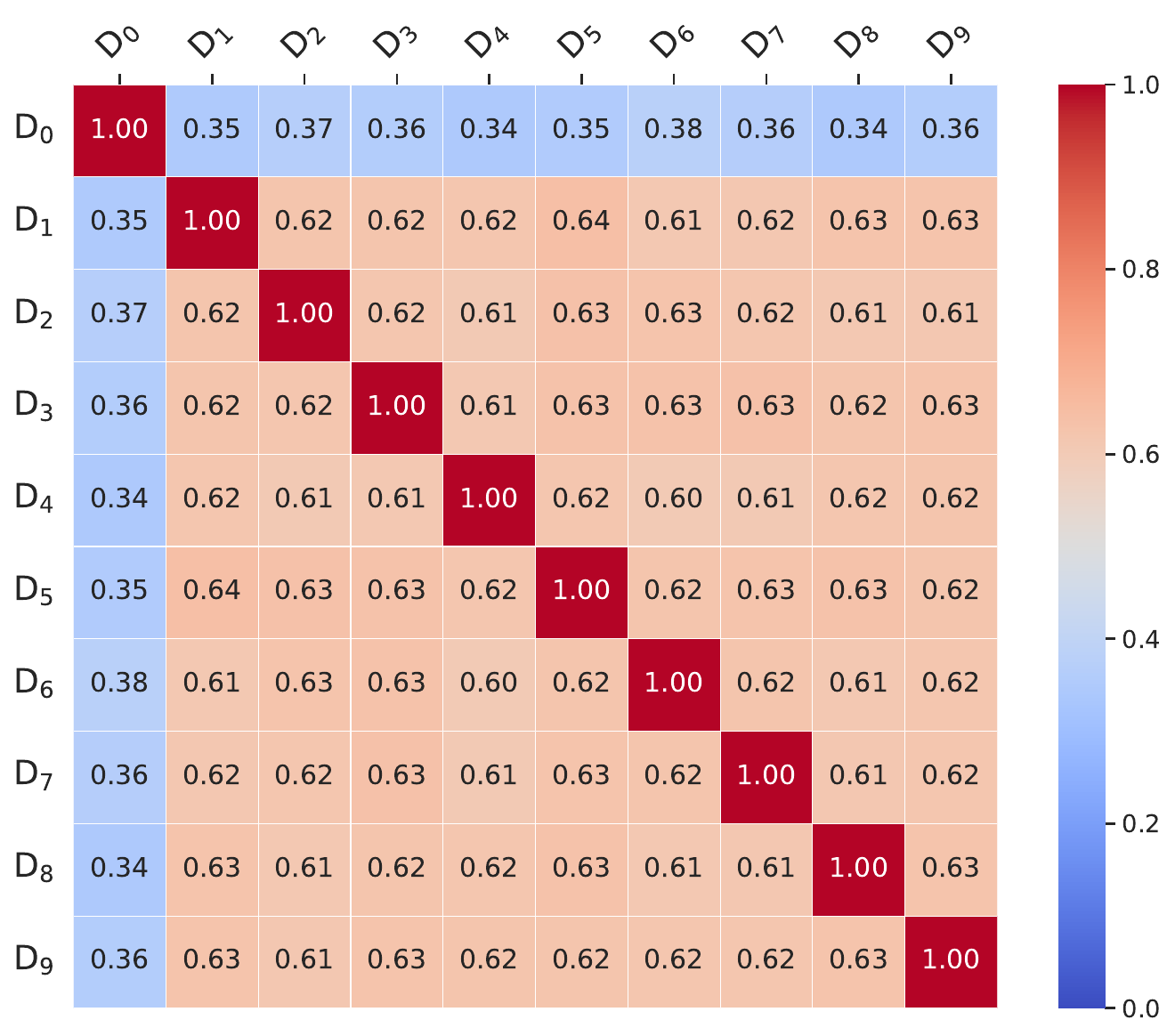}
    \caption{Utility-isolated safety region overlap analysis using SNIP on Llama-2-13B-Chat. The thresholds used to determine top safety-critical parameters within each weight matrix are: $q\%=3\%, p\%=8\%$. Left: Iso-Utility IoU vs $n$ (forward order). We begin with $\mathcal{D}_0$ and gradually add one dataset at a time, in the order from $\mathcal{D}_1$ to $\mathcal{D}_9$. Next, we isolate each identified safety region with the utility region identified by $\mathcal{D}_u$;
    Middle: Iso-Utility IoU vs $n$ (backward order). We begin with $\mathcal{D}_9$ and gradually add one dataset at a time, in the order from $\mathcal{D}_8$ to $\mathcal{D}_0$. Next, we isolate each identified safety region with the utility region identified by $\mathcal{D}_u$;
    Right: pairwise Iso-Utility IoU for $\{\mathcal{D}_i\}_{i=0}^9$. The matrix is symmetric. Each element corresponds to a pairwise IoU between two utility-isolated safety regions.}
    \label{fig:results_barplot_pairwise_multi_cat_snip_wanda_12}
    \end{subfigure}

\end{figure*}

\begin{figure*}[t]\ContinuedFloat

    \begin{subfigure}[b]{\textwidth}
    \includegraphics[width=0.34\linewidth]{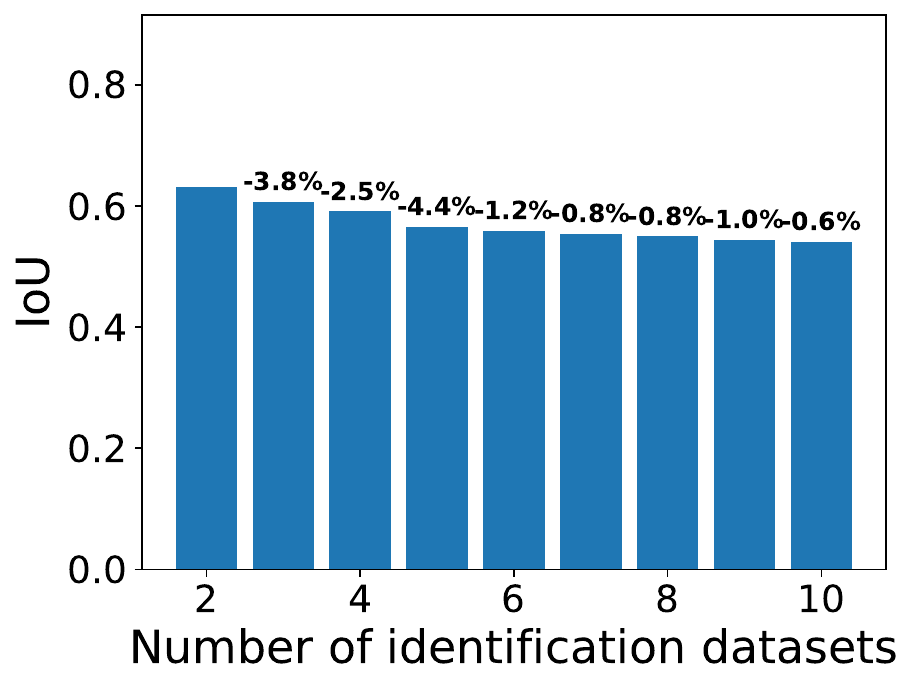}
    \hfill
    \includegraphics[width=0.34\linewidth]{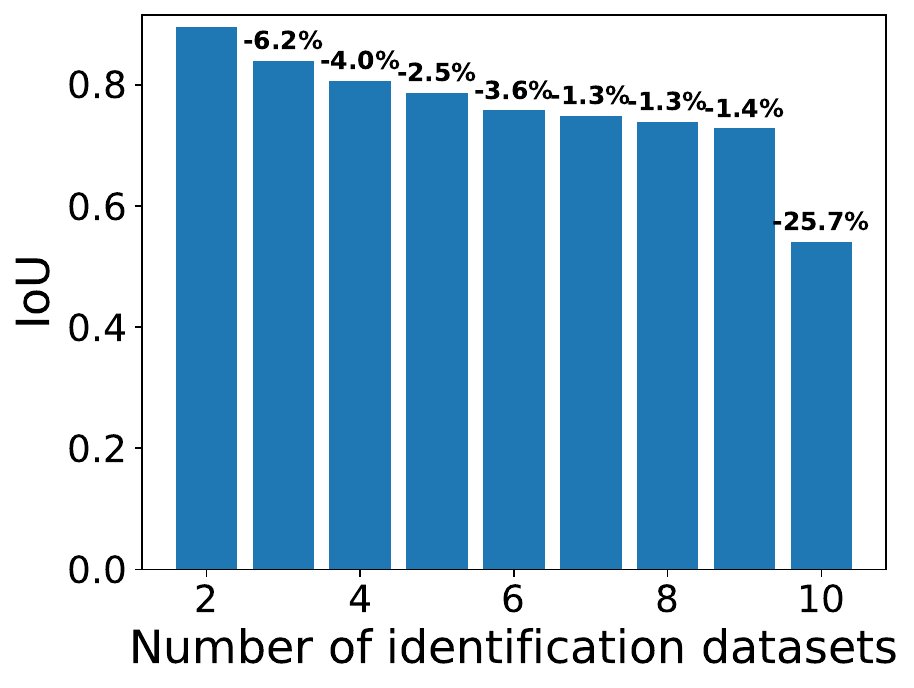}
    \hfill
    \includegraphics[width=0.3\linewidth]{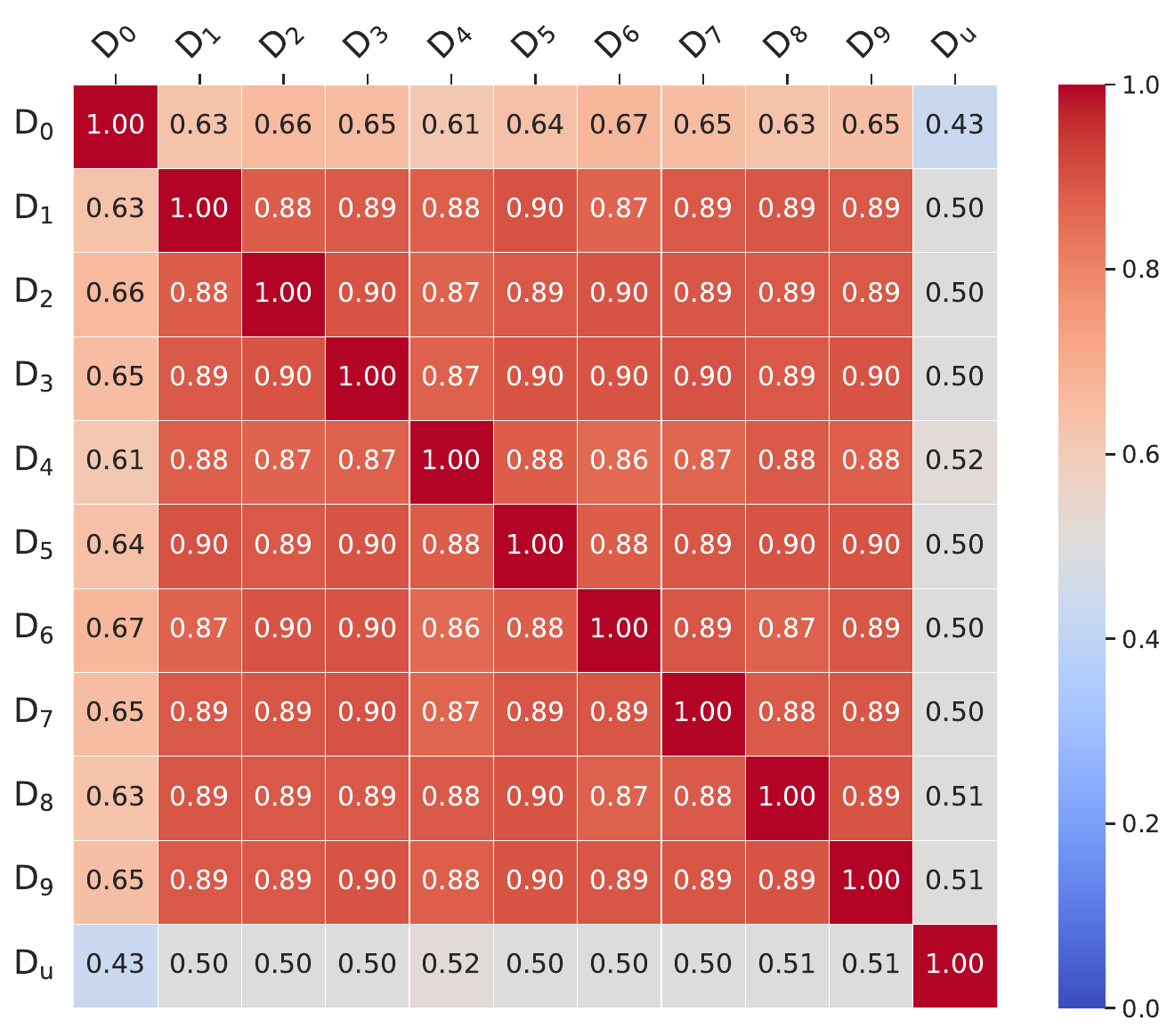}
    \caption{Safety region overlap analysis using Wanda on Llama-2-13B-Chat. The threshold used to determine top safety-critical parameters within each weight matrix is: $q\%=1\%$. Left: IoU vs $n$ (forward order). We begin with $\mathcal{D}_0$ and gradually add one dataset at a time, in the order from $\mathcal{D}_1$ to $\mathcal{D}_9$;
    Middle: IoU vs $n$ (backward order). We begin with $\mathcal{D}_9$ and gradually add one dataset at a time, in the order from $\mathcal{D}_8$ to $\mathcal{D}_0$;
    Right: pairwise IoU for $\{\mathcal{D}_i\}_{i=0}^9$. The matrix is symmetric. Each element corresponds to a pairwise IoU between two safety regions.}
    \label{fig:results_barplot_pairwise_multi_cat_snip_wanda_13}
    \end{subfigure}

    \vspace{0.5cm}

    \begin{subfigure}[b]{\textwidth}
    \includegraphics[width=0.34\linewidth]{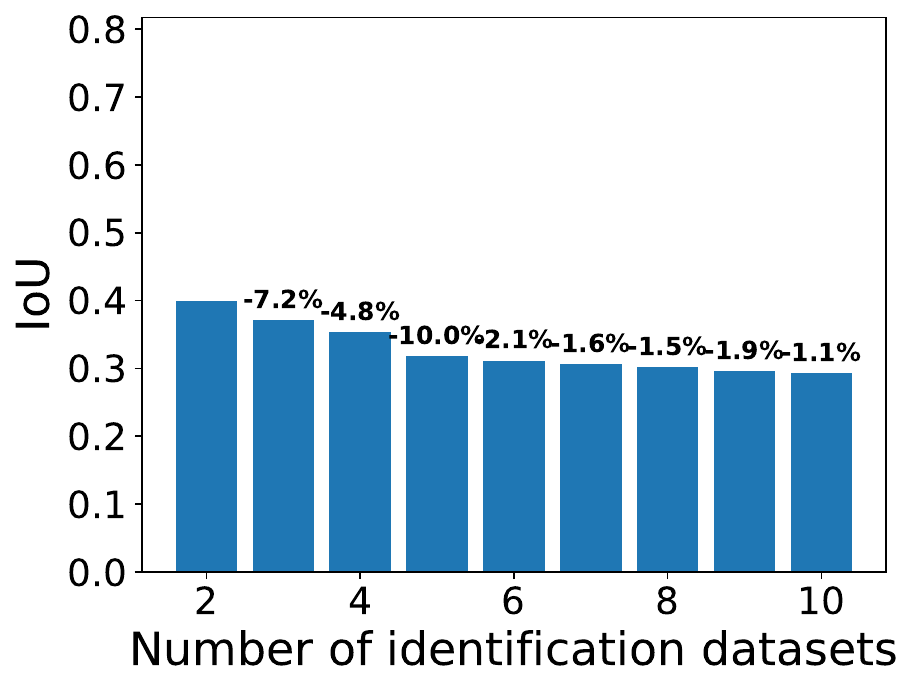}
    \hfill
    \includegraphics[width=0.34\linewidth]{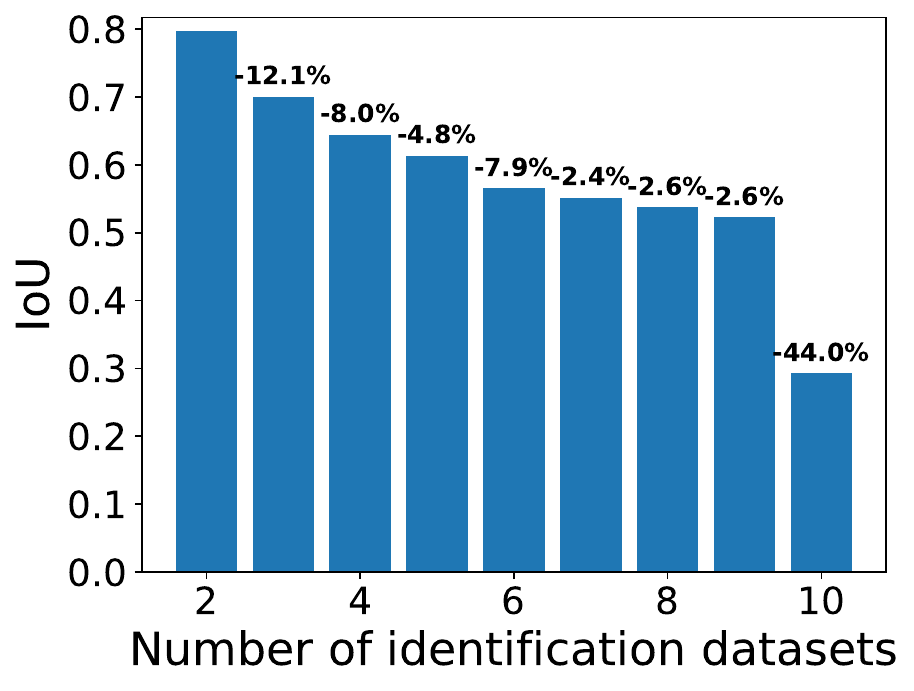}
    \hfill
    \includegraphics[width=0.3\linewidth]{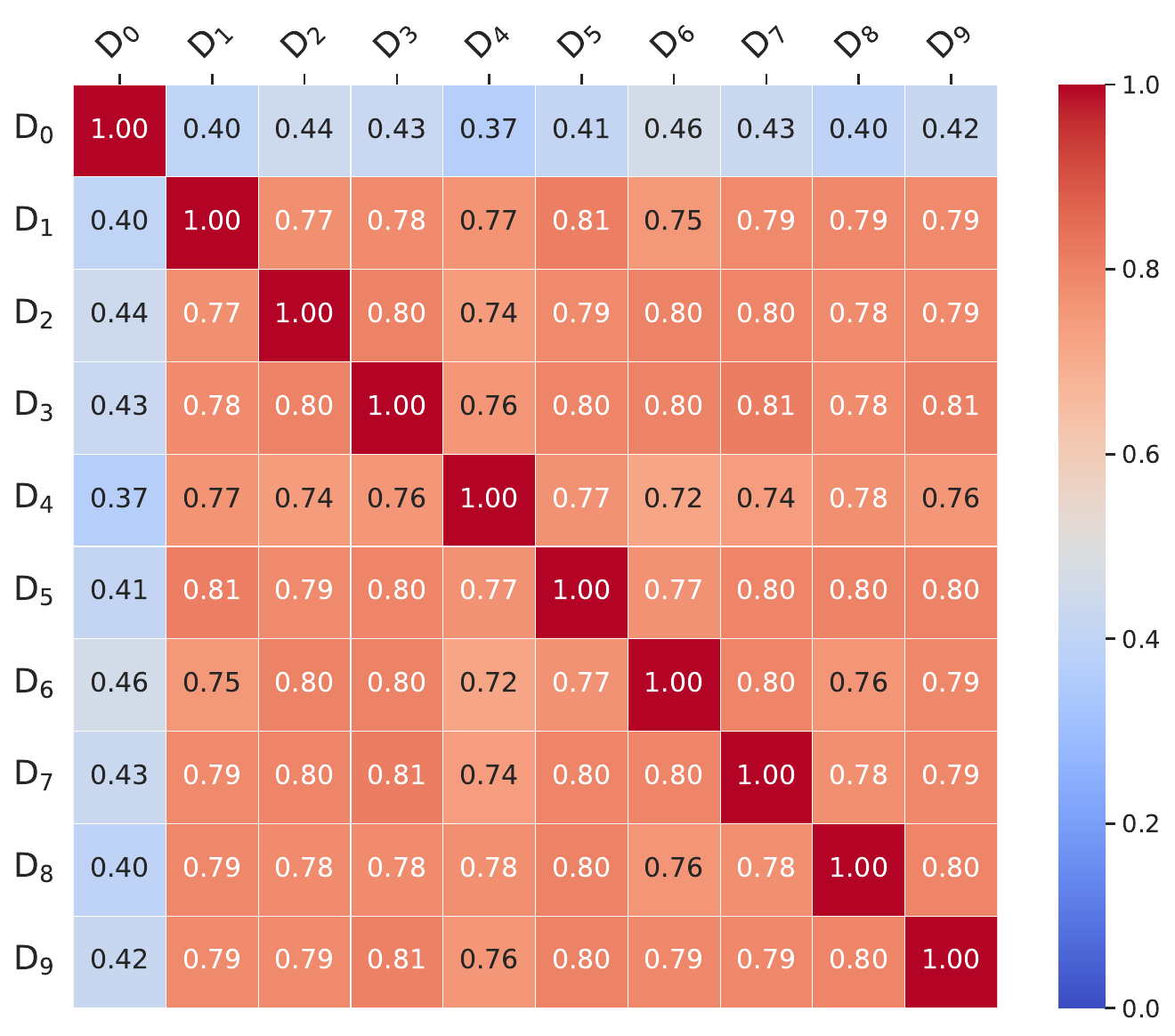}
    \caption{Utility-isolated safety region overlap analysis using Wanda on Llama-2-13B-Chat. The thresholds used to determine top safety-critical parameters within each weight matrix are: $q\%=1\%, p\%=1\%$. Left: Iso-Utility IoU vs $n$ (forward order). We begin with $\mathcal{D}_0$ and gradually add one dataset at a time, in the order from $\mathcal{D}_1$ to $\mathcal{D}_9$. Next, we isolate each identified safety region with the utility region identified by $\mathcal{D}_u$;
    Middle: Iso-Utility IoU vs $n$ (backward order). We begin with $\mathcal{D}_9$ and gradually add one dataset at a time, in the order from $\mathcal{D}_8$ to $\mathcal{D}_0$. Next, we isolate each identified safety region with the utility region identified by $\mathcal{D}_u$;
    Right: pairwise Iso-Utility IoU for $\{\mathcal{D}_i\}_{i=0}^9$. The matrix is symmetric. Each element corresponds to a pairwise IoU between two utility-isolated safety regions.}
    \label{fig:results_barplot_pairwise_multi_cat_snip_wanda_14}
    \end{subfigure}

    \vspace{0.5cm}

    \begin{subfigure}[b]{\textwidth}
    \includegraphics[width=0.34\linewidth]{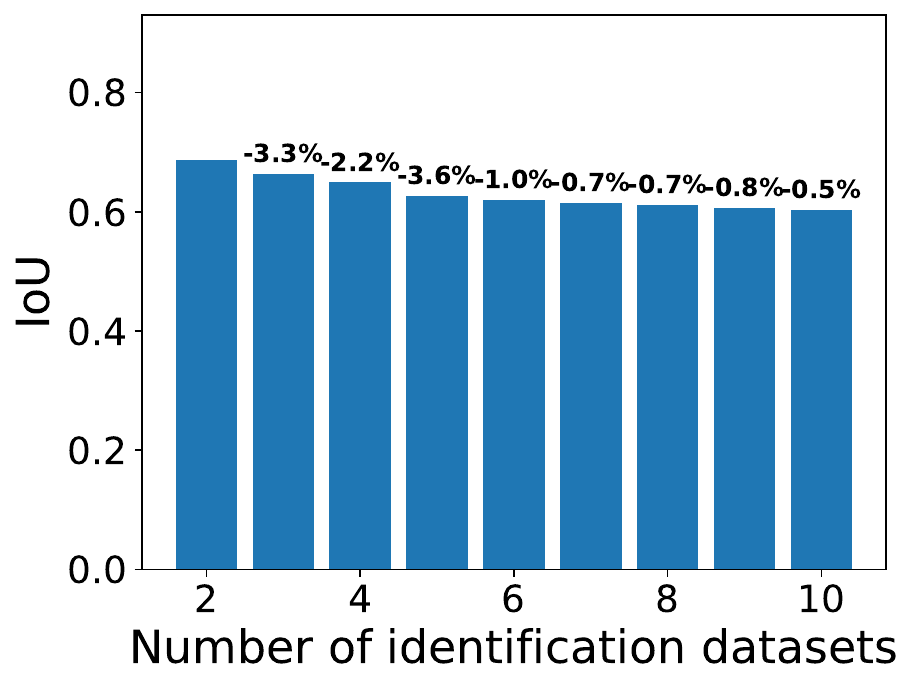}
    \hfill
    \includegraphics[width=0.34\linewidth]{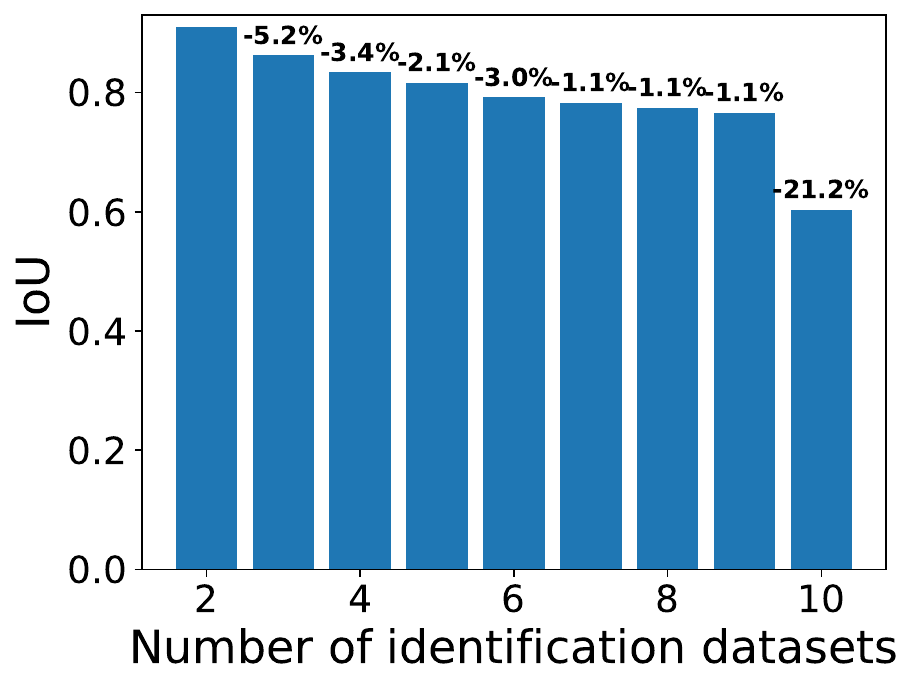}
    \hfill
    \includegraphics[width=0.3\linewidth]{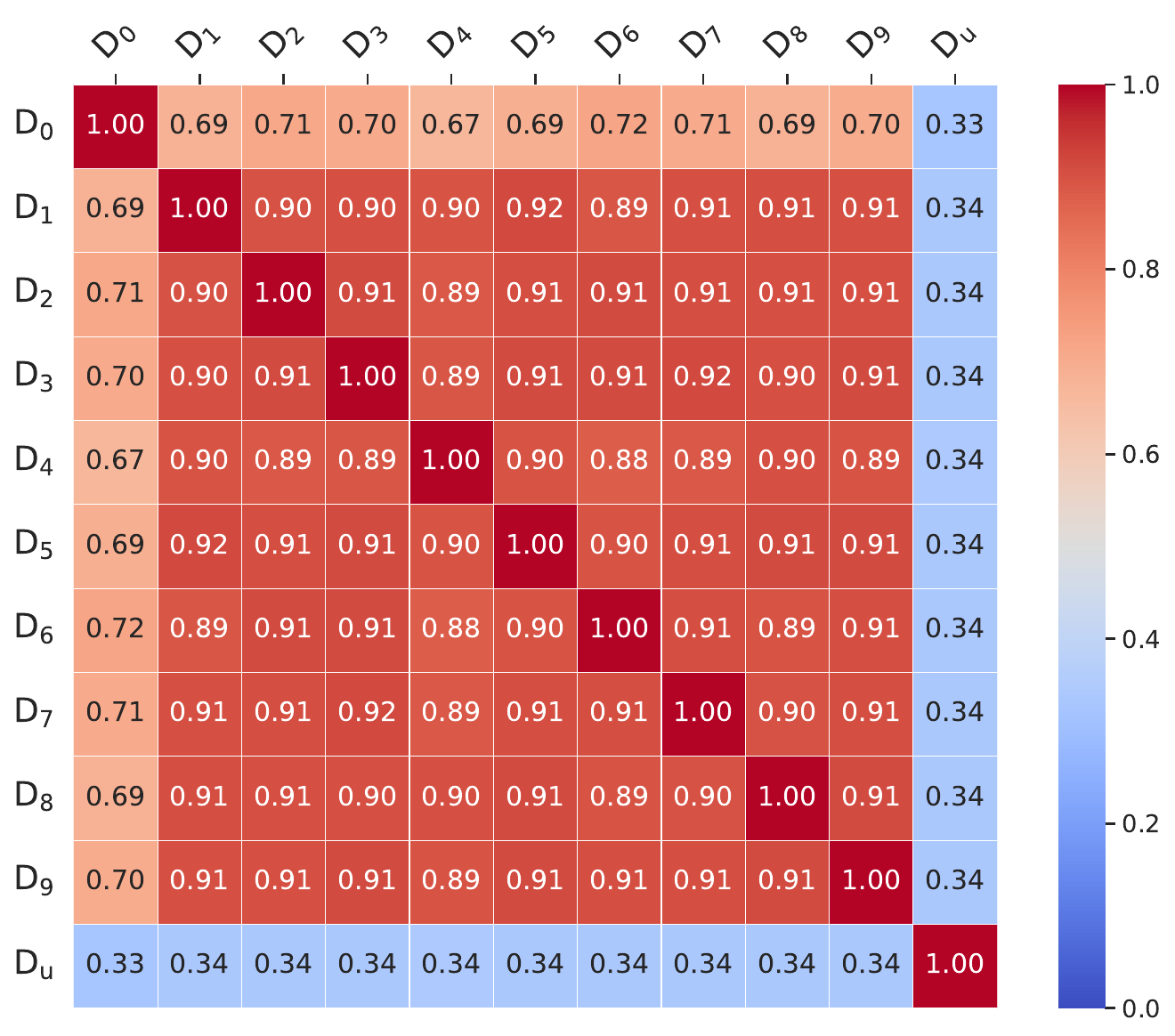}
    \caption{Safety region overlap analysis using Wanda on Llama-2-13B-Chat. The threshold used to determine top safety-critical parameters within each weight matrix is: $q\%=3\%$. Left: IoU vs $n$ (forward order). We begin with $\mathcal{D}_0$ and gradually add one dataset at a time, in the order from $\mathcal{D}_1$ to $\mathcal{D}_9$;
    Middle: IoU vs $n$ (backward order). We begin with $\mathcal{D}_9$ and gradually add one dataset at a time, in the order from $\mathcal{D}_8$ to $\mathcal{D}_0$;
    Right: pairwise IoU for $\{\mathcal{D}_i\}_{i=0}^9$. The matrix is symmetric. Each element corresponds to a pairwise IoU between two safety regions.}
    \label{fig:results_barplot_pairwise_multi_cat_snip_wanda_15}
    \end{subfigure}

\end{figure*}

\begin{figure*}[t]\ContinuedFloat

    \begin{subfigure}[b]{\textwidth}
    \includegraphics[width=0.34\linewidth]{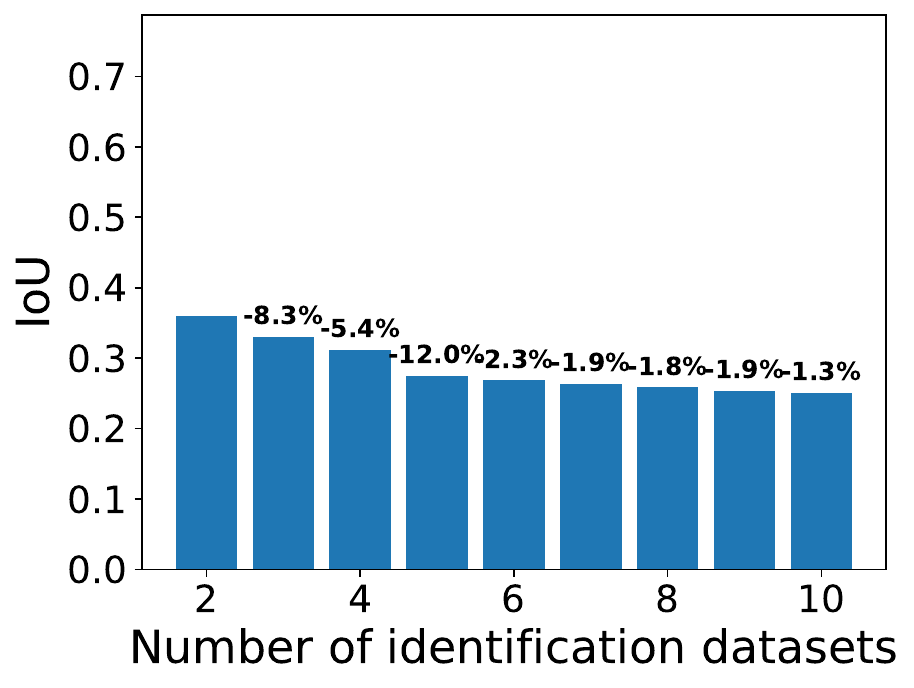}
    \hfill
    \includegraphics[width=0.34\linewidth]{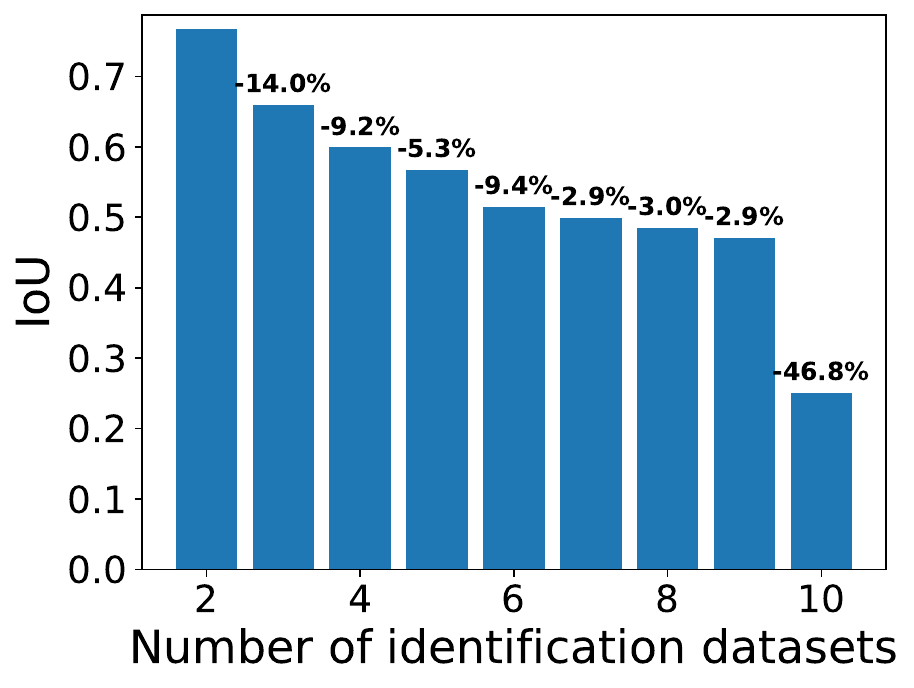}
    \hfill
    \includegraphics[width=0.3\linewidth]{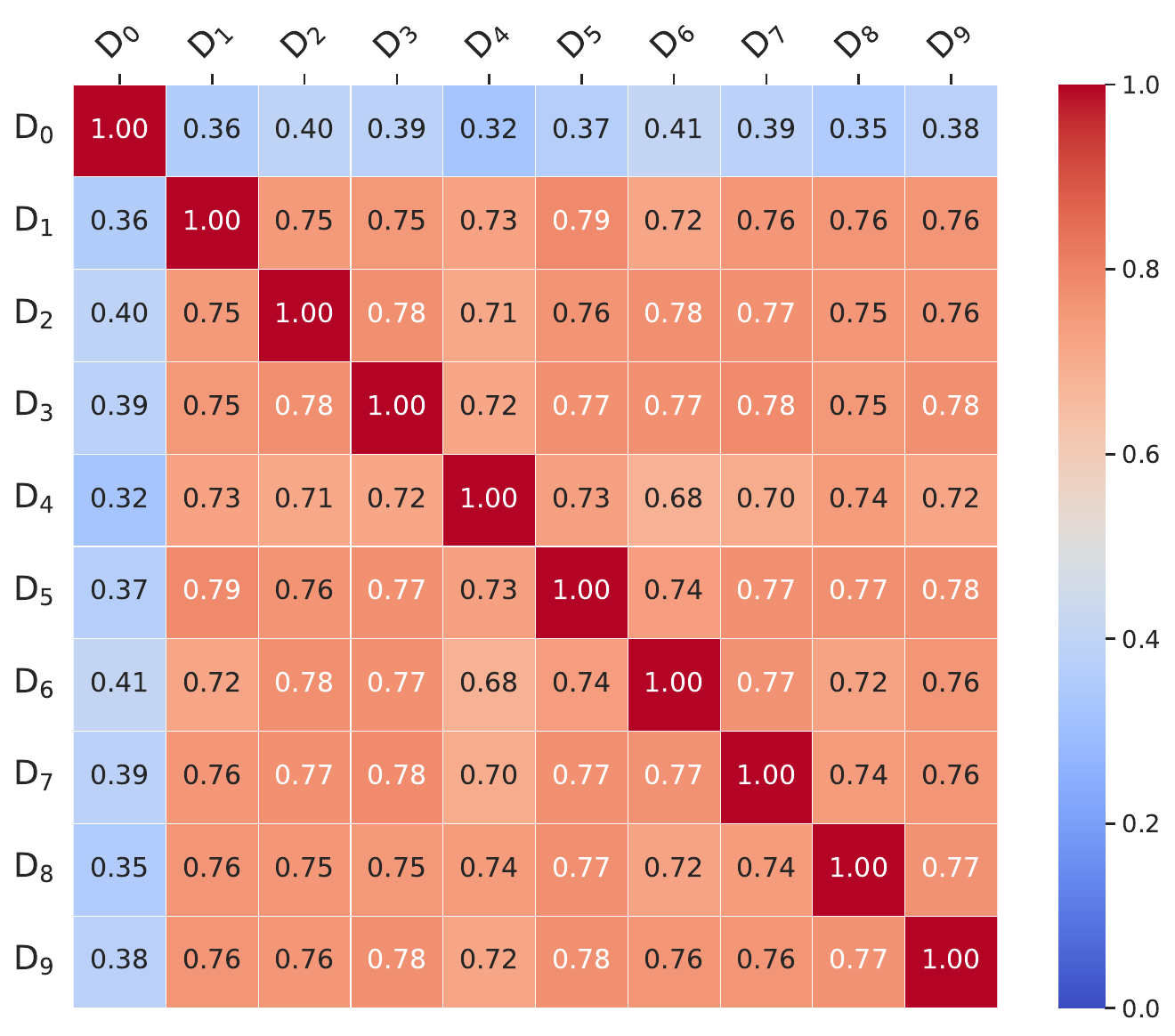}
    \caption{Utility-isolated safety region overlap analysis using Wanda on Llama-2-13B-Chat. The thresholds used to determine top safety-critical parameters within each weight matrix are: $q\%=3\%, p\%=8\%$. Left: Iso-Utility IoU vs $n$ (forward order). We begin with $\mathcal{D}_0$ and gradually add one dataset at a time, in the order from $\mathcal{D}_1$ to $\mathcal{D}_9$. Next, we isolate each identified safety region with the utility region identified by $\mathcal{D}_u$;
    Middle: Iso-Utility IoU vs $n$ (backward order). We begin with $\mathcal{D}_9$ and gradually add one dataset at a time, in the order from $\mathcal{D}_8$ to $\mathcal{D}_0$. Next, we isolate each identified safety region with the utility region identified by $\mathcal{D}_u$;
    Right: pairwise Iso-Utility IoU for $\{\mathcal{D}_i\}_{i=0}^9$. The matrix is symmetric. Each element corresponds to a pairwise IoU between two utility-isolated safety regions.}
    \label{fig:results_barplot_pairwise_multi_cat_snip_wanda_16}
    \end{subfigure}

\caption{Safety region overlap analysis using SNIP or Wanda on Llama-2-7B-Chat and Llama-2-13B-Chat}
\label{fig:results_barplot_pairwise_multi_cat_snip_wanda}
\end{figure*}

\begin{figure*}[htbp!]
    \centering
    
    \begin{subfigure}[b]{\textwidth}
    \includegraphics[width=0.34\linewidth]{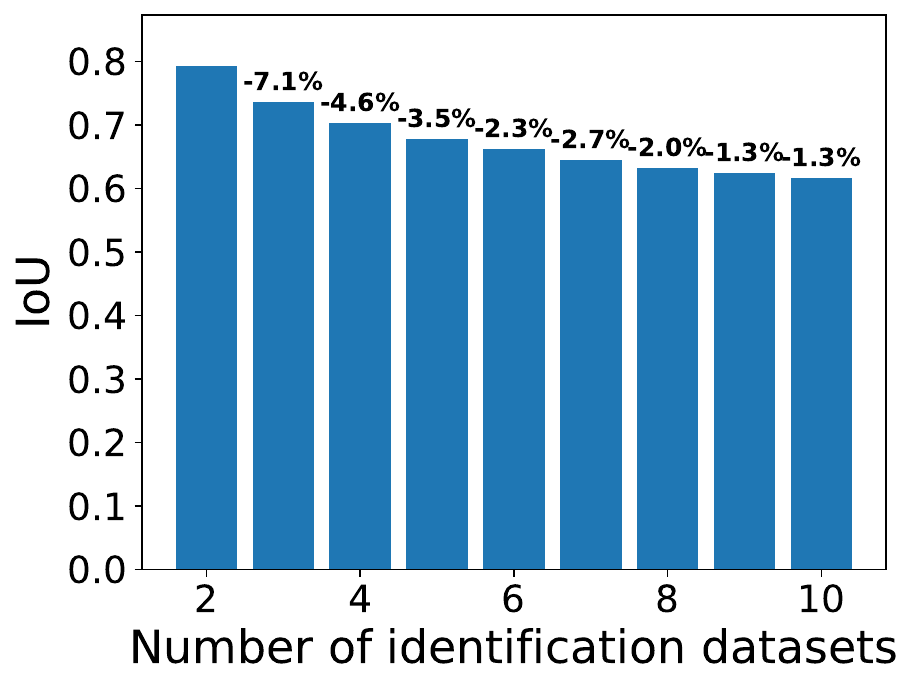}
    \hfill
    \includegraphics[width=0.34\linewidth]{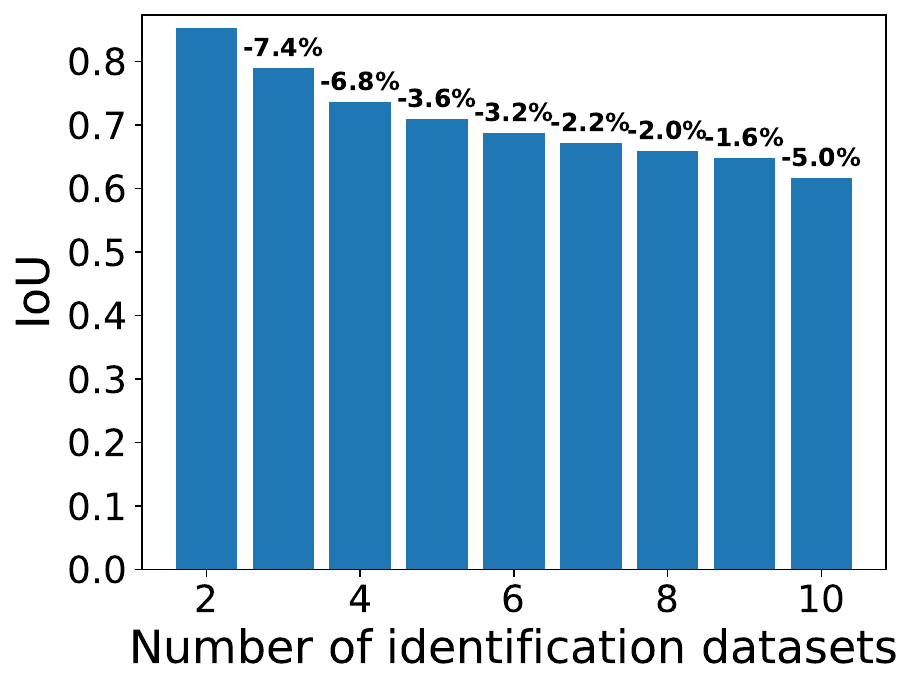}
    \hfill
    \includegraphics[width=0.3\linewidth]{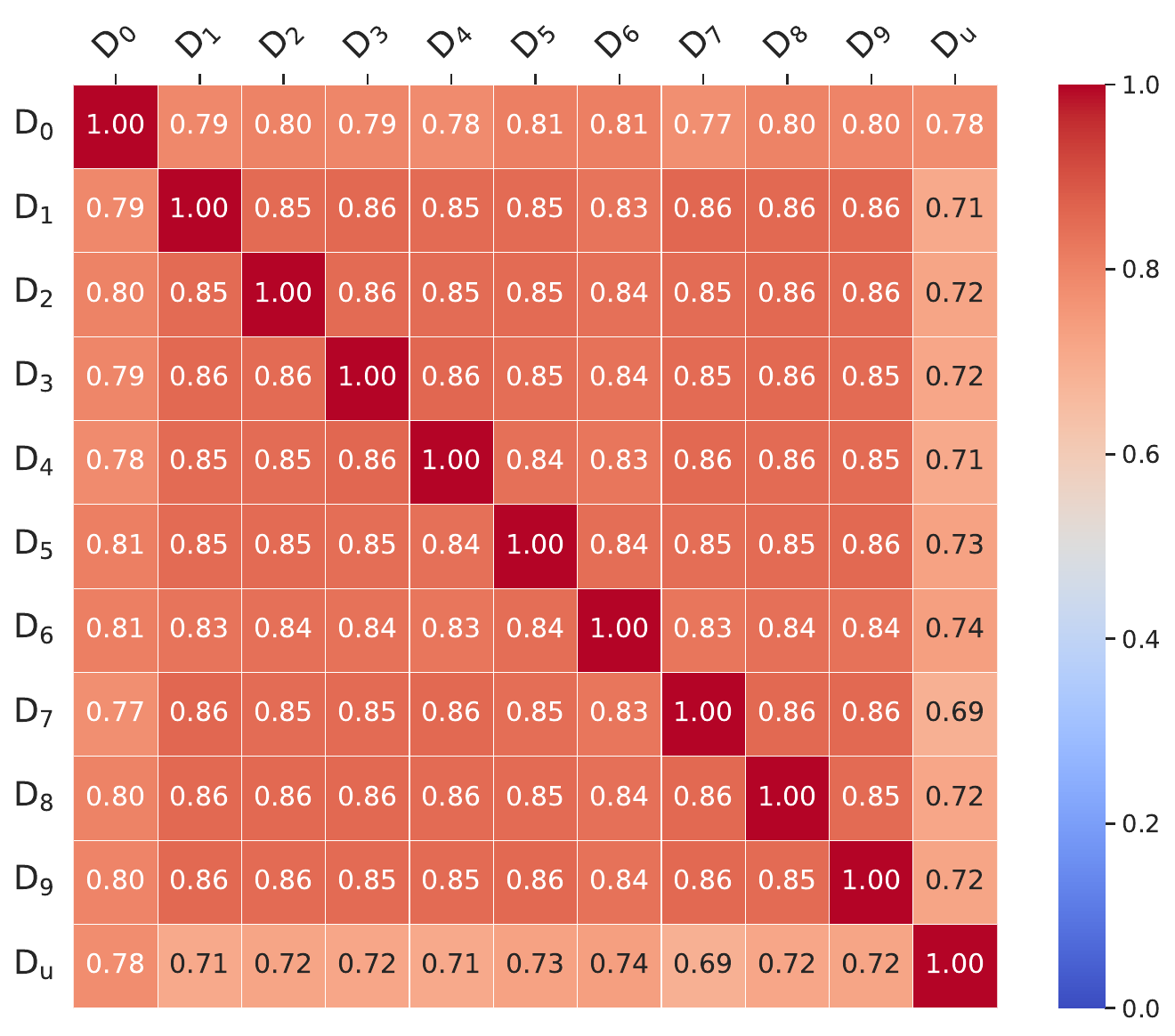}
    \caption{Safety region overlap analysis using SafeNeuron on Llama-2-7B-Chat. Left: IoU vs $n$ (forward order). We begin with $\mathcal{D}_0$ and gradually add one dataset at a time, in the order from $\mathcal{D}_1$ to $\mathcal{D}_9$;
    Middle: IoU vs $n$ (backward order). We begin with $\mathcal{D}_9$ and gradually add one dataset at a time, in the order from $\mathcal{D}_8$ to $\mathcal{D}_0$;
    Right: pairwise IoU for $\{\mathcal{D}_i\}_{i=0}^9$. The matrix is symmetric. Each element corresponds to a pairwise IoU between two safety regions.}
    \label{fig:results_barplot_pairwise_multi_cat_safeneuron_1}
    \end{subfigure}

    \vspace{0.5cm}

    \begin{subfigure}[b]{\textwidth}
    \includegraphics[width=0.34\linewidth]{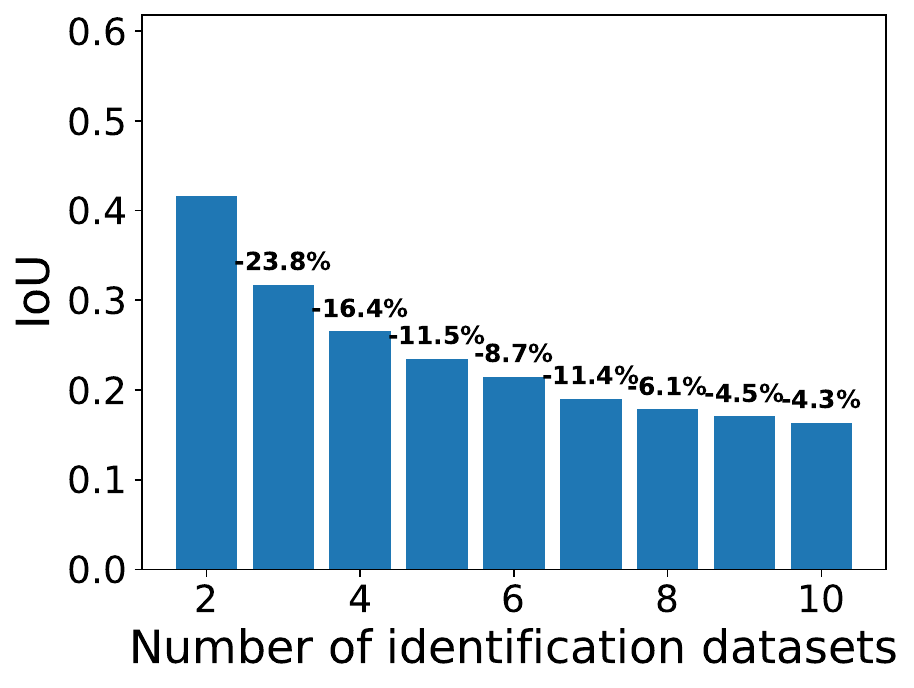}
    \hfill
    \includegraphics[width=0.34\linewidth]{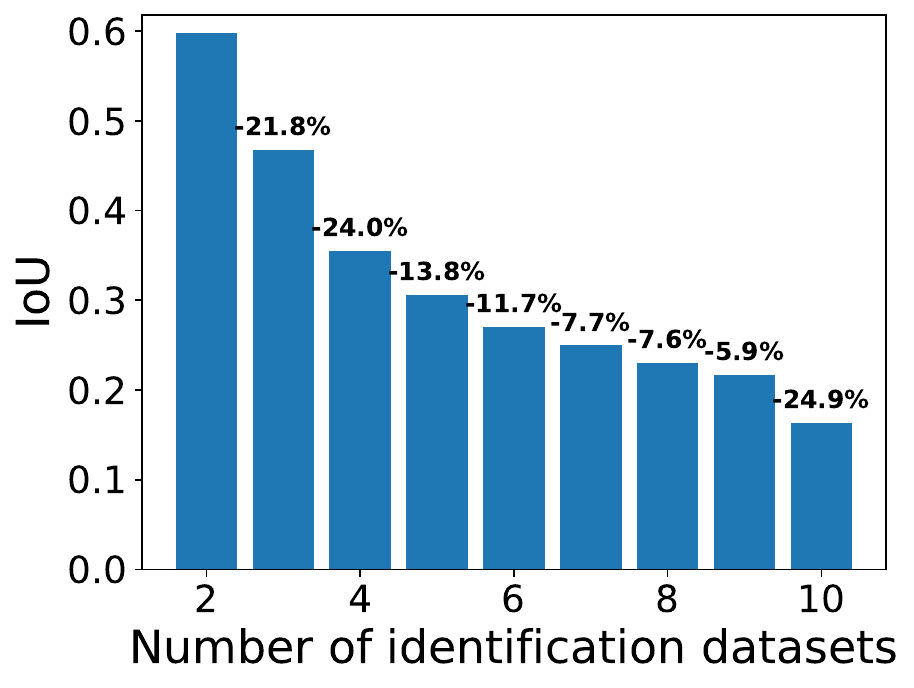}
    \hfill
    \includegraphics[width=0.3\linewidth]{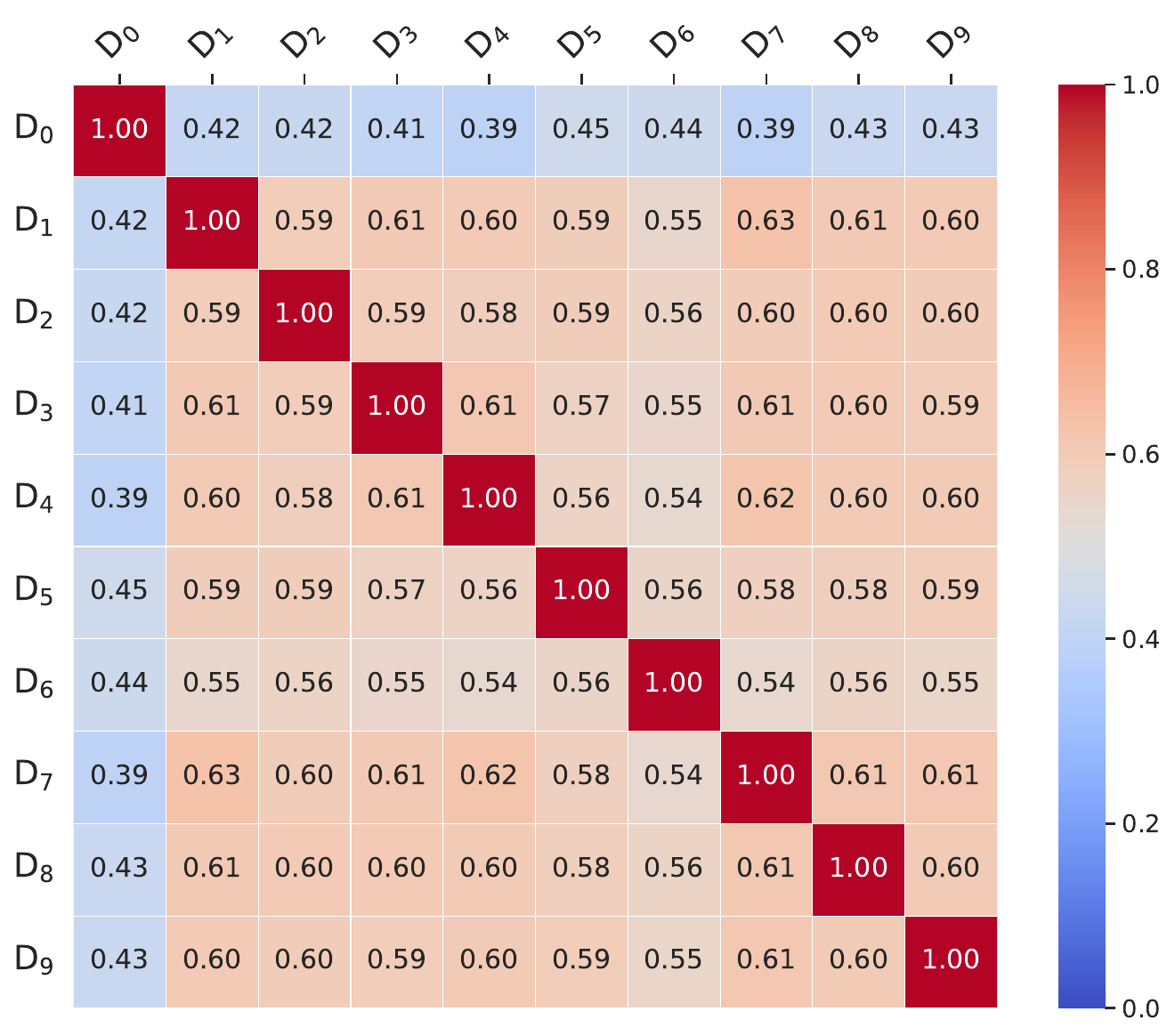}
    \caption{Utility-isolated safety region overlap analysis using SafeNeuron on Llama-2-7B-Chat. Left: Iso-Utility IoU vs $n$ (forward order). We begin with $\mathcal{D}_0$ and gradually add one dataset at a time, in the order from $\mathcal{D}_1$ to $\mathcal{D}_9$. Next, we isolate each identified safety region with the utility region identified by $\mathcal{D}_u$;
    Middle: Iso-Utility IoU vs $n$ (backward order). We begin with $\mathcal{D}_9$ and gradually add one dataset at a time, in the order from $\mathcal{D}_8$ to $\mathcal{D}_0$. Next, we isolate each identified safety region with the utility region identified by $\mathcal{D}_u$;
    Right: pairwise Iso-Utility IoU for $\{\mathcal{D}_i\}_{i=0}^9$. The matrix is symmetric. Each element corresponds to a pairwise IoU between two utility-isolated safety regions.}
    \label{fig:results_barplot_pairwise_multi_cat_safeneuron_2}
    \end{subfigure}

    \vspace{0.5cm}

    \begin{subfigure}[b]{\textwidth}
    \includegraphics[width=0.34\linewidth]{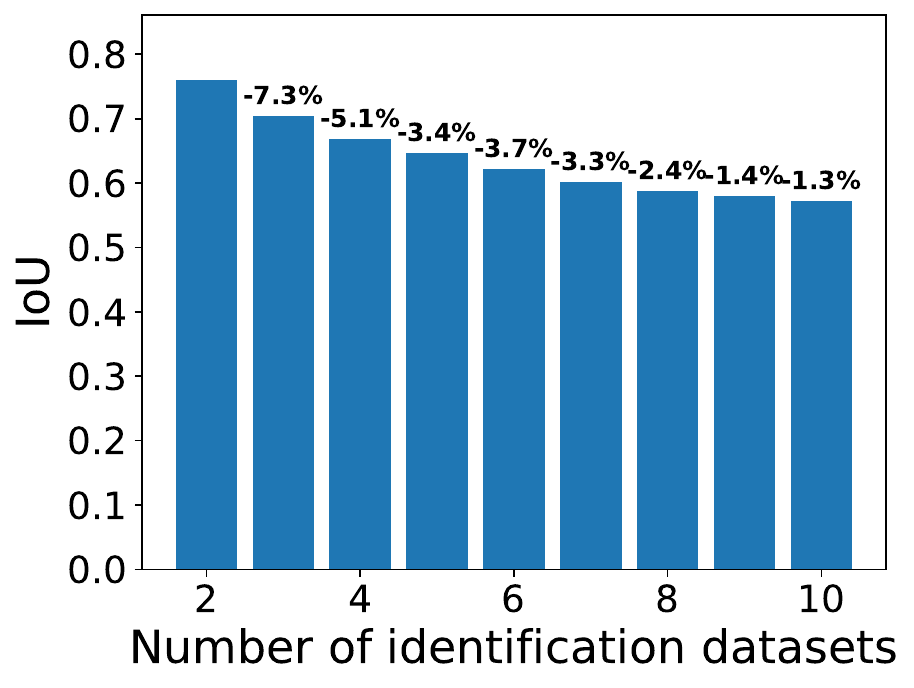}
    \hfill
    \includegraphics[width=0.34\linewidth]{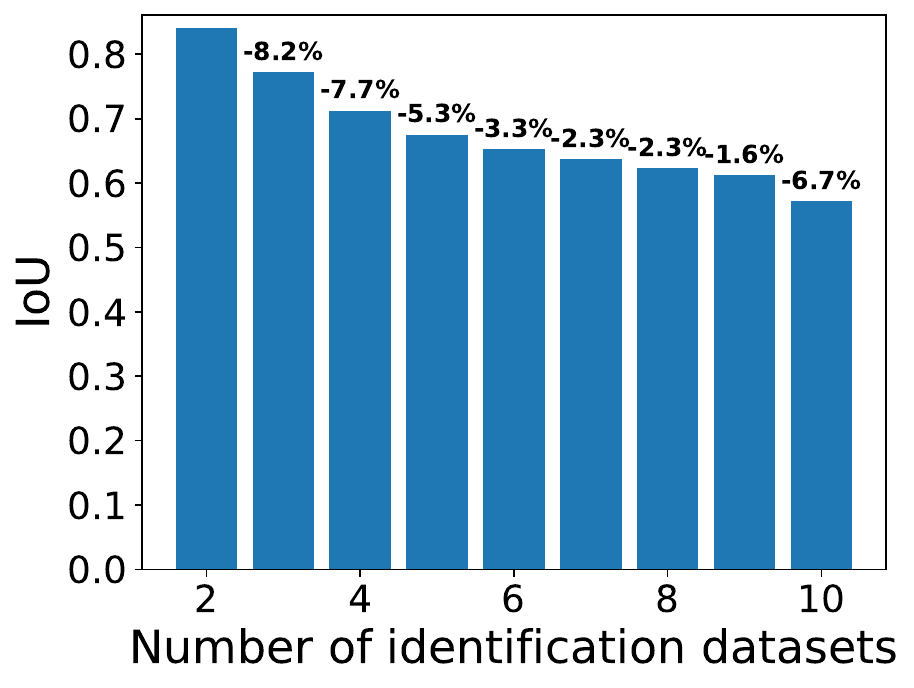}
    \hfill
    \includegraphics[width=0.3\linewidth]{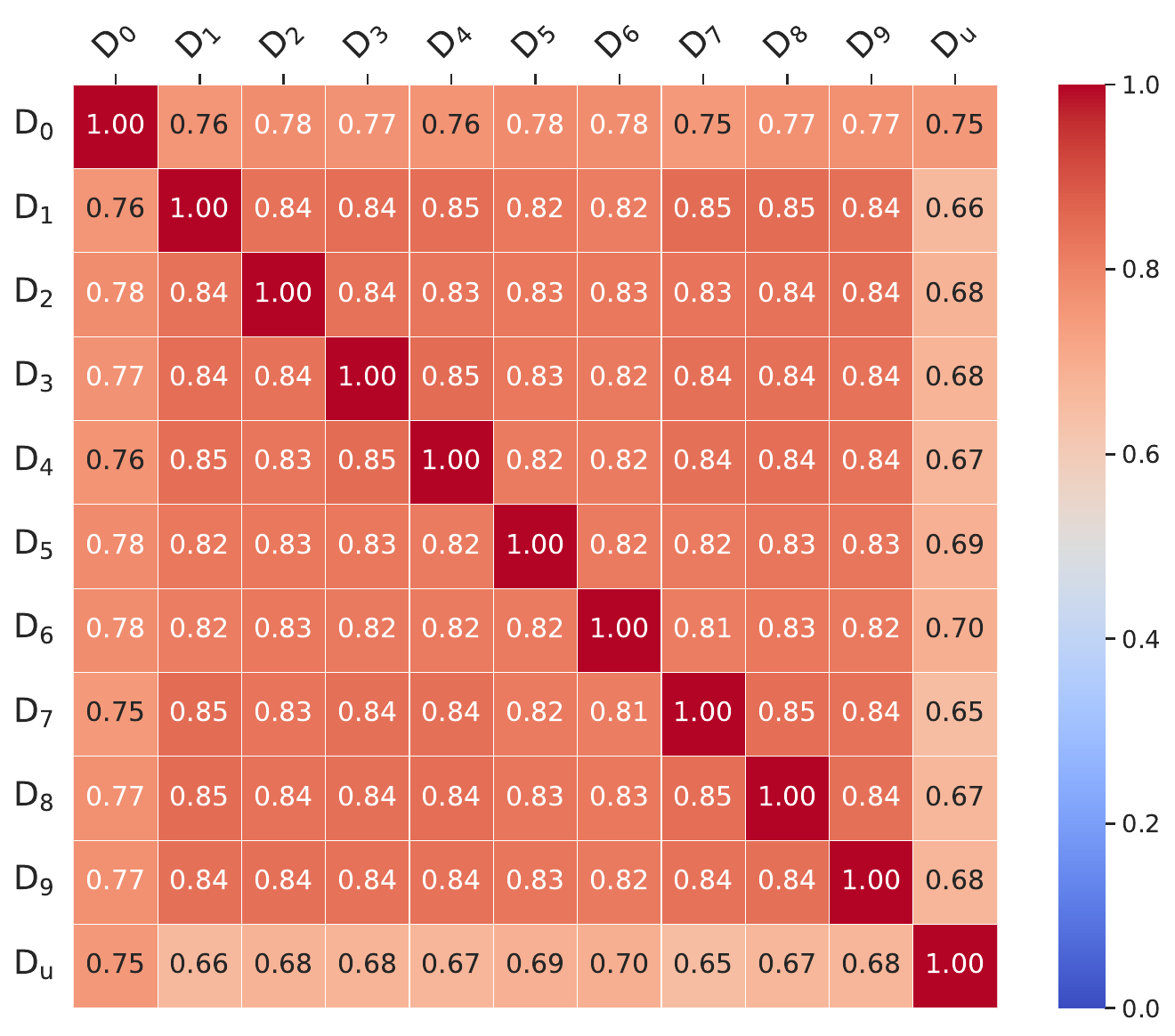}
    \caption{Safety region overlap analysis using SafeNeuron on Mistral-7B-Instruct-v0.2. Left: IoU vs $n$ (forward order). We begin with $\mathcal{D}_0$ and gradually add one dataset at a time, in the order from $\mathcal{D}_1$ to $\mathcal{D}_9$;
    Middle: IoU vs $n$ (backward order). We begin with $\mathcal{D}_9$ and gradually add one dataset at a time, in the order from $\mathcal{D}_8$ to $\mathcal{D}_0$;
    Right: pairwise IoU for $\{\mathcal{D}_i\}_{i=0}^9$. The matrix is symmetric. Each element corresponds to a pairwise IoU between two safety regions.}
    \label{fig:results_barplot_pairwise_multi_cat_safeneuron_3}
    \end{subfigure}
\end{figure*}

\begin{figure*}[t]\ContinuedFloat

    \begin{subfigure}[b]{\textwidth}
    \includegraphics[width=0.34\linewidth]{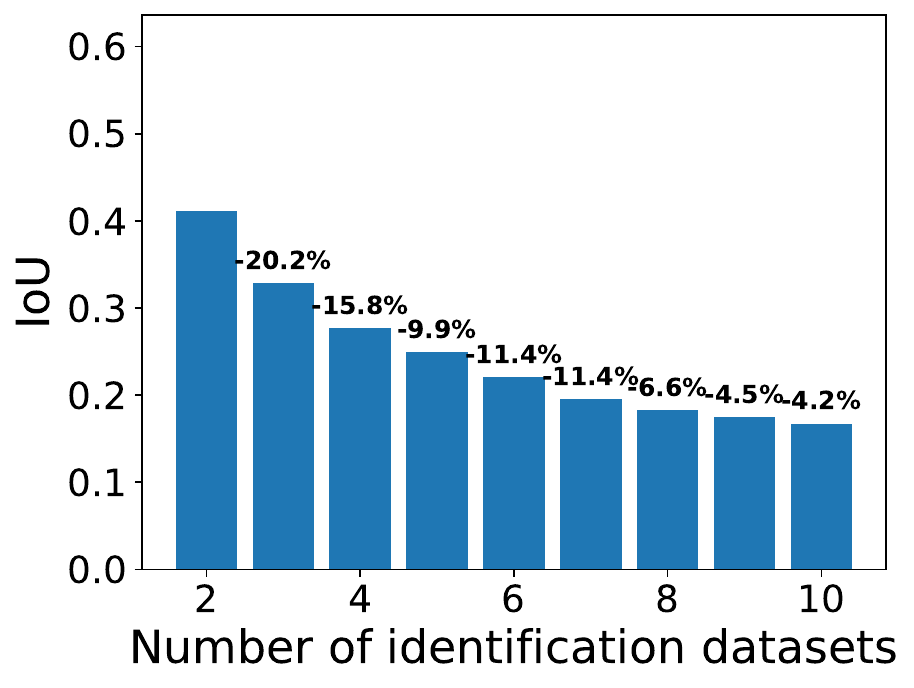}
    \hfill
    \includegraphics[width=0.34\linewidth]{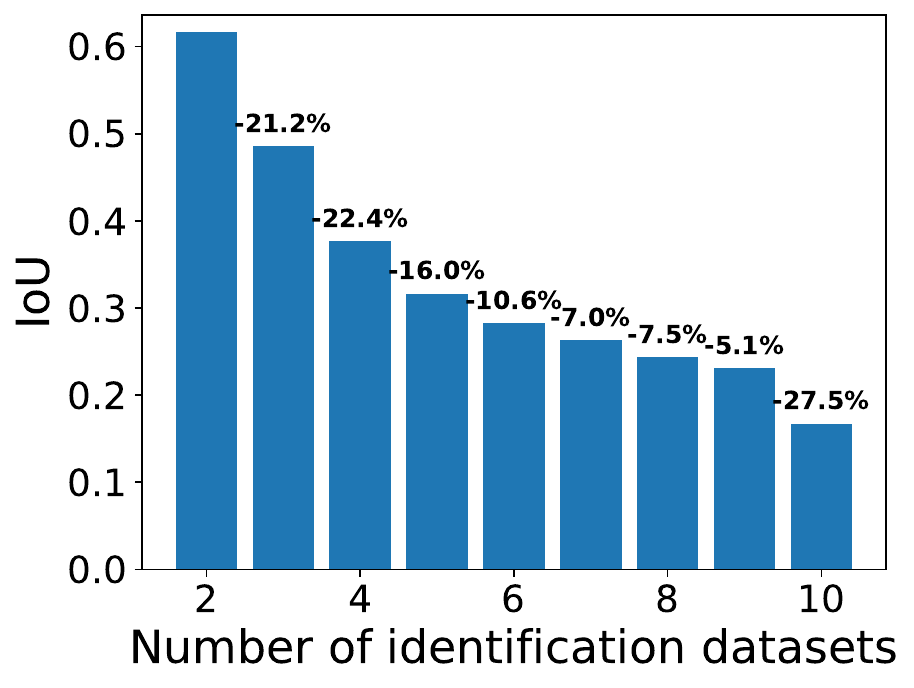}
    \hfill
    \includegraphics[width=0.3\linewidth]{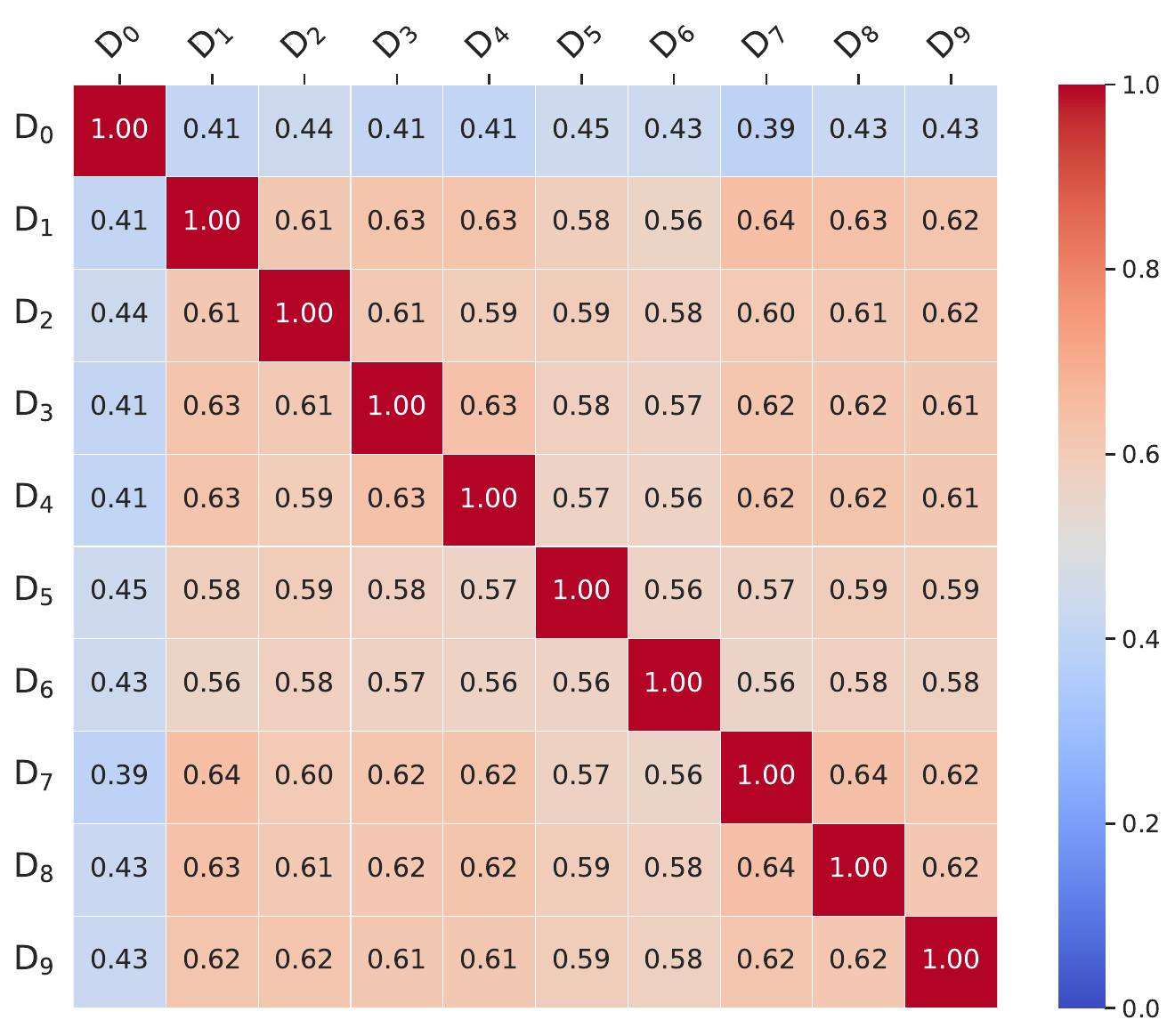}
    \caption{Utility-isolated safety region overlap analysis using SafeNeuron on Mistral-7B-Instruct-v0.2. Left: Iso-Utility IoU vs $n$ (forward order). We begin with $\mathcal{D}_0$ and gradually add one dataset at a time, in the order from $\mathcal{D}_1$ to $\mathcal{D}_9$. Next, we isolate each identified safety region with the utility region identified by $\mathcal{D}_u$;
    Middle: Iso-Utility IoU vs $n$ (backward order). We begin with $\mathcal{D}_9$ and gradually add one dataset at a time, in the order from $\mathcal{D}_8$ to $\mathcal{D}_0$. Next, we isolate each identified safety region with the utility region identified by $\mathcal{D}_u$;
    Right: pairwise Iso-Utility IoU for $\{\mathcal{D}_i\}_{i=0}^9$. The matrix is symmetric. Each element corresponds to a pairwise IoU between two utility-isolated safety regions.}
    \label{fig:results_barplot_pairwise_multi_cat_safeneuron_4}
    \end{subfigure}

\vspace{0.5cm}

    \begin{subfigure}[b]{\textwidth}
    \includegraphics[width=0.34\linewidth]{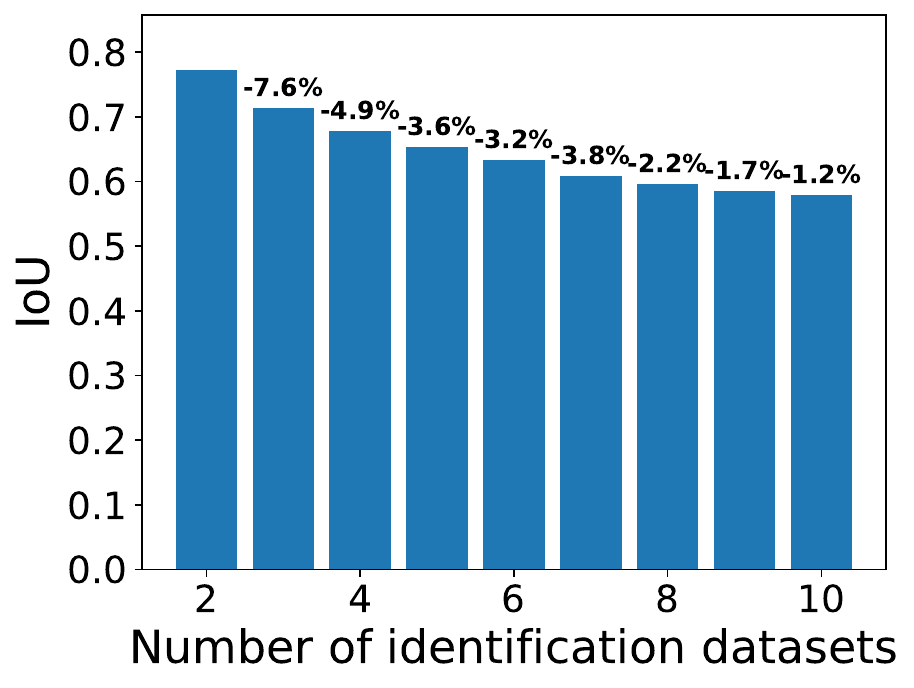}
    \hfill
    \includegraphics[width=0.34\linewidth]{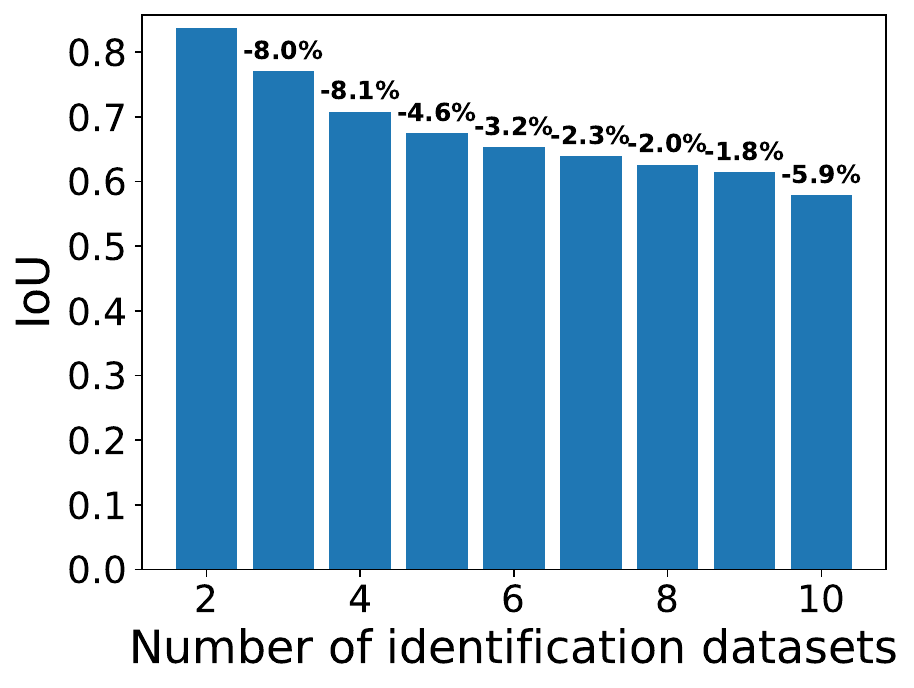}
    \hfill
    \includegraphics[width=0.3\linewidth]{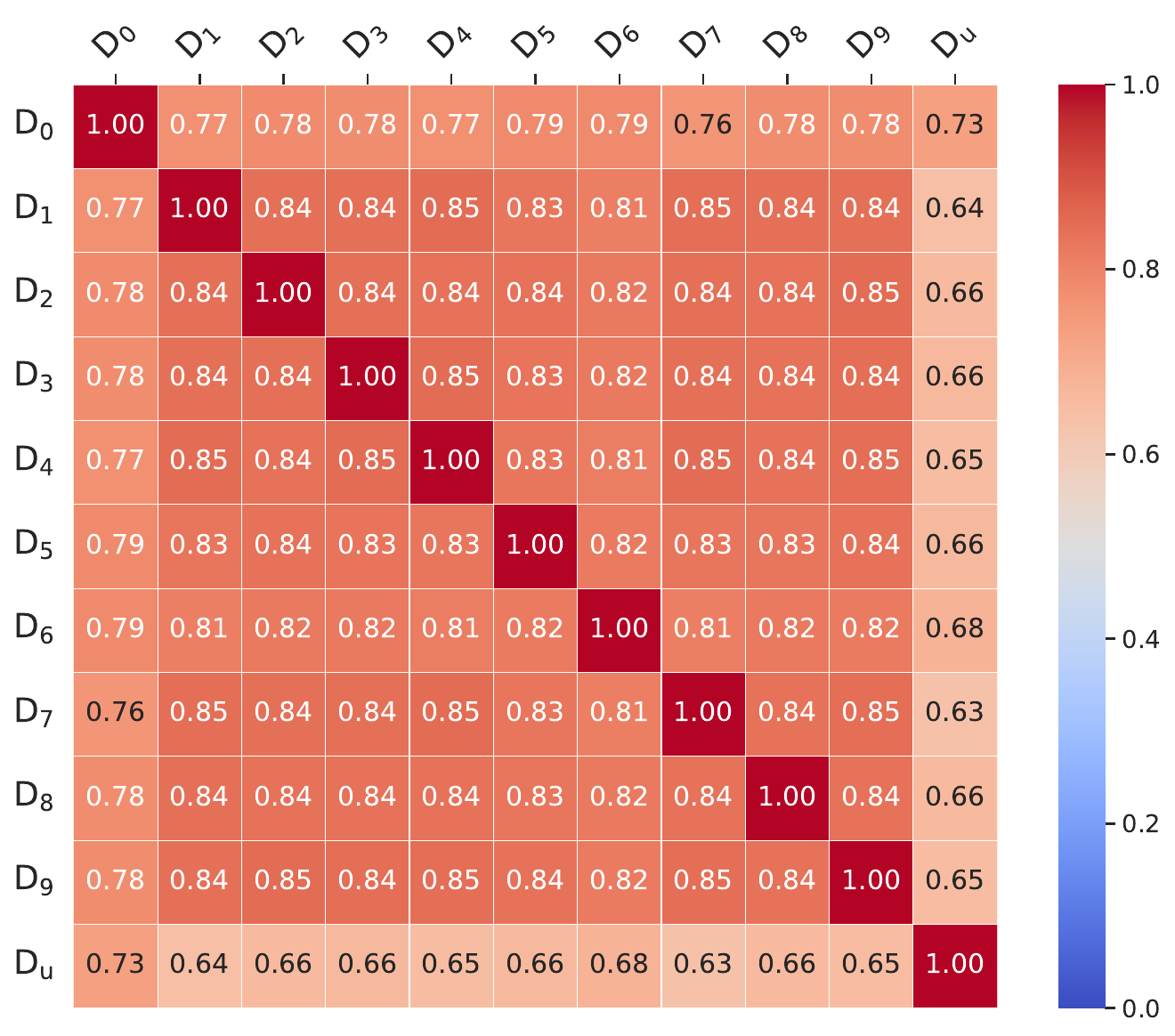}
    \caption{Safety region overlap analysis using SafeNeuron on Llama-3-8B-Instruct. Left: IoU vs $n$ (forward order). We begin with $\mathcal{D}_0$ and gradually add one dataset at a time, in the order from $\mathcal{D}_1$ to $\mathcal{D}_9$;
    Middle: IoU vs $n$ (backward order). We begin with $\mathcal{D}_9$ and gradually add one dataset at a time, in the order from $\mathcal{D}_8$ to $\mathcal{D}_0$;
    Right: pairwise IoU for $\{\mathcal{D}_i\}_{i=0}^9$. The matrix is symmetric. Each element corresponds to a pairwise IoU between two safety regions.}
    \label{fig:results_barplot_pairwise_multi_cat_safeneuron_5}
    \end{subfigure}

\caption{Safety region overlap analysis using SafeNeuron on Llama-2-7B-Chat, Llama-3-8B-Instruct and Mistral-7B-Instruct-v0.2.
}
\label{fig:results_barplot_pairwise_multi_cat_safeneuron}
\end{figure*}

\section{Details about Cosine Similarity Analysis}
\label{appendix: cos_sim}

\begin{figure*}[htbp!]
    \centering
    \begin{subfigure}[b]{0.325\textwidth}
        \includegraphics[width=\linewidth]{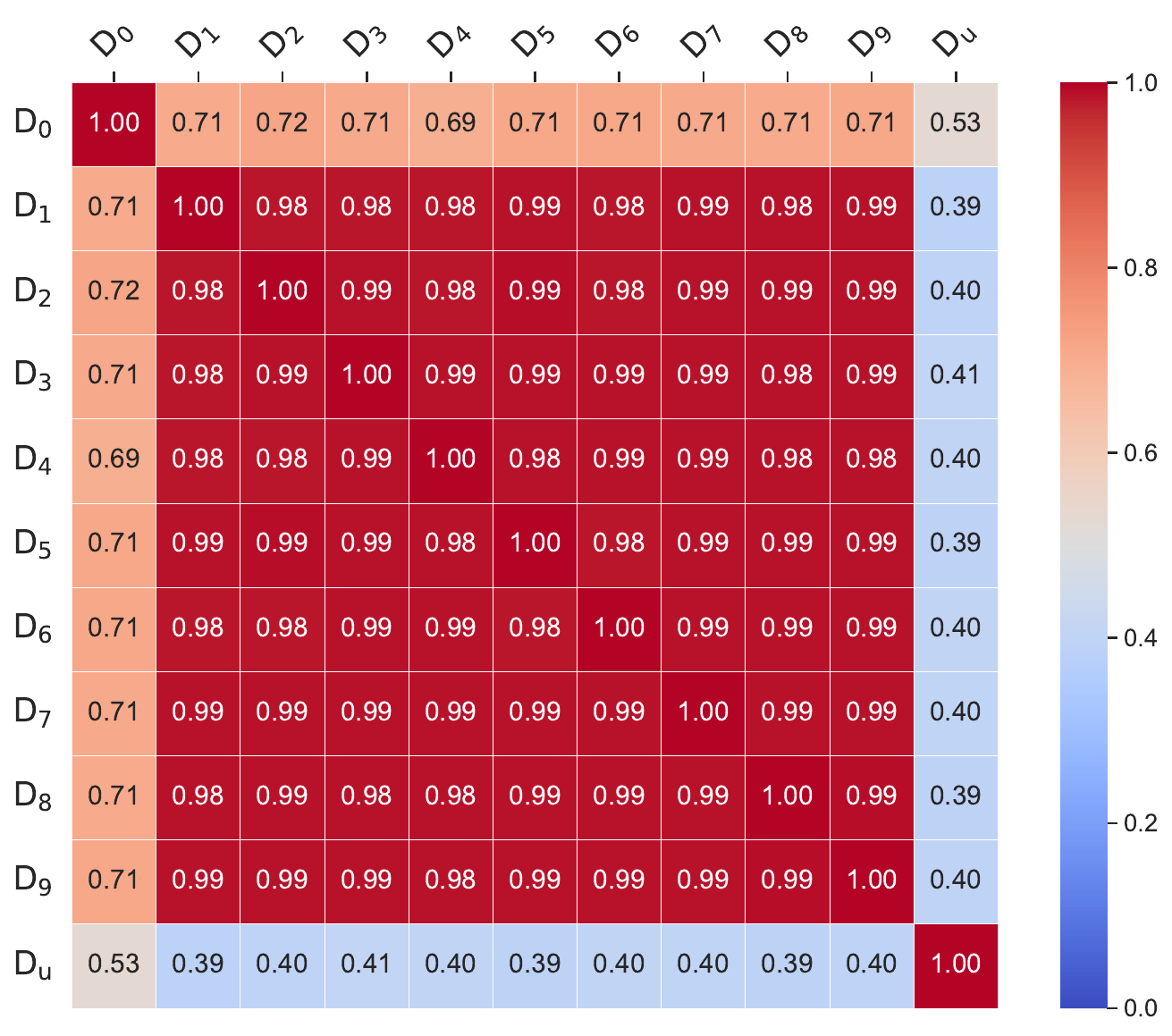}
        \caption{SNIP \& Wanda}
        \label{fig:appendix_cos_sim_uniform_snip_wanda}
    \end{subfigure}
    \hfill
    \begin{subfigure}[b]{0.325\textwidth}
        \includegraphics[width=\linewidth]{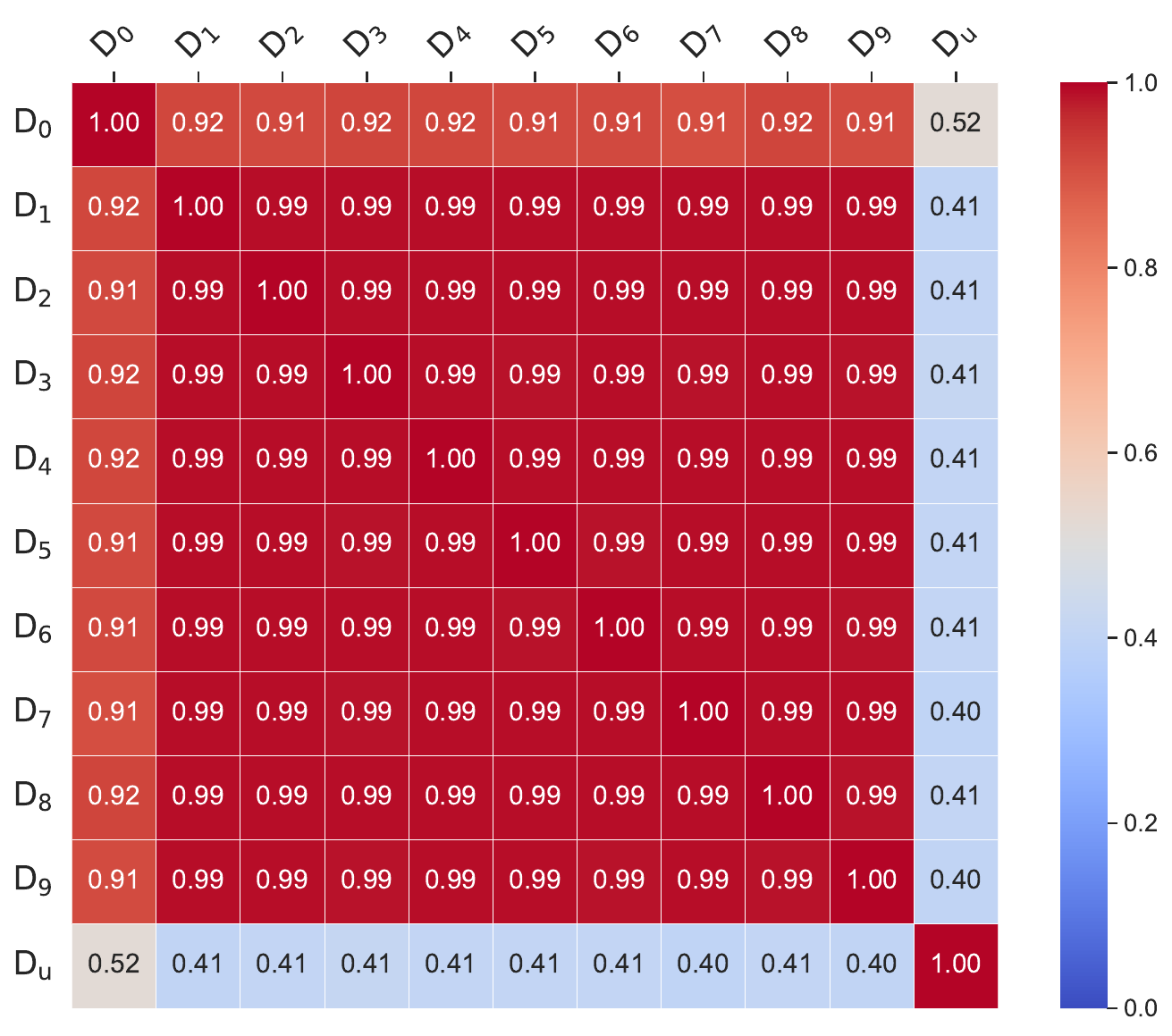}
        \caption{SafeNeuron}
        \label{fig:appendix_cos_sim_uniform_safeneuron}
    \end{subfigure}
    \hfill
    \begin{subfigure}[b]{0.325\textwidth}
        \includegraphics[width=\linewidth]{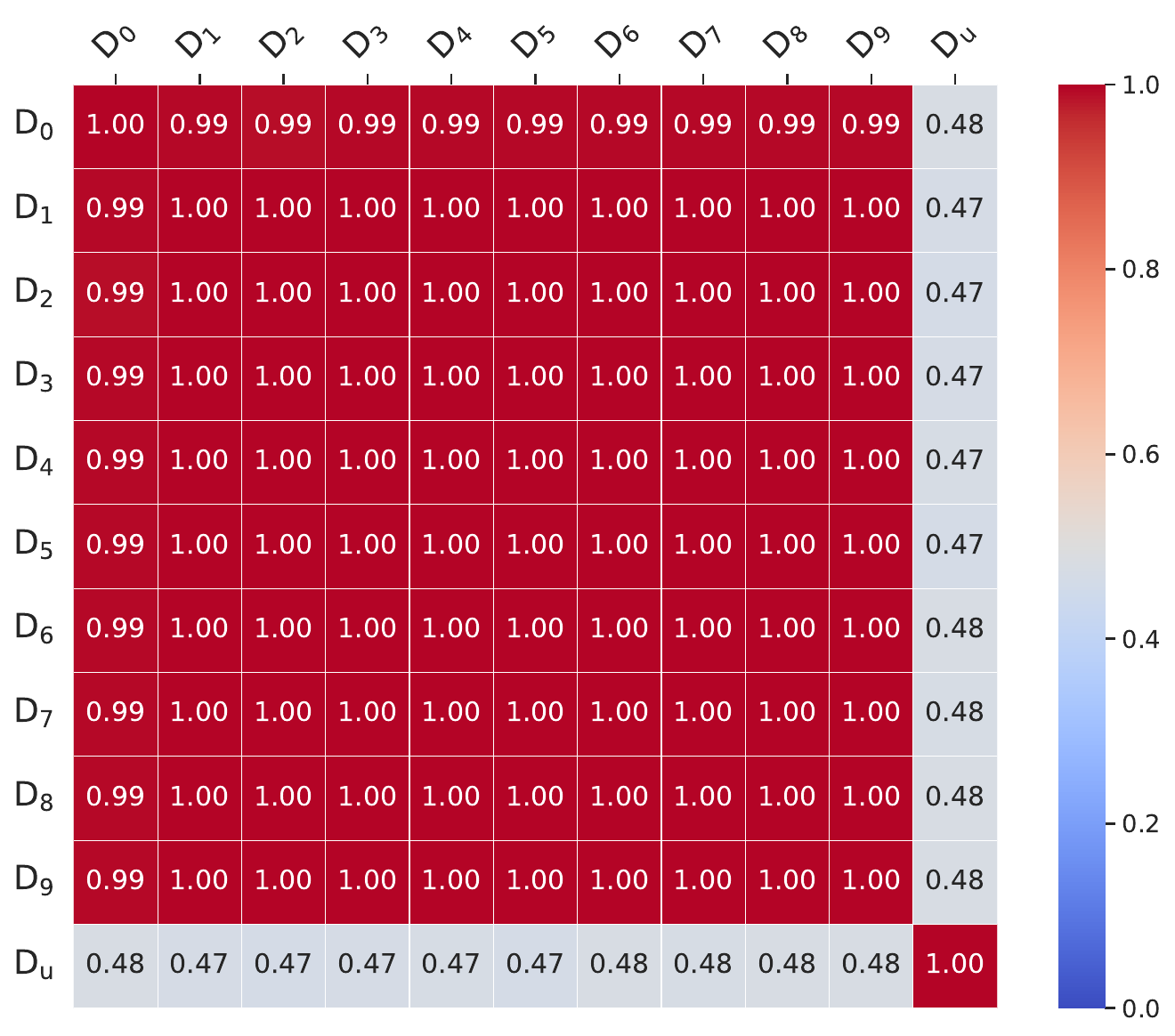}
        \caption{NLSR}
        \label{fig:appendix_cos_sim_uniform_nlsr}
    \end{subfigure}
\caption{
Pairwise cosine similarity for the multi-category datasets of SNIP \& Wanda, SafeNeuron, and NLSR. For SNIP \& Wanda, $\mathcal{D}_0$ was sampled using random seed equal to $0$.}
\label{fig:appendix_cos_sim_uniform}
\end{figure*}

\begin{figure*}[htbp!]
    \centering
    \begin{subfigure}[b]{0.48\textwidth}
        \includegraphics[width=\linewidth]{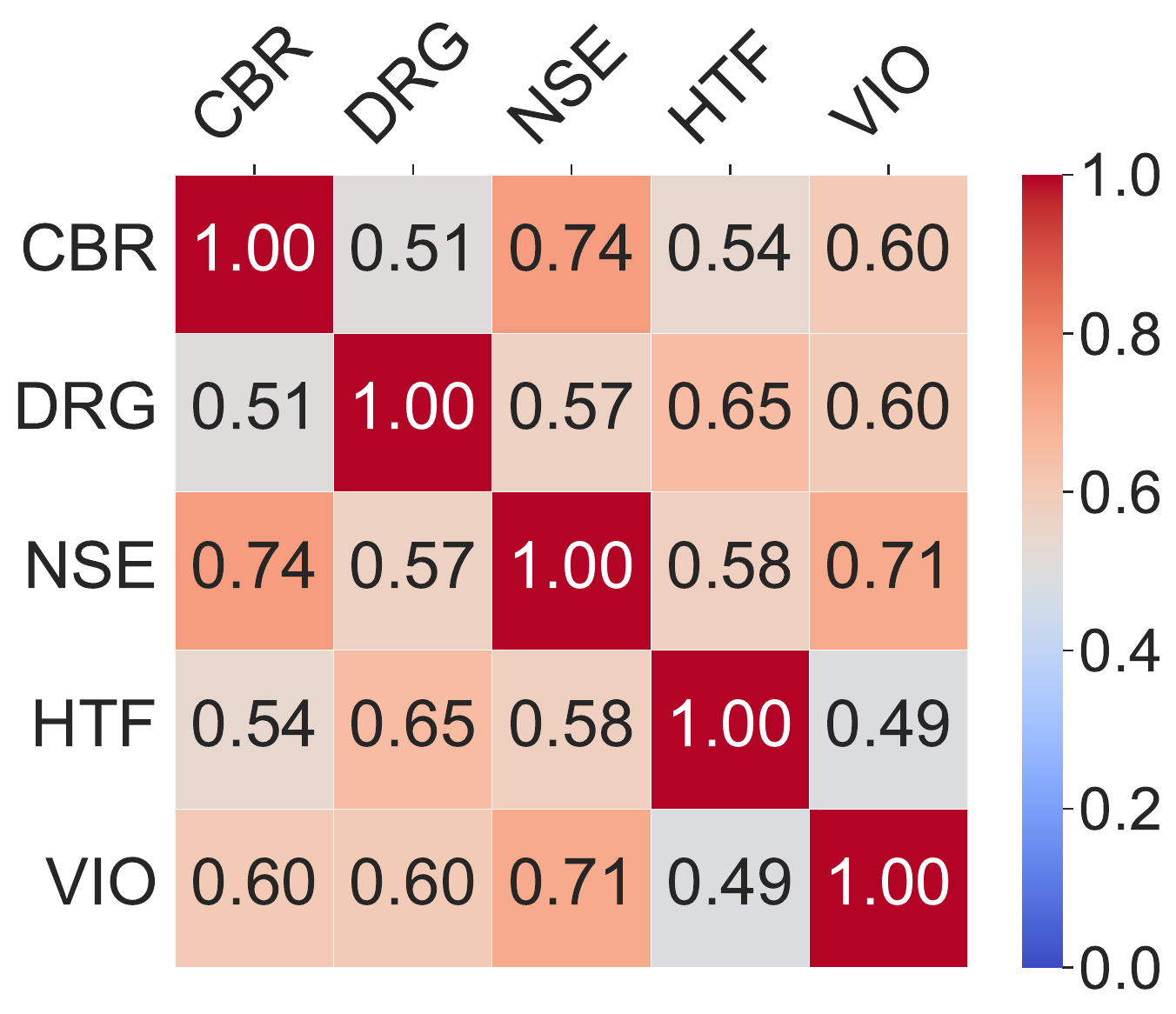}
        \caption{SNIP \& Wanda}
        \label{fig:appendix_cos_sim_single_snip_wanda}
    \end{subfigure}
    \hfill
    \begin{subfigure}[b]{0.48\textwidth}
        \includegraphics[width=\linewidth]{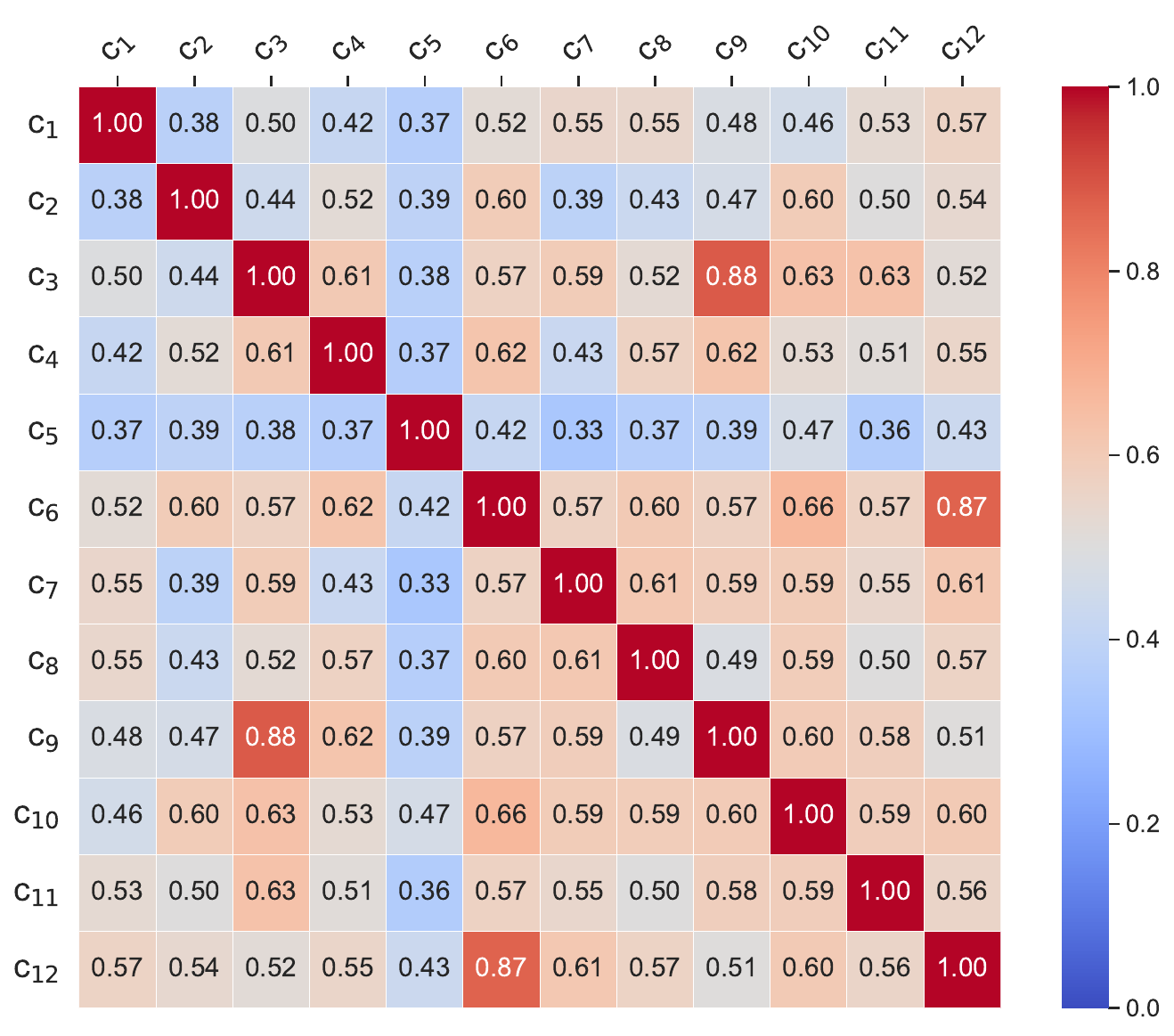}
        \caption{SafeNeuron}
        \label{fig:appendix_cos_sim_single_safeneuron}
    \end{subfigure}
\caption{
Pairwise cosine similarity for the single-category datasets of SNIP \& Wanda, SafeNeuron. Note that they show the same matrix as the pairwise cosine similarity matrix in Fig.~\ref{fig:results_second_category}.
}
\label{fig:appendix_cos_sim_single}
\end{figure*}

Given two identification datasets $\mathcal{D}_i$ and $\mathcal{D}_j$, the cosine similarity between them is defined as the cosine similarity between their centroid embeddings, \ie the mean of the embeddings of all samples in the dataset. We use the embedding model \emph{text-embedding-3-large} of OpenAI. Specifically, for the datasets of SNIP \& Wanda and NLSR, each data sample consists of a harmful query paired with a refusal answer. Since refusal answers are largely uniform in semantics, we only get the embedding of each harmful query to compute the centroids. For SafeNeuron, since the datasets consist solely of harmful queries, we directly embed these queries to obtain the centroids. Finally, Fig.~\ref{fig:appendix_cos_sim_uniform} shows the pairwise cosine similarity for the multi-category datasets of SNIP \& Wanda, SafeNeuron, and NLSR, while Fig.~\ref{fig:appendix_cos_sim_single} shows the results for the single-category datasets of SNIP \& Wanda and SafeNeuron.

\section{More Pairwise Overlap Analysis with Single-Category Datasets}
\label{appendix:pairwise_single_cat}

\begin{figure*}[t]
    
    \begin{subfigure}{0.475\textwidth}
    \centering
    \includegraphics[width=0.49\linewidth]{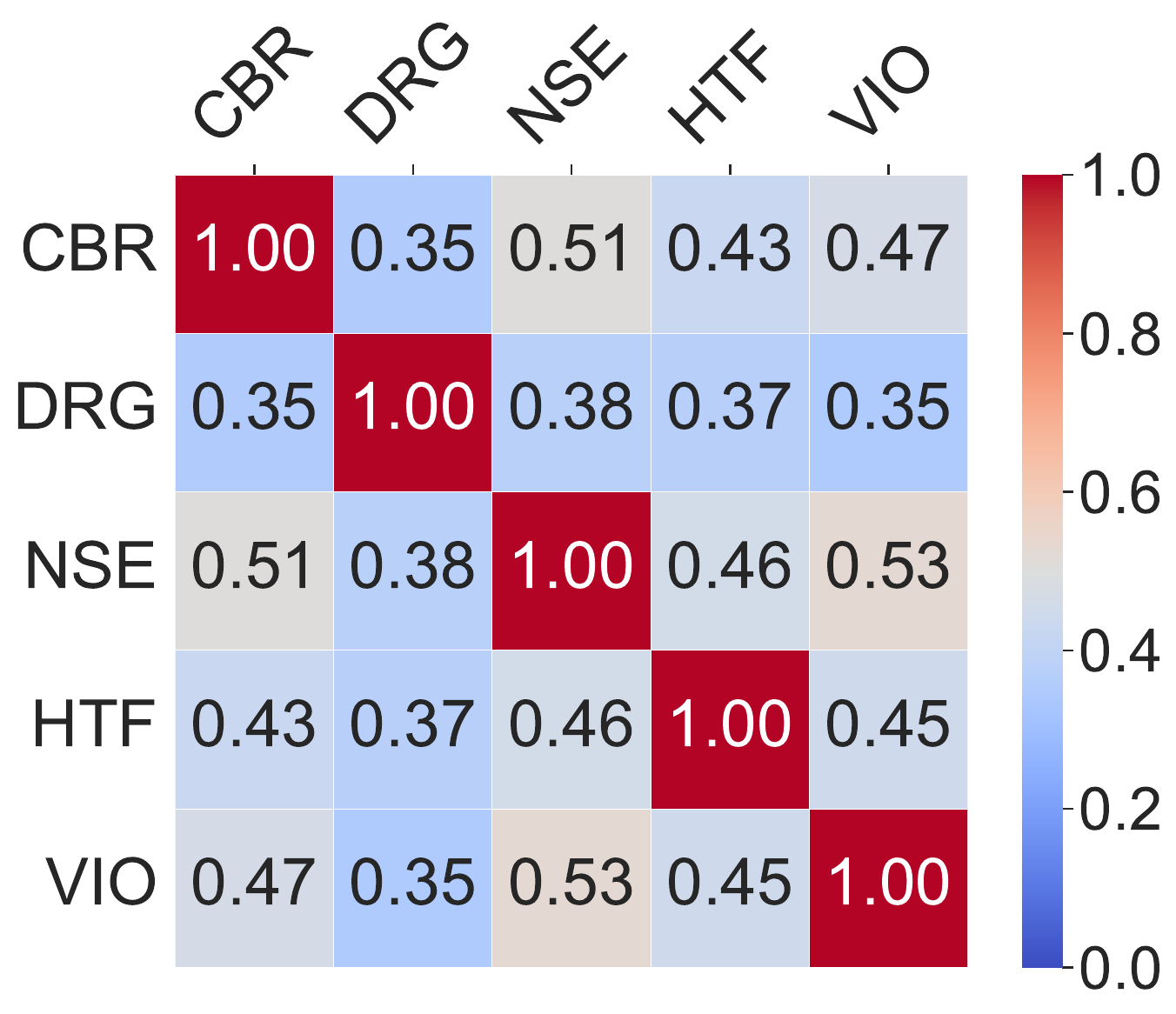}
    \hfill
    \includegraphics[width=0.49\linewidth]{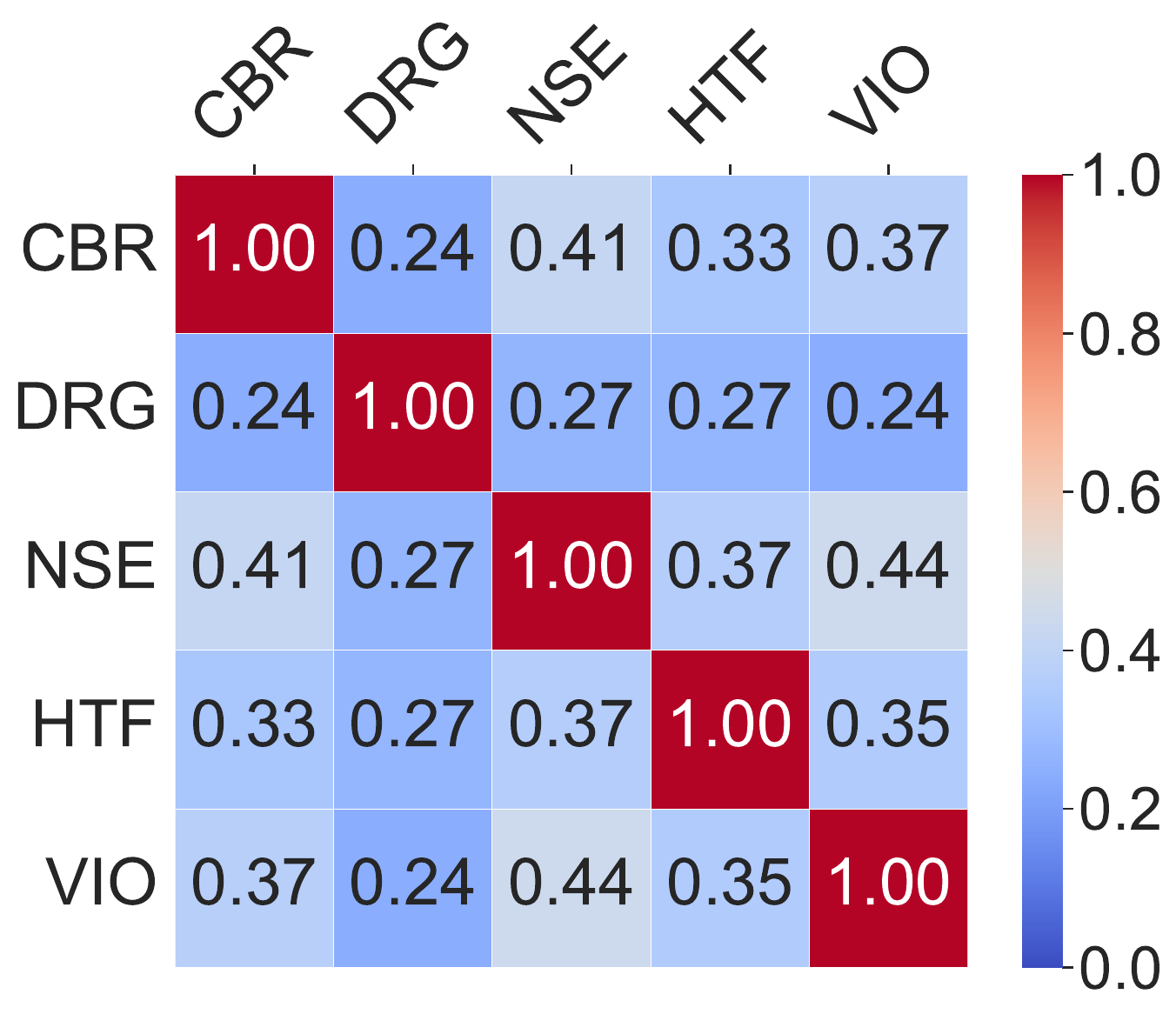}
    \caption{Method: SNIP; Targeted model: Llama2-7B-Chat. Left: $q\%=1\%, r=0.53$, with $p$-value $0.11$; right: $q\%=1\%, p\%=1\%, r=0.52$, with $p$-value $0.12$.}
    \label{fig:appendix_pairwise_single_cat_snip_wanda_1}
    \end{subfigure}
    \hfill
    \begin{subfigure}{0.475\textwidth}
    \centering
    \includegraphics[width=0.49\linewidth]{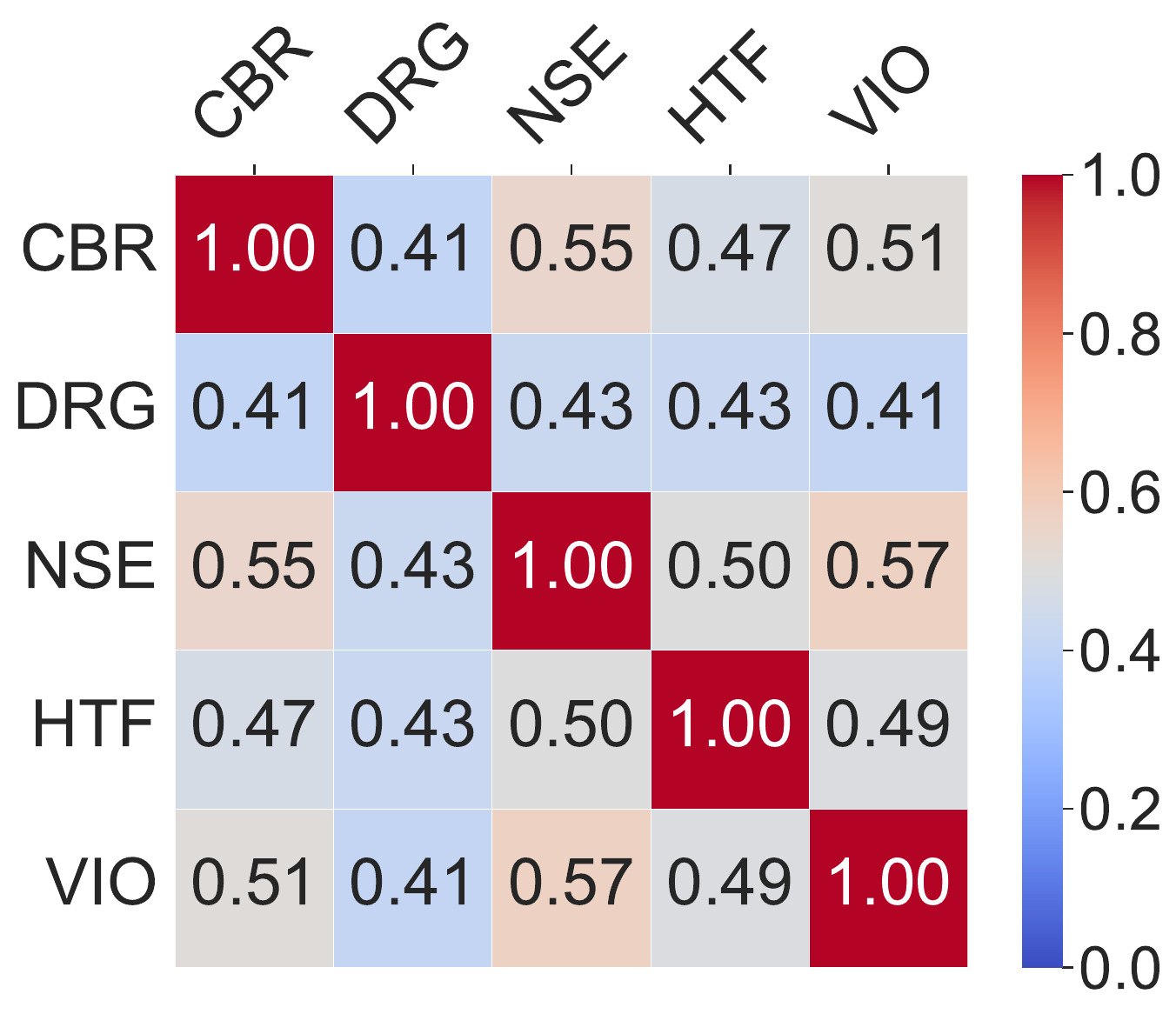}
    \hfill
    \includegraphics[width=0.49\linewidth]{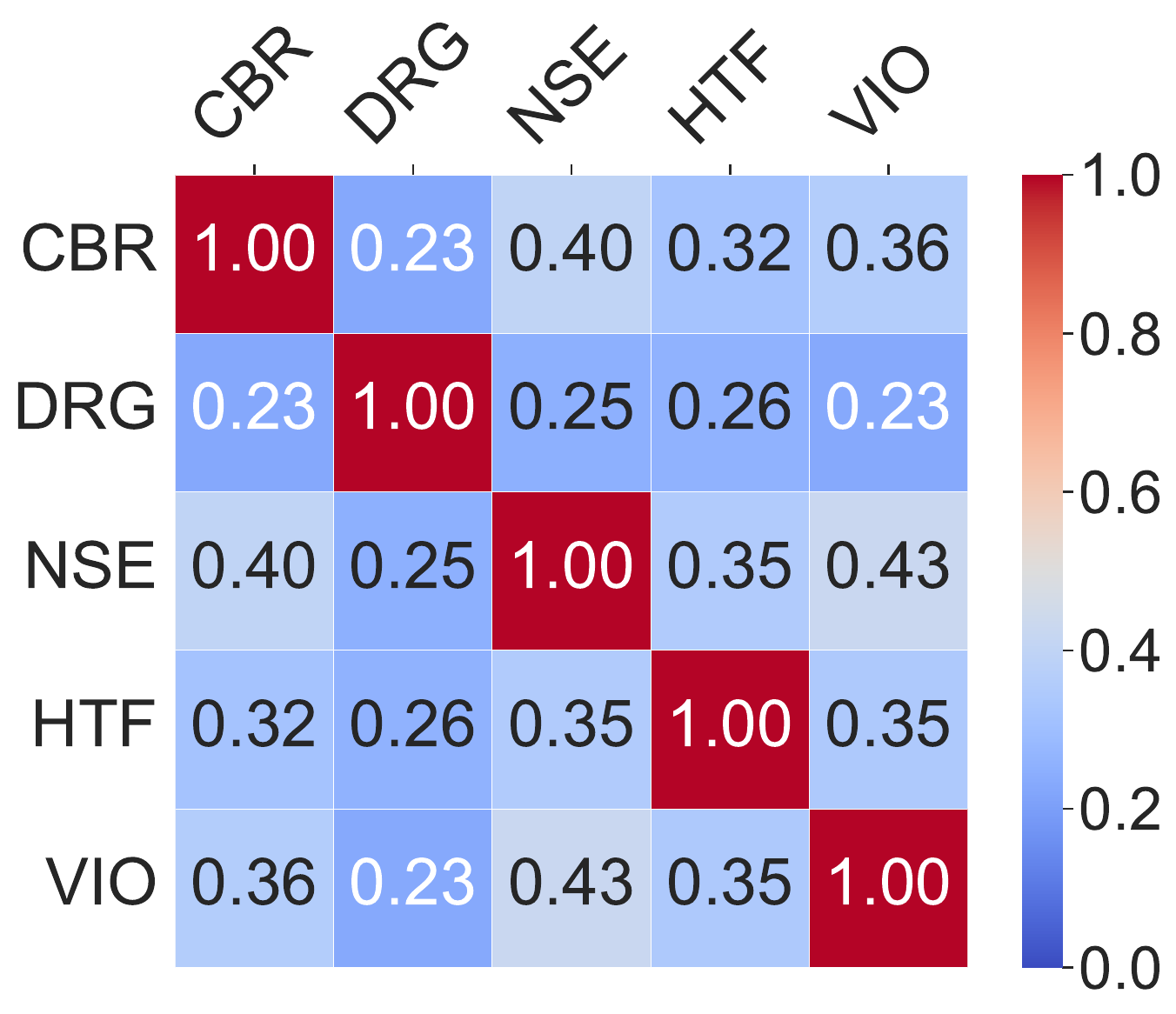}
    \caption{Method: SNIP; Targeted model: Llama2-7B-Chat. Left: $q\%=3\%, r=0.55$, with $p$-value $0.1$; right: $q\%=3\%, p\%=7\%, r=0.51$, with $p$-value $0.13$.}
    \label{fig:appendix_pairwise_single_cat_snip_wanda_2}
    \end{subfigure}

    \vspace{0.2cm}

    \begin{subfigure}{0.475\textwidth}
    \includegraphics[width=0.49\linewidth]{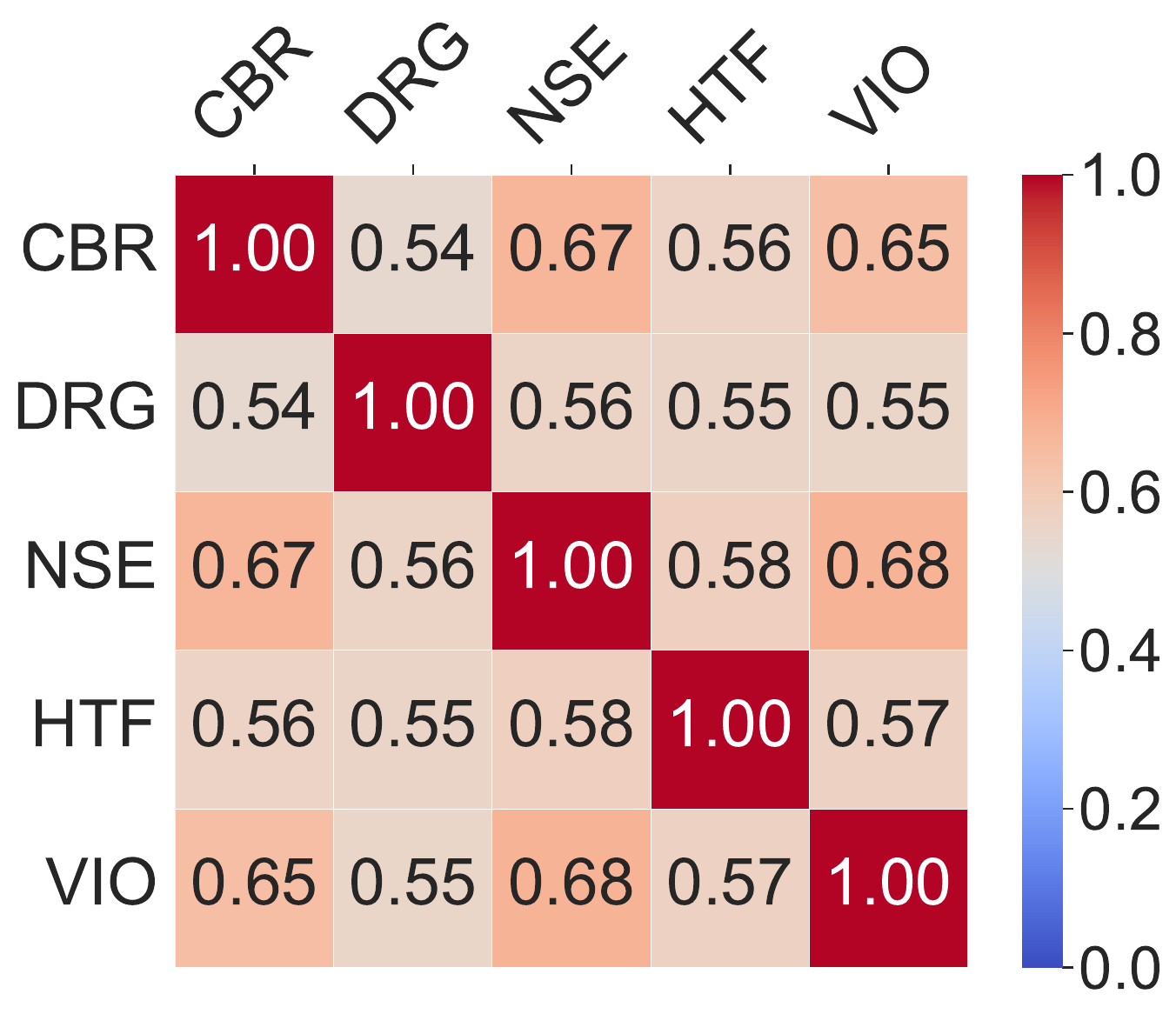}
    \hfill
    \includegraphics[width=0.49\linewidth]{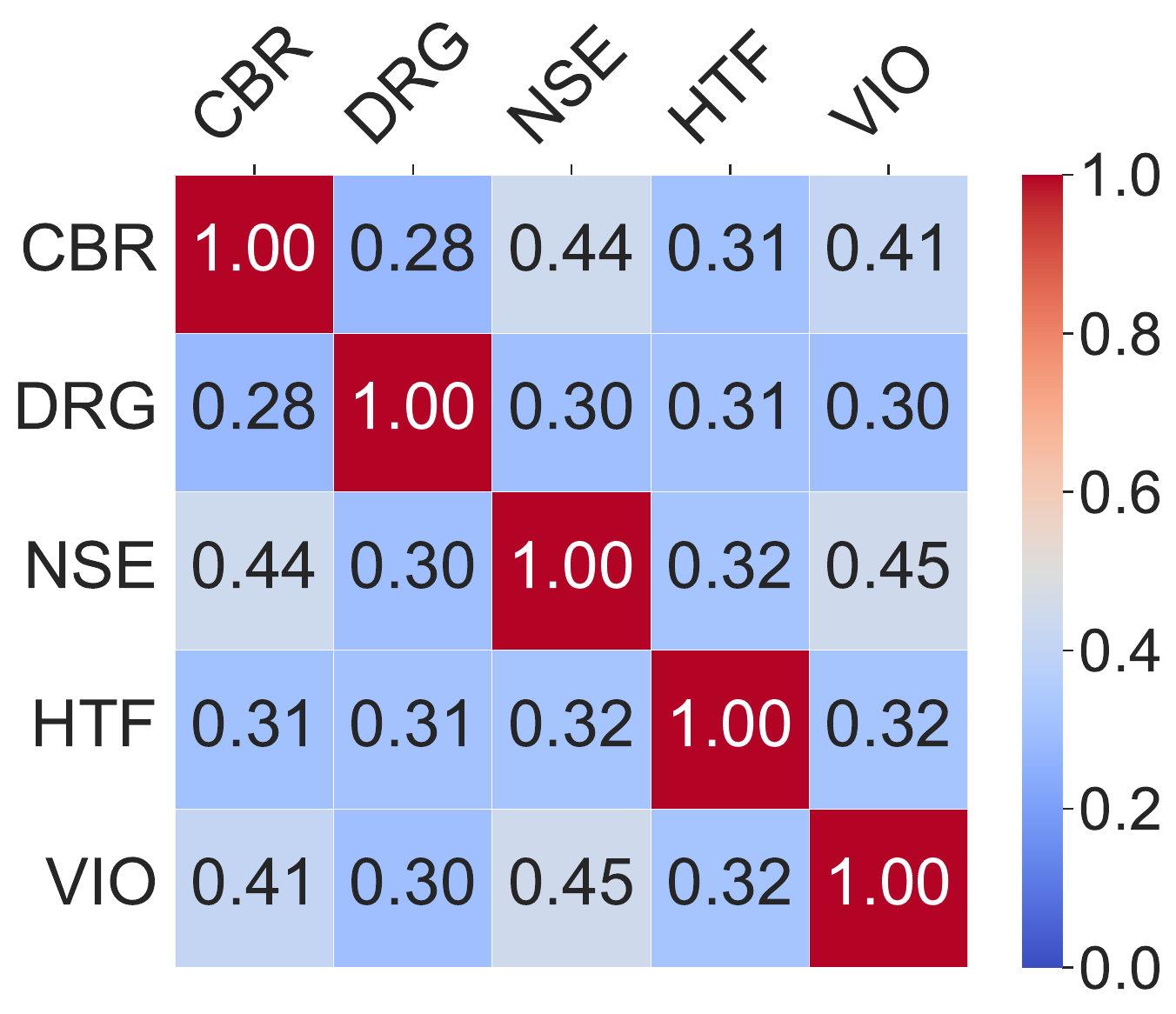}
    \caption{Method: Wanda; Targeted model: Llama2-7B-Chat. Left: $q\%=1\%, r=0.75$, with $p$-value $0.01$; right: $q\%=1\%, p\%=1\%, r=0.77$, with $p$-value $0.01$.}
    \label{fig:appendix_pairwise_single_cat_snip_wanda_3}
    \end{subfigure}
    \hfill
    \begin{subfigure}{0.475\textwidth}
    \includegraphics[width=0.49\linewidth]{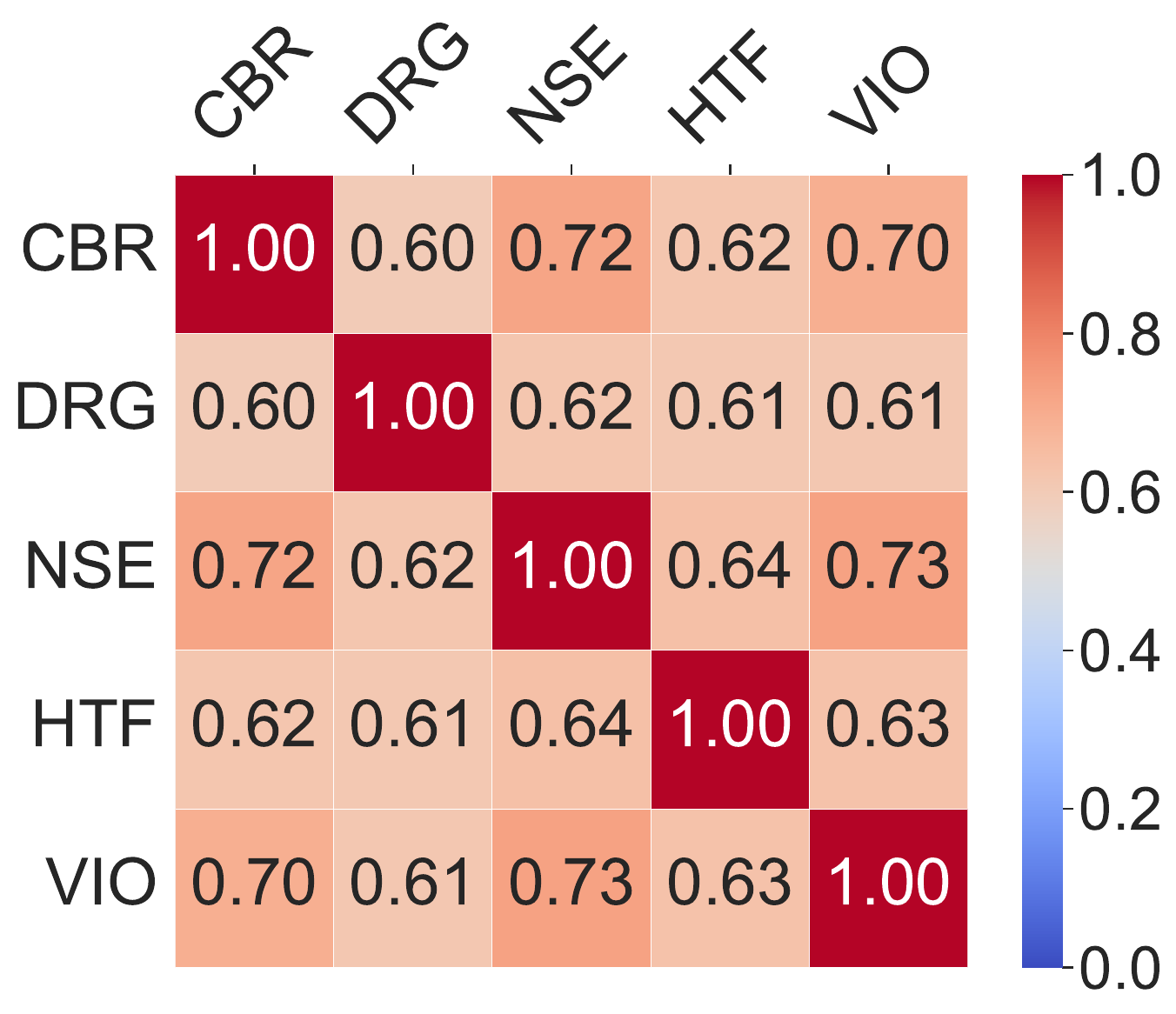}
    \hfill
    \includegraphics[width=0.49\linewidth]{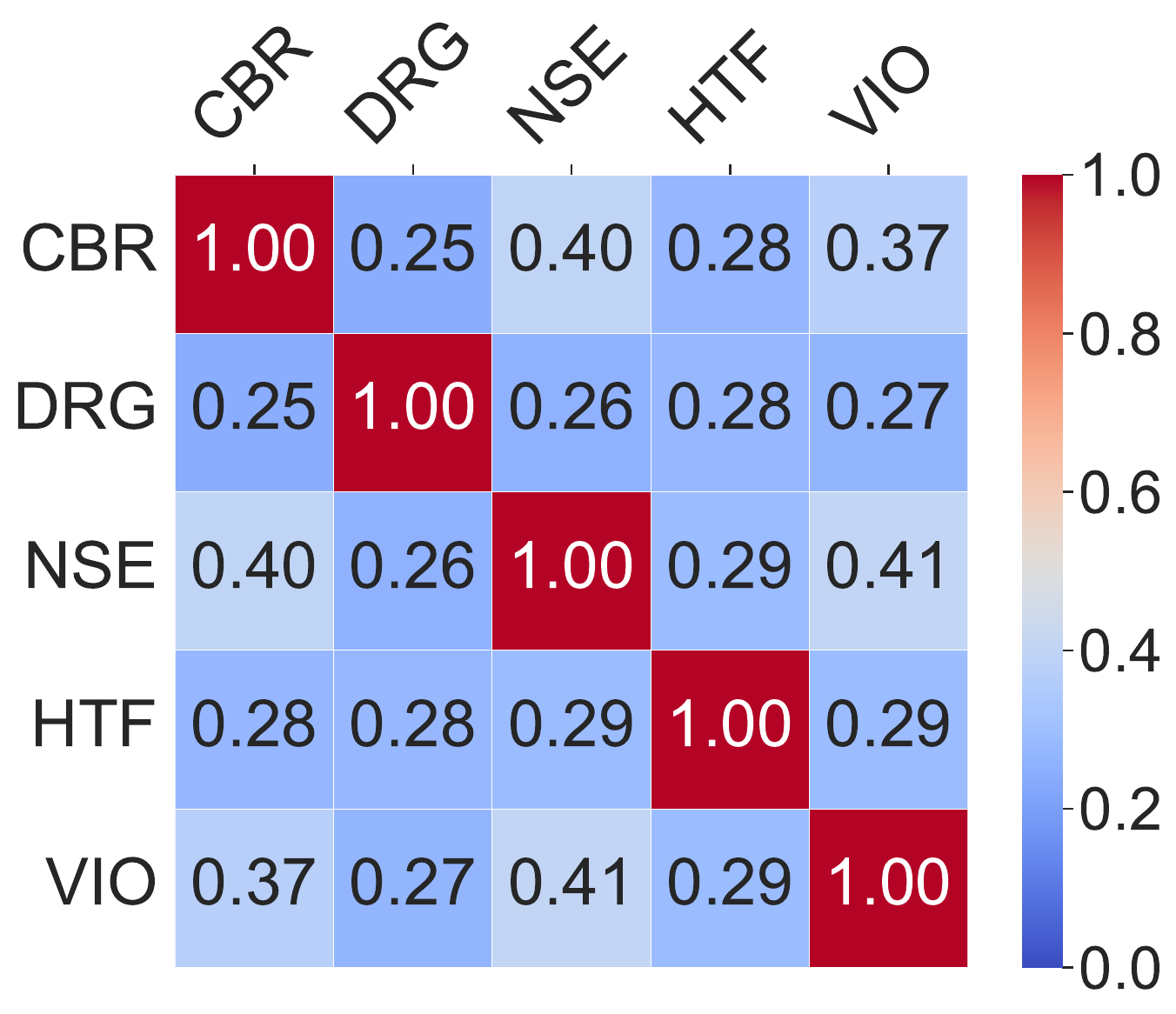}
    \caption{Method: Wanda; Targeted model: Llama2-7B-Chat. Left: $q\%=3\%, r=0.74$, with $p$-value $0.01$; right: $q\%=3\%, p\%=7\%, r=0.75$, with $p$-value $0.01$.}
    \label{fig:appendix_pairwise_single_cat_snip_wanda_4}
    \end{subfigure}

    \vspace{0.2cm}

    \begin{subfigure}{0.475\textwidth}
    \includegraphics[width=0.49\linewidth]{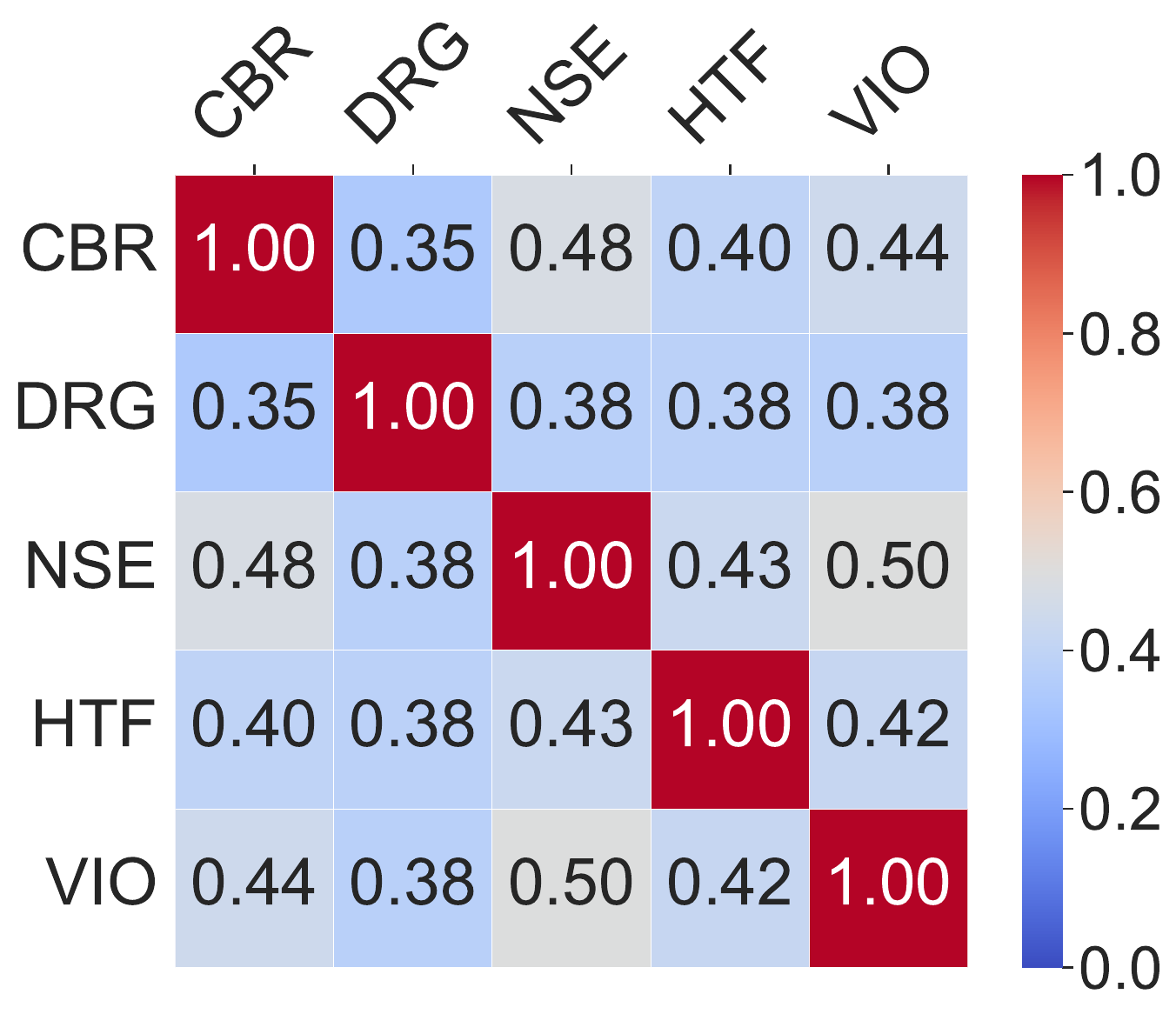}
    \hfill
    \includegraphics[width=0.49\linewidth]{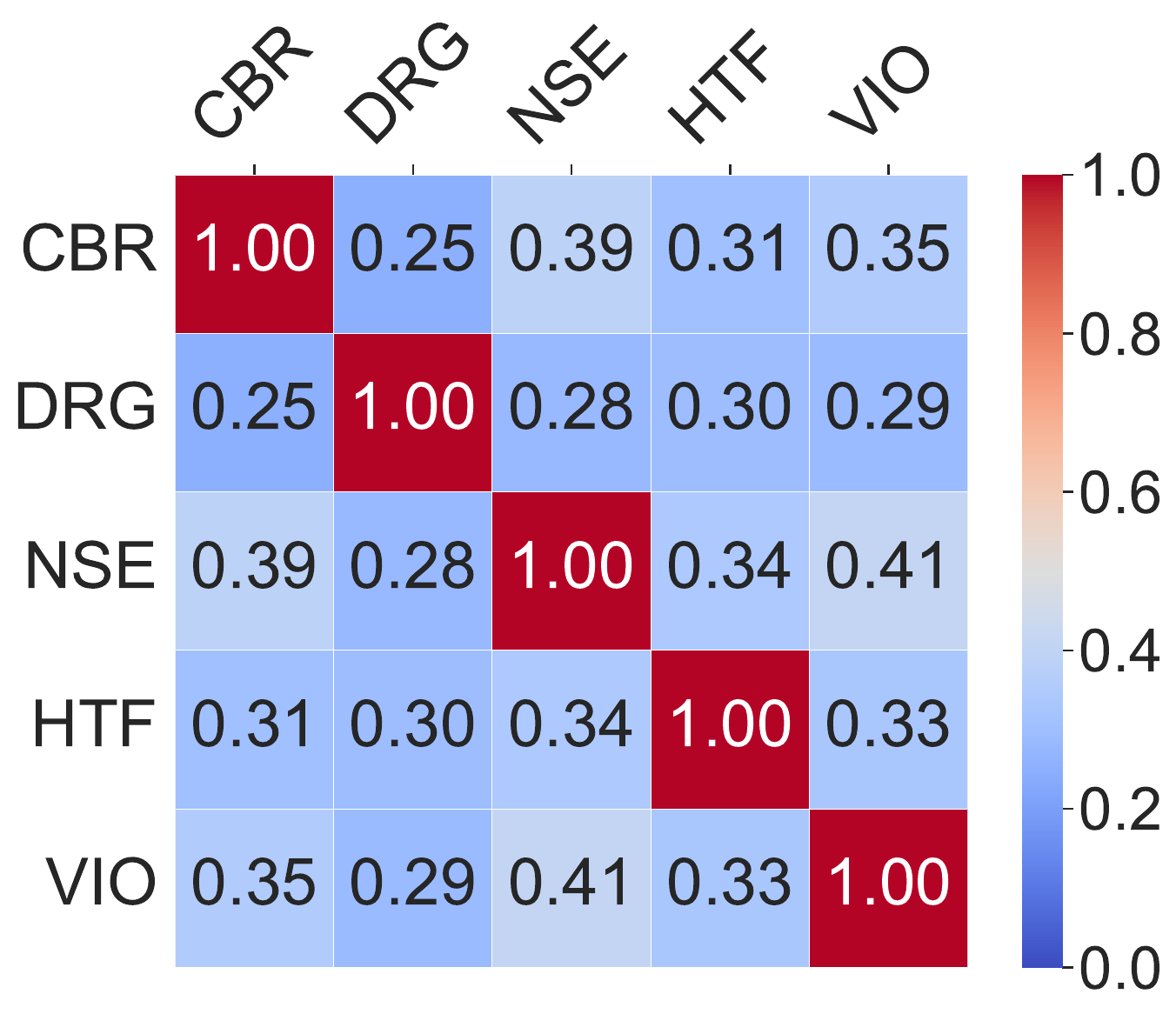}
    \caption{Method: SNIP; Targeted model: Llama2-13B-Chat. Left: $q\%=1\%, r=0.67$, with $p$-value $0.03$; right: $q\%=1\%, p\%=1\%, r=0.68$, with $p$-value $0.03$.}
    \label{fig:appendix_pairwise_single_cat_snip_wanda_5}
    \end{subfigure}
    \hfill
    \begin{subfigure}{0.475\textwidth}
    \includegraphics[width=0.49\linewidth]{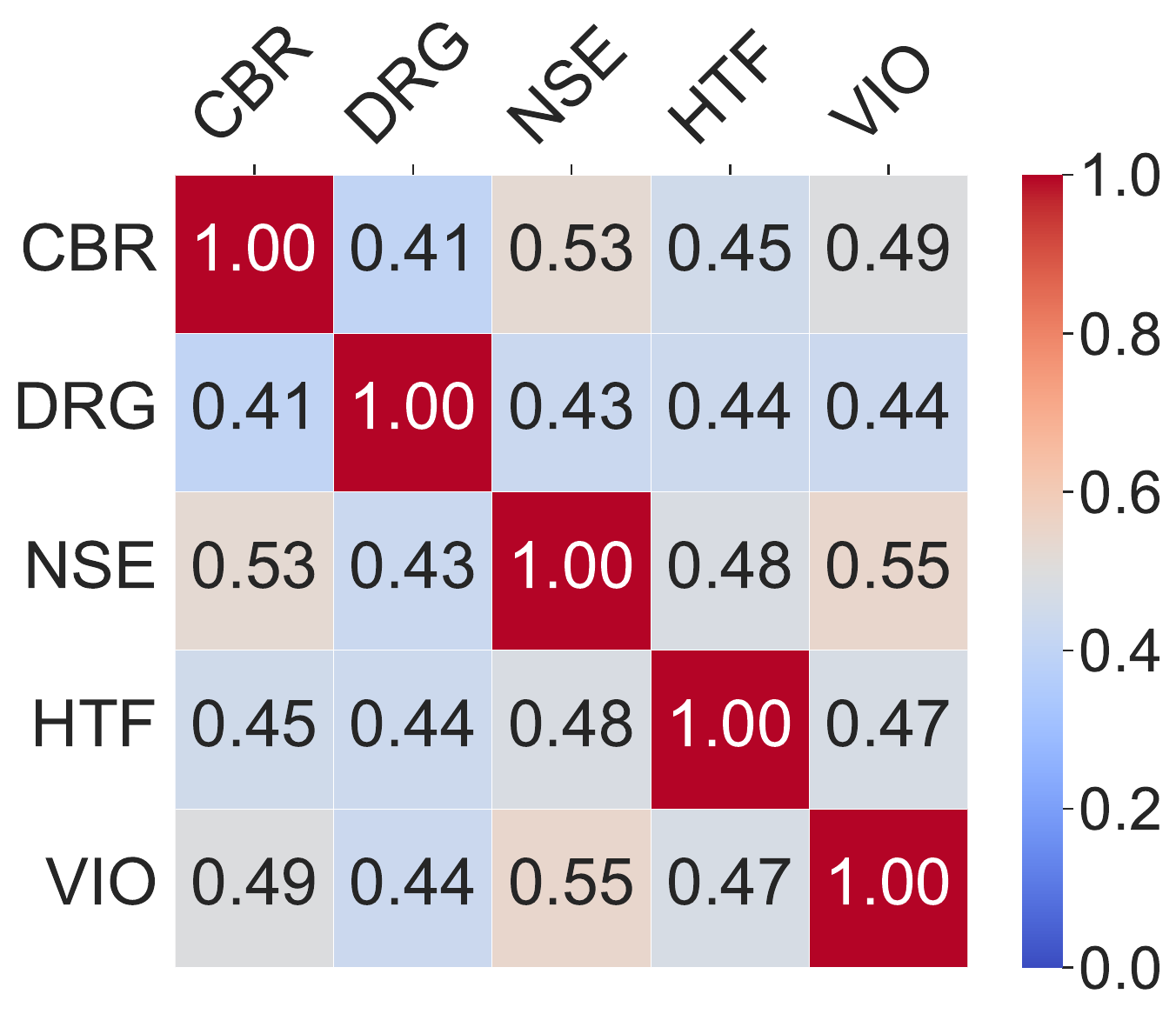}
    \hfill
    \includegraphics[width=0.49\linewidth]{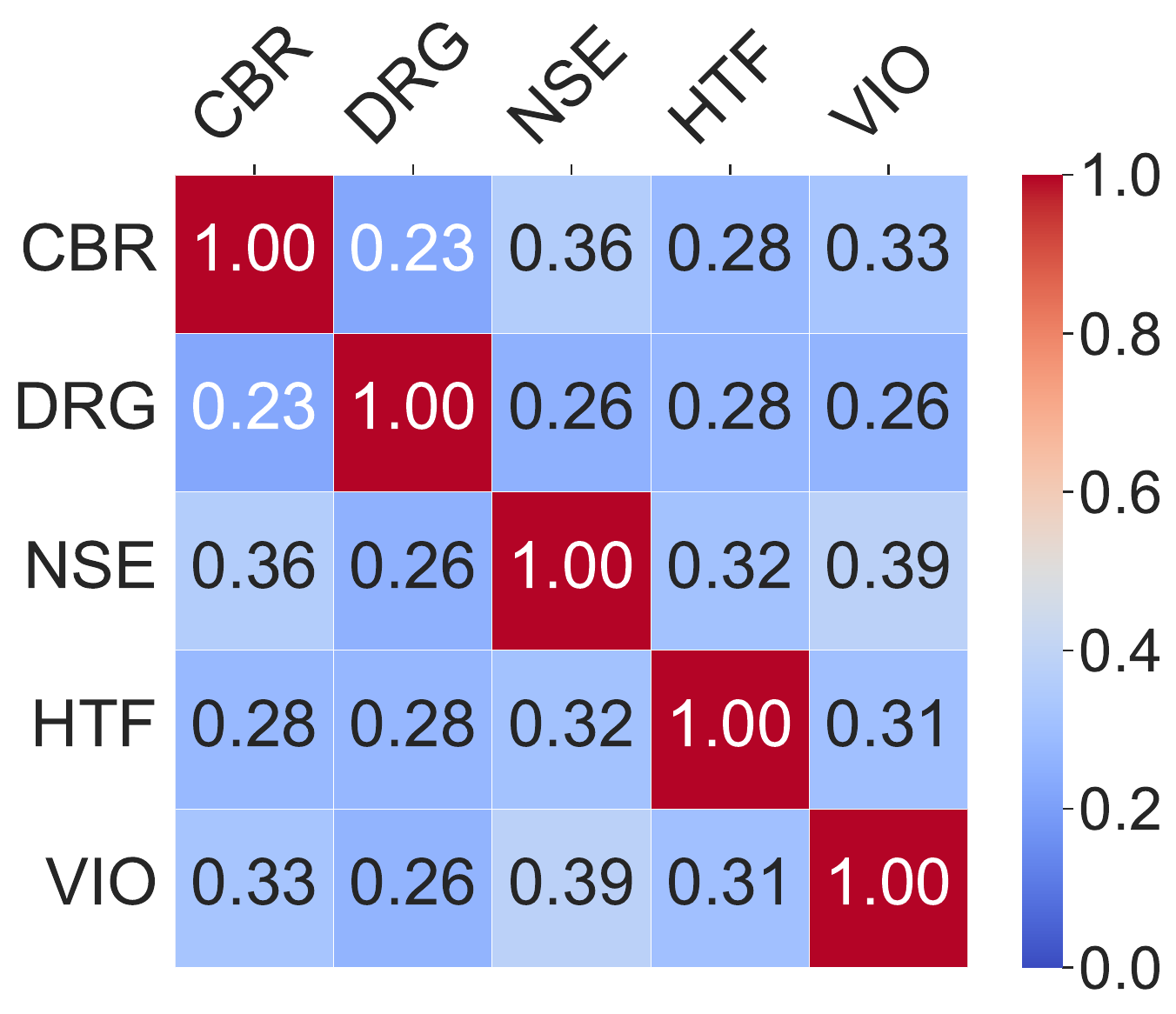}
    \caption{Method: SNIP; Targeted model: Llama2-13B-Chat. Left: $q\%=3\%, r=0.68$, with $p$-value $0.03$; right: $q\%=3\%, p\%=8\%, r=0.68$, with $p$-value $0.03$.}
    \label{fig:appendix_pairwise_single_cat_snip_wanda_6}
    \end{subfigure}

    \vspace{0.2cm}

    \begin{subfigure}{0.475\textwidth}
    \includegraphics[width=0.49\linewidth]{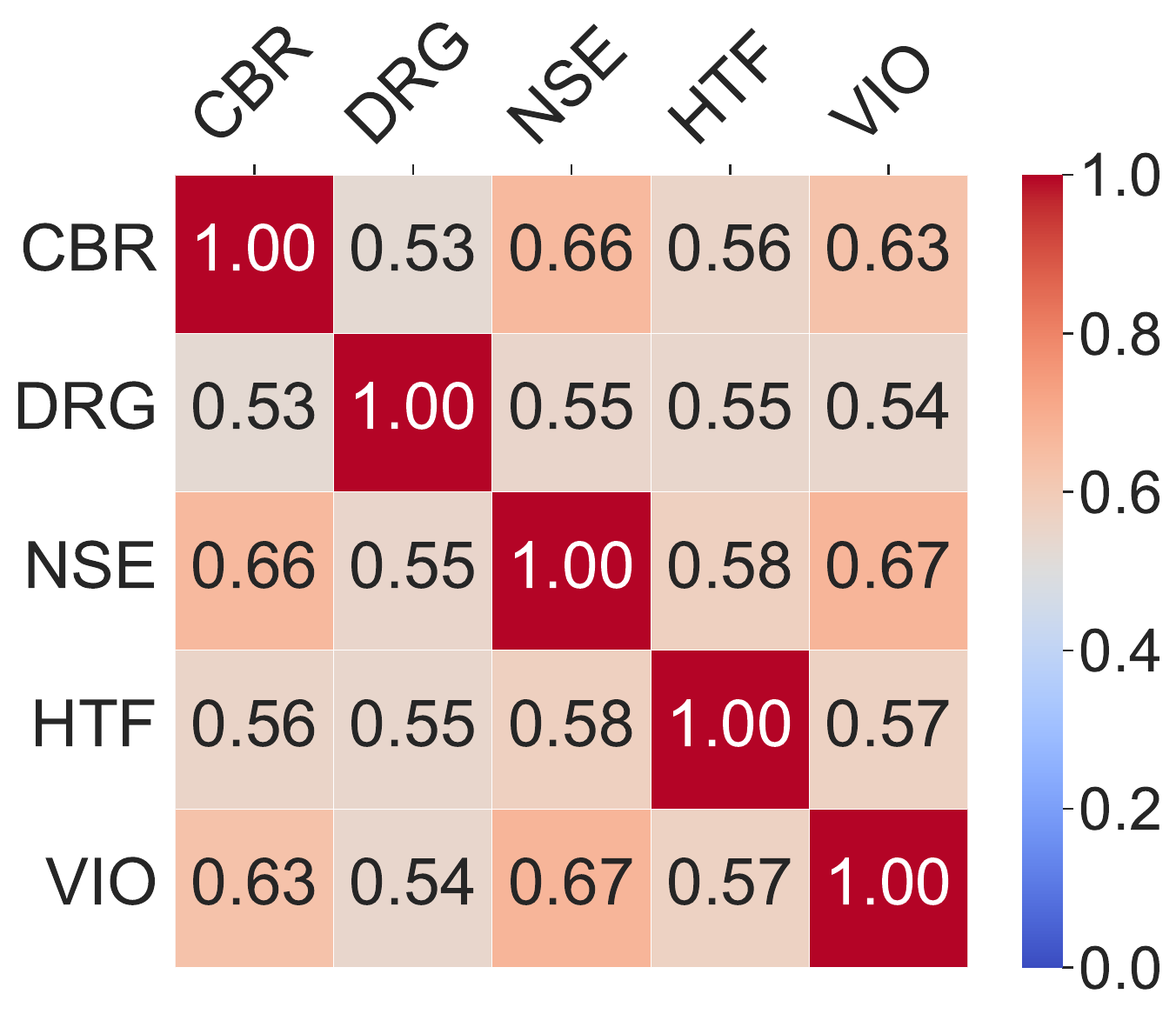}
    \hfill
    \includegraphics[width=0.49\linewidth]{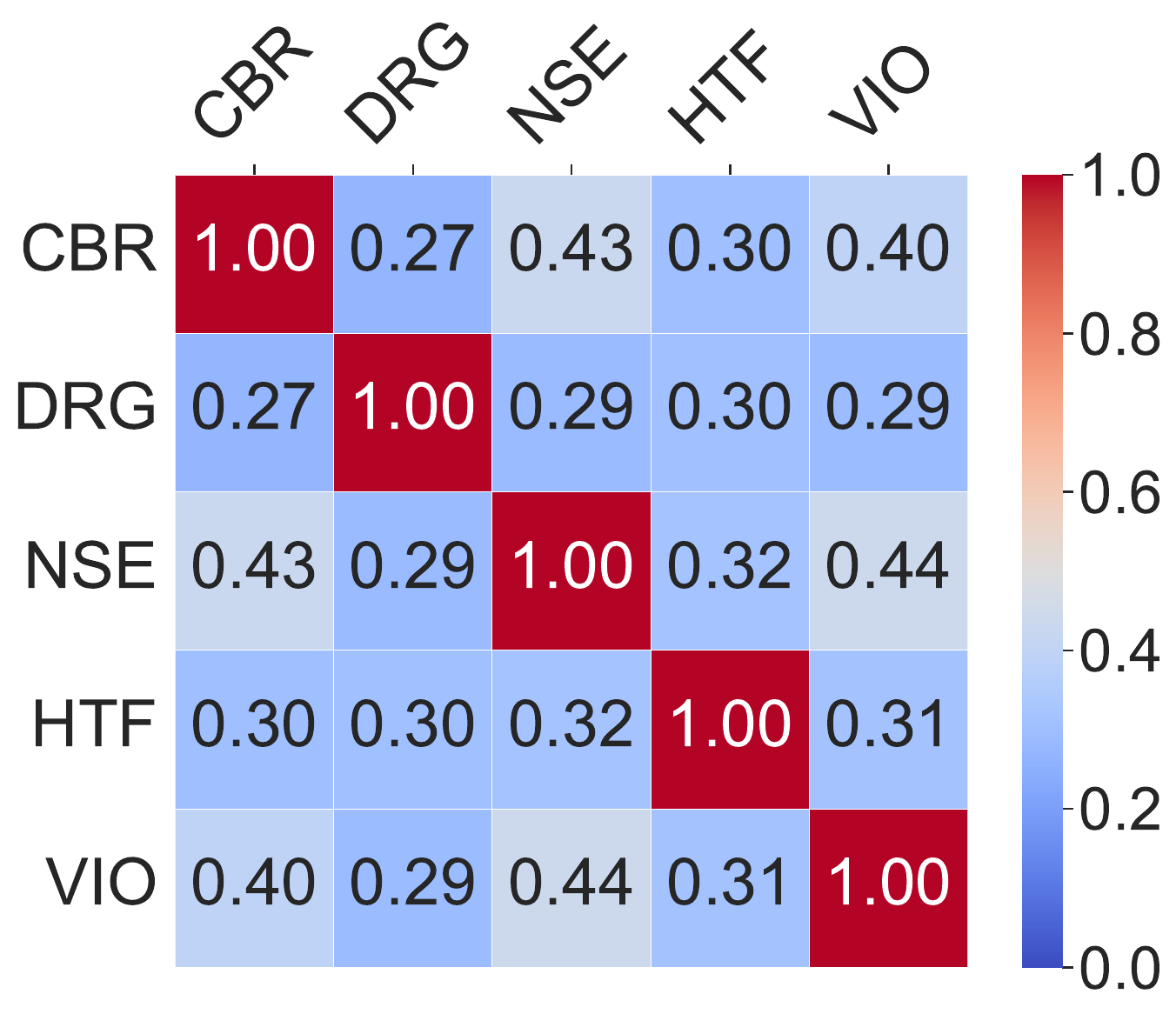}
    \caption{Method: Wanda; Targeted model: Llama2-13B-Chat. Left: $q\%=1\%, r=0.74$, with $p$-value $0.01$; right: $q\%=1\%, p\%=1\%, r=0.77$, with $p$-value $0.01$.}
    \label{fig:appendix_pairwise_single_cat_snip_wanda_7}
    \end{subfigure}
    \hfill
    \begin{subfigure}{0.475\textwidth}
    \includegraphics[width=0.49\linewidth]{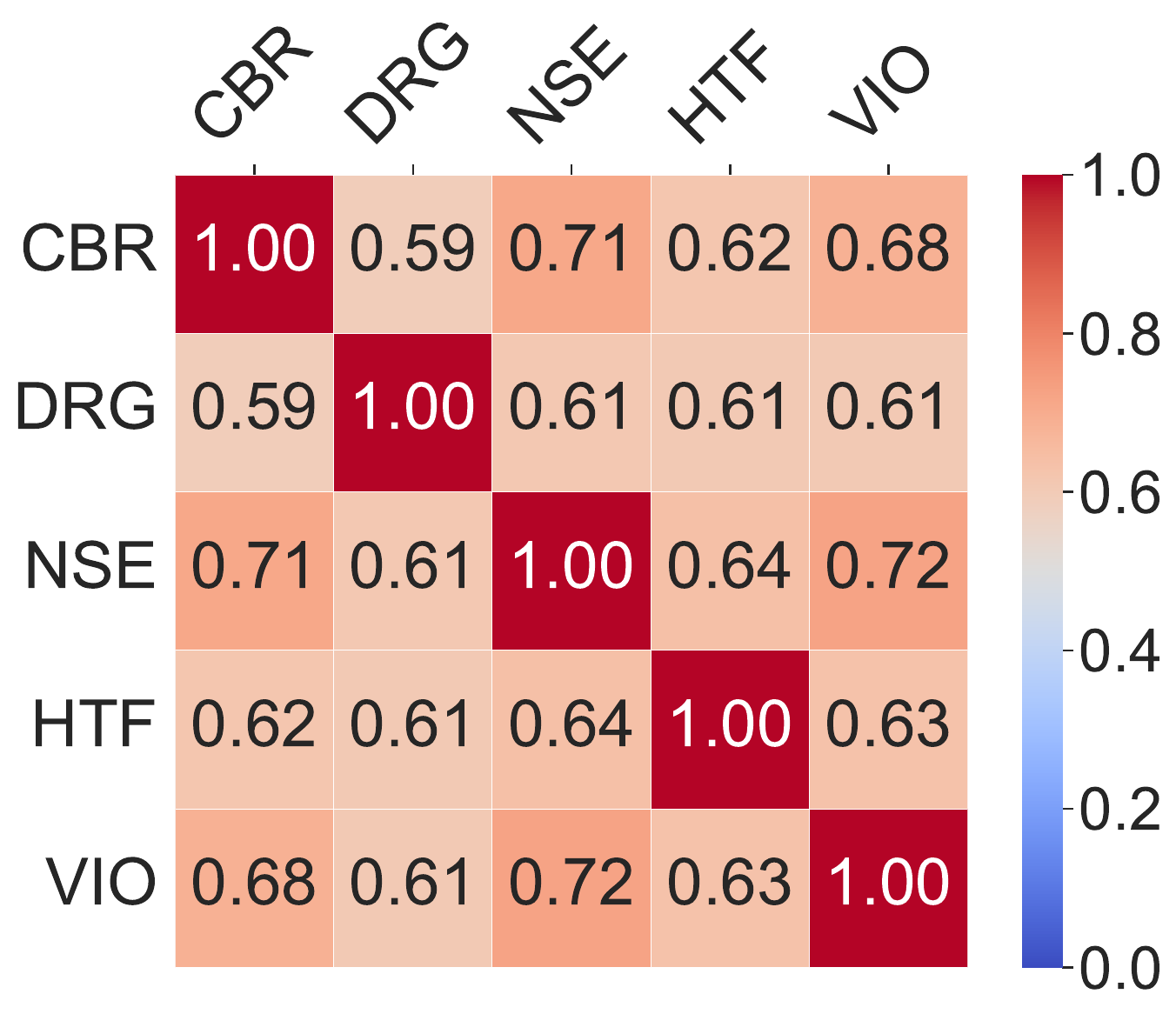}
    \hfill
    \includegraphics[width=0.49\linewidth]{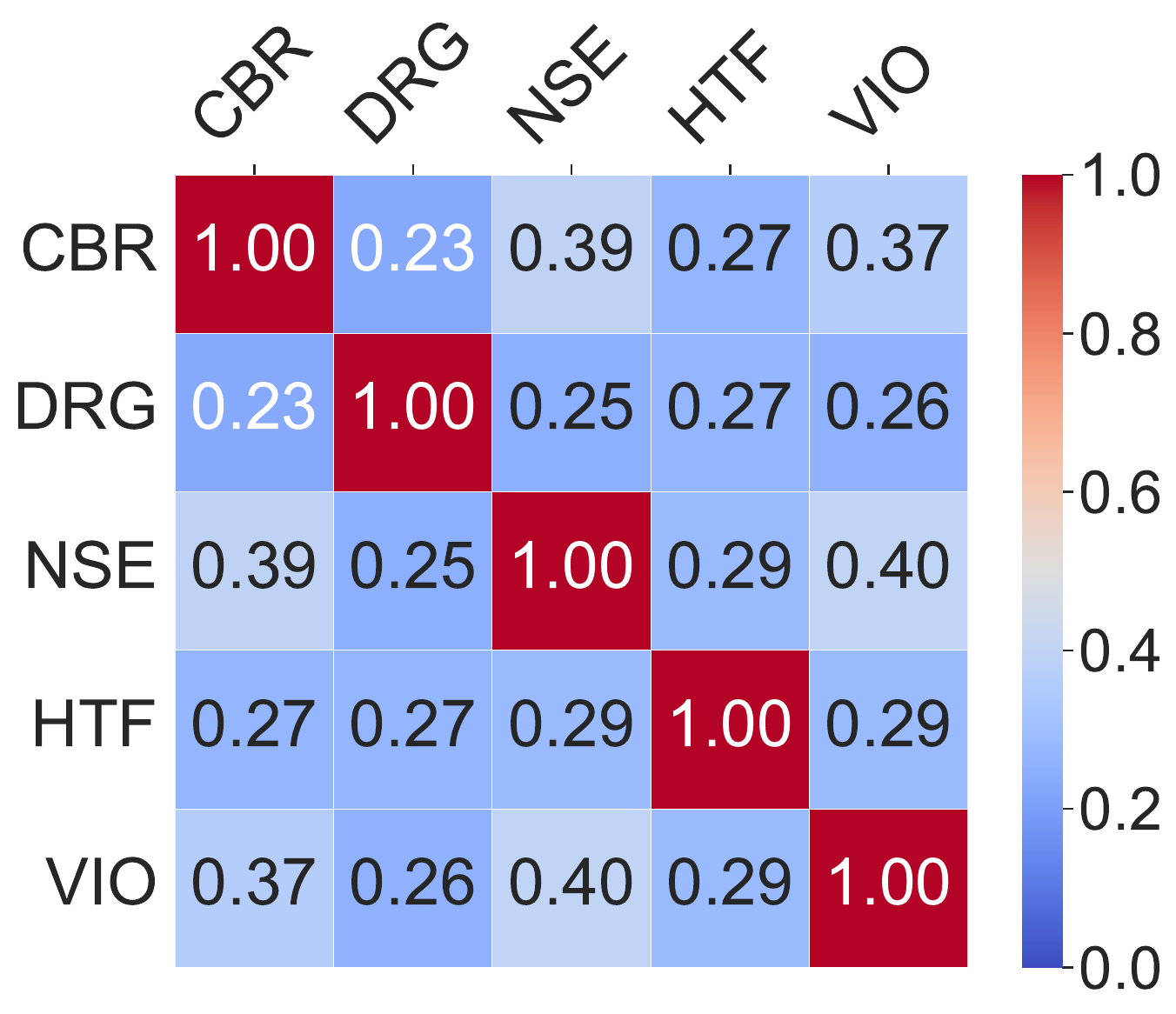}
    \caption{Method: Wanda; Targeted model: Llama2-13B-Chat. Left: $q\%=3\%, r=0.74$, with $p$-value $0.01$; right: $q\%=3\%, p\%=8\%, r=0.75$, with $p$-value $0.01$.}
    \label{fig:appendix_pairwise_single_cat_snip_wanda_8}
    \end{subfigure}

\caption{Pairwise safety region overlap with single-category identification datasets for SNIP \& Wanda. For each subfigure, the left corresponds to the pairwise overlap between safety regions; the right corresponds to the pairwise overlap between utility-isolated safety regions. We report the Pearson correlation $r$ between dataset cosine similarity and pairwise overlap. 
}
\label{fig:appendix_pairwise_single_cat_snip_wanda}
\end{figure*}

\begin{figure*}[htbp!]
    \centering
    \begin{subfigure}{\textwidth}
    \includegraphics[width=0.48\linewidth]{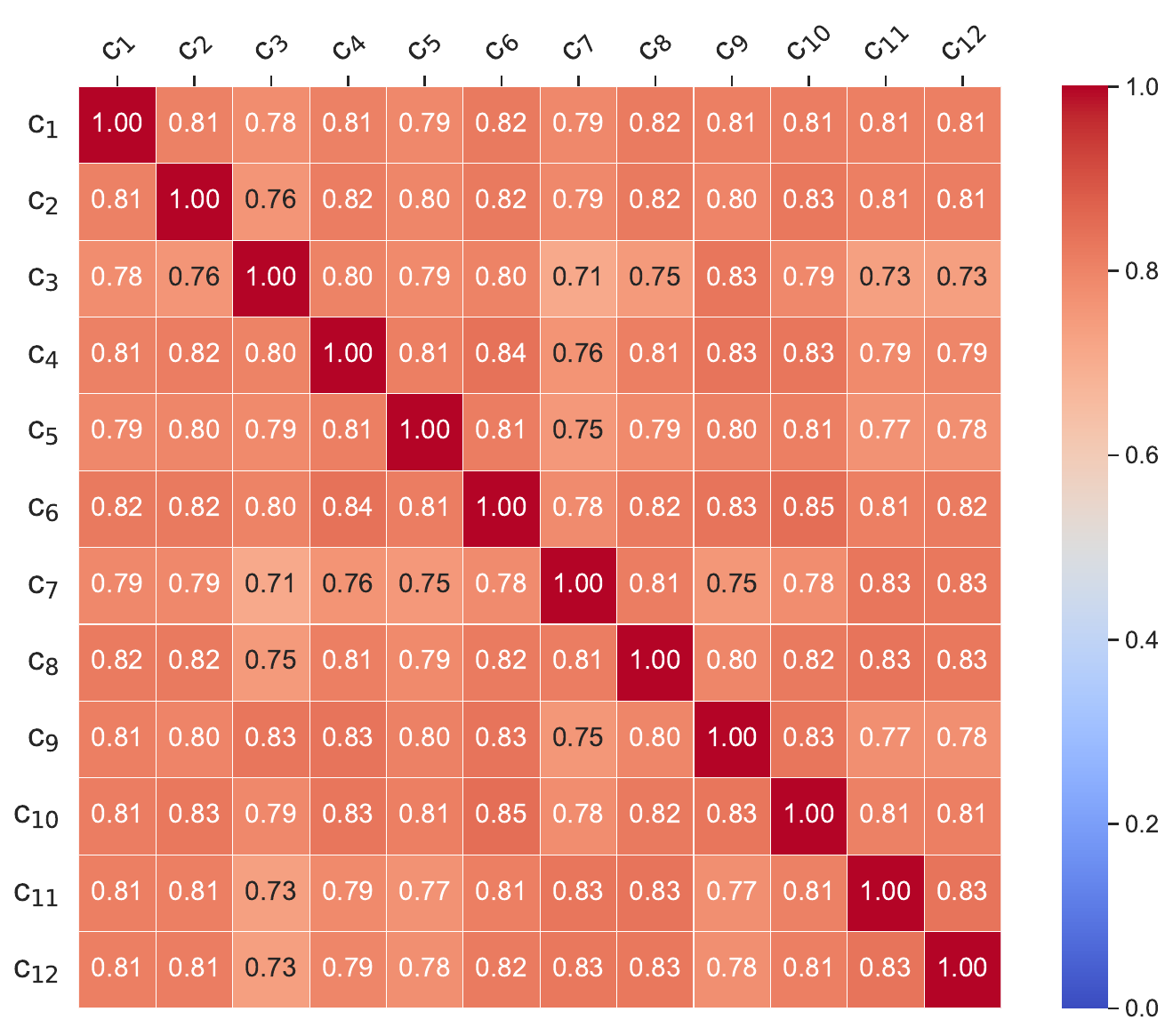}
    \hfill
    \includegraphics[width=0.48\linewidth]{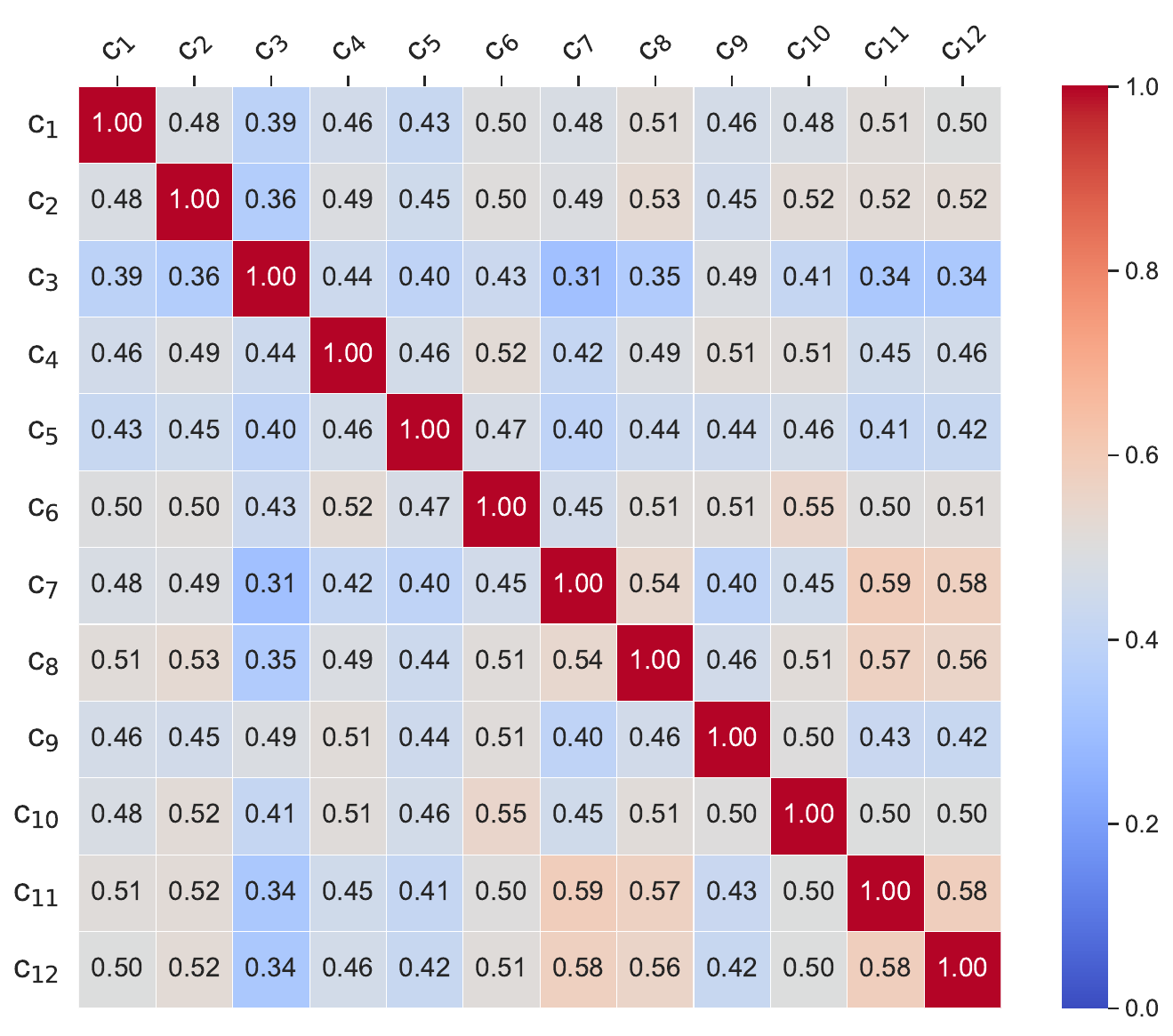}
    \caption{Llama-2-7B-Chat. Left: $r=0.28$, with $p$-value $0.02$; right: $r=0.28$, with $p$-value $0.02$.}
    \label{fig:appendix_pairwise_single_cat_safeneuron_1}
    \end{subfigure}

    \vspace{0.2cm}

    \begin{subfigure}{\textwidth}
    \includegraphics[width=0.48\linewidth]{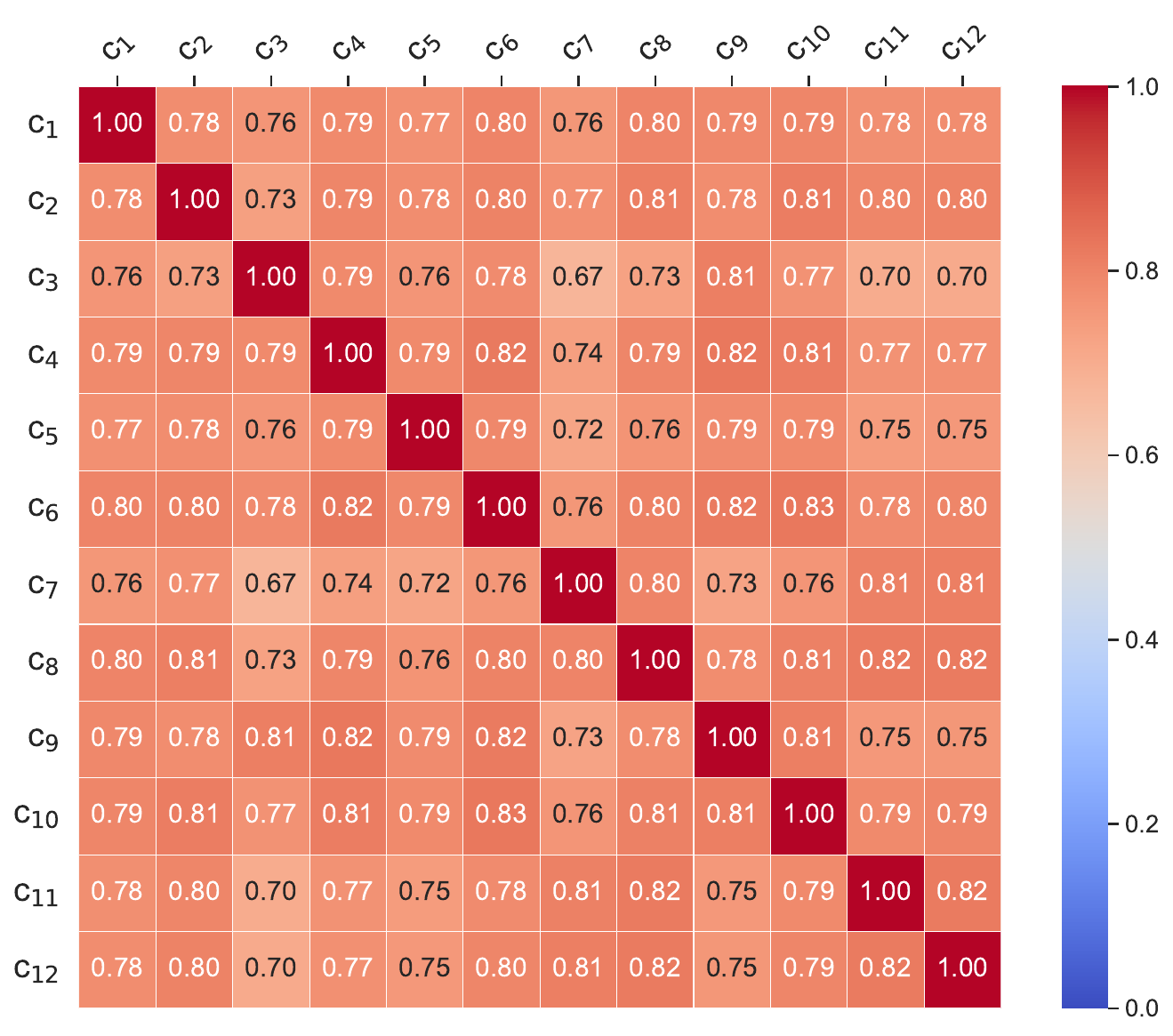}
    \hfill
    \includegraphics[width=0.48\linewidth]{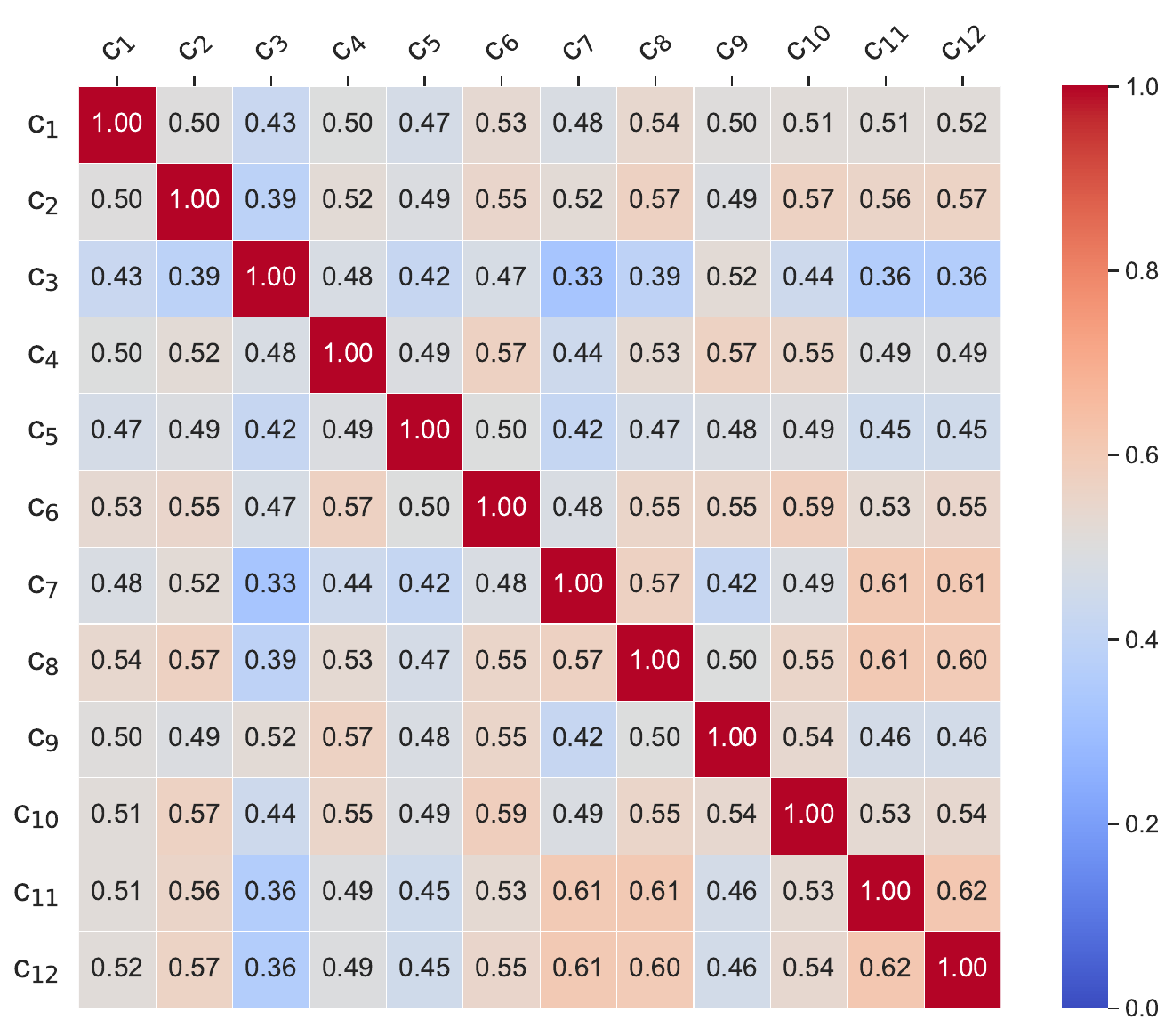}
    \caption{Llama-3-8B-Instruct. Left: $r=0.28$, with $p$-value $0.02$; right: $r=0.28$, with $p$-value $0.02$.}
    \label{fig:appendix_pairwise_single_cat_safeneuron_2}
    \end{subfigure}

    \vspace{0.2cm}

    \begin{subfigure}{\textwidth}
    \includegraphics[width=0.48\linewidth]{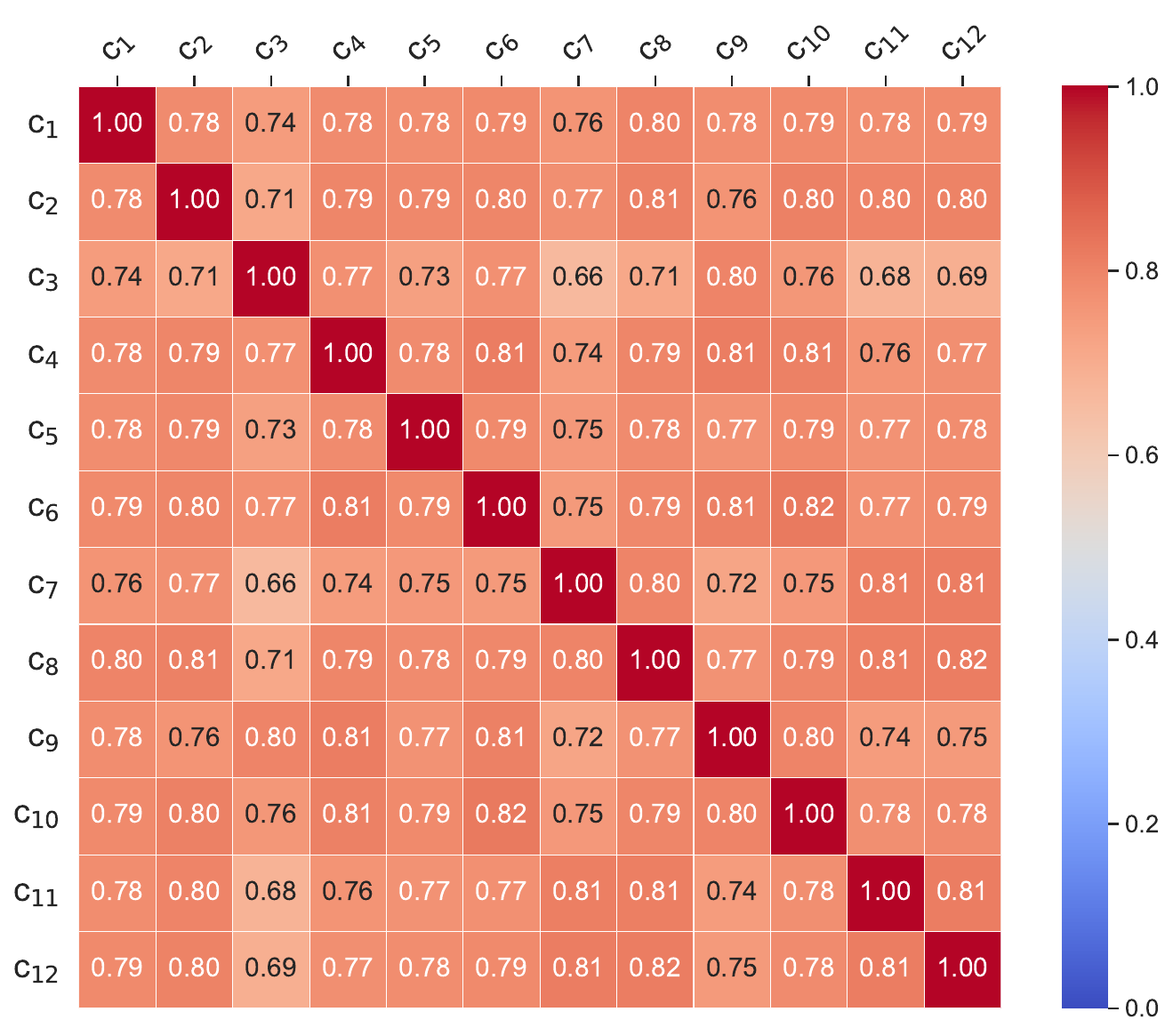}
    \hfill
    \includegraphics[width=0.48\linewidth]{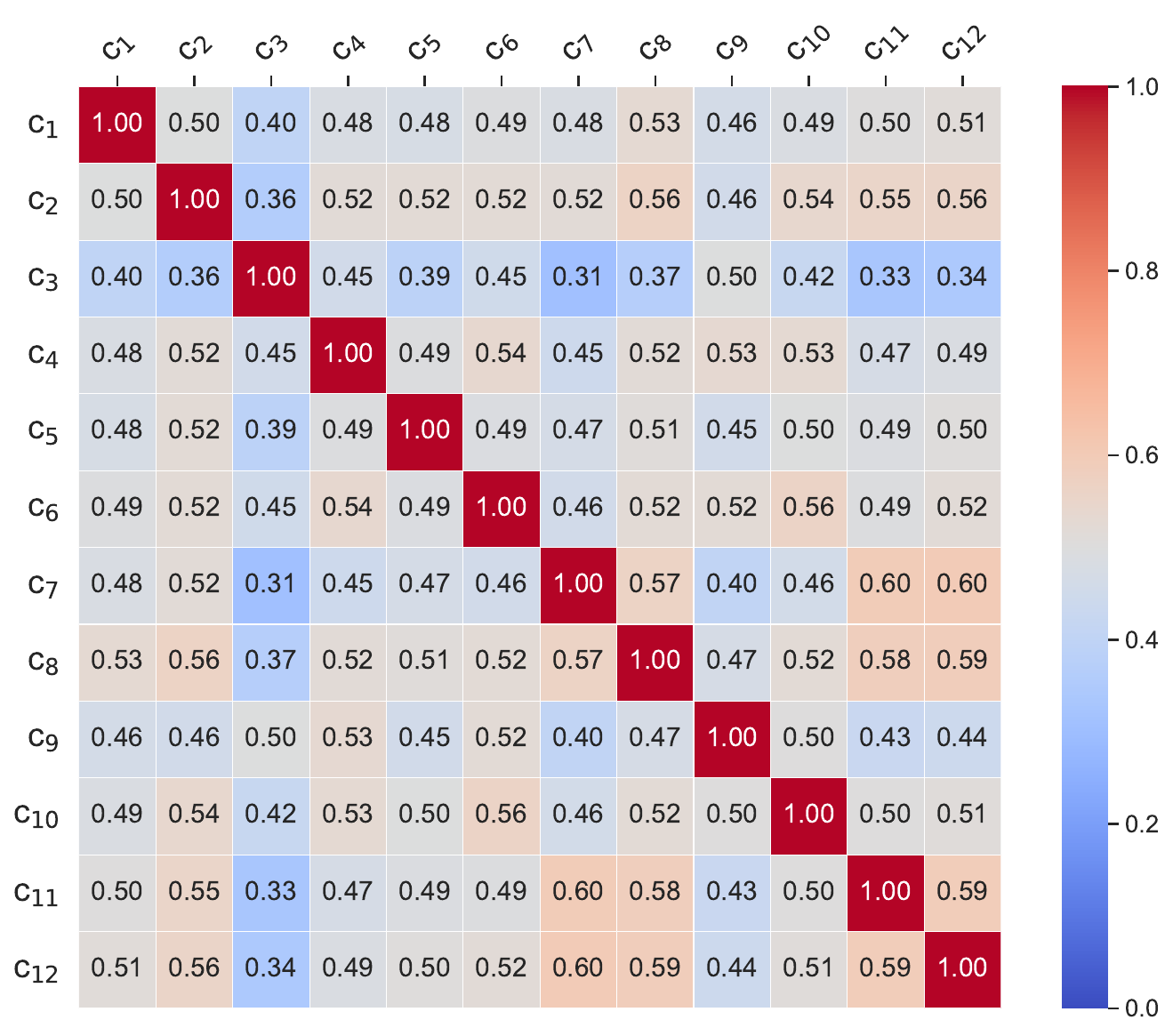}
    \caption{Mistral-7B-Instruct-v0.2. Left: $r=0.12$, with $p$-value $0.35$; right: $r=0.12$, with $p$-value $0.35$.}
    \label{fig:appendix_pairwise_single_cat_safeneuron_3}
    \end{subfigure}

\caption{Pairwise safety region overlap with single-category identification datasets for SafeNeuron. For each subfigure, the left corresponds to the pairwise overlap between safety regions; the right corresponds to the pairwise overlap between utility-isolated safety regions. We report the Pearson correlation $r$ between dataset cosine similarity and pairwise overlap.
}

\label{fig:appendix_pairwise_single_cat_safeneuron}
\end{figure*}

Fig.~\ref{fig:appendix_pairwise_single_cat_snip_wanda} and~\ref{fig:appendix_pairwise_single_cat_safeneuron} show additional safety region pairwise overlaps with single-category datasets and the Pearson correlation between the pairwise overlap and dataset semantic cosine similarity. For the single-category datasets used in each method, the corresponding semantic cosine similarity analysis is presented in Fig.~\ref{fig:appendix_cos_sim_single}.

\section{Details about $\mathcal{D}_0$ for each method}
\label{appendix:d0}

For each method, we denote by $\mathcal{D}_0$ the identification dataset used in its original paper.
Specifically, the sources of $\mathcal{D}_0$ for each method are as follows: 
\begin{itemize}
    \item \textbf{SNIP \& Wanda}: AdvBench by~\citet{zou2023universaltransferableadversarialattacks}.
    \item \textbf{SafeNeuron}: the training split from~\citet{zou2024improving}.
    \item \textbf{NLSR}: the test split of PKU-SafeRLHF-30K~\cite{beavertails, pkusaferlhf}.
\end{itemize}

\end{document}